\documentclass{article}

\PassOptionsToPackage{numbers}{natbib}

\usepackage[final]{neurips_2022}

\usepackage[utf8]{inputenc} 
\usepackage[T1]{fontenc}    
\usepackage{hyperref}       
\usepackage{url}            
\usepackage{booktabs}       
\usepackage{amsfonts}       
\usepackage{nicefrac}       
\usepackage{microtype}      
\usepackage[table]{xcolor}         

\usepackage{graphicx}
\usepackage{subcaption}

\usepackage{comment}
\usepackage{amsmath}
\usepackage{makecell}
\usepackage{amssymb}
\usepackage{mathtools}
\usepackage{amsthm}
\usepackage{bbm}
\usepackage{enumitem}
\usepackage[capitalize,noabbrev]{cleveref}
\usepackage{kotex} 
\usepackage{multicol}
\usepackage{multirow}
\usepackage{csquotes}
\usepackage{wrapfig}

\usepackage{pifont}
\theoremstyle{plain}
\newtheorem{theorem}{Theorem}[section]

\theoremstyle{definition}

\theoremstyle{remark}

\newtheorem{observation}[theorem]{\underline{\scshape Observation}}

\DeclareFixedFont{\ttb}{T1}{txtt}{bx}{n}{12} 
\DeclareFixedFont{\ttm}{T1}{txtt}{m}{n}{12}  

\usepackage{color}
\definecolor{deepblue}{rgb}{0,0,0.5}
\definecolor{deepred}{rgb}{0.6,0,0}
\definecolor{deepgreen}{rgb}{0,0.5,0}

\usepackage{listings}

\newcommand\pythonstyle{\lstset{
language=Python,
basicstyle=\ttm\footnotesize,
morekeywords={self},              
keywordstyle=\ttb\color{deepblue}\footnotesize,
emph={parse_transform, get_composed_transform, transforms},          
emphstyle=\ttb\color{deepred}\footnotesize,    
stringstyle=\color{deepgreen}\footnotesize,
frame=tb,                         
showstringspaces=false,
commentstyle=\itshape,
morestring=[s]{`}{'},
frame=single,
frameround={t}{t}{t}{t},
numberstyle=\tiny\color{gray},
}}

\lstnewenvironment{python}[1][]
{
\pythonstyle
\lstset{#1}
}
{}

\newcommand\pythoninline[1]{{\pythonstyle\lstinline!#1!}}

\newcolumntype{L}[1]{>{\raggedright\let\newline\\\arraybackslash\hspace{0pt}}m{#1}}
\newcolumntype{X}[1]{>{\centering\let\newline\\\arraybackslash\hspace{0pt}}p{#1}}

\usepackage[textsize=tiny]{todonotes}

\title{Understanding Cross-Domain Few-Shot Learning\\Based on Domain Similarity and Few-Shot Difficulty}

\author{
  Jaehoon Oh\thanks{Equal contribution.} \\
  KAIST DS \\
  Daejeon, South Korea \\
  \texttt{\small jhoon.oh@kaist.ac.kr} \\
  \And
  Sungnyun Kim\footnotemark[1] \\
  KAIST AI \\
  Seoul, South Korea \\
  \texttt{\small ksn4397@kaist.ac.kr} \\
  \And
  Namgyu Ho\footnotemark[1] \\
  KAIST AI \\
  Seoul, South Korea \\
  \texttt{\small itsnamgyu@kaist.ac.kr} \\
  \AND
  \!\!\!\!Jin-Hwa Kim \\
  \!\!\!\!NAVER AI Lab, SNU AIIS \\
  \!\!\!\!Seongnam, South Korea \\
  \!\!\!\!\texttt{\small j1nhwa.kim@navercorp.com}  \\
  \And
  \!\!\!\!\!\!Hwanjun Song\thanks{Corresponding authors.} \\
  \!\!\!\!\!\!NAVER AI Lab \\
  \!\!\!\!\!\!Seongnam, South Korea \\
  \!\!\!\!\!\!\texttt{\small hwanjun.song@navercorp.com} \\
  \And
  \!\!\!\!\!\!Se-Young Yun\footnotemark[2] \\
  \!\!\!\!\!\!KAIST AI \\
  \!\!\!\!\!\!Seoul, South Korea \\
  \!\!\!\!\!\!\texttt{\small yunseyoung@kaist.ac.kr}
}

\begin{document}

\maketitle

\begin{abstract}
Cross-domain few-shot learning\,(CD-FSL) has drawn increasing attention for handling large differences between the source and target domains--an important concern in real-world scenarios. To overcome these large differences, recent works have considered exploiting small-scale unlabeled data from the target domain during the pre-training stage. This data enables self-supervised pre-training on the target domain, in addition to supervised pre-training on the source domain. In this paper, we empirically investigate which pre-training is preferred based on \emph{domain similarity} and \emph{few-shot difficulty} of the target domain. We discover that the performance gain of self-supervised pre-training over supervised pre-training becomes large when the target domain is dissimilar to the source domain, or the target domain itself has low few-shot difficulty. We further design two pre-training schemes, mixed-supervised and two-stage learning, that improve performance. In this light, we present six findings for CD-FSL, which are supported by extensive experiments and analyses on three source and eight target benchmark datasets with varying levels of domain similarity and few-shot difficulty. Our code is available at \url{https://github.com/sungnyun/understanding-cdfsl}.
\end{abstract}
\section{Introduction}\label{sec:intro}
Few-shot learning (FSL) is a machine learning paradigm to learn novel classes from \emph{few} examples with supervised information \citep{matchingnet, wang2020generalizing}. 
Unlike standard supervised learning, a model is pre-trained on the source dataset consisting of \emph{base} classes and then transferred into the target dataset consisting of \emph{novel} classes with few examples, where base and novel classes are disjoint but share similar data domains.
However, this underlying assumption is not applicable to real-world scenarios because source\,(base classes) and target\,(novel classes) domains are different in general. This leads to poor generalization performance because of the change in feature and label distributions, posing a new challenge in FSL\,\citep{bscd_fsl, tseng2020cross}. \looseness=-1

In this regard, \emph{cross-domain few-shot learning}\,(CD-FSL) is gaining immense attention with the BSCD-FSL (Broader Study of CD-FSL) benchmark\,\cite{bscd_fsl}, which enables us to evaluate real-world few-shot learning tasks. The BSCD-FSL benchmark is a collection of four different datasets with varying levels of domain similarity to large-scale natural image collections, such as ImageNet \citep{imagenet}.
Although there are two possible directions for FSL, meta-learning\,\citep{maml, lee2019meta, tseng2020cross} and transfer learning\,\citep{chen2018a, Dhillon2020A, tian2020rethinking}, transfer learning has been reported to have higher performance than meta-learning approaches in cross-domain scenarios. Therefore, following the transfer learning pipeline, recent studies for CD-FSL\,\citep{phoo2021selftraining, islam2021dynamic} have mainly focused on improving the pre-training phase before fine-tuning on the target labeled data with novel classes.

To address the challenge of different domains, there have been recent efforts to leverage \emph{unlabeled} examples from the target domain as auxiliary data for pre-training, in addition to labeled examples from the source domain.
For example, along with the supervised cross-entropy loss, STARTUP \citep{phoo2021selftraining} and Dynamic Distillation \citep{islam2021dynamic} incorporate distillation loss and FixMatch-like loss for self-supervision, respectively.
In other words, they develop sophisticated pre-training approaches that can leverage source and target data together.
However, the basic pre-training schemes, supervised learning\,(SL) on the source domain and self-supervised learning\,(SSL) on the target domain, have not been thoroughly studied with respect to their pros and cons in CD-FSL.

\begin{wrapfigure}{R}{0.5\textwidth}
    \centering
    \includegraphics[width=0.5\textwidth]{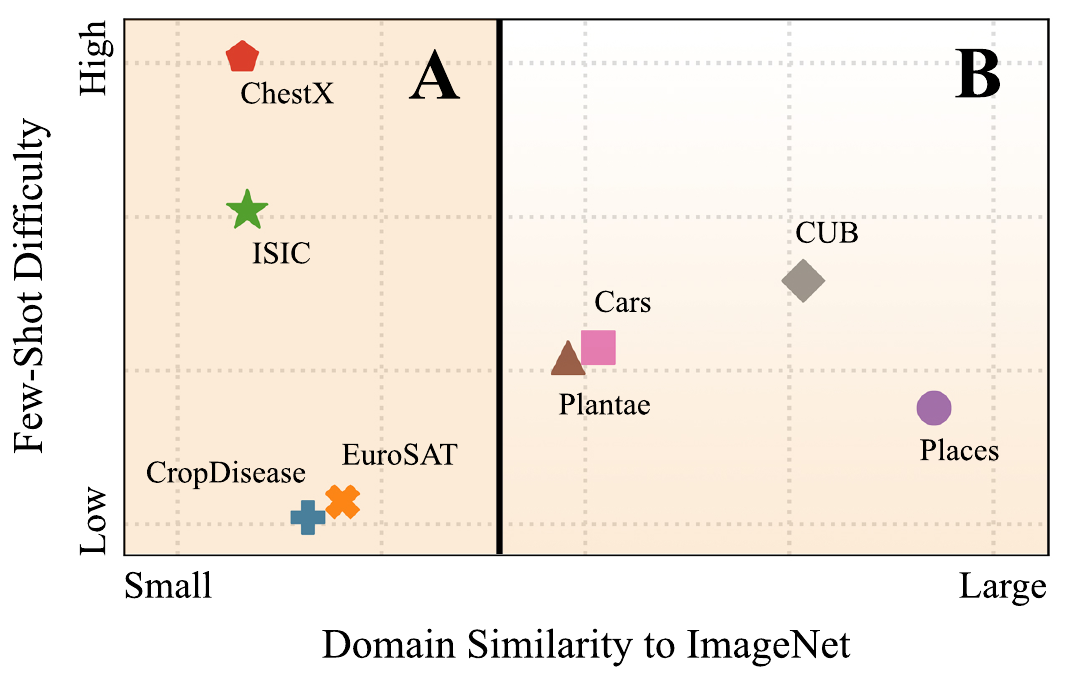}
    \caption{Our insights on the pre-training approaches. (A) SSL is preferred for all datasets with small domain similarity. (B) SL is preferred for high-difficulty datasets with large domain similarity. The formal definitions of similarity and difficulty are explained in Section \ref{subsec:metrics}.
    }
    \label{fig:similarity_difficulty}
 \end{wrapfigure}

In this paper, we establish an \emph{empirical understanding} of the effectiveness of SL and SSL for a better pre-training process in CD-FSL.
To this end, we begin by scrutinizing an opposing finding of the previous works\,\citep{phoo2021selftraining, islam2021dynamic}. We discover that readily available SSL methods, \emph{e.g.}, SimCLR \citep{chen2020simple}, can outperform the standard SL method for pre-training, even when the amount of unlabeled target data for SSL is much smaller than that of labeled source data for SL\,(see \cref{sec:analysis1}).

Next, we investigate why the CD-FSL performance depends on different pre-training schemes using the two properties: \emph{domain similarity} and \emph{few-shot difficulty}.
\textbf{Domain Similarity} is the similarity between the source and target domains, which is known to affect the transferability of the source domain features into the target domain\,\citep{cui2018large, li2020rethinking}.
However, we find it insufficient to identify the effectiveness of SL and SSL based on domain similarity alone.
To solve this conundrum, we propose \textbf{Few-Shot Difficulty} as a measure of the inherent hardness of a dataset, based on the upper bound of empirical FSL performance.
By grounding our analysis on these two metrics, we discover coherent insights on CD-FSL pre-training schemes, depicted in Figure \ref{fig:similarity_difficulty}.
Our analyses point to two conclusions: (A) When domain similarity is small, SSL is preferred due to the limited transferability of source information. On the other hand, (B) SL is preferred when domain similarity is large and few-shot difficulty is high, because supervision from the source dataset achieves stronger performance compared to self-supervision on difficult target data\,(see Section \ref{sec:analysis2}).

Finally, to investigate whether SL and SSL can synergize, we design a joint learning scheme using both SL and SSL, coined as \emph{mixed-supervised learning} (MSL). It is observed that SL and SSL can synergize when they have similar performances.
Furthermore, we extend our analysis to a \textit{two-stage} pre-training scheme, motivated by recent works on CD-FSL \citep{phoo2021selftraining, islam2021dynamic}.
We observe that this generally improves performance because the SL pre-trained model provides a good initialization for the second phase of pre-training\,(see \cref{sec:analysis3}).
\section{Related Work}\label{sec:related}

\subsection{Few-Shot Learning (FSL)}
FSL has been mainly studied in the literature based on two approaches, meta-learning and transfer learning.
In the meta-learning approach, a model is trained on the meta-train set (\emph{i.e.}, source data) in an episodic manner, mimicking the evaluation procedure, such that fast adaptation is possible on the meta-test set (\emph{i.e.}, few-shot target data).
This family of approaches include learning a good initialization \citep{gradient_hpo, maml, pmaml, raghu2019rapid, oh2021boil}, learning a metric space \citep{matchingnet, snell2017prototypical, relationnet}, and learning an update rule \citep{metasgd, learntolearn, Flennerhag2020Meta-Learning}. By contrast, in the transfer-based approach \citep{chen2018a, Dhillon2020A, tian2020rethinking}, a model is pre-trained on the source dataset following the general supervised learning procedure in a mini-batch manner, and subsequently fine-tuned on the target dataset for evaluation.

\subsection{Cross-Domain Few-Shot Learning (CD-FSL)}
CD-FSL has addressed a more challenging and realistic scenario where the source and target domains are dissimilar \citep{bscd_fsl, tseng2020cross}. Such a cross-domain setting makes it difficult to transfer source information into the target domain owing to large domain differences \citep{pan2009survey, li2020rethinking, neyshabur2020being, zhang2020overcoming}.
In general, the most recent methods have been developed on top of the fine-tuning paradigm because this paradigm outperforms the traditional meta-learning approach such as FWT\,\cite{bscd_fsl}.
STARTUP \citep{phoo2021selftraining} and Dynamic Distillation \citep{islam2021dynamic} are the two representative algorithms, and they suggested using small-scale unlabeled data from the target domain in pre-training such that a pre-trained model can be well-adaptable for the target domain.
%
Specifically, both algorithms first train a teacher network with cross-entropy loss on labeled source data. Then, STARTUP trains the student network with cross-entropy loss on the source data together with two unsupervised losses on the target data: distillation loss \citep{hinton2015distilling} and self-supervised loss (\emph{i.e.}, SimCLR \citep{chen2020simple}).
Dynamic Distillation trains the student network with cross-entropy loss on labeled source data and KL loss based on FixMatch\,\citep{sohn2020fixmatch} on unlabeled target data.

\subsection{Self-Supervised Learning (SSL)}
SSL has attracted attention as a method of learning useful representations from unlabeled data\,\citep{dosovitskiy2015discriminative, doersch2015unsupervised, zhang2016colorful, pathak2016context, noroozi2017representation}.
When this field first emerged, hand-crafted pretext tasks, such as solving jigsaw puzzles\,\citep{noroozi2016unsupervised} and predicting rotations\,\citep{gidaris2018unsupervised}, were designed and utilized for training.
In recent times, there has been an effort to use contrastive loss, which enhances representation learning based on augmentation and negative samples\,\citep{chen2020simple, tian2020contrastive, he2020momentum, caron2020unsupervised}. This contrastive loss encourages the alignment of positive pairs and uniformity of data distribution on the hypersphere\,\citep{wang2020understanding}.
This improves the transferability of a model by encouraging it to contain lower-level semantics compared to supervised approaches\,\citep{islam2021broad}. However, this advantage is conditional on the availability of numerous negative samples. To alleviate such constraint, non-contrastive approaches that do not use negative samples have been proposed\,\citep{grill2020bootstrap, chen2021exploring, zbontar2021barlow, tian2021understanding}.
In our empirical study, we use two contrastive approaches, SimCLR\,\citep{chen2020simple} and MoCo\,\citep{he2020momentum}, and two non-contrastive approaches, BYOL \citep{grill2020bootstrap} and SimSiam \citep{chen2021exploring}.
The details of each algorithm are described in Appendix \ref{appx:ssl_methods}.

For the completeness of our survey, we include prior works that address SSL for cross-domain and/or few-shot learning.
\citet{kim2021cds} addressed self-supervised pre-training under label-shared cross-domain, while our setting does not share the label space between domains. \citet{ericsson2021well} observed that SSL on the source data improves performance on the BSCD-FSL dataset. However, domain-specific SSL\,(\emph{i.e.}, SSL on target data) was not addressed.
\citet{cole2021does} showed that adding data from different domains can lead to performance degradation when data is numerous. \citet{phoo2021selftraining} and \citet{islam2021dynamic} argued that plain SSL methods struggle to outperform SL for CD-FSL. We investigate domain-specific SSL and demonstrate its superiority, which opposes the finding from previous studies.
\section{Overview}\label{sec:preliminaries}


\begin{wraptable}[11]{r}{8.3cm}
\small
\vspace*{-1.0cm}
\caption{Summary of the notations.}\label{tab:notation}
\begin{tabular}{L{1.4cm} L{6.1cm}}
\toprule
Notation & Description \\
\midrule
\!$\mathcal{D}_{B}, \mathcal{D}_N$ & Source and target datasets, $\mathcal{D}_{B} \cap \mathcal{D}_{N} = \emptyset$ \\ 
\!$\mathcal{C}_{B}, \mathcal{C}_{N}$ & Base classes for $\mathcal{D}_B$ and novel classes for $\mathcal{D}_N$ \\ 
\!$\mathcal{D}_{U}~(\subset \!\mathcal{D}_N$)\!\!\! & Unlabeled target data for SSL \\
\!$\mathcal{D}_{L}~(\subset \!\mathcal{D}_N$)\!\!\! & Labeled target data for evaluation, $\mathcal{D}_{U}\cap\mathcal{D}_{L}=\emptyset$\!\!\!\!\!\!\!\!\!\\
\!$n, k$          & $\#$ classes and examples for $n$-way $k$-shot \\
\!$\mathcal{D}_{S}~(\subset \!\mathcal{D}_L)$\!\!\! & A support set with size $nk$ for fine-tuning \\
\!$\mathcal{D}_{Q}~(\subset \!\mathcal{D}_L)$\!\!\! & A query set for evaluation, $\mathcal{D}_{S} \cap \mathcal{D}_{Q}=\emptyset$ \\\midrule
\!$f$        & A feature extractor (backbone network) \\ 
\!$h_{\sf sl}$   & A classification head for SL during pre-training\\
\!$h_{\sf ssl}$  & A projection head for SSL during pre-training \\
\!$g$        & A classification head during fine-tuning\\\bottomrule
\end{tabular}
\vspace*{-35pt}
\end{wraptable}

\vspace*{-6pt}
We clarify the scope of our empirical study, propose formal definitions of domain similarity and few-shot difficulty, and describe experimental configurations. Table \ref{tab:notation} summarizes the notations used in this paper.

\subsection{Scope of the Empirical Study\label{subsec:pre_statement}}
Our objective is to learn a feature extractor $f$ on base classes $\mathcal{C}_B$ in source data $\mathcal{D}_B$, which can extract informative representations for novel classes $\mathcal{C}_N$ in target data $\mathcal{D}_N$. Typically, a classifier $g$ is fine-tuned and the model $g \circ f$ is evaluated using labeled target examples $\mathcal{D}_{L}\,(\subset \mathcal{D}_{N})$ after pre-training $f$ on the source data $\mathcal{D}_B$ under the condition that the base classes are largely different from the novel classes.

Following the recent literature\,\citep{phoo2021selftraining, islam2021dynamic}, we further assume that additional unlabeled data $\mathcal{D}_{U}\,(\subset \mathcal{D}_{N})$ is available in the pre-training phase.
We follow the split strategy used in \citet{phoo2021selftraining}, where 20\% of the target data $\mathcal{D}_{N}$ is used as the unlabeled data $\mathcal{D}_{U}$ for pre-training.
Note that the size of the unlabeled portion is very small\,(\emph{e.g.}, only a few thousand examples) compared to large-scale datasets typically considered for self-supervised learning.
In this problem setup, the pre-training phase of CD-FSL can be carried out based on \emph{three} learning strategies:

\begin{itemize}[leftmargin=*]
    \item \textbf{Supervised Learning}: Let $f$ and $h_{\sf sl}$ be the feature extractor and linear classifier for the base classes $\mathcal{C}_B$, respectively. Then, a model $h_{\sf sl}\!\circ\!f$ is pre-trained only for the labeled source data $\mathcal{D}_B$ by minimizing the standard cross-entropy loss $\ell_{\sf ce}$ in a mini-batch manner,\footnote{The batch loss on the entire data is used for ease of exposition.}
    \begin{equation}\label{eq:loss_sl}
    \mathcal{L}_{\sf sl}(f, h_{\sf sl}; \mathcal{D}_B) = \frac{1}{|\mathcal{D}_B|} \sum_{(x, y) \in \mathcal{D}_B} \!\!\!\!\ell_{\sf ce}(h_{\sf sl} \circ f(x), y).
    \end{equation}
    
    \item \textbf{Self-Supervised Learning}: Let $h_{\sf ssl}$ be the projection head. Then, a model $h_{\sf ssl} \circ f$ is pre-trained only for the unlabeled target data $\mathcal{D}_U$, which is much smaller than the labeled source data, by minimizing (non-)contrastive self-supervised loss $\ell_{\sf self}$\,(\textit{e.g.}, NT-Xent),\footnotemark[1] 
    \begin{equation}\label{eq:loss_ssl}
    \mathcal{L}_{\sf ssl}(f, h_{\sf ssl} ; \mathcal{D}_U) = \frac{1}{2|\mathcal{D}_U|} \sum_{x \in \mathcal{D}_U}\!\! \Big[ \ell_{\sf self} \big(z_1; z_2; \{z^-\} \big) + \ell_{\sf self} \big(z_2; z_1; \{z^-\} \big) \Big]
    \end{equation}
    \begin{equation*}
    {\rm where}~~ z_i = h_{\sf ssl} \circ f(A_i(x)),
    \end{equation*}
    and $A_i(x)$ is the $i$-th augmentation of the same input $x$. This training loss forces $z_1$ to be similar to $z_2$ and dissimilar to the set of negative features $\{z^-\}$. In addition, there are non-contrastive SSL methods that do not rely on negative examples, \emph{i.e.}, $\{z^-\}\!=\!\emptyset$. We provide a more detailed explanation of SSL losses, including multiple (non-)constrastive approaches in Appendix \ref{appx:ssl_methods}.
    
    \item \textbf{Mixed-Supervised Learning}: MSL exploits labeled as well as unlabeled data from different domains simultaneously. MSL can be intuitively formulated by minimizing the interpolation of their losses in Eqs.\,\eqref{eq:loss_sl} and \eqref{eq:loss_ssl},
    \begin{equation}\label{eq:loss_msl}
    \mathcal{L}_{\sf msl}(f, h_{\sf sl}, h_{\sf ssl}; \mathcal{D}_B, \mathcal{D}_U) = (1-\gamma)\cdot\mathcal{L}_{\sf sl}(f, h_{\sf sl}; \mathcal{D}_B) + \gamma\cdot\mathcal{L}_{\sf ssl}(f, h_{\sf ssl}; \mathcal{D}_U),\!\!\!
    \end{equation}
    where $0\,<\,\gamma\,<\,1$ and the feature extractor $f$ is hard-shared and trained through SL and SSL losses with a balancing hyperparameter $\gamma$. 
    This can be a generalization of STARTUP and Dynamic Distillation, which use Eq.\,\eqref{eq:loss_msl} in the second pre-training phase with a moderate modification after the typical pre-training phase using SL.
\end{itemize}

Our analysis focuses on pre-training and fine-tuning schemes due to the superiority of transfer-based methods over typical FSL algorithms such as MAML\,\citep{maml}, which is shown in\,\citep{bscd_fsl}.
Based on the three learning strategies above, we conduct an empirical study to gain an in-depth understanding of their effectiveness in the pre-training phase, providing deep insight into the following questions:

\begin{enumerate}[leftmargin=*]
\item Which is more effective for pre-training, using only SL or SSL? \hspace*{2.7cm} $\vartriangleright$ Section \ref{sec:analysis1}
\item How to apply domain similarity and few-shot difficulty to identify the more effective pre-training scheme between SL and SSL, for CD-FSL? \hspace*{5.62cm} $\vartriangleright$ Section \ref{sec:analysis2}
\item Can MSL, a combination of SL and SSL, as well as a two-stage scheme improve performance? \hspace*{11.78cm} $\vartriangleright$ Section \ref{sec:analysis3} \looseness=-1
\end{enumerate}

\subsection{Domain Similarity and Few-Shot Difficulty}\label{subsec:metrics}

We present a procedure for estimating the two metrics on datasets, which are used to analyze the pre-training schemes.
First, we use \textit{domain similarity} introduced in \citep{cui2018large}, which is based on Earth Mover's Distance\,(EMD \citep{rubner1998metric}) because the distance between the two domains can be considered as the cost of \textit{moving} images from one domain to the other in the transfer learning context\,\citep{cui2018large, li2020rethinking}. Further details on this metric, \emph{e.g.}, advantages of EMD, are explained in Appendix~\ref{appx:domain_similarity}.

We can easily compute EMD using the retrieved sample representations.\footnote{To extract the representation of images, we follow \citet{li2020rethinking} by using a large model trained on a large-scale dataset, ResNet101 pre-trained on ImageNet. Note that \citet{cui2018large} used JFT dataset \citep{sun2017revisiting}, which is not released for public use. Furthermore, we measure domain similarity using different feature extractors, described in Table~\ref{tab:similarity_diff_model} of Appendix~\ref{appx:domain_similarity}. Our analysis is consistent regardless of the feature extractor used.}
We create the prototype vector ${\bf p}_i$, which is an averaged representation for all examples belonging to class $i$. Next, let $i \in \mathcal{C}_{B}$ and $j \in \mathcal{C}_{N}$ be a class in base\,(source) and novel\,(target) classes, respectively. Then, the domain similarity between the source and target data is formulated as
\begin{align}
\label{eq:domain_sim}
{\rm Sim}&(\mathcal{D}_{B}, \mathcal{D}_{N}) = {\rm exp}\big({-\alpha ~{\rm EMD}(\mathcal{D}_{B}, \mathcal{D}_{N})}\big) \quad
{\rm where} \,\,\, {\rm EMD}(\mathcal{D}_{B}, \mathcal{D}_{N}) = \frac{\sum_{i \in \mathcal{C}_{B},\, j \in \mathcal{C}_{N}} f_{i,j}\,d_{i,j}}{\sum_{i \in \mathcal{C}_{B},\, j \in \mathcal{C}_{N}} f_{i,j}} \nonumber\\
&{\rm subject~to} ~~  f_{i,j} \geq 0, ~\sum_{i \in \mathcal{C}_{B},\, j \in \mathcal{C}_{N}} \!\!\!\!f_{i,j} = 1, ~\sum_{j \in \mathcal{C}_{N}} f_{i,j} \leq \frac{|\mathcal{D}_{B}[i]|}{|\mathcal{D}_{B}|},\,\,
\sum_{i \in \mathcal{C}_{B}} f_{i,j} \leq \frac{|\mathcal{D}_{N}[j]|}{|\mathcal{D}_{N}|},
\end{align}
where $d_{i,j}=||{\bf p}_{i} - {\bf p}_{j}||_2$; $f_{i,j}$ is the optimal flow between ${\bf p}_i$ and ${\bf p}_j$ subject to the constraints for EMD; $\mathcal{D}[i]$ returns all examples of the specified class $i$ in $\mathcal{D}$; and $\alpha$ is typically set to $0.01$\,\cite{cui2018large}.
Namely, EMD can be interpreted as the weighted distance of all combinations between the base and novel classes. The larger similarity indicates that source and target data share similar domains.
%

Next, we propose \emph{few-shot difficulty}, which quantifies the difficulty of a dataset based on the empirical upper bound of few-shot performance in our problem setup, regardless of its relationship to the source dataset.
To capture the upper bound of FSL performance, we use 20\% of the target dataset as labeled data to pre-train the model in a supervised manner.
Then, the pre-trained model is evaluated on the remaining unseen target data for the $5$-way $k$-shot classification task.\footnote{We use a few-shot learning task instead of classification on the entire data, preventing the performance from being distorted by other factors, such as data imbalance and the number of classes.}
As the generalization capability indicates the hardness\,\cite{song2020carpe}, the classification accuracy for unseen data is used and converted into the few-shot difficulty using an exponential function with a hyperparameter $\beta$\,(the default value is 0.01),
\begin{equation}
{\rm Diff}(\mathcal{D}, k) = {\rm exp} (-\beta~{\rm Acc}(\mathcal{D}, k)),
\label{eq:domain_dif}
\end{equation}
where ${\rm Acc}(\mathcal{D}, k)$ returns the average of $5$-way $k$-shot classification accuracy over 600 episodes for the given data $\mathcal{D}$. Note that in our paper, $k$ is set to 5, but the order of difficulty is the same regardless of $k$.
High few-shot difficulty implies that the achievable accuracy is low even when there is no domain difference between pre-training and evaluation.

\subsection{Experimental Configurations}\label{subsec:eval_setting}

\noindent\textbf{Cross-Domain Datasets.} We use ImageNet, tieredImageNet, and miniImageNet as source datasets for generality. Regarding the target domain, we prepare eight datasets with varying domain similarity and few-shot difficulty; domain similarity is computed based on both the source and target datasets, while few-shot difficulty is computed based on the target dataset. 
To summarize their order in Figure \ref{fig:similarity_difficulty}, {\slshape \textbf{domain similarity} to ImageNet: Places $>$ CUB $>$ Cars $>$ Plantae $>$ EuroSAT $>$ CropDisease $>$ ISIC $>$ ChestX}, and {\slshape \textbf{few-shot difficulty}: ChestX $>$ ISIC $>$ CUB $>$ Cars $>$ Plantae $>$ Places $>$ EuroSAT $>$ CropDisease}.
For instance, Places data has the largest domain similarity to ImageNet, while ChestX has the highest few-shot difficulty. Appendix \ref{appx:detail} provides the details of each dataset.
The detailed values for domain similarity and few-shot difficulty are reported in Appendix \ref{appx:domain_similarity} and \ref{appx:difficulty}, respectively. These are visualized in Appendix \ref{appx:similarity_difficulty_variants}.
We also provide the results on the case when source and target domains are the same, \emph{i.e.}, the standard FSL setting, in Appendix \ref{appx:same-domain-fsl-results}. 

\noindent\textbf{Evaluation Pipeline.} We follow the standard evaluation pipeline of CD-FSL\,\cite{bscd_fsl}.
The evaluation process is performed in an episodic manner, where each episode represents a distinct few-shot task. Each episode is comprised of a support set $\mathcal{D}_S$ and a query set $\mathcal{D}_Q$, which are sampled from the entire labeled target data $\mathcal{D}_{L}$. The support set $\mathcal{D}_S$ and query set $\mathcal{D}_Q$ consist of $n$ classes that are randomly selected among the entire set of novel classes $\mathcal{C}_{N}$.
For the $n$-way $k$-shot setting, $k$ examples are randomly drawn from each class for the support set $\mathcal{D}_S$, while $k_{q}$\,(typically $15$) examples for the query set $\mathcal{D}_Q$. Thus, the support and query set are defined as,
\begin{equation}
\mathcal{D}_{S} = \{(x_i^s,y_i^s)\}_{i=1}^{n \times k} ~~{\rm and}~~ \mathcal{D}_{Q} = \{(x_i^q,y_i^q)\}_{i=1}^{n \times k_q}.
\end{equation}
For evaluation, a classifier $g$ is fine-tuned on the support set $\mathcal{D}_S$, using features extracted from the fixed pre-trained backbone $f$.
Note that $g$ is for the evaluation purpose different from $h_{\sf sl}$ and $h_{\sf ssl}$ for pre-training.
The fine-tuned model $g \circ f$ is then tested on the query set $\mathcal{D}_Q$. We set $n=5$ and $k=\{1, 5\}$, and the accuracy is averaged over 600 episodes following convention\,\cite{bscd_fsl, phoo2021selftraining}.

\noindent\textbf{Implementation.} We use different backbone networks depending on the source data. For ImageNet and tieredImageNet, ResNet18 is used as the backbone, while ResNet10 is used for miniImageNet. For ResNet18 pre-trained on ImageNet with SL, we use the model provided by PyTorch \citep{PyTorch} repository. This setup is exactly the same for all pre-training schemes.
Additional details on the training setup are provided in Appendix \ref{appx:implement_detail}.
\begin{table*}[!t]
\caption{$5$-way $k$-shot CD-FSL performance\,(\%) of the models pre-trained by SL and SSL. We report the average accuracy and its 95\% confidence interval over 600 few-shot episodes. B and S indicate base and strong augmentation, respectively. 
The best accuracy is marked in bold for each backbone.
}\label{tab:tlssl_aug}
\centering
\footnotesize\addtolength{\tabcolsep}{-3.0pt}
\resizebox{\linewidth}{!}{%
\begin{tabular}{c|c|c|c|cc|cc|cc|cc}
    \toprule
    \multirow{2}{*}{\makecell[l]{\!Source\!\\~Data\!}} & Pre-train & \multirow{2}{*}{Method} & \multirow{2}{*}{Aug.} & \multicolumn{2}{c|}{EuroSAT} & \multicolumn{2}{c|}{CropDisease} & \multicolumn{2}{c|}{ISIC} & \multicolumn{2}{c}{ChestX} \\
    & Scheme & & & $k$=1 & $k$=5 & $k$=1 & $k$=5 & $k$=1 & $k$=5 & $k$=1 & $k$=5 \\
    \midrule
    ImageNet & SL & Default      & B & 66.14{\scriptsize$\pm$.83} & 84.73{\scriptsize$\pm$.51} & 74.18{\scriptsize$\pm$.82} & 92.81{\scriptsize$\pm$.45} & 31.11{\scriptsize$\pm$.55} & 44.10{\scriptsize$\pm$.58} & 22.48{\scriptsize$\pm$.39} & 25.51{\scriptsize$\pm$.44} \\
    \midrule
    \multirow{2}{*}{\makecell[l]{\!~~~tiered\\\!ImageNet}} & \multirow{2}{*}{SL} & \multirow{2}{*}{Default} & B & 61.81{\scriptsize$\pm$.88} & 79.87{\scriptsize$\pm$.67} & 66.82{\scriptsize$\pm$.90} & 87.19{\scriptsize$\pm$.59} & 30.35{\scriptsize$\pm$.60} & 41.67{\scriptsize$\pm$.55} & 22.34{\scriptsize$\pm$.38} & 25.08{\scriptsize$\pm$.45} \\
    & & & S & 60.07{\scriptsize$\pm$.88} & 79.95{\scriptsize$\pm$.66} & 65.70{\scriptsize$\pm$.94} & 86.34{\scriptsize$\pm$.60} & 29.75{\scriptsize$\pm$.56} & 40.60{\scriptsize$\pm$.58} & 22.11{\scriptsize$\pm$.42} & 25.20{\scriptsize$\pm$.41} \\
    \midrule
    \multirow{8}{*}{-} & \multirow{8}{*}{SSL} & \multirow{2}{*}{SimCLR}  & B & 70.37{\scriptsize$\pm$.86} & 87.80{\scriptsize$\pm$.46} & 90.94{\scriptsize$\pm$.69} & 97.44{\scriptsize$\pm$.29} & 34.13{\scriptsize$\pm$.69} & 44.37{\scriptsize$\pm$.66} & 21.41{\scriptsize$\pm$.41} & 25.05{\scriptsize$\pm$.42} \\
    &   &                          & S &\textbf{84.30}{\scriptsize$\pm$.73} &\textbf{94.12}{\scriptsize$\pm$.32} &\textbf{91.00}{\scriptsize$\pm$.76} &\textbf{97.46}{\scriptsize$\pm$.34} &\textbf{36.39}{\scriptsize$\pm$.66} &47.85{\scriptsize$\pm$.65} &21.55{\scriptsize$\pm$.41} &25.26{\scriptsize$\pm$.44} \\
    &   & \multirow{2}{*}{MoCo}    & B & 51.21{\scriptsize$\pm$.93} & 68.19{\scriptsize$\pm$.74} & 70.22{\scriptsize$\pm$.95} & 87.11{\scriptsize$\pm$.60} & 27.79{\scriptsize$\pm$.53} & 36.60{\scriptsize$\pm$.59} & 21.44{\scriptsize$\pm$.43} & 24.28{\scriptsize$\pm$.43} \\
    &   &                          & S & 69.11{\scriptsize$\pm$.98} & 81.01{\scriptsize$\pm$.73} & 80.08{\scriptsize$\pm$.97} & 92.48{\scriptsize$\pm$.52} & 29.54{\scriptsize$\pm$.59} & 39.28{\scriptsize$\pm$.58} & 21.74{\scriptsize$\pm$.42} & 24.58{\scriptsize$\pm$.44} \\
    &   & \multirow{2}{*}{BYOL}    & B & 60.98{\scriptsize$\pm$.91} & 84.88{\scriptsize$\pm$.56} & 81.58{\scriptsize$\pm$.78} & 96.82{\scriptsize$\pm$.27} & 35.31{\scriptsize$\pm$.64} & \textbf{49.26}{\scriptsize$\pm$.64} & 22.65{\scriptsize$\pm$.42} & \textbf{28.80}{\scriptsize$\pm$.49} \\
    &   &                          & S & 66.16{\scriptsize$\pm$.86} & 87.83{\scriptsize$\pm$.48} & 85.77{\scriptsize$\pm$.73} & 96.93{\scriptsize$\pm$.30} & 34.53{\scriptsize$\pm$.62} & 47.59{\scriptsize$\pm$.63} & \textbf{22.75}{\scriptsize$\pm$.41} & 28.36{\scriptsize$\pm$.46} \\
    &    & \multirow{2}{*}{SimSiam} & B & 44.06{\scriptsize$\pm$.86} & 61.03{\scriptsize$\pm$.72} & 75.36{\scriptsize$\pm$.82} & 92.31{\scriptsize$\pm$.44} & 26.99{\scriptsize$\pm$.52} & 35.68{\scriptsize$\pm$.52} & 22.02{\scriptsize$\pm$.41} & 26.06{\scriptsize$\pm$.46} \\
    &   &                          & S & 70.80{\scriptsize$\pm$.88} & 85.10{\scriptsize$\pm$.57} & 84.72{\scriptsize$\pm$.80} & 96.05{\scriptsize$\pm$.36} & 30.17{\scriptsize$\pm$.56} & 39.51{\scriptsize$\pm$.55} & 22.17{\scriptsize$\pm$.40} & 26.56{\scriptsize$\pm$.46} \\\bottomrule
    \multicolumn{12}{c}{(a) ResNet18 is used as a backbone.}\\ \toprule
     \multirow{2}{*}{\makecell[l]{\!~~~~mini\\\!ImageNet}} & \multirow{2}{*}{SL} & \multirow{2}{*}{Default}       & B & 64.03{\scriptsize$\pm$.91} & 82.72{\scriptsize$\pm$.59} & 73.38{\scriptsize$\pm$.87} & 91.53{\scriptsize$\pm$.49} & 30.68{\scriptsize$\pm$.58} & 41.77{\scriptsize$\pm$.59} & 22.64{\scriptsize$\pm$.40} & 26.26{\scriptsize$\pm$.45} \\
    &                  &                   & S & 65.03{\scriptsize$\pm$.88} & 84.00{\scriptsize$\pm$.56} & 72.82{\scriptsize$\pm$.87} & 91.32{\scriptsize$\pm$.49} & 29.91{\scriptsize$\pm$.54} & 40.84{\scriptsize$\pm$.56} & 22.88{\scriptsize$\pm$.42} & 27.01{\scriptsize$\pm$.44} \\
    \midrule
    \multirow{8}{*}{-} & \multirow{8}{*}{SSL} & \multirow{2}{*}{SimCLR}  & B & 66.77{\scriptsize$\pm$.84} & 86.39{\scriptsize$\pm$.48} & 89.33{\scriptsize$\pm$.66} & 96.82{\scriptsize$\pm$.32} & 33.32{\scriptsize$\pm$.63} & 44.50{\scriptsize$\pm$.64} & 22.26{\scriptsize$\pm$.42} & 24.34{\scriptsize$\pm$.42} \\
    &                     &                          & S &\textbf{79.50}{\scriptsize$\pm$.78} &\textbf{92.36}{\scriptsize$\pm$.37} &\textbf{89.49}{\scriptsize$\pm$.74} &\textbf{97.24}{\scriptsize$\pm$.33} &\textbf{34.90}{\scriptsize$\pm$.64} &\textbf{46.76}{\scriptsize$\pm$.61} &21.97{\scriptsize$\pm$.41} &25.62{\scriptsize$\pm$.43} \\
    &                     & \multirow{2}{*}{MoCo}   & B & 48.70{\scriptsize$\pm$.92} & 66.85{\scriptsize$\pm$.72} & 68.77{\scriptsize$\pm$.92} & 87.67{\scriptsize$\pm$.57} & 27.76{\scriptsize$\pm$.54} & 38.03{\scriptsize$\pm$.57} & 21.55{\scriptsize$\pm$.42} & 24.48{\scriptsize$\pm$.44} \\
    &                    &                          & S & 76.20{\scriptsize$\pm$.89} & 89.54{\scriptsize$\pm$.46} & 80.19{\scriptsize$\pm$.99} & 93.41{\scriptsize$\pm$.53} & 30.20{\scriptsize$\pm$.55} & 41.14{\scriptsize$\pm$.57} & 21.64{\scriptsize$\pm$.40} & 24.49{\scriptsize$\pm$.43} \\
    &                     & \multirow{2}{*}{BYOL}    & B & 61.18{\scriptsize$\pm$.82} & 83.11{\scriptsize$\pm$.57} & 80.50{\scriptsize$\pm$.75} & 94.85{\scriptsize$\pm$.35} & 33.02{\scriptsize$\pm$.62} & 46.72{\scriptsize$\pm$.65} & 22.90{\scriptsize$\pm$.41} & 27.40{\scriptsize$\pm$.47} \\
    &                    &                          & S & 66.45{\scriptsize$\pm$.80} & 86.55{\scriptsize$\pm$.50} & 80.10{\scriptsize$\pm$.76} & 94.53{\scriptsize$\pm$.41} & 33.50{\scriptsize$\pm$.59} & 45.99{\scriptsize$\pm$.63} & 23.11{\scriptsize$\pm$.42} & 27.71{\scriptsize$\pm$.44} \\
    &                    & \multirow{2}{*}{SimSiam} & B & 44.57{\scriptsize$\pm$.82} & 63.67{\scriptsize$\pm$.67} & 82.83{\scriptsize$\pm$.73} & 95.37{\scriptsize$\pm$.34} & 30.74{\scriptsize$\pm$.60} & 41.28{\scriptsize$\pm$.62} & 22.76{\scriptsize$\pm$.42} & 27.50{\scriptsize$\pm$.47} \\
    &                    &                          & S & 71.66{\scriptsize$\pm$.88} & 85.21{\scriptsize$\pm$.59} & 81.25{\scriptsize$\pm$.77} & 95.13{\scriptsize$\pm$.37} & 31.80{\scriptsize$\pm$.59} & 41.44{\scriptsize$\pm$.59} & \textbf{23.22}{\scriptsize$\pm$.41} & \textbf{27.83}{\scriptsize$\pm$.46} \\
    \bottomrule
    \multicolumn{12}{c}{(b) ResNet10 is used as a backbone.}\\
\end{tabular}}
\vspace*{-0.5cm}
\label{tab:analysis1}
\end{table*}

\section{Supervised Learning on Source vs. Self-Supervised Learning on Target}\label{sec:analysis1}
We begin by investigating the superiority of SSL on the target dataset over SL on the source dataset for pre-training. We compare the CD-FSL performance of pre-trained models using four representative (widely cited) SSL methods\,(SimCLR \citep{chen2020simple}, MoCo \citep{he2020momentum}, BYOL \citep{grill2020bootstrap}, and SimSiam \citep{chen2021exploring}) with that of an SL method\,(Default) in Table \ref{tab:analysis1}. Four different domain datasets from the BSCD-FSL benchmarks (EuroSAT, CropDisease, ISIC, and ChestX) are used as target data.
Table \ref{tab:analysis1} provides empirical evidence of the findings in this section.
Recent literature has reported that SSL pre-training does \emph{not} work better than SL for the CD-FSL task because of insufficient unlabeled examples in the target domain\,\citep{phoo2021selftraining, islam2021dynamic}. However, our observation contradicts this previous finding.

\vspace*{0.15cm}
\begin{observation}
\emph{SSL on the target domain can achieve remarkably higher performance over SL on the labeled source domain, even with small-scale ({i.e.}, a few thousand) unlabeled target data.}
\label{obs:obs4_1}
\end{observation}

\noindent{\scshape Evidence.} SSL methods are observed to outperform SL in most cases, even though SSL does not leverage source data for pre-training. In particular, SSL methods show much higher performance compared to the model pre-trained on the entire ImageNet dataset, which has more than 1.2M training examples. This leads to the conclusion that SSL on the target domain can be better than SL on the source domain for CD-FSL pre-training. In other words, unlabeled target data available at the pre-training phase is worth more than labeled source data, even if the unlabeled target data is much smaller (\emph{e.g.}, 8k examples for CropDisease) than the labeled source data.
In Appendix \ref{appx:ratio_resnet10}, we show that SSL can outperform SL using even smaller portions of unlabeled target data.

\vspace*{0.15cm}
\begin{observation}
\emph{SSL achieves significant performance gains with strong data augmentation.}
\label{obs:obs4_2}
\end{observation}

\noindent{\scshape Evidence.} In addition, the results in Table \ref{tab:analysis1} provide the performance sensitivity to data augmentation. For this study, two types of augmentation are used: (1)\,base augmentation from \citep{chen2020simple}, which consists of random resized crop, color jitter, horizontal flip, and normalization, and (2)\,strong augmentation from \citep{islam2021dynamic}, which adds Gaussian blur and random gray scale to the base (see the detail of the augmentations in Appendix \ref{appx:implement_detail}). With strong augmentation, SSL methods exhibit significant performance gains of up to $27.50\%$p compared to base augmentation, \emph{i.e.}, MoCo on EuroSAT in Table \ref{tab:analysis1}(b).
However, SL does not benefit from strong augmentation as SSL does. This has also been observed in the literature\,\cite{chen2020simple}. Therefore, the performance of SSL can be further improved for CD-FSL if more suitable augmentation is applied. Based on this observation, we use strong augmentation for SSL as the default setup in the rest of our paper.

Meanwhile, the superiority among SSL algorithms varies with target dataset. In Table \ref{tab:analysis1}, we observe that SimCLR performs best in EuroSAT and CropDisease, while in ISIC, SimCLR and BYOL both perform well. For ChestX, BYOL and SimSiam show good performance.
The SSL methods can be categorized into two groups: contrastive (SimCLR and MoCo) and non-contrastive (BYOL and SimSiam).
For the rest of our paper, we focus our analysis on SimCLR and BYOL, which are representative methods from each group with robust performance.
The results for other target datasets are presented in Appendix \ref{appx:other_dataset_slssl}.

\section{Closer Look at Domain Similarity and Few-Shot Difficulty}\label{sec:analysis2}

\begin{figure}[t]
     \centering
     \begin{subfigure}[h]{.49\linewidth}
         \centering
         \includegraphics[width=\linewidth]{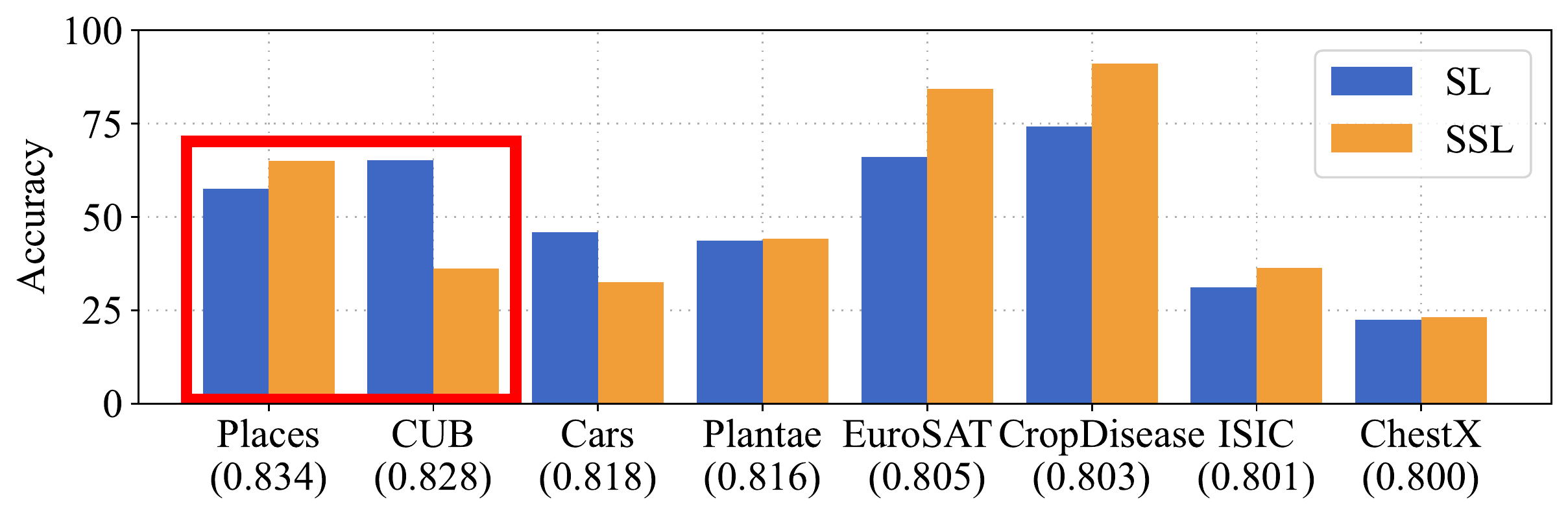}
         \caption{5-way 1-shot}
         \label{fig:sim_miniIN_1shot}
     \end{subfigure}
     \hfill
     \begin{subfigure}[h]{.49\linewidth}
         \centering
         \includegraphics[width=\linewidth]{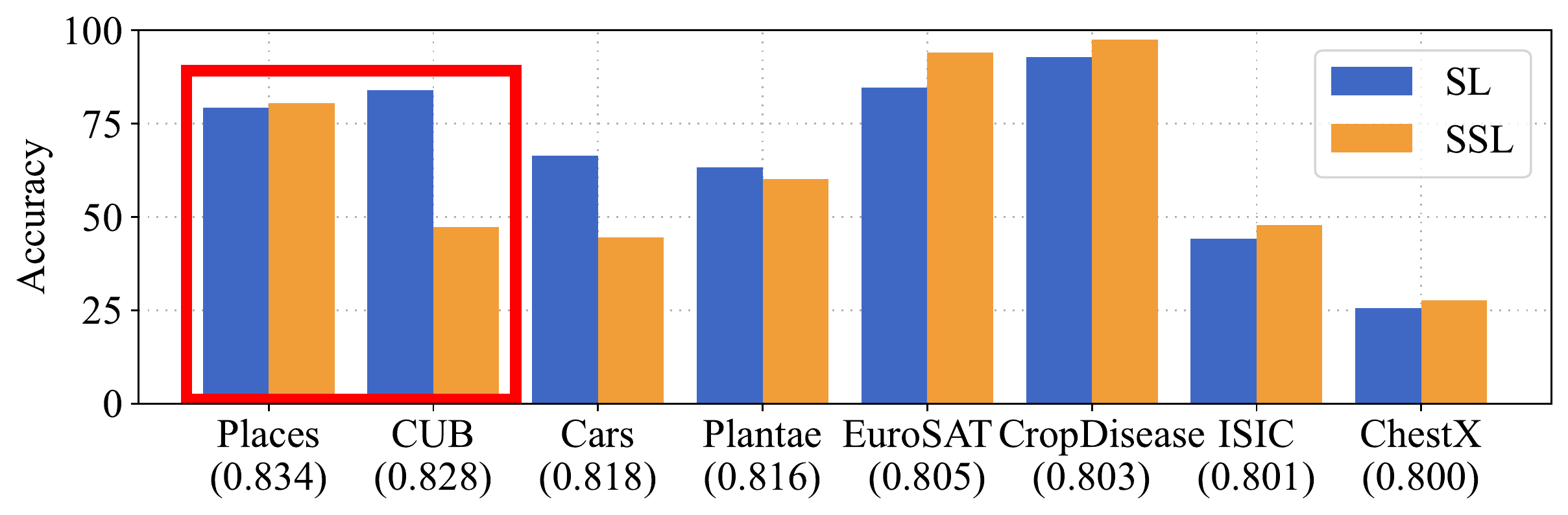}
         \caption{5-way 5-shot}
         \label{fig:sim_miniIN_5shot}
     \end{subfigure}
     \caption{$5$-way $k$-shot CD-FSL performance\,(\%) of SL and SSL according to domain similarity\,(values in x-axis), with ImageNet source data. 
     The red box shows that SL outperforms SSL in the second largest domain similarity, while SSL outperforms SL in the largest domain similarity. \looseness=-1}
     \label{fig:sim_miniIN}
\end{figure}

We investigate why the CD-FSL performance depends on different pre-training schemes, \emph{i.e.,} SL or SSL, based on the two metrics: domain similarity and few-shot difficulty in Eqs. \eqref{eq:domain_sim} and \eqref{eq:domain_dif}.
We analyze the relationship between few-shot performance and the two metrics on various target datasets and provide insights for developing a more effective pre-training approach.

Including BSCD-FSL, we consider four additional datasets from different domains: Places, CUB, Cars, and Plantae. Note that these additional datasets are known to be more similar to ImageNet than the BSCD-FSL datasets are \citep{ericsson2021well}, and our estimated similarity shows the same trend. We mainly use ImageNet as the source dataset to make our analysis more reliable. We analyze their domain similarity and few-shot difficulty and display them in {Figure \ref{fig:similarity_difficulty}}, where ImageNet is used as source data for domain similarity. In this section, to select the SSL method for each dataset, we use SimCLR for all datasets except ChestX, where BYOL is used, based on the performance observed in Section~\ref{sec:analysis1}.


\subsection{Domain Similarity}\label{subsec:insight_pre}

Figure \ref{fig:sim_miniIN} shows the CD-FSL performance of the pre-trained models using SL and SSL for eight target datasets with varying domain similarity, where all the datasets are sorted by domain similarity.
A common belief about domain similarity is that, as domain similarity increases, it is more beneficial for pre-training to use a large amount of labeled source data\,\citep{cui2018large, li2020rethinking, ericsson2021well}. Our analysis shows that this belief is partially true.

\vspace*{0.15cm}
\begin{observation}
\emph{SL does not consistently benefit from large domain similarity.}
\label{obs:obs5_2}
\end{observation}
\noindent{\scshape Evidence.} For the aforementioned belief to be true, the performance gain of SL over SSL should be greater as domain similarity increases.
However, although SL outperforms SSL in the CUB dataset with the second largest domain similarity, in the Places dataset with the largest domain similarity, SSL rather exhibits higher CD-FSL accuracy than SL\,(see the red box in Figure \ref{fig:sim_miniIN}).
%
Furthermore, in the ChestX dataset with the smallest domain similarity, SL and SSL have similar performances. 
These results demonstrate that unlike prior belief, large domain similarity does not always guarantee the superiority of SL.
In other words, there is an inconsistency that cannot be explained solely by domain similarity, and we explore why this inconsistency occurs by taking few-shot difficulty into account.

\subsection{Few-Shot Difficulty}\label{subsec:insight_difficulty}

\begin{figure}[t]
    \centering
    \begin{subfigure}[h]{.49\linewidth}
         \centering
         \includegraphics[width=\linewidth]{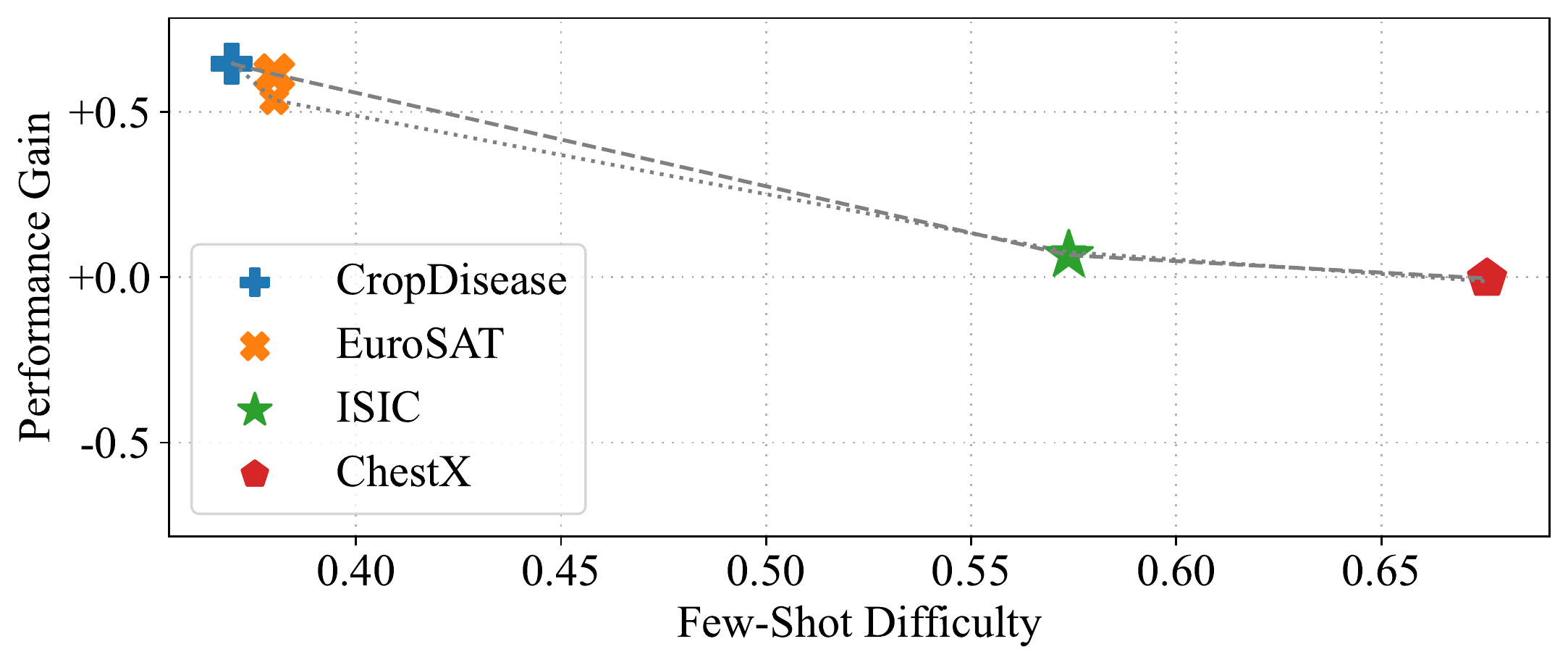}
         \caption{Small Similarity (ImageNet)}
    \end{subfigure}
    \begin{subfigure}[h]{.49\linewidth}
         \centering
         \includegraphics[width=\linewidth]{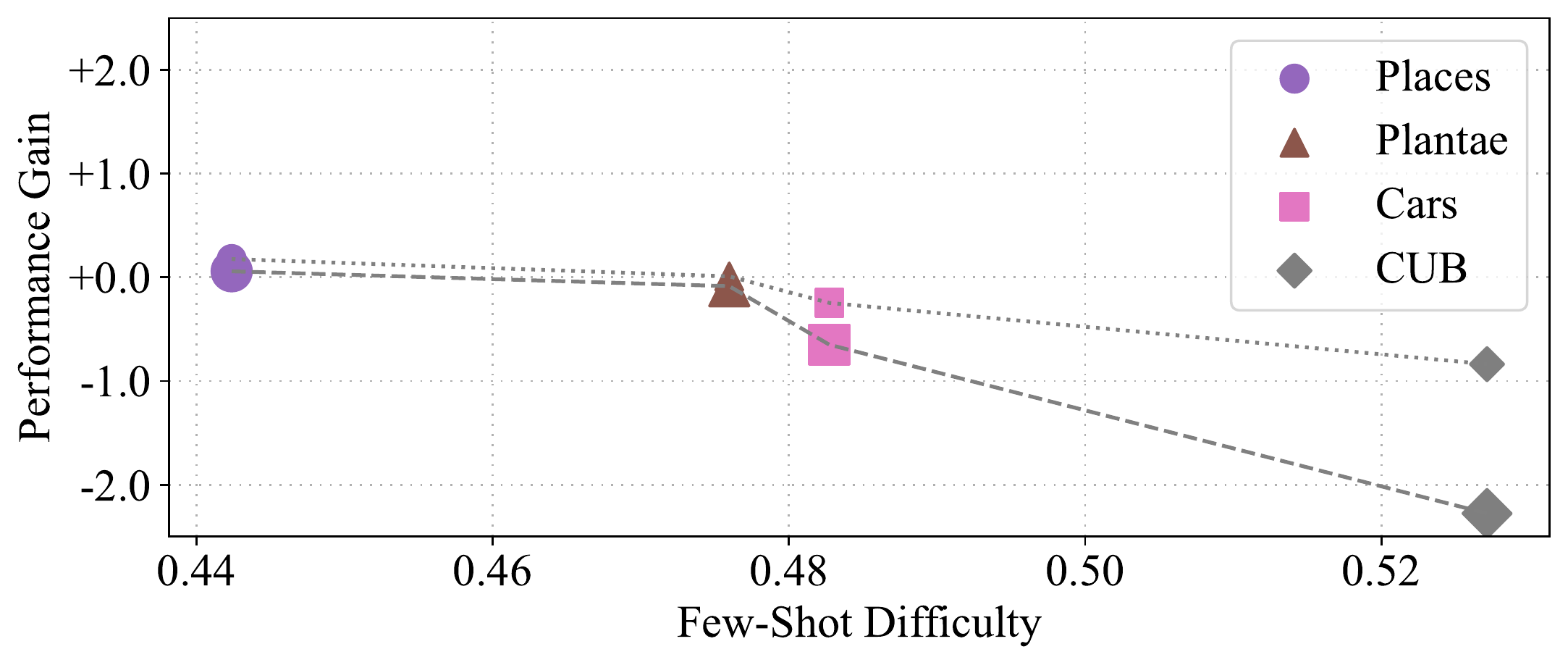}
         \caption{Large Similarity (ImageNet)}
    \end{subfigure}
    
    \begin{subfigure}[h]{.49\linewidth}
         \centering
         \includegraphics[width=\linewidth]{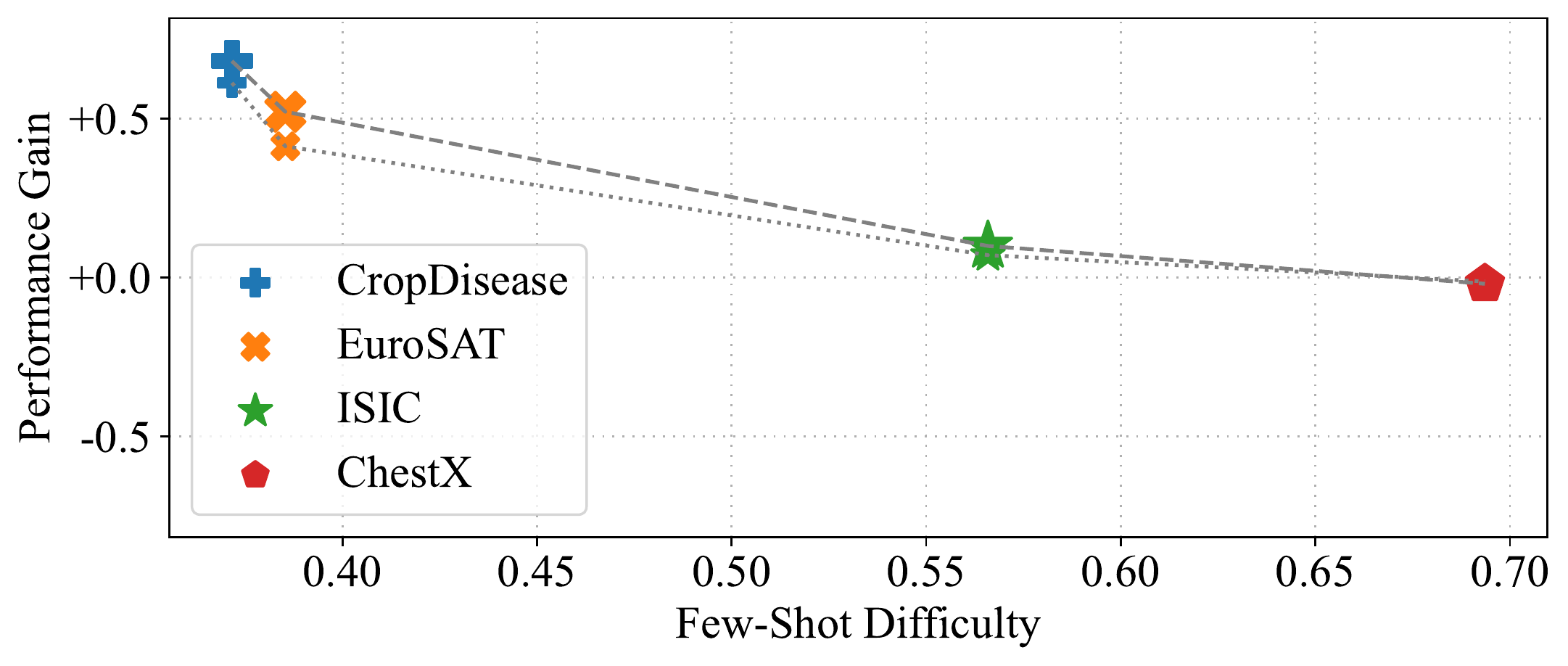}
         \caption{Small Similarity (miniImageNet)}
    \end{subfigure}
    \begin{subfigure}[h]{.49\linewidth}
         \centering
         \includegraphics[width=\linewidth]{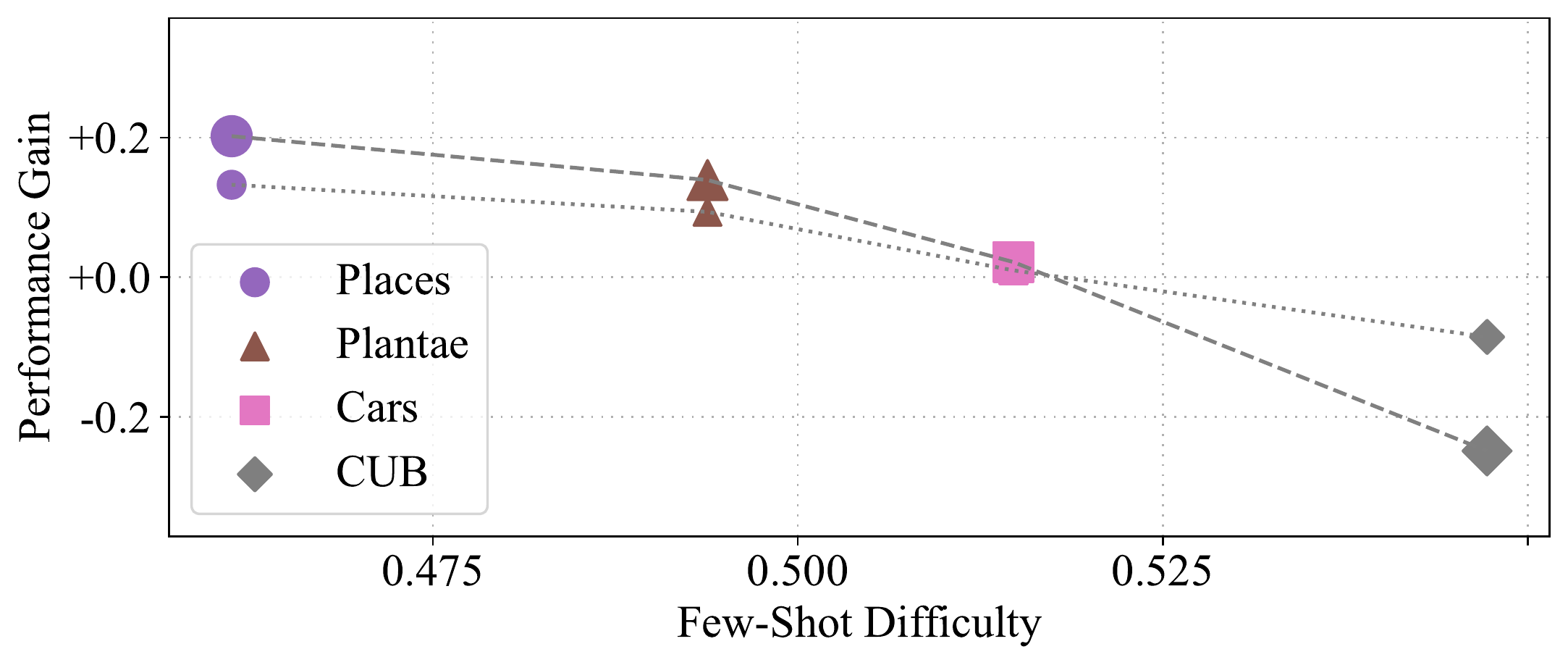}
         \caption{Large Similarity (miniImageNet)}
    \end{subfigure}
     \caption{5-way $k$-shot performance gain of SSL over SL for the two dataset groups according to the few-shot difficulty (small markers: $k$=1, large markers: $k$=5). Results are shown for two source datasets: ImageNet and miniImageNet, each with their corresponding backbones.
     }
     \label{fig:difficulty_main}
\end{figure}

In this sense, we study the impact of few-shot difficulty by categorizing the eight datasets into two groups: one with small domain similarity (\emph{i.e.}, BSCD-FSL) and another with large domain similarity (\emph{i.e.}, other datasets).
Figure \ref{fig:difficulty_main} shows the performance gain of SSL over SL for datasets with varying few-shot difficulty for each group. The performance gain of SSL over SL is defined as ${( {\rm Error}_{\sf sl}-{\rm Error}_{\sf ssl}} )/{{\rm Error}_{\sf sl}}$, which indicates the relative improvement of the classification error.

\vspace*{0.15cm}
\begin{observation}
\emph{Performance gain of SSL over SL becomes greater at smaller domain similarity or lower few-shot difficulty.}
\label{obs:obs5_3}
\end{observation}

\noindent{\scshape Evidence.} For both groups, the performance gain of SSL over SL becomes greater as few-shot difficulty decreases. In particular, the performance gain is the greatest on the CropDisease and Places datasets with the lowest few-shot difficulty in each group, while the performance gain is the least on the ChestX and CUB datasets with the highest few-shot difficulty in each group. For the target data with higher few-shot difficulty, \emph{it may not be easy to learn discriminative representations by solely using SSL without label supervision}.

Meanwhile, comparing the two groups (BSCD-FSL vs. other datasets), it is observed that the performance gain of SSL over SL is significantly worse for the group with large domain similarity. Namely, the performance gain is near or less than zero when domain similarity is large because features learned from SL with label supervision can be better transferred. Note that the negative value of performance gain means that SL outperforms SSL. Furthermore, the performance gain is closely related to the source dataset size for the datasets with large similarity (see Figures \ref{fig:difficulty_main}(b) and \ref{fig:difficulty_main}(d)). For instance, on the CUB dataset, the performance gain ($k$=5) is $-2.276$ and $-0.249$ for ImageNet and miniImageNet, respectively. However, when domain similarity is small (see Figures \ref{fig:difficulty_main}(a) and \ref{fig:difficulty_main}(c)), the source dataset size does not significantly affect the performance gain of SSL over SL.

In summary, we first conclude that SSL is advantageous to SL when the target domain is extremely dissimilar to the source domain (\emph{i.e.}, the performance gain is greater than 0), which is in line with Observation \ref{obs:obs4_1}.
This implies supervision with a huge amount of source data cannot overcome domain differences.
However, when domain similarity is large, the few-shot difficulty must be considered to determine a better strategy between SSL and SL.
Namely, SL becomes more preferable as few-shot difficulty increases due to the benefits from supervision on the source dataset. The same trend is observed when tieredImageNet is used as the source dataset (Appendix \ref{appx:other_source}).
\section{Advanced Scheme: MSL and Two-Stage}\label{sec:analysis3}

In this section, we further study SL and SSL in a more advanced scheme from the domain similarity and few-shot difficulty perspective, in line with previous observations.
We first investigate whether SL and SSL can synergize by studying MSL. Next, we analyze the two-stage pre-training scheme used in recent works \citep{phoo2021selftraining, islam2021dynamic}.

\subsection{Can SL and SSL Synergize?}
To identify whether SL and SSL can complement each other, we first consider a mixed-loss pre-training scheme, MSL, described in Eq.\,\eqref{eq:loss_msl}.
We define that synergy between SL and SSL occurs when MSL is superior to both SL and SSL.
Table \ref{tab:main_msl}(a) summarizes the performance of the models under each pre-training scheme on eight target datasets, grouped by their domain similarity (BSCD-FSL vs. other datasets) and then sorted by the few-shot difficulty in ascending order.
In MSL, the hyperparameter $\gamma$ is set to be 0.875 found by a grid search, detailed in Appendix \ref{appx:gamma_abla}.

\begin{observation}
\label{obs:obs6_1}
\vspace*{0.15cm}
\emph{SL and SSL can synergize when SL and SSL have similar performances.}
\end{observation} 

\noindent{\scshape Evidence.} In Table \ref{tab:main_msl}(a), it is observed that SL and SSL can synergize (\emph{i.e.}, MSL > SL, SSL) on four datasets: ISIC, ChestX, Places, and Plantae.
SL and SSL have similar performances on these datasets, as shown by the large markers ($k$=5) in Figures \ref{fig:difficulty_main}(a) and \ref{fig:difficulty_main}(b).
MSL can learn diverse features, owing to differences in training domains (i.e, source vs. target) and learning frameworks (i.e., supervised vs. unsupervised), which allows for synergy\,\citep{ericsson2021well, liu2021self, grigg2021self, gontijo-lopes2022no}.
However, when either SL or SSL significantly outperforms the other, MSL does not perform best.
In addition, MSL performance can be improved further in the large similarity group by emphasizing the SL component through a larger batch size (Appendix \ref{appx:batchsize}).

\begin{table*}[!t]
\caption{5-way 5-shot CD-FSL performance\,(\%) of the models pre-trained by SL, SSL, and MSL including their two-stage versions. ResNet18 is used as the backbone model, and ImageNet is used as the source data for SL and MSL. The balancing coefficient $\gamma$ in Eq.\,\eqref{eq:loss_msl} of MSL is set to be 0.875. Datasets are grouped by domain similarity and sorted by few-shot difficulty in ascending order in each group (CropDisease $<$ ChestX $\vert$ Places $<$ CUB). The {best} results are marked in bold.}\label{tab:main_msl}
\centering
\footnotesize\addtolength{\tabcolsep}{-3.0pt}
\resizebox{\linewidth}{!}{
\begin{tabular}{c|c|c|cccc|cccc}
    \toprule 
    &\,\,Pre-train\,\, & \multirow{2}{*}{Method} & \multicolumn{4}{c|}{Small Similarity} & \multicolumn{4}{c}{Large Similarity} \\
    & Scheme & & \!\!CropDisease\!\! & \!\!EuroSAT\!\! & ISIC & ChestX & Places & Plantae & Cars & CUB \\
    \midrule
    &SL & Default & 92.81{\scriptsize$\pm$.45} &  84.73{\scriptsize$\pm$.51} &  44.10{\scriptsize$\pm$.58} &  25.51{\scriptsize$\pm$.44} & 79.22{\scriptsize$\pm$.64} &  63.21{\scriptsize$\pm$.82} &  \textbf{66.38}{\scriptsize$\pm$.80} & \textbf{83.93}{\scriptsize$\pm$.66} \\ \cmidrule{2-11}
    &\multirow{2}{*}{SSL} & SimCLR & \textbf{97.46}{\scriptsize$\pm$.34} &  \textbf{94.12}{\scriptsize$\pm$.32} &  {47.85}{\scriptsize$\pm$.65} &  25.26{\scriptsize$\pm$.44} & 80.43{\scriptsize$\pm$.61} &  60.07{\scriptsize$\pm$.84} &  44.55{\scriptsize$\pm$.74} & 47.36{\scriptsize$\pm$.79} \\
    \rotatebox[origin=c]{90}{\makecell[l]{\!\!\!\!\!\!\!\!\!\!\!\!\!\!Single-Stage}\!\!\!\!\!\!\!\!\!\!\!\!\!\!}&& BYOL & {96.93}{\scriptsize$\pm$.30} &  87.83{\scriptsize$\pm$.48} &  47.59{\scriptsize$\pm$.63} &  {28.36}{\scriptsize$\pm$.46} & 72.47{\scriptsize$\pm$.63} &  61.02{\scriptsize$\pm$.82} &  48.56{\scriptsize$\pm$.76} & 51.31{\scriptsize$\pm$.78} \\\cmidrule{2-11}
    &\multirow{2}{*}{MSL} & \,\,SimCLR\,\, &  96.50{\scriptsize$\pm$.35} & 90.11{\scriptsize$\pm$.40} & 45.38{\scriptsize$\pm$.63} & 26.05{\scriptsize$\pm$.44} & \textbf{82.56}{\scriptsize$\pm$.58} & {64.76}{\scriptsize$\pm$.83} & {51.84}{\scriptsize$\pm$.79} & 64.53{\scriptsize$\pm$.80} \\
    && BYOL & 96.74{\scriptsize$\pm$.31} & {90.82}{\scriptsize$\pm$.40} & \textbf{49.14}{\scriptsize$\pm$.70} & \textbf{29.58}{\scriptsize$\pm$.47}& {81.27}{\scriptsize$\pm$.59} & \textbf{67.39}{\scriptsize$\pm$.81} & 46.76{\scriptsize$\pm$.73} & {69.67}{\scriptsize$\pm$.82} \\
    \bottomrule
    \end{tabular}\vspace*{0.05cm}}\\
{\small (a) Performance comparison for single-stage schemes.} \\
\resizebox{\linewidth}{!}{
\begin{tabular}{c|c|c|cccc|p{12mm}p{12mm}p{12mm}p{12mm}} \toprule
    &\multirow{2}{*}{\!SL$\rightarrow$SSL\!} & SimCLR & \textbf{97.88}{\scriptsize$\pm$.30} & \textbf{95.28}{\scriptsize$\pm$.27} & 48.38{\scriptsize$\pm$.60} & 25.25{\scriptsize$\pm$.44} & {84.40}{\scriptsize$\pm$.53} & 66.35{\scriptsize$\pm$.82} & 51.31{\scriptsize$\pm$.84} & 57.11{\scriptsize$\pm$.88} \\
    && BYOL & {97.58}{\scriptsize$\pm$.26} & {91.82}{\scriptsize$\pm$.39} & {49.32}{\scriptsize$\pm$.63} & {28.27}{\scriptsize$\pm$.48} & 78.87{\scriptsize$\pm$.60} & 67.83{\scriptsize$\pm$.82} & 54.70{\scriptsize$\pm$.84} & 60.60{\scriptsize$\pm$.82} \\ \cmidrule{2-11}
    \rotatebox[origin=c]{90}{\makecell[l]{\!\!\!\!\!\!\!\!\!\!\!\!\!\!\!\!Two-Stage\!\!\!\!\!\!\!\!\!\!\!\!\!\!}}&\multirow{2}{*}{\!SL$\rightarrow$MSL\!} & SimCLR & 97.49{\scriptsize$\pm$.30} & 91.70{\scriptsize$\pm$.35} & 47.43{\scriptsize$\pm$.62} & 26.24{\scriptsize$\pm$.44} & \textbf{85.76}{\scriptsize$\pm$.52} & {69.24}{\scriptsize$\pm$.81} & {58.97}{\scriptsize$\pm$.82} & {81.51}{\scriptsize$\pm$.72} \\
    && BYOL &  97.09{\scriptsize$\pm$.31} & 90.89{\scriptsize$\pm$.40} & \textbf{50.72}{\scriptsize$\pm$.67} & \textbf{30.20}{\scriptsize$\pm$.48} & 83.29{\scriptsize$\pm$.55} & \textbf{74.16}{\scriptsize$\pm$.77} & 68.87{\scriptsize$\pm$.80} & 84.34{\scriptsize$\pm$.67} \\
    \cmidrule{2-11}
    & \multirow{2}{*}{\!SL$\rightarrow$MSL\!$^\mathbf{+}$\!\!\!} & STARTUP & 96.06{\scriptsize$\pm$.33} & 89.70{\scriptsize$\pm$.41} & 46.02{\scriptsize$\pm$.59} & 27.24{\scriptsize$\pm$.46} & 85.00{\scriptsize$\pm$.52} & 69.40{\scriptsize$\pm$.84} & 68.43{\scriptsize$\pm$.82} & \textbf{89.60}{\scriptsize$\pm$.55} \\
    && DynDistill & 97.60{\scriptsize$\pm$.35} & 92.28{\scriptsize$\pm$.46} & 50.06{\scriptsize$\pm$.86} & 29.65{\scriptsize$\pm$.67} & 82.22{\scriptsize$\pm$.81}& 71.49{\tiny$\pm$1.06} & \textbf{69.45}{\tiny$\pm$1.12} & 86.54{\tiny$\pm$1.88} \\
    \bottomrule
\end{tabular}\vspace*{0.05cm}}\\
{\small (b) Performance comparison for two-stage schemes.}
\vspace*{-0.2cm}
\end{table*}

\subsection{Extension to Two-Stage Approach}

We extend the single-stage to two-stage approaches, extracting more sophisticated target representations. In two-stage pre-training, a model is pre-trained \emph{in prior} with labeled source data in the first phase and further trained through SSL or MSL in the second phase, \emph{i.e.}, SL\,$\rightarrow$\,SSL or SL\,$\rightarrow$\,MSL.
This pipeline has been adopted by recent algorithms, such as STARTUP\,\cite{phoo2021selftraining} and DynDistill\,\cite{islam2021dynamic}, but they additionally maintain an extra network or incorporate the knowledge distillation in the second phase, \emph{i.e.}, SL\,$\rightarrow$\,MSL$^{+}$.
Table \ref{tab:main_msl}(b) summarizes the CD-FSL performance of two-stage schemes.

\begin{observation}
\label{obs:obs6_2}
\vspace*{0.15cm}
\emph{Two-stage pre-training schemes are better than their single-stage counterparts.}
\end{observation} 

\noindent{\scshape Evidence.} Two-stage pre-training approaches generally achieve much higher performance than their single-stage counterparts, \emph{i.e.}, SL\,$\rightarrow$\,SSL outperforms SSL, and SL\,$\rightarrow$\,MSL outperforms MSL. When SL is used separately in the first phase, it appears to provide a good initialization for the second phase because a converged extractor on the source data is better than a random extractor \citep{neyshabur2020being}. 
Also, the benefit of the two-stage pre-training is significant when domain similarity is large. This observation is promising for practitioners because pre-trained models on ImageNet or bigger datasets are readily accessible.
In addition, our simple two-stage methods, without any additional techniques, are shown to achieve comparable performance to the meticulously designed two-stage approaches such as STARTUP, even though our main goal is analysis of basic pre-training methods. Appendix \ref{appx:summary} summarizes the full results including meta-learning based algorithms.
\vspace*{-7pt}
\section{Conclusion}\label{sec:conclusion}
\vspace*{-7pt}

We established a thorough empirical understanding of CD-FSL. Our work is a pioneering study that unveils hidden findings in the empirical use of CD-FSL. We believe it can inspire subsequent studies like theoretical analysis, which our paper did not cover.
In particular, we focused on the effectiveness of SL, SSL, and MSL, which can be realized with single- and two-stage pre-training schemes.
We (1)\,observed that their performances are closely related to domain similarity between the source and target datasets and few-shot difficulty of the target dataset, and (2)\,proposed how they can be effectively combined for pre-training.
Through our empirical study, we presented six findings that have been either misunderstood or unexplored. To justify all the findings, extensive experiments were conducted on benchmarks with varying degrees of domain similarity and few-shot difficulty. 


\section*{Acknowledgements}
This work was supported by Institute of Information \& communications Technology Planning \& Evaluation (IITP) grant funded by the Korea government(MSIT) (No.2019-0-00075, Artificial Intelligence Graduate School Program(KAIST), 10\%) and Institute of Information \& communications Technology Planning \& Evaluation(IITP) grant funded by the Korea government(MSIT) (No. 2022-0-00871, Development of AI Autonomy and Knowledge Enhancement for AI Agent Collaboration, 90\%)

\bibliography{reference}
\bibliographystyle{abbrvnat}

\newpage
\section*{Checklist}


\begin{enumerate}

\item For all authors...
\begin{enumerate}
  \item Do the main claims made in the abstract and introduction accurately reflect the paper's contributions and scope?
    \answerYes{In the abstract and introduction, we established the contributions and scope of our paper as the six findings for CD-FSL, which are also described in Figure \ref{fig:similarity_difficulty} in a compact manner.}
  \item Did you describe the limitations of your work?
    \answerYes{Refer to Conclusion. As our work is a pioneering study, we exhaustively analyzed CD-FSL and provided several insightful observations; however, there was a lack of theoretical analysis.}
  \item Did you discuss any potential negative societal impacts of your work?
    \answerNo{We have checked the ethics guidelines and think no corresponding aspects were found.}
  \item Have you read the ethics review guidelines and ensured that your paper conforms to them?
    \answerYes{We read and ensure it.}
\end{enumerate}

\item If you are including theoretical results...
\begin{enumerate}
  \item Did you state the full set of assumptions of all theoretical results?
    \answerNA{}
        \item Did you include complete proofs of all theoretical results?
    \answerNA{}
\end{enumerate}

\item If you ran experiments...
\begin{enumerate}
  \item Did you include the code, data, and instructions needed to reproduce the main experimental results (either in the supplemental material or as a URL)?
    \answerYes{Refer to Abstract. We provided the code URL, where we also described how to set up the data.}
  \item Did you specify all the training details (e.g., data splits, hyperparameters, how they were chosen)?
    \answerYes{Refer to Appendix \ref{appx:detail} and \ref{appx:implement_detail} for dataset and implementation details.}
        \item Did you report error bars (e.g., with respect to the random seed after running experiments multiple times)?
    \answerYes{We added the 95\% confidence interval of 600 episodes, following other few-shot learning works.}
        \item Did you include the total amount of compute and the type of resources used (e.g., type of GPUs, internal cluster, or cloud provider)?
    \answerYes{Refer to Appendix \ref{appx:training_setup} for the training details, including the amount of resources used.}
\end{enumerate}

\item If you are using existing assets (e.g., code, data, models) or curating/releasing new assets...
\begin{enumerate}
  \item If your work uses existing assets, did you cite the creators?
    \answerYes{We provided the full details on the existing algorithms and datasets in Appendix \ref{appx:ssl_methods} and \ref{appx:detail}.}
  \item Did you mention the license of the assets?
    \answerYes{We included the license information of datasets and our own code assets in the code URL attached in Abstract.}
  \item Did you include any new assets either in the supplemental material or as a URL?
    \answerYes{Refer to Abstract. We attached our modified code asset as a form of URL, and this will be open via GitHub.}
  \item Did you discuss whether and how consent was obtained from people whose data you're using/curating?
    \answerNA{}
  \item Did you discuss whether the data you are using/curating contains personally identifiable information or offensive content?
    \answerNA{}
\end{enumerate}

\item If you used crowdsourcing or conducted research with human subjects...
\begin{enumerate}
  \item Did you include the full text of instructions given to participants and screenshots, if applicable?
    \answerNA{}
  \item Did you describe any potential participant risks, with links to Institutional Review Board (IRB) approvals, if applicable?
    \answerNA{}
  \item Did you include the estimated hourly wage paid to participants and the total amount spent on participant compensation?
    \answerNA{}
\end{enumerate}

\end{enumerate}


\newpage
\appendix

\section{Self-Supervised Learning Methods}\label{appx:ssl_methods}

\subsection{SimCLR}
SimCLR \citep{chen2020simple} is one of the simplest yet high-performance contrastive learning methods. Its key idea is mapping the semantically similar examples to be close in the representation space while dissimilar examples to be distant. The similar examples are often called positive samples, and the dissimilar ones are called negative samples. Formally, all examples in the current batch $\!\{x_k\}_{k=1:B}$ with size $B$ are augmented to generated an augmented batch $\{\tilde{x}_{2k-1}, \tilde{x}_{2k}\}_{k=1:B}$, where ${\tilde{x}}_{2k-1}$ and ${\tilde{x}}_{2k}$ are the examples differently augmented from the same input $x_k$. Then, the representations $\{z_{2k-1}, z_{2k}\}_{k=1:B}$ are extracted from a feature extractor with projection layers. Based on the representations, SimCLR performs contrastive learning such that it minimizes the contrastive loss:
\vspace*{-3pt}
\begin{gather*}
\mathcal{L}_{\text{SimCLR}} = \frac{1}{2B} \sum_{k=1}^{B} \Big[\ell(2k-1, 2k)+\ell(2k,2k-1)\Big] \\
\text{where} \,\, \ell(i,j) = -\log \frac{\exp({\sf sim}(z_i, z_j) / \tau)}{\sum_{n=1}^{2B} \mathbf{1}_{[n \neq i]} \exp({\sf sim}(z_i, z_n) / \tau)}
\end{gather*}
\noindent where $\mathbf{1}$ is an indicator function, $\tau$ is a temperature hyperparameter, and ${\sf sim}(\mathbf{u},\mathbf{v})=\mathbf{u}^\top \mathbf{v}/\|\mathbf{u}\|\|\mathbf{v}\|$ measures cosine similarity between two vectors $\mathbf{u}$ and $\mathbf{v}$.

\subsection{MoCo}
MoCo (Momentum Contrast \citep{he2020momentum}) is a variant of SimCLR method, which leverages the memory bank and the momentum update of an encoder. Similar to SimCLR, MoCo also minimizes the contrastive loss with positive and negative samples; the positive sample is the other augmentation (view) from the same instance, but the negative samples are not those from the current batch. Instead, MoCo fetches the negative samples from the memory bank, which has been enqueued from the previous batches. To emphasize the usage of the memory bank, the anchor sample, which is contrasted by positive and negative samples, is called query $q$, while the others are called keys $\{k_0,k_1,\dots\,k_K\}$. A positive key $k_0$ is the augmentation from the same sample as $q$. Then, MoCo minimizes the following loss:
\vspace*{-3pt}
\begin{equation*}
\mathcal{L}_{\text{MoCo}} = - \frac{1}{B}\sum_{i=1}^B \log \frac{\exp({\sf sim}(q(i), k_0(i)) / \tau)}{\sum_{j=0}^{K} \exp({\sf sim}(q(i), k_j(i)) / \tau)}
\end{equation*}
where the query representation and the key representations are extracted from different models. That is, $q=f_q(x_q)$ where $f_q$ is a main encoder, while $k=f_k(x_k)$ where $f_k$ is a momentum encoder that is updated by the moving average of its previous state and that of $f_q$. MoCo overcomes the dependency of the negative sample size on batch size, efficiently achieving the objective of SimCLR using the small memory bank and the additional network.

There are two versions of MoCo: MoCo-v1 \citep{he2020momentum} and MoCo-v2 \citep{chen2020improved}. Since MoCo-v2 is a simple improvement of MoCo-v1, such as cosine annealing, MLP projector, and different hyperparameters, we only considered the MoCo-v1 version in this paper.

\subsection{BYOL}
While SimCLR and MoCo used positive and negative samples to construct a contrastive task, BYOL (Bootstrap Your Own Latent \citep{grill2020bootstrap}) achieves higher performances than state-of-the-art contrastive learning models without using the negative samples. That is, BYOL is a non-contrastive SSL method, completely free from the need for negative samples. To this end, BYOL minimizes a similarity loss between the two augmented views using two networks. 

There are two networks involved: online network $f_\theta$ and target network $f_\xi$. This is a similar setting to MoCo; the online network is a main encoder and the target network is an encoder that is updated by weighted moving average. Given an image $x$, it augments $x$ into two views $\tilde{x}$ and $\tilde{x}'$. Each view is represented by the encoder with a projector $g_\theta$ and $g_\xi$: $z_\theta = g_\theta(f_\theta(\tilde{x}))$ and $z_\xi' = g_\xi(f_\xi(\tilde{x}'))$. Then, by prediction layers $q_\theta$, a prediction $q_\theta(z_\theta)$ is output and it is compared with the target projection. BYOL uses a mean squared error between the normalized prediction and target projection:
\vspace*{-3pt}
\begin{equation*}
\mathcal{L}_{\text{BYOL}} = 2 - 2\cdot \frac{\langle q_\theta(z_\theta), z_\xi' \rangle}{\| q_\theta(z_\theta) \|_2 \cdot \| z_\xi' \|_2}
\end{equation*}
BYOL also uses a symmetric loss function that passes $\tilde{x}'$ through the online network and $\tilde{x}$ through the target network. The two losses are summed, and the same thing is done for every sample in a batch. 

\subsection{SimSiam}
SimSiam (Simple Siamese \citep{chen2021exploring}) basically shares a similar idea to the BYOL model. The loss form is exactly the same, but SimSiam does not use an extra target network that is updated by momentum. Instead, SimSiam uses the same online network $f_\theta$ to output the representation of the two views $\tilde{x},\tilde{x}'$, but blocks the gradient flow for the target projection. While \citet{grill2020bootstrap} insisted in BYOL on the importance of a momentum encoder since it can prevent collapsing, \citet{chen2021exploring} found that a stop-gradient operation is a key to avoiding collapsing. Thus, SimSiam loss is described as follows:
\vspace*{-3pt}
\begin{equation*}
\mathcal{L}_{\text{SimSiam}} = 2 - 2\cdot \frac{\langle q_\theta(z_\theta), {\sf sg}(q_\theta(z_\theta')) \rangle}{\| q_\theta(z_\theta) \|_2 \cdot \| {\sf sg}(q_\theta(z_\theta')) \|_2}
\end{equation*}
where ${\sf sg}$ indicates the stop-gradient operation.

\section{Datasets Details}\label{appx:detail}

\subsection{Datasets}

In this paper, we used two source domain datasets and eight target domain datasets. Table \ref{tab:dataset_summary} summarizes the referenced papers, number of classes, and number of samples of each dataset. For source domain datasets, we used miniImageNet and tieredImageNet, which are two different subsets of the ImageNet-1k dataset \citep{imagenet}. The source dataset for miniImageNet (miniImageNet-train) includes 64 base classes, while the target dataset for miniImageNet (miniImageNet-test) include 20 classes that are disjoint from miniImageNet-train, following Appendix \ref{appx:same-domain-fsl-results}. Similarly, tieredImageNet is partitioned into a train and test set for the source data and target data, respectively. In our FSL experiments, we also reported the performance of SL model pre-trained on ImageNet. However, we did not actually pre-train with the ImageNet dataset, but fine-tuned from the pre-trained model offered by an official PyTorch \citep{PyTorch} library.

The target domain datasets can be separated into two groups: BSCD-FSL benchmark \citep{bscd_fsl} and non-BSCD-FSL. First, the BSCD-FSL benchmark includes CropDisease, EuroSAT, ISIC, and ChestX. These datasets are \textit{supposed} to be distant from the miniImageNet source, with CropDisease most similar and ChestX most dissimilar. The criteria are perspective distortion, semantic content, and color depth. We followed \citet{phoo2021selftraining} for the splitting procedure of the target dataset into a pre-training unlabeled set and a few-shot evaluation set. A short description of each dataset is provided below.
\vspace*{-3pt}

\begin{itemize}[leftmargin=*]
    \item \textbf{CropDisease} is a set of diseased plant images.
    \item \textbf{EuroSAT} is a set of satellite images of the landscapes.
    \item \textbf{ISIC} is a set of dermoscopy images of human skin lesions.
    \item \textbf{ChestX} is a set of X-Ray images on the human chest.
\end{itemize}

In addition to the BSCD-FSL benchmark, we introduced four target datasets that are more commonly used in the (CD-)FSL literature. They are Places, Plantae, Cars, and CUB. However, there is no standard rule to separate the pre-training set and the evaluation set for these four datasets. Thus, we sampled the images from each dataset. A short description and the sampling strategy of a dataset are provided below. Also, for the reproducibility of our work, we provide the code for the sampling procedure and the list of images we used.
\vspace*{-3pt}

\begin{itemize}[leftmargin=*]
    \item \textbf{Places} contains the images designed for scene recognition, such as bedrooms and streets, etc. However, because Places is an enormous dataset to use in the FSL context, we sampled 16 classes out of 365 classes (in a total of train, val, and test). Also, to make the dataset size smaller, we sampled 1,715 images per class, which is a reduced amount from the original 4,941 images per class on average.
    \item \textbf{Plantae} contains the plant images. Similar to Places, we sampled some images to reduce the dataset size. However, unlike Places, Plantae is a highly class-imbalanced set. Therefore, we sampled the top 69 classes that have many samples out of 2,917 classes. 
    \item \textbf{Cars} contains the images of 196 car models. We used the entire images that the Cars dataset has (train and test).
    \item \textbf{CUB} contains the images of 200 species of birds. We used the entire images that the CUB dataset has (train, val, and test).
\end{itemize}

\vspace{-0.5cm}
\begin{table}[h]
\caption{Summary of datasets we used in this paper. Note that we used a subset of images for Places and Plantae dataset.}
\vskip 0.10in
\centering
\small
\begin{tabular}{ccccc}
    \toprule
    Datasets  & miniImageNet-train & miniImageNet-test & tieredImageNet-train & tieredImageNet-test \\
    \midrule
    Reference & \citet{matchingnet} &  \citet{matchingnet} & \citet{tieredimagenet} & \citet{tieredimagenet} \\
    \# of classes & 64 & 20 & 351 & 160 \\
    \# of samples & 38,400 & 12,000 & 448,695 & 206,209 \\
    \midrule
    \midrule
    Datasets  & CropDisease & EuroSAT & ISIC & ChestX \\
    \midrule
    Reference & \citet{cropdisease} &  \citet{eurosat} & \citet{isic} & \citet{chestx} \\
    \# of classes & 38 & 10 & 7 & 7 \\
    \# of samples & 43,456 & 27,000 & 10,015 & 25,848 \\
    \midrule
    \midrule
    Datasets & Places & Plantae & Cars & CUB \\
    \midrule
    Reference & \citet{zhou2017places} & \citet{van2018inaturalist} & \citet{krause20133d} & \citet{WelinderEtal2010} \\
    \# of classes & 16 & 69 & 196 & 200 \\
    \# of samples & 27,440 & 26,650 & 16,185 & 11,788  \\
    \bottomrule
    \end{tabular}\label{tab:dataset_summary}
\end{table}

\begin{figure}[h]
     \centering
     \begin{subfigure}[h]{0.22\linewidth}
         \centering
         \includegraphics[width=\linewidth]{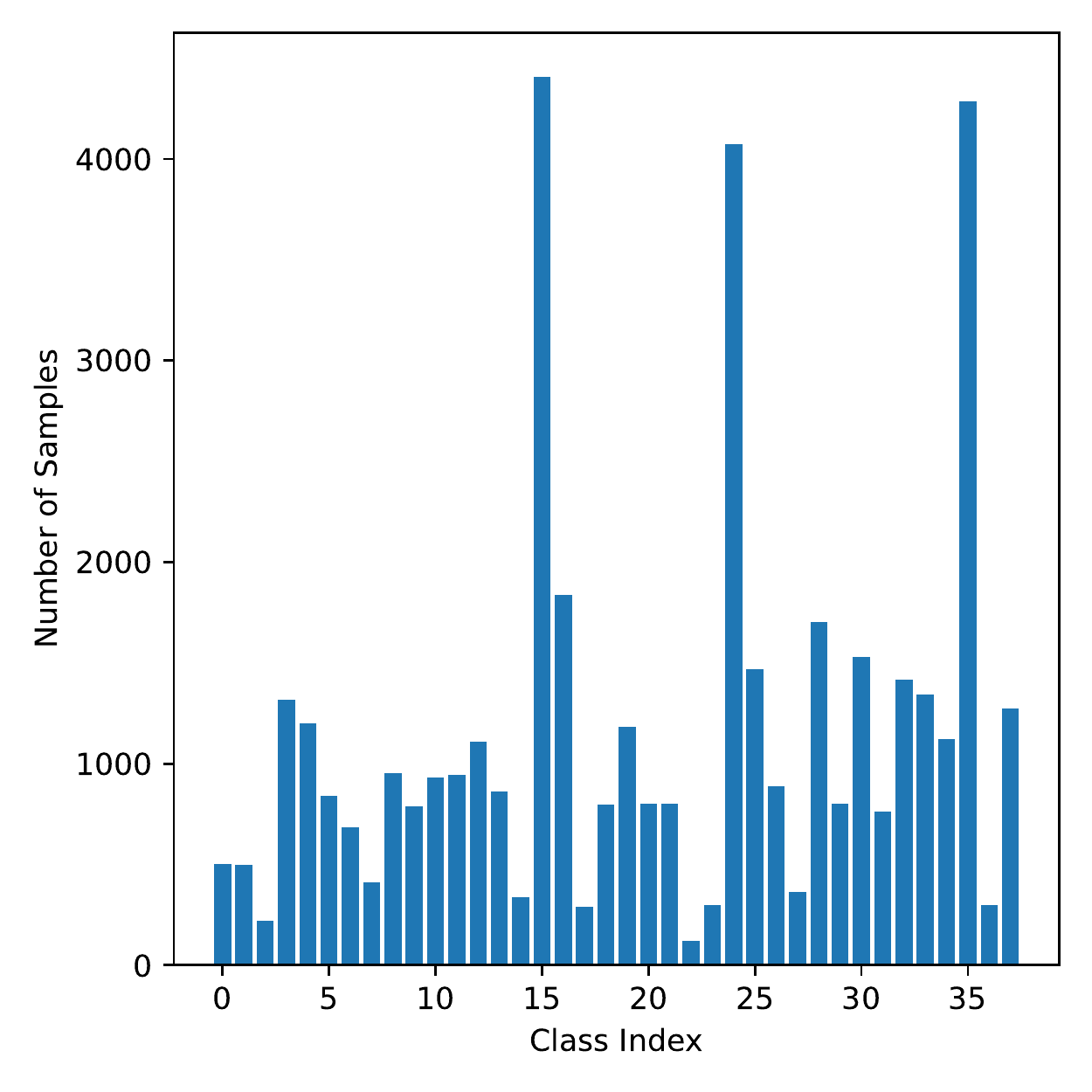}
         \caption{CropDisease}
         \label{fig:dataset_distribution_crop}
     \end{subfigure}
     \hfill
     \begin{subfigure}[h]{0.22\linewidth}
         \centering
         \includegraphics[width=\linewidth]{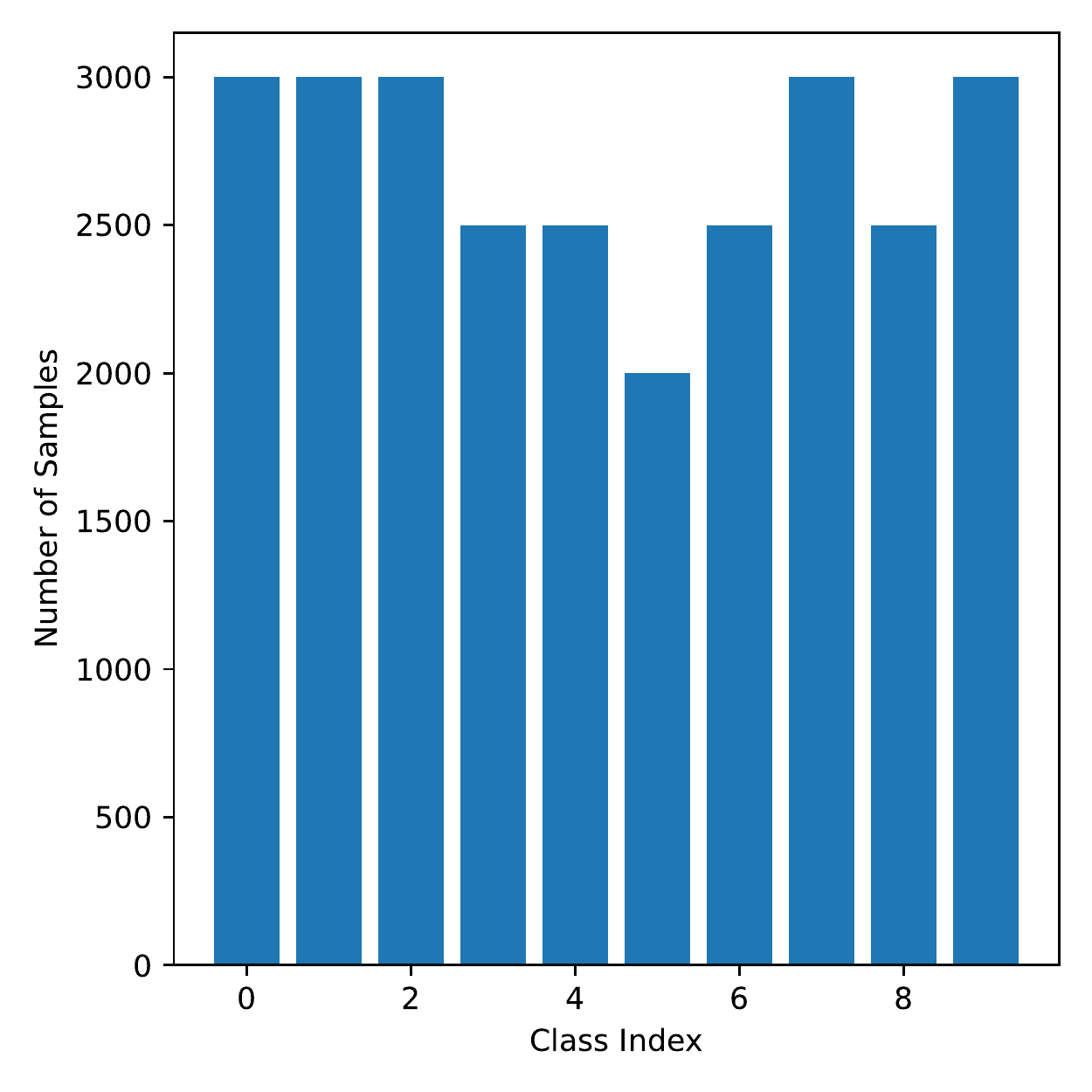}
         \caption{EuroSAT}
         \label{fig:dataset_distribution_euro}
     \end{subfigure}
     \hfill
     \begin{subfigure}[h]{0.22\linewidth}
         \centering
         \includegraphics[width=\linewidth]{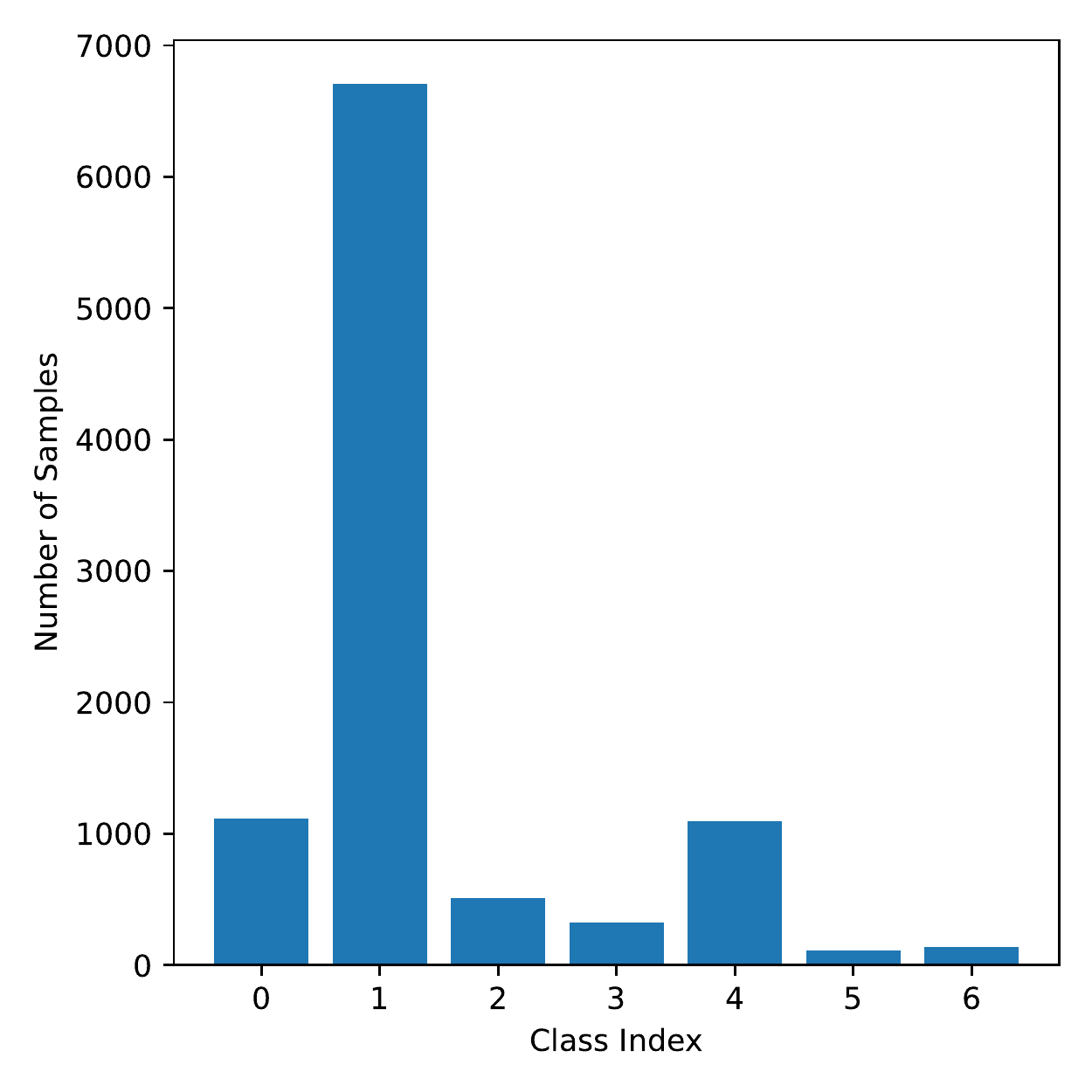}
         \caption{ISIC}
         \label{fig:dataset_distribution_isic}
     \end{subfigure}
     \hfill
     \begin{subfigure}[h]{0.22\linewidth}
         \centering
         \includegraphics[width=\linewidth]{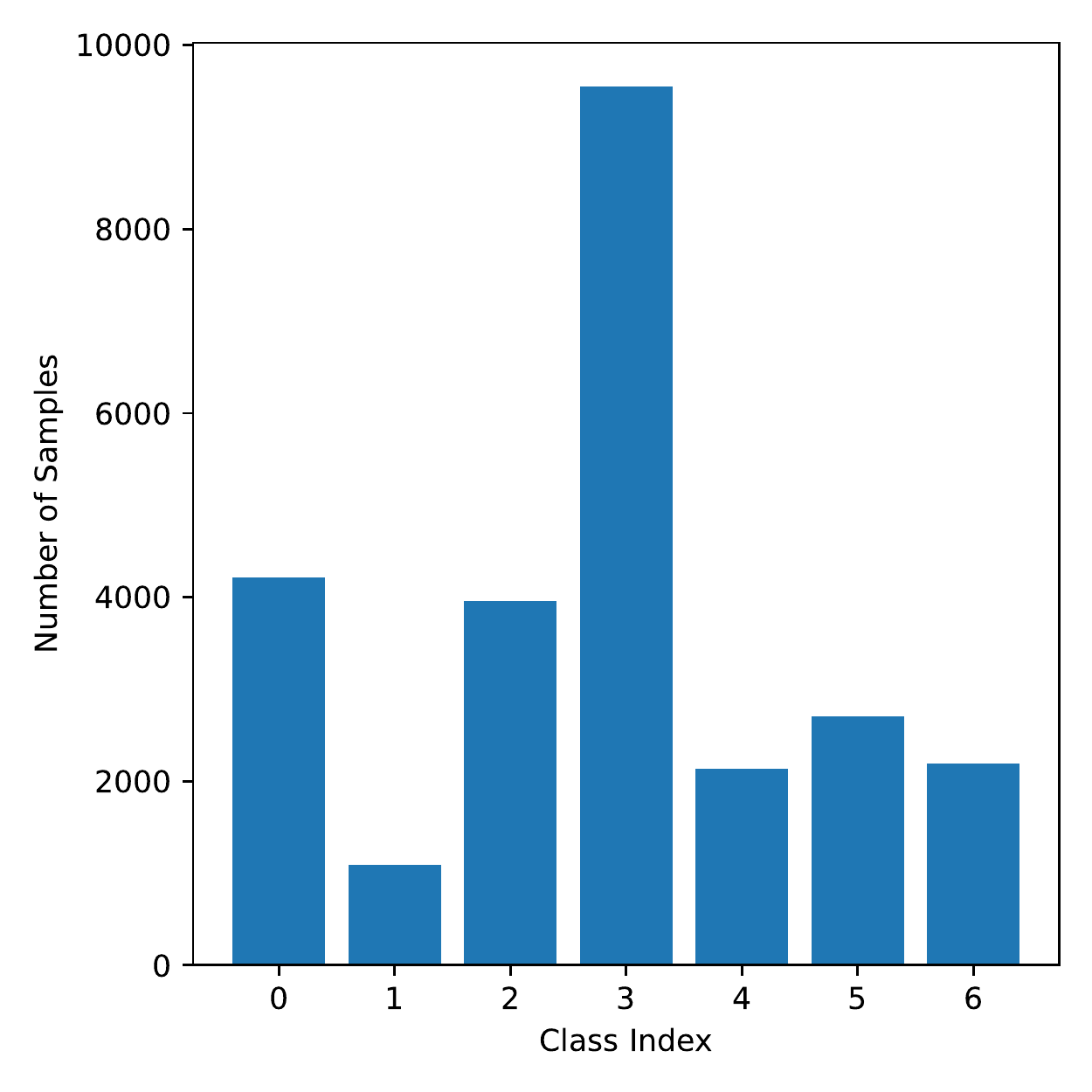}
         \caption{ChestX}
         \label{fig:dataset_distribution_chest}
     \end{subfigure}
     
     \begin{subfigure}[h]{0.22\linewidth}
         \centering
         \includegraphics[width=\linewidth]{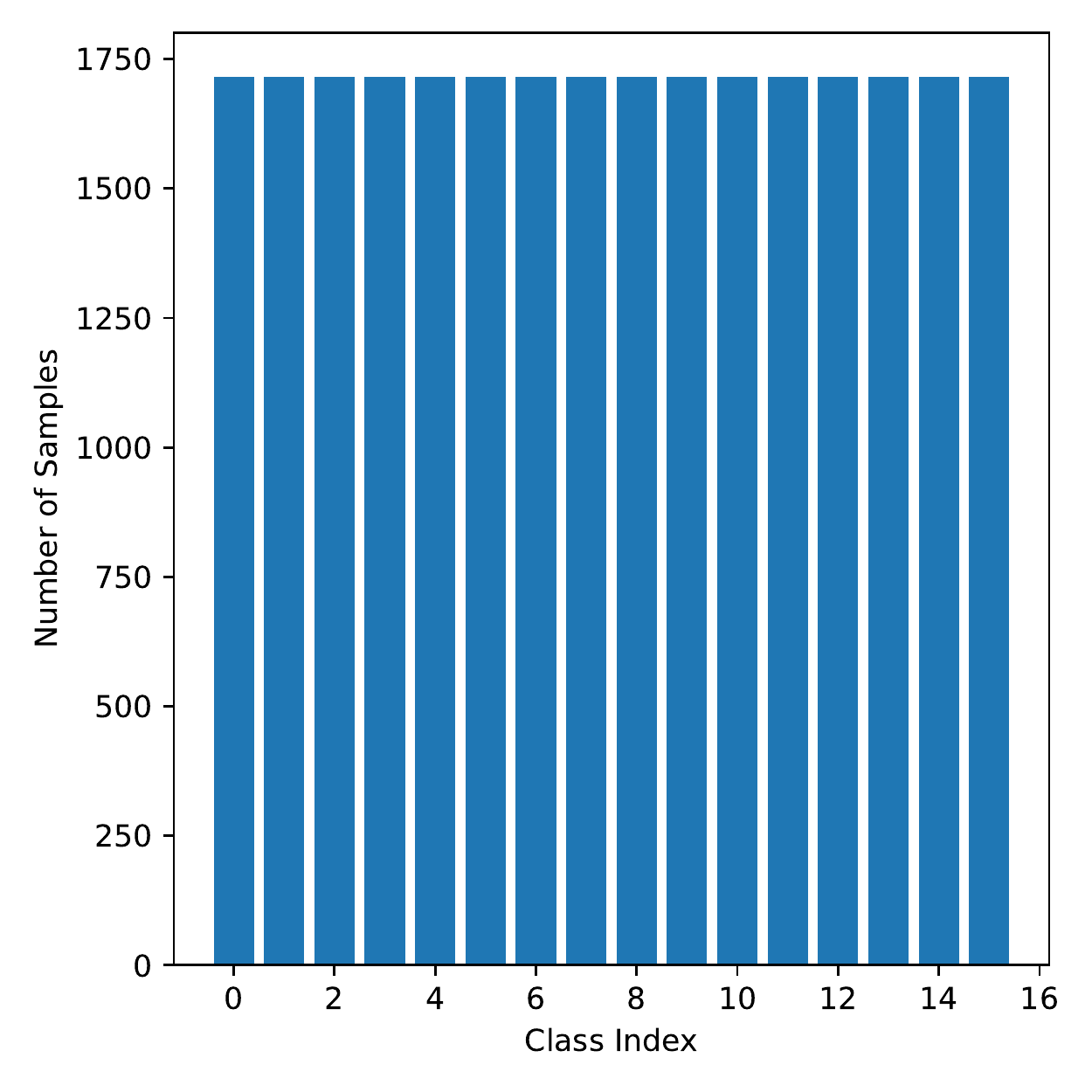}
         \caption{Places}
         \label{fig:dataset_distribution_places}
     \end{subfigure}
     \hfill
     \begin{subfigure}[h]{0.22\linewidth}
         \centering
         \includegraphics[width=\linewidth]{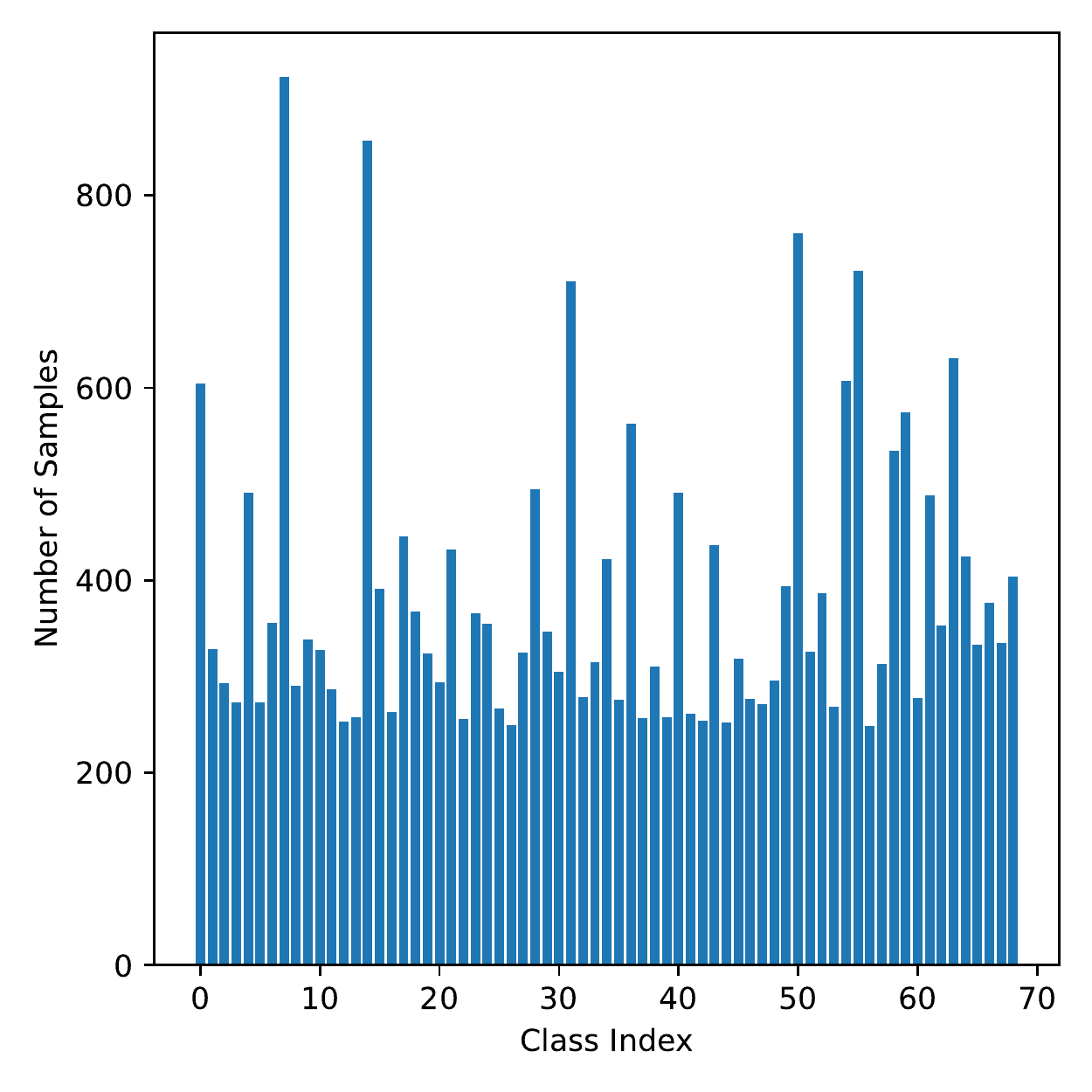}
         \caption{Plantae}
         \label{fig:dataset_distribution_plantae}
     \end{subfigure}
     \hfill
     \begin{subfigure}[h]{0.22\linewidth}
         \centering
         \includegraphics[width=\linewidth]{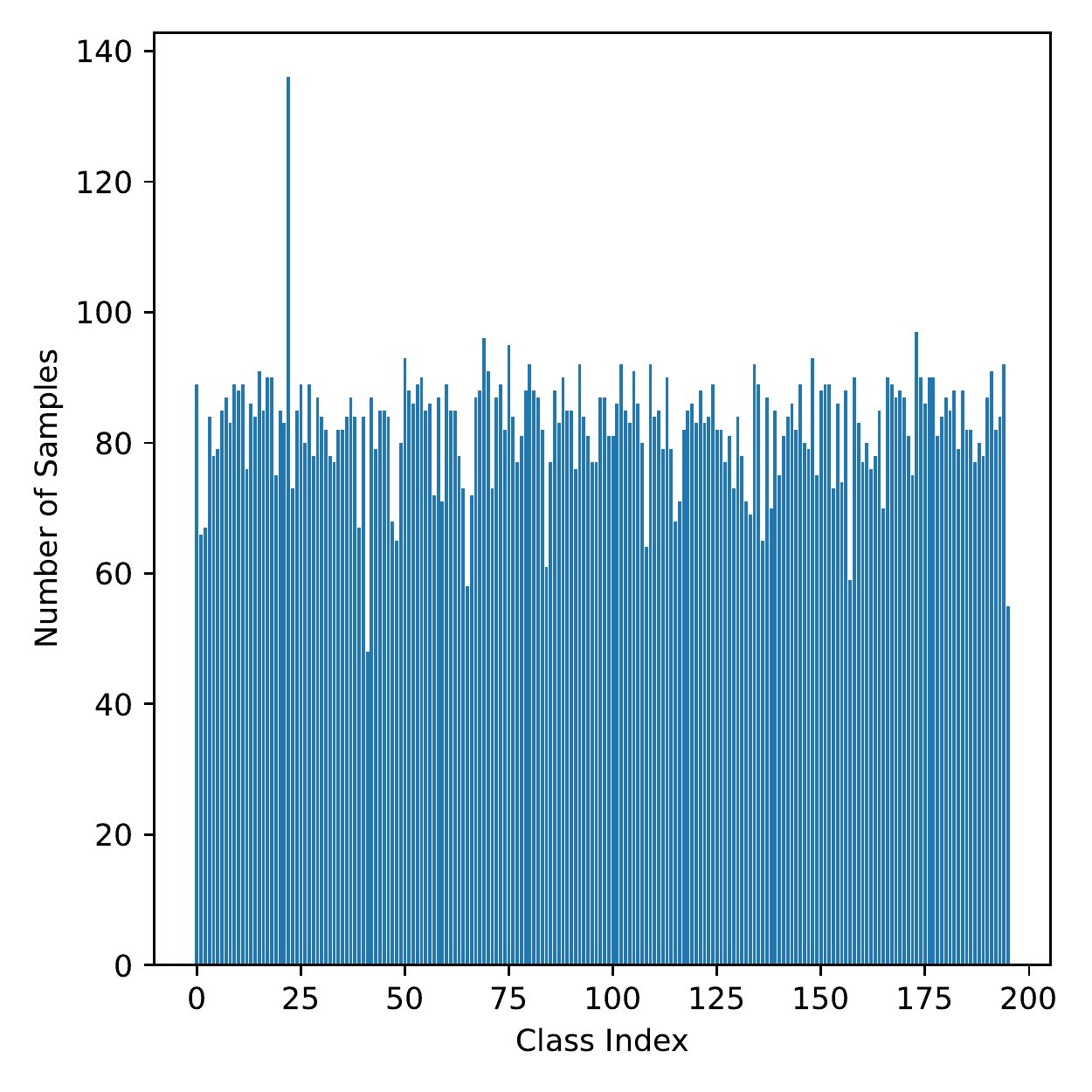}
         \caption{Cars}
         \label{fig:dataset_distribution_cars}
     \end{subfigure}
     \hfill
     \begin{subfigure}[h]{0.22\linewidth}
         \centering
         \includegraphics[width=\linewidth]{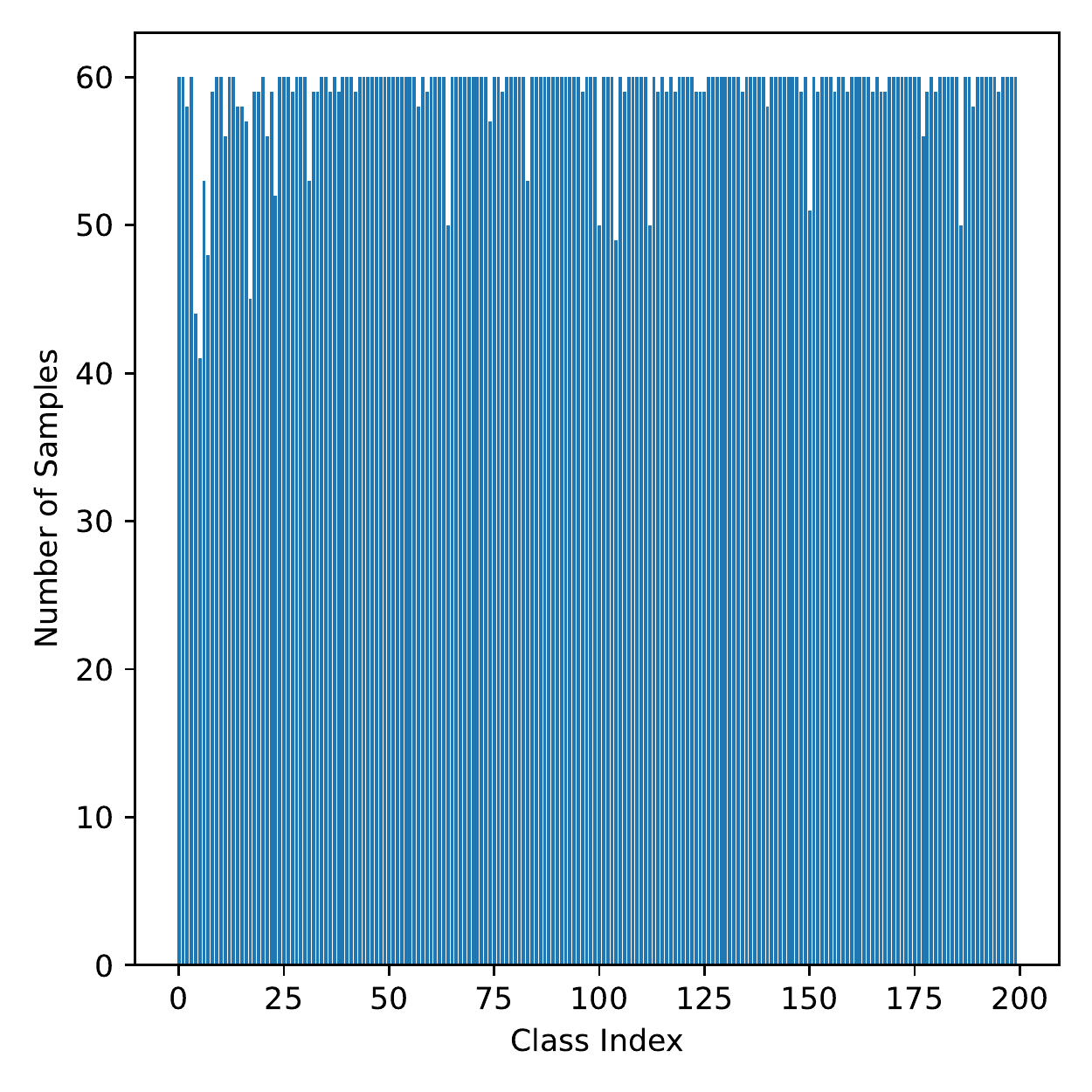}
         \caption{CUB}
         \label{fig:dataset_distribution_cub}
     \end{subfigure}
     \caption{Class distributions of eight target datasets considered in our study.}
     \label{fig:dataset_distribution}
\end{figure}

Figure \ref{fig:dataset_distribution} shows the class distribution of each target dataset considered in our study. We observe major differences in the class distributions. For example, the EuroSAT, Places, and CUB datasets have overall balanced class distributions, while the ISIC dataset is extremely unbalanced, with the number of samples per class ranging from 115 to 9,547. We also see that the number of samples per class varies over the eight datasets. The average number of samples per class for the ChestX dataset is 3,693, while for the CUB dataset, this number goes down to only 59.

We posit that the class distribution contributes to the difficulty of each dataset, thus implicitly considered as part of our analysis of target datasets. However, we note that class imbalance is not the deciding factor in dataset difficulty. For example, the CropDisease dataset has a relatively imbalanced class distribution yet is shown to have very low difficulty in our study. Explicitly, the effects of class distribution on CD-FSL have not been studied in our paper.

\clearpage
\subsection{Image Examples}
\vspace*{-5pt}

\begin{figure}[h!]
     \centering

     \begin{subfigure}[h]{0.24\linewidth}
         \centering
         \includegraphics[width=\linewidth]{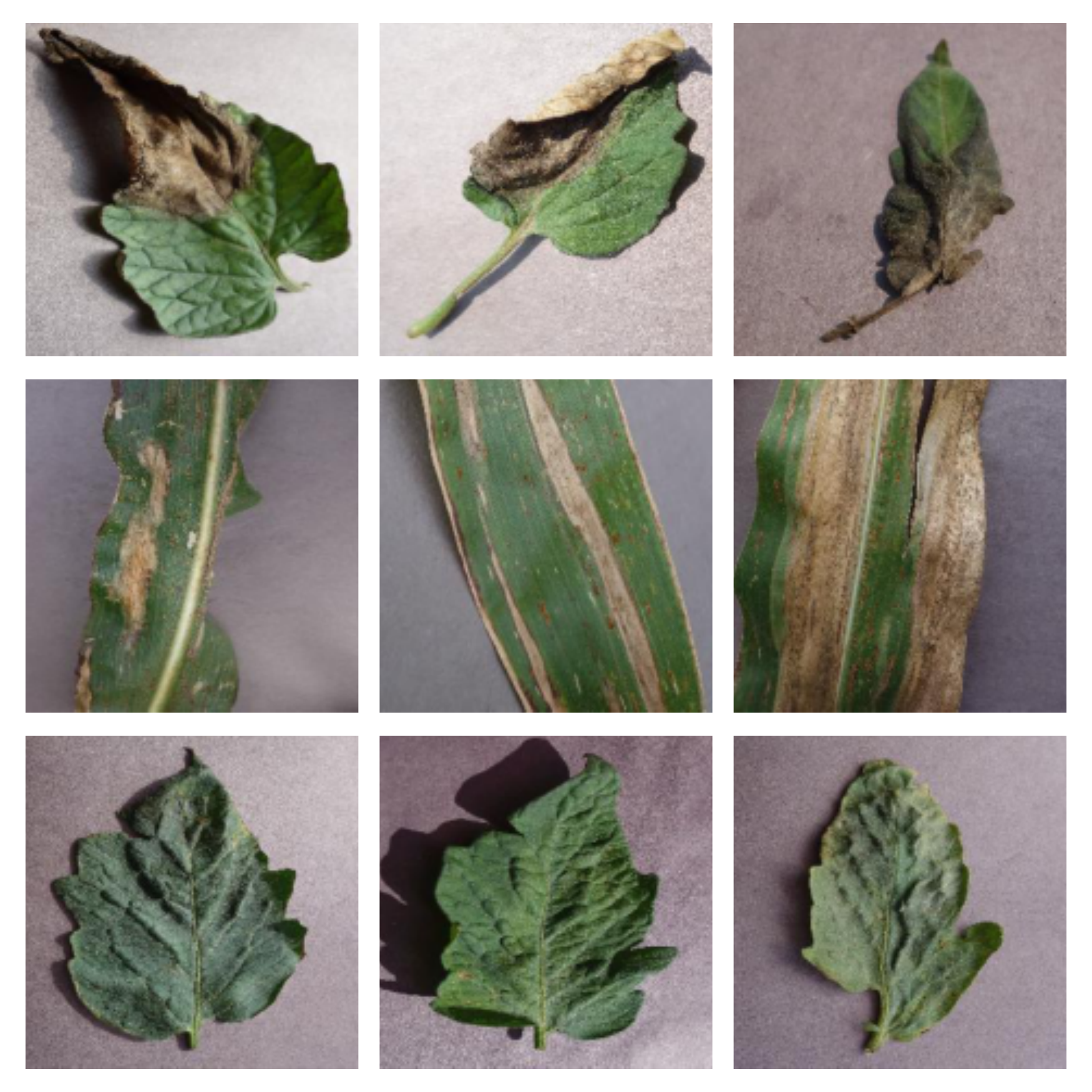}
         \caption{CropDisease}
         \label{fig:image_examples_crop}
     \end{subfigure}
     \hfill
     \begin{subfigure}[h]{0.24\linewidth}
         \centering
         \includegraphics[width=\linewidth]{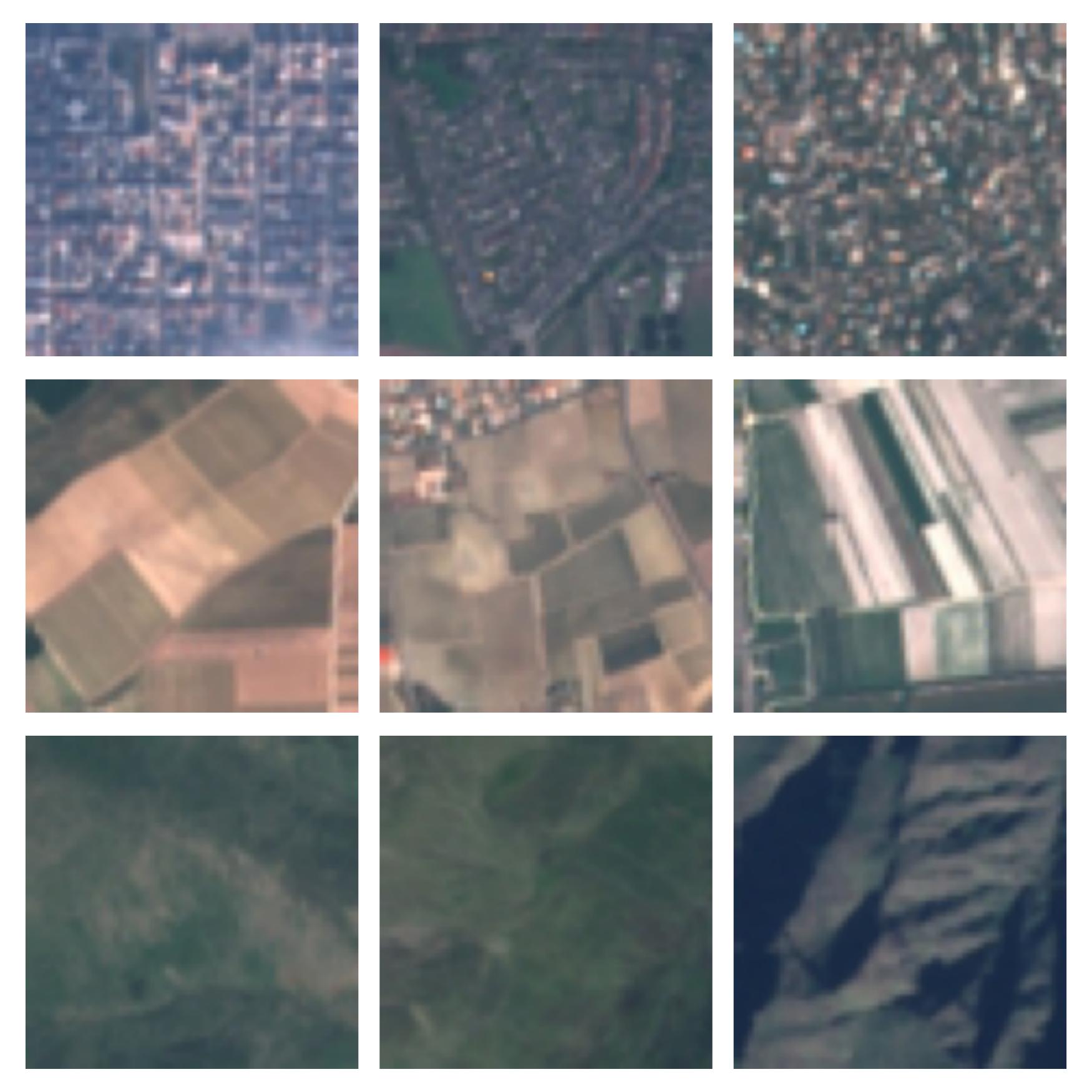}
         \caption{EuroSAT}
         \label{fig:image_examples_euro}
     \end{subfigure}
     \hfill
     \begin{subfigure}[h]{0.24\linewidth}
         \centering
         \includegraphics[width=\linewidth]{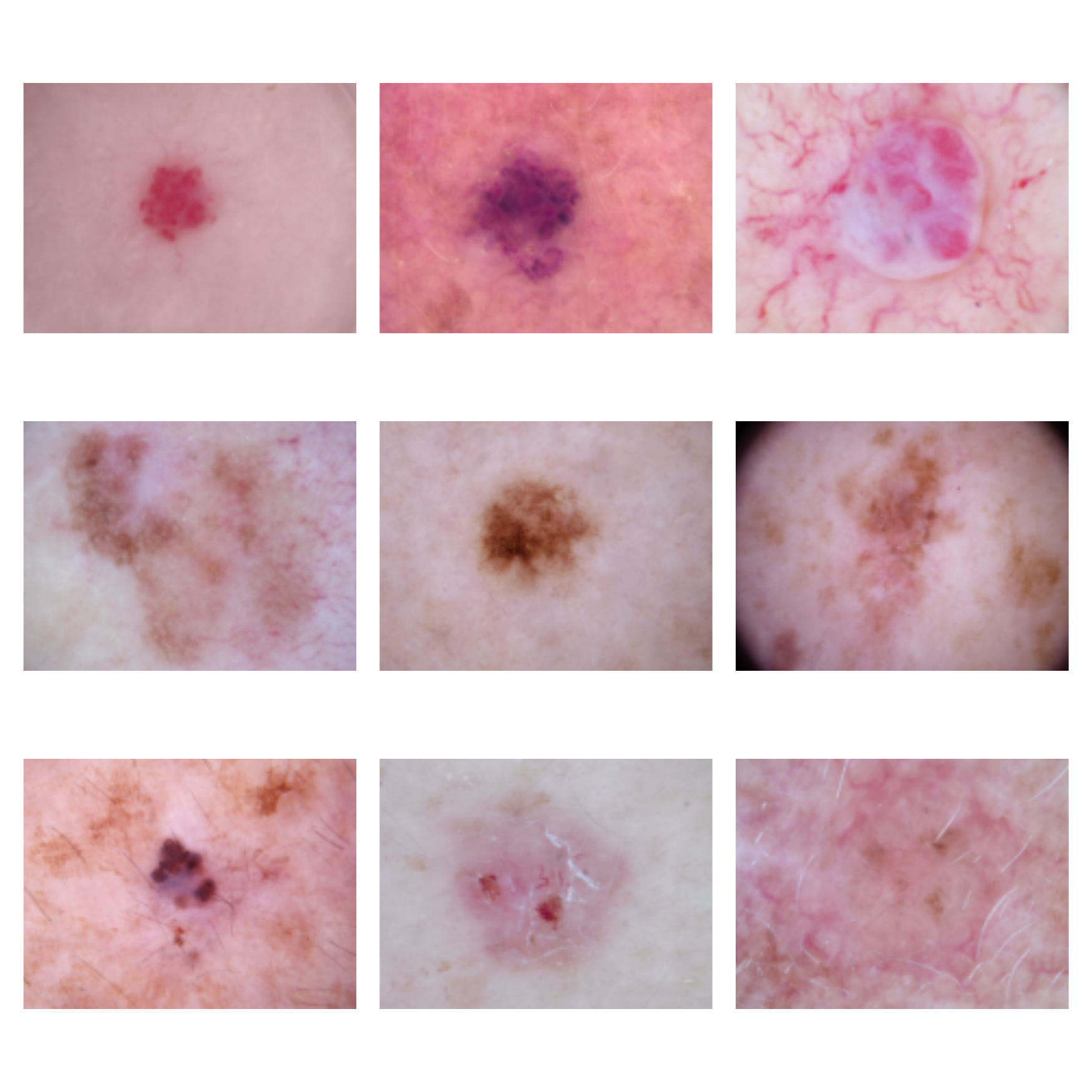}
         \caption{ISIC}
         \label{fig:image_examples_isic}
     \end{subfigure}
     \hfill
     \begin{subfigure}[h]{0.24\linewidth}
         \centering
         \includegraphics[width=\linewidth]{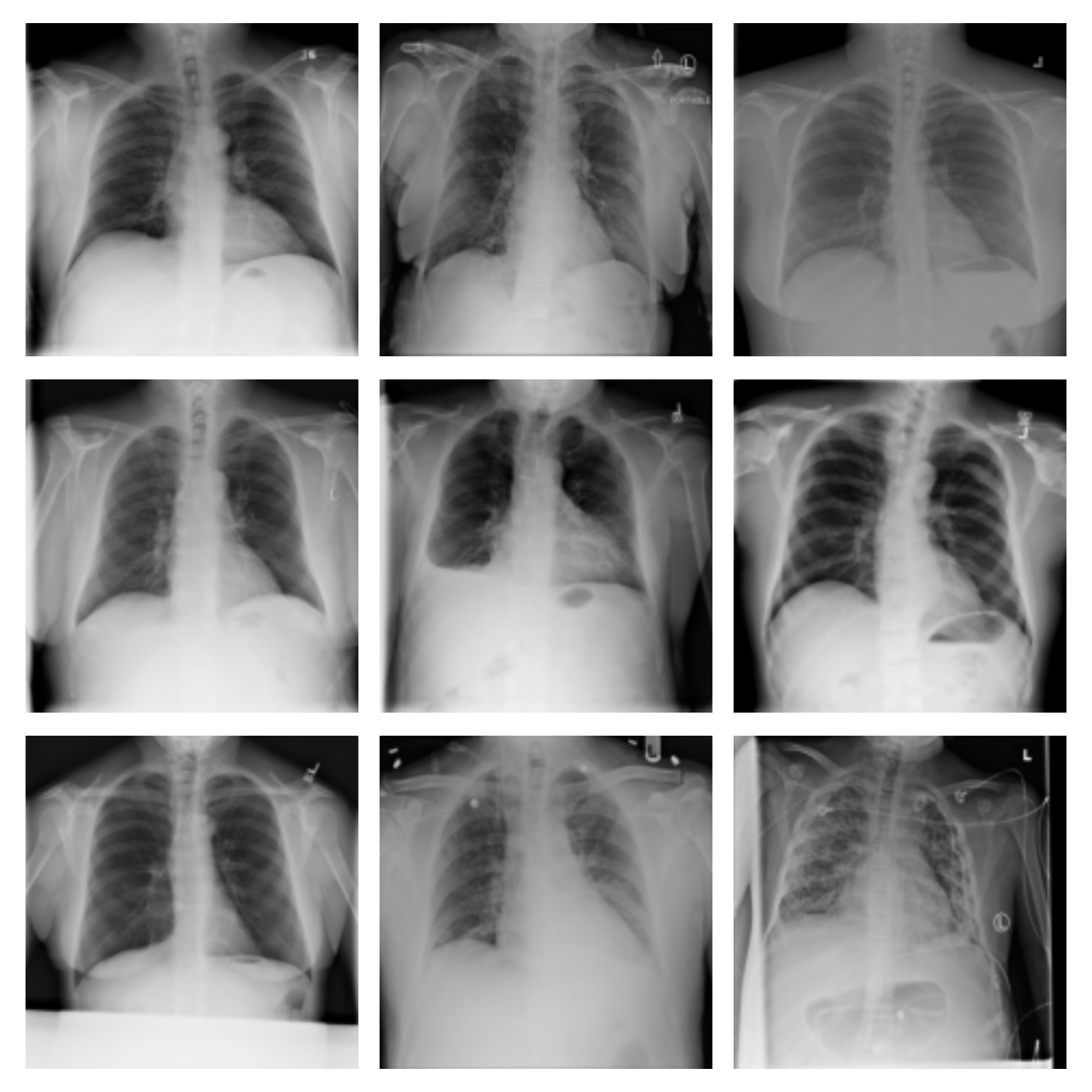}
         \caption{ChestX}
         \label{fig:image_examples_chest}
     \end{subfigure}
     
     \begin{subfigure}[h]{0.24\linewidth}
         \centering
         \includegraphics[width=\linewidth]{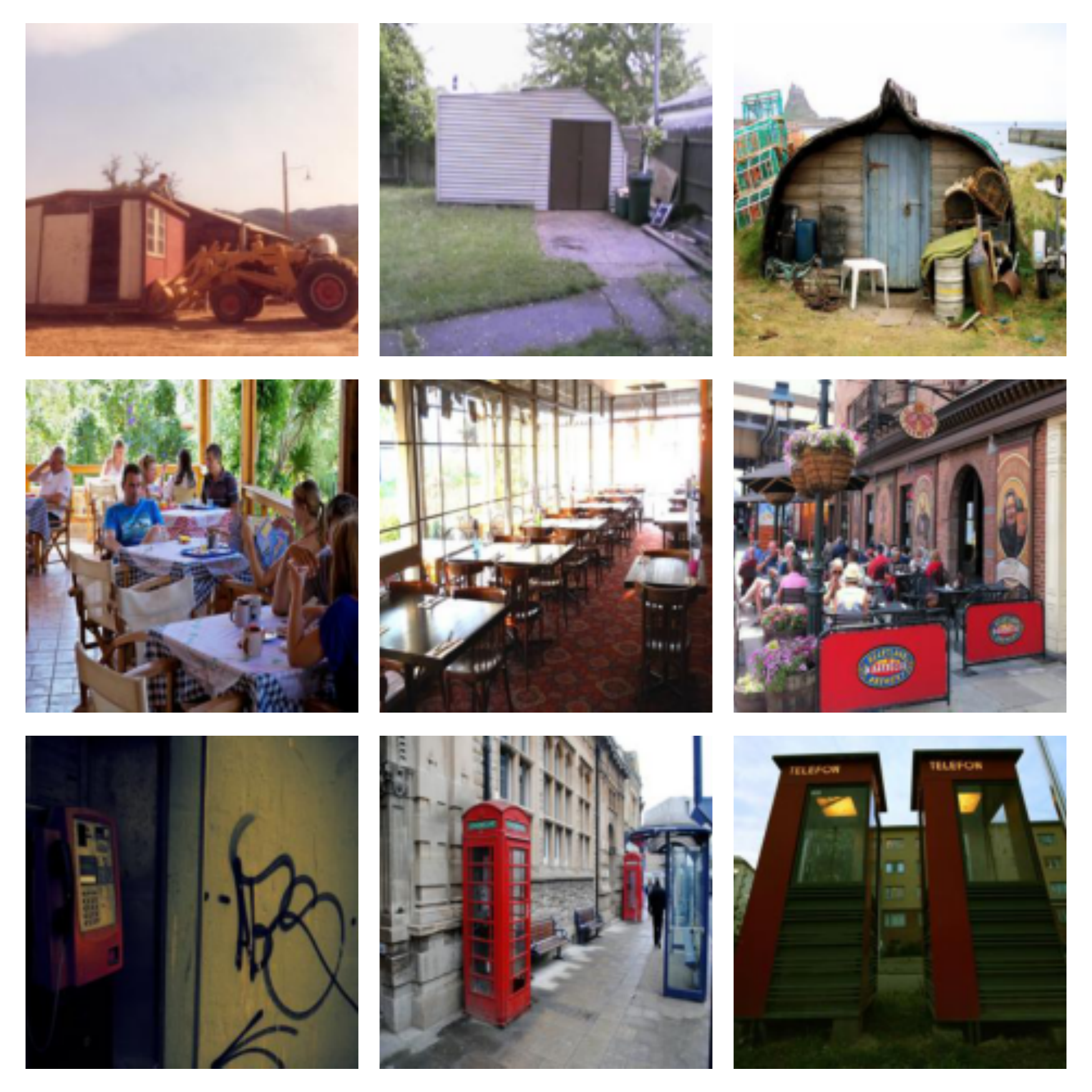}
         \caption{Places}
         \label{fig:image_examples_places}
     \end{subfigure}
     \hfill
     \begin{subfigure}[h]{0.24\linewidth}
         \centering
         \includegraphics[width=\linewidth]{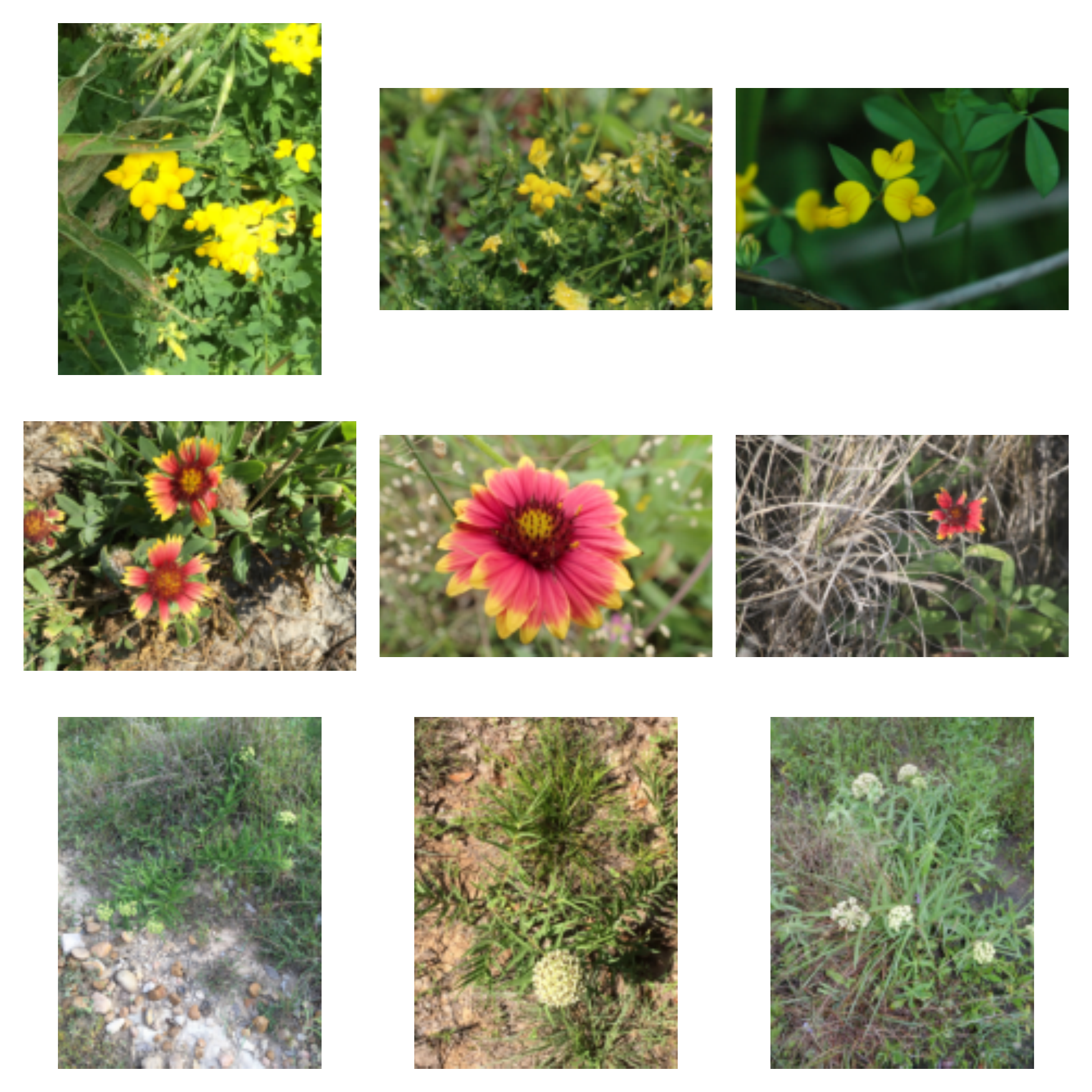}
         \caption{Plantae}
         \label{fig:image_examples_plantae}
     \end{subfigure}
     \hfill
     \begin{subfigure}[h]{0.24\linewidth}
         \centering
         \includegraphics[width=\linewidth]{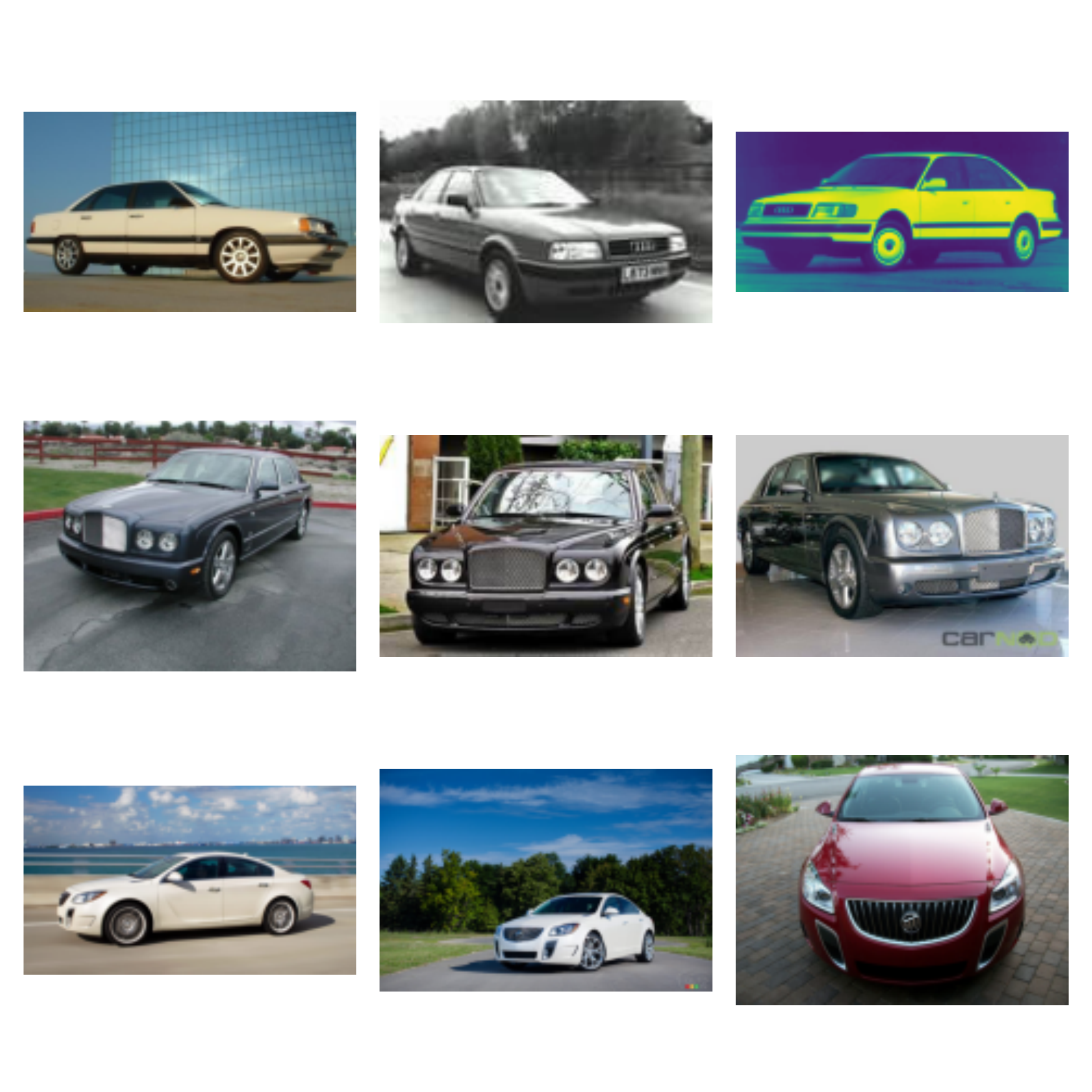}
         \caption{Cars}
         \label{fig:image_examples_cars}
     \end{subfigure}
     \hfill
     \begin{subfigure}[h]{0.24\linewidth}
         \centering
         \includegraphics[width=\linewidth]{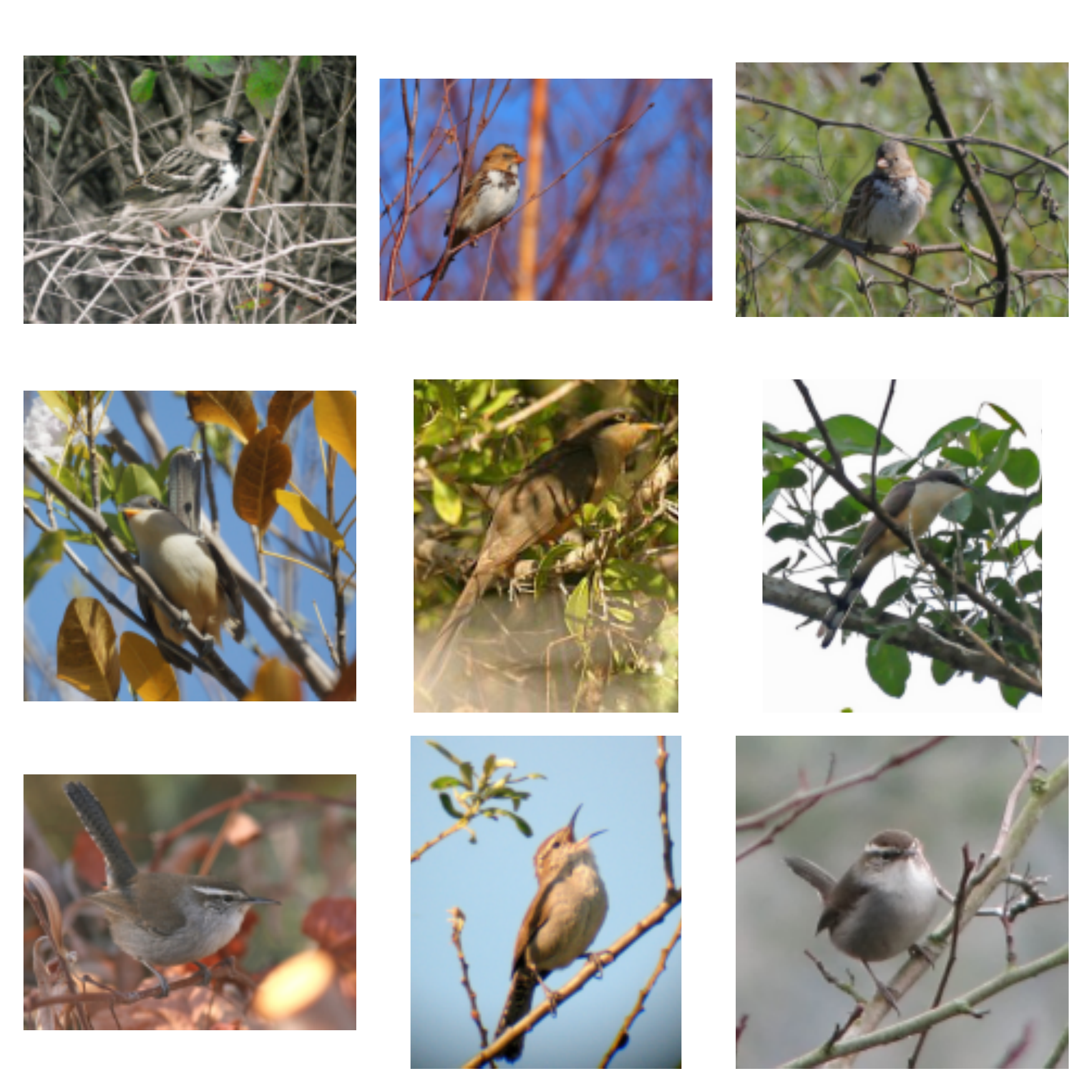}
         \caption{CUB}
         \label{fig:image_examples_cub}
     \end{subfigure}
    \vspace*{-3pt}
     \caption{Image examples from eight target datasets considered in our study. Each row displays three samples from a distinct class randomly sampled from each target dataset.}
     \label{fig:image_examples}
\end{figure}

We illustrate the qualitative characteristics of eight target domains for CD-FSL by showing nine randomly sampled examples from three distinct classes for each target dataset in Figure \ref{fig:image_examples}.
The previous work \citep{bscd_fsl} defined domain similarity to miniImageNet source with respect to perspective distortion, semantic content, and color depth.
This can be seen in Figure \ref{fig:image_examples}(a)-(d); CropDisease consists of natural images regarding agriculture, EuroSAT contains satellite images taken from a fixed perspective, ISIC and ChestX contain images with fixed perspective and unique semantics, with ChestX being grayscale.
On the other hand, non-BSCD-FSL datasets in Figure \ref{fig:image_examples}(e)-(h) depict familiar scenes or objects to human eyes. 

Using the EMD (Earth Mover's Distance) analysis, we discovered that domain similarity is mainly determined by the semantic content and color depth of the images.
For example, we find that ChestX and ISIC, which exhibit highly distinct semantic content, have high domain similarities with all three target datasets. CropDisease and EuroSAT have relatively higher domain similarity within the BSCD-FSL benchmark, and this can be attributed to the fact that the image subjects are from the natural image setting, albeit with fixed perspective and lack of either background or foreground. Places shows the highest domain similarity, which can be attributed to the existence of diverse subjects from the natural image domain, similar to the source datasets.

We can also observe that each target dataset has varying levels of difficulty. For example, for ChestX, it appears challenging to detect the small differences between the grayscale images and nearly impossible to distinguish any prominent features between classes to the untrained eye. On the other hand, classes from CropDisease are shown to have distinct features that are easily distinguishable.

\section{Implementation Details}\label{appx:implement_detail}

\subsection{Training Setup}\label{appx:training_setup}

We generally follow the training setup of previous works without validation dataset. Each model was pre-trained on a single RTX A5000, and each pre-training stage of 1000 epochs took 2.5--13.6 hours for ResNet18. SSL pre-training on CropDisease took 6.1, 8.1, 10.3, and 13.6 hours for SimCLR, MoCo, BYOL, and SimSiam, respectively. MSL pre-training took approximately $\times1.5$ more time compared to SSL, and training time scaled linearly with the size of the target dataset for SSL and MSL. 

We explain training details below:

\paragraph{SL Pre-Training} We use an SGD optimizer with an initial learning rate of 0.1, the momentum of 0.9, and weight decay coefficient of $10^{-4}$ is used. When using miniImageNet as source data, we train the ResNet10 model for 1,000 epochs with batch size 64, learning rate decayed by 1/10 at epoch \{400, 600, 800\}. When using tieredImageNet as source data, we train the ResNet18 model for 90 epochs with batch size 256. For ImageNet, the pre-trained ResNet18 model offered by an official PyTorch \citep{PyTorch} library is used.

\paragraph{SimCLR Pre-Training} We follow the setting in \citet{phoo2021selftraining} except batch size, learning rate, and augmentation method; SGD optimizer with momentum 0.9 and weight decay $10^{-4}$ is used. 1000 epochs are trained with batch size 32. Because SimCLR (including other SSL methods) uses a multi-viewed batch, it has an effective batch size of 64 by augmentations. The learning rate starts with 0.1 and is decayed by 1/10 times at epoch \{400, 600, 800\}. For the SimCLR loss, a two-layer projection head (i.e., Linear-ReLU-Linear) is added on top of the extractor. The projection head uses a hidden dimension of 512 and an output feature dimension of 128. The temperature value of NT-Xent loss (normalized temperature-scaled cross-entropy loss, \citet{chen2020simple}) is set to 1.0.

\paragraph{MoCo Pre-Training} We use the same optimizer, epochs, and batch size as SimCLR pre-training. The projector of both query and key is one fully-connected layer with a feature of dimension 128. Also, we used a moving average coefficient of 0.999 for the momentum encoder and the memory bank size of 1,024. Note that the original MoCo \citep{he2020momentum} uses a considerable size of a memory bank (i.e., 65,536) because a large number of negative samples is required in self-supervised learning of ImageNet data. However, in the case of our self-supervised learning on small-size target data, a large memory bank is neither needed nor recommended. Moreover, a large number of negative samples can make the contrastive task too hard to optimize for extremely fine-grained images, as we observed in ChestX FSL performances. This is the main reason why MoCo rarely surpasses SimCLR in our experiments. Also, note that the hyperparameters are mainly suited to the SimCLR model, but we did not further search or tune the hyperparameters.

\paragraph{BYOL Pre-Training} We use a different optimizer for BYOL; Adam \citep{kingma2014adam} optimizer with the initial learning rate of $3\times10^{-4}$. The online and target projector are both composed of two-layer MLP (i.e., Linear-BatchNorm1D-ReLU-Linear) with a hidden dimension of 4,096 and a projection dimension of 256. The moving average coefficient for the target network is 0.99. A predictor after the online projector is also a two-layer MLP with a hidden dimension of 4,096 and an output prediction dimension of 256.

\paragraph{SimSiam Pre-Training} SimSiam uses the same network structure as BYOL, except there is no auxiliary target encoder and target projector. Every other training setup is the same.

\paragraph{MSL Pre-Training} When training the MSL model, the batch size of source data is 64 and that of target data is 32, because target data are augmented twice to make positive pairs. Although we pre-trained for 1,000 epochs, one epoch corresponds to an entire sweep over the target data. The source batch is randomly sampled at every iteration, independently from the epoch. A conflicting setting is that we used an SGD optimizer for SL pre-training and an Adam optimizer for BYOL pre-training. Therefore, in MSL (BYOL) pre-training, there were two choices of an optimizer. We confirmed with some experiments that the SGD optimizer better works for MSL (BYOL).

\paragraph{Two-Stage Pre-Training} In Section \ref{sec:analysis3}, we extended the single-stage pre-training to the two-stage approaches. The initial model for the second stage is the SL model, which is exactly the same model as the above \textbf{SL Pre-Training}. The second stage of pre-training also follows the same procedure as the single-stage, both for SSL and MSL.

\paragraph{Fine-Tuning} We follow the setting in \citet{bscd_fsl}; SGD optimizer with learning rate 0.01, momentum 0.9, and weight decay 0.001 is used. Only the linear classifier is trained with a frozen pre-trained extractor, and 100 epochs are trained with batch size 4. Note that, for a fair comparison, we removed a projector or predictor that is additionally introduced in SSL pre-training. 

\subsection{Data Augmentations}
We provide below the PyTorch-style code for the base and strong augmentation. A short description for each transform with our set parameter is as follows:
\vspace*{-5pt}
\begin{itemize}[leftmargin=*]
    \item RandomResizedCrop: Randomly crop a portion of an image and then resize it to 224x224.
    \item RandomColorJitter: Randomly change the brightness, contrast, and saturation, with a probability of 1.0.
    \item RandomHorizontalFlip: Randomly flip an image on a vertical axis, with a probability of 0.5.
    \item RandomGrayscale: Randomly convert image into grayscale, with a probability of 0.1.
    \item RandomGaussianBlur: Randomly blur an image with Gaussian blur of kernel size (5,5), with a probability of 0.3.
\end{itemize}

\begin{python}
import torchvision.transforms as transforms

def parse_transform(transform, image_size=224):
    if transform == `RandomResizedCrop':
        return transforms.RandomResizedCrop(image_size)
        
    elif transform == `RandomColorJitter':
        return transforms.RandomApply(
               [transforms.ColorJitter(0.4,0.4,0.4,0.0)], p=1.0)
        
    elif transform == `RandomGrayscale':
        return transforms.RandomGrayscale(p=0.1)
        
    elif transform == `RandomGaussianBlur':
        return transforms.RandomApply(
               [transforms.GaussianBlur(kernel_size=(5,5))], p=0.3)
        
    elif transform == `Resize':
        return transforms.Resize([image_size, image_size])
        
    elif transform == `Normalize':
        return transforms.Normalize(mean=[0.485,0.456,0.406],
                                    std=[0.229,0.224,0.225])
    elif transform == `ToTensor':
        return transforms.ToTensor()

def get_composed_transform(augmentation: str, image_size=224):
    if augmentation == `base':
        transform_list = [`RandomResizedCrop', `RandomColorJitter',
                          `RandomHorizontalFlip', `ToTensor', `Normalize']
                          
    elif augmentation == `strong':
        transform_list = [`RandomResizedCrop', `RandomColorJitter',
                          `RandomGrayscale', `RandomGaussianBlur',
                          `RandomHorizontalFlip', `ToTensor', `Normalize']
                          
    elif augmentation == `none':
        transform_list = [`Resize', `ToTensor', `Normalize']


    transform_funcs = [parse_transform(x, image_size=image_size) 
                       for x in transform_list]
    transform = transforms.Compose(transform_funcs)
    return transform
\end{python}

\clearpage
\section{Domain Similarity}\label{appx:domain_similarity}

To estimate the domain similarity, we follow \citet{cui2018large} and \citet{li2020rethinking} by calculating EMD as the distance between two domains. EMD is informally defined as the minimum cost of moving one accumulation into another. EMD has advantages compared to other metric choices, such as Kullback-Leibler divergence (KLD), Jensen-Shannon divergence (JSD), or maximum mean discrepancy (MMD). We can compute EMD directly from the samples, whereas KLD and JSD require explicit expressions for the densities \citep{cover1999elements}. MMD can also be considered but is less powerful in high dimensions and highly dependent on the kernel and its hyperparameters \citep{ramdas2015decreasing}.

The BSCD-FSL benchmark contains four datasets with varying levels of domain similarity: CropDisease, EuroSAT, ISIC, and ChestX. These datasets are known to be distant from the source dataset, miniImageNet.
\citet{bscd_fsl} provided the order of domain similarity for the BSCD-FSL benchmark based on three qualitative factors: perspective distortion, semantic contents, and color depth. However, our quantitative metric in Eq. \eqref{eq:domain_sim} shows a somewhat different order of domain similarity between the four datasets in BSCD-FSL.
The known similarity order for BSCD-FSL was {\slshape ``CropDisease $>$ EuroSAT $>$ ISIC $>$ ChestX"} under the assumption that a dataset has domain similar to ImageNet if it has perspective distortion\,(\emph{i.e.}, CropDisease), is natural\,(\emph{i.e.}, CropDisease, EuroSAT),  and has RGB color depth\,(\emph{i.e.}, CropDisease, EuroSAT, ISIC).
%

In contrast, we observe a different order of {\slshape ``EuroSAT $>$ CropDisease $\approx$ ISIC $>$ ChestX"} in Table \ref{tab:similarity_summary} and Table \ref{tab:similarity_diff_model} when using our quantitative metric.
It turns out that semantic content and color depth are significant factors in deciding domain similarity; thus, ChestX is always the most dissimilar to the source domain. On the other hand, perspective distortion is less important than semantic content and color depth for determining domain similarity, considering that the order of CropDisease and EuroSAT is reversed. Therefore, we observe that EuroSAT is the closest dataset to the source domain. The change in the domain similarity rank of EuroSAT, when tieredImageNet is used as the source dataset, is discussed below.

\vspace{-10pt}
\begin{table}[!h]
\caption{Domain Similarity. Earth Mover's Distance (EMD) and similarity (calculated by $\exp(-\alpha \times \text{EMD})$) are reported. The feature extractor used is ResNet101 provided by PyTorch \citep{PyTorch}. Rank 1 dataset indicates that the source and target datasets are the most similar.}\label{tab:similarity_summary}
\vskip 0.10in
\centering
\footnotesize\addtolength{\tabcolsep}{-2pt}
\resizebox{\linewidth}{!}{
\begin{tabular}{cccccccccc}
    \toprule
    & & Places & CUB & Cars & Plantae & EuroSAT & CropDisease & ISIC & ChestX \\
    \midrule
    \multirow{3}{*}{EMD} & IN & 18.14 & 18.91 & 20.13 & 20.31 & 21.67 & 21.88 & 22.25 & 22.28 \\
    & tieredIN & 17.26 & 19.90 & 20.23 & 20.63 & 19.20 & 22.07 & 22.94 & 23.19 \\
    & miniIN & 17.49 & 19.38 & 20.34 & 20.29 & 21.10 & 21.66 & 22.20 & 22.33 \\
    \midrule
    \multirow{3}{*}{Sim} & IN & 0.834 (1) & 0.828 (2) & 0.818 (3) & 0.816 (4) & 0.805 (5) & 0.803 (6) & 0.801 (7) & 0.800 (8) \\
    & tieredIN & 0.841 (1) & 0.820 (3) & 0.817 (4) & 0.814 (5) & 0.825 (2) & 0.802 (6) & 0.795 (7) & 0.793 (8) \\
    & miniIN & 0.840 (1) & 0.824 (2) & 0.816 (4) & 0.816 (3) & 0.810 (5) & 0.805 (6) & 0.801 (7) & 0.800 (8) \\
    \bottomrule
\end{tabular}}
\end{table}

Domain similarity can differ according to the feature extractor used because it is based on representations. In the main paper, we use ResNet101 to extract representations because we use ResNet-like models for our few-shot classification tasks. However, other architectures (\emph{e.g.}, DenseNet and ViT) can also be used. Table \ref{tab:similarity_diff_model} shows the domain similarity to the ImageNet source dataset, measured using different feature extractors. To do this, we used the open-source library \texttt{timm}.\footnote{\url{http://github.com/rwightman/pytorch-image-models/}} For the details of each architecture, please refer to the original papers: ResNet\,\citep{he2016deep}, MobileNetV2\,\citep{sandler2018mobilenetv2}, EfficientNet\,\citep{tan2019efficientnet, tan2021efficientnetv2}, DenseNet\,\citep{huang2017densely}, and ViT\,\citep{dosovitskiy2021an}.

Although the exact ordering of domain similarity can change, it does not undermine the consistency of our analysis. In particular, we explain that:
\begin{itemize}[leftmargin=*]
    \vspace{-5pt}
    \item \textbf{Important point for Obs.~\ref{obs:obs5_2}.} From Obs.~\ref{obs:obs5_2}, we argue that larger domain similarity does not always guarantee the superiority of SL. To demonstrate this, we compare the performance of SL and SSL on Places and CUB. Namely, SSL is better than SL on Places, while SL is better than SSL on CUB, despite Places being closer to the source dataset. 
    As we can see in Table \ref{tab:similarity_diff_model}, this observation does not change, even when using other feature extractors.
    In fact, when EfficientNet-b4 is used, the domain similarity ranking of Cars and EuroSAT are changed, exacerbating the inconsistency.
    Furthermore, when using ViT models, the similarity ranking of CUB moves down to fifth place, despite it showing the largest margin between SL and SSL performance, in favor of SL on source data (refer to Figure \ref{fig:sim_miniIN}).
    \item \textbf{Important point for Obs.~\ref{obs:obs5_3}.} From Obs.~\ref{obs:obs5_3}, we argue that for both groups, the performance gain of SSL over SL becomes greater as few-shot difficulty decreases. We first divide the eight target datasets into two groups according to the domain similarity.
    Within each similarity group, the performance gain of SSL over SL is highly related to the few-shot difficulty.
    As shown in Table \ref{tab:similarity_diff_model}, EuroSAT can be categorized into the large similarity group when using EfficientNet-b4, ViT-B/16, or ViT-L/16.
    However, because EuroSAT has lower few-shot difficulty than Places (refer to Figure \ref{fig:similarity_difficulty}), the superiority of SSL over SL on EuroSAT is consistently explained with few-shot difficulty, even when inside the large similarity group.
    In addition, the domain similarity ranking of CropDisease and ISIC based on ResNet101 is different from that based on other extractors. However, both datasets remain inside the small similarity group, hence does not affect Obs.~\ref{obs:obs5_3}. 
\end{itemize}

\vspace{-10pt}
\begin{table}[h]
\caption{Domain Similarity to ImageNet measured across different architectures. Similarities (calculated by $\exp(-\alpha \times \text{EMD})$) are reported. The feature extractors used are ResNet101, provided by PyTorch\,\citep{PyTorch}, and others, by \texttt{timm} open-source library. Rank 1 dataset indicates that the source and target datasets are the most similar.}\label{tab:similarity_diff_model}
\vskip 0.10in
\centering
\footnotesize\addtolength{\tabcolsep}{-2pt}
\resizebox{\linewidth}{!}{
\begin{tabular}{ccccccccc}
    \toprule
    Extractor & Places & CUB & Cars & Plantae & EuroSAT & CropDisease & ISIC & ChestX \\
    \midrule
    ResNet101 (main) & 0.834 (1) & 0.828 (2) & 0.818 (3) & 0.816 (4) & 0.805 (5) & 0.803 (6) & 0.801 (7) & 0.800 (8) \\
    \midrule
    ResNet18        & 0.866 (1) & 0.847 (2) & 0.845 (3) & 0.843 (4) & 0.829 (5) & 0.823 (7) & 0.828 (6) & 0.815 (8) \\
    MobileNetV2     & 0.919 (1) & 0.918 (2) & 0.908 (4) & 0.910 (3) & 0.903 (5) & 0.889 (7) & 0.893 (6) & 0.884 (8) \\
    EfficientNet-b0 & 0.913 (1) & 0.910 (2) & 0.903 (3) & 0.901 (4) & 0.901 (5) & 0.873 (7) & 0.877 (6) & 0.873 (8) \\
    EfficientNet-b4 & 0.969 (1) & 0.967 (2) & 0.963 (5) & 0.965 (4) & 0.966 (3) & 0.956 (7) & 0.961 (6) & 0.953 (8) \\
    EfficientNetV2  & 0.930 (1) & 0.927 (2) & 0.924 (3) & 0.924 (4) & 0.924 (5) & 0.906 (7) & 0.914 (6) & 0.896 (8) \\
    DenseNet121     & 0.818 (1) & 0.802 (2) & 0.791 (3) & 0.787 (4) & 0.785 (5) & 0.761 (7) & 0.767 (6) & 0.752 (8) \\
    ViT-B/16        & 0.508 (1) & 0.415 (5) & 0.438 (4) & 0.442 (3) & 0.444 (2) & 0.386 (7) & 0.409 (6) & 0.390 (8) \\
    ViT-L/16        & 0.478 (1) & 0.395 (5) & 0.396 (4) & 0.426 (2) & 0.422 (3) & 0.372 (7) & 0.391 (6) & 0.367 (8) \\
    \bottomrule
\end{tabular}}
\end{table}

\clearpage
\section{Few-Shot Difficulty}\label{appx:difficulty}
We quantify few-shot difficulty using our empirical upper bound on each dataset following in Eq. \eqref{eq:domain_dif}. The few-shot difficulty depends on a backbone network and $k$. Table \ref{tab:difficulty_summary} describes few-shot difficulty according to the combination of our backbone (ResNet10 and ResNet18) and $k$.
It is observed that the order of few-shot difficulty remains the same, except between ISIC and CUB when ResNet10 is used as a backbone and $k$=1.
We point out that Obs.~\ref{obs:obs5_3} still stands under this variation.

\begin{table}[h]
\caption{Few-shot difficulty (ranking). 5-way $k$-shot performances are reported. To quantify the data difficulty, we designed the upper performance case, where we use SL pre-training with 20\% of target data as labeled data. The $k$-shot difficulty  is calculated by $\text{Diff@}k = \exp(-\beta\times\text{Perf@}k)$. Rank 1 dataset is the most difficult one.}\label{tab:difficulty_summary}
\vskip 0.10in
\centering
\footnotesize\addtolength{\tabcolsep}{-2pt}
\resizebox{\linewidth}{!}{
\begin{tabular}{cccccccccc}
    \toprule
    Backbone & $k$ & CropDisease & EuroSAT & ISIC & ChestX & Places & Plantae & Cars & CUB \\
    \midrule
    \multirow{3}{*}{RN18} & 1 & 96.92{\scriptsize$\pm$.32} & 90.51{\scriptsize$\pm$.55} & 42.83{\scriptsize$\pm$.80} & 31.00{\scriptsize$\pm$.60}  & 63.97{\scriptsize$\pm$.87} & 52.83{\scriptsize$\pm$.89} & 48.71{\scriptsize$\pm$.82} & 42.96{\scriptsize$\pm$.76} \\
     & 5 & 99.51{\scriptsize$\pm$.10} & 96.74{\scriptsize$\pm$.21} & 55.55{\scriptsize$\pm$.67} & 39.19{\scriptsize$\pm$.58} & 81.56{\scriptsize$\pm$.57} & 74.24{\scriptsize$\pm$.71} & 72.83{\scriptsize$\pm$.67} & 64.03{\scriptsize$\pm$.77} \\
     & 20 & 99.69{\scriptsize$\pm$.07} & 97.45{\scriptsize$\pm$.17} & 61.32{\scriptsize$\pm$.62} & 42.11{\scriptsize$\pm$.56} & 86.10{\scriptsize$\pm$.47} & 82.17{\scriptsize$\pm$.64} & 82.08{\scriptsize$\pm$.53} & 74.14{\scriptsize$\pm$.66} \\
     \midrule
     \multicolumn{2}{c}{Diff@1} & 0.379 (8) & 0.405 (7) & 0.652 (2) & 0.733 (1) & 0.527 (6) & 0.590 (5) & 0.614 (4) & 0.651 (3)    \\
     \multicolumn{2}{c}{Diff@5} & 0.370 (8) & 0.380 (7) & 0.574 (2) & 0.676 (1) & 0.442 (6) & 0.476 (5) & 0.483 (4) & 0.527 (3)    \\
     \multicolumn{2}{c}{Diff@20} & 0.369 (8) & 0.377 (7) & 0.542 (2) & 0.656 (1) & 0.423 (6) & 0.440 (5) & 0.440 (4) & 0.476 (3) \\
     \midrule
    \multirow{3}{*}{RN10} & 1& 92.44{\scriptsize$\pm$.55} & 83.34{\scriptsize$\pm$.68} & 42.89{\scriptsize$\pm$.77} & 28.89{\scriptsize$\pm$.55}  & 57.25{\scriptsize$\pm$.82} & 49.08{\scriptsize$\pm$.83} & 43.32{\scriptsize$\pm$.72} & 40.72{\scriptsize$\pm$.73} \\
     & 5 & 99.00{\scriptsize$\pm$.15} & 95.37{\scriptsize$\pm$.26} & 56.94{\scriptsize$\pm$.65} & 36.59{\scriptsize$\pm$.56}  & 77.39{\scriptsize$\pm$.63} & 70.56{\scriptsize$\pm$.76} & 66.40{\scriptsize$\pm$.68} & 60.29{\scriptsize$\pm$.78} \\
     & 20 & 99.54{\scriptsize$\pm$.07} & 97.28{\scriptsize$\pm$.18} & 63.93{\scriptsize$\pm$.58} & 42.03{\scriptsize$\pm$.55} & 84.62{\scriptsize$\pm$.49} & 80.50{\scriptsize$\pm$.65} & 78.56{\scriptsize$\pm$.57} & 71.71{\scriptsize$\pm$.67} \\
     \midrule
     \multicolumn{2}{c}{Diff@1} & 0.397 (8) & 0.435 (7) & 0.651 (3) & 0.749 (1) & 0.564 (6) & 0.612 (5) & 0.648 (4) & 0.666 (2)    \\
     \multicolumn{2}{c}{Diff@5} & 0.372 (8) & 0.385 (7) & 0.566 (2) & 0.694 (1) & 0.461 (6) & 0.494 (5) & 0.515 (4) & 0.547 (3)    \\
     \multicolumn{2}{c}{Diff@20} & 0.370 (8) & 0.378 (7) & 0.528 (2) & 0.657 (1) & 0.429 (6) & 0.447 (5) & 0.456 (4) & 0.488 (3)    \\
    \bottomrule
\end{tabular}}
\end{table}

\paragraph{Few-shot Difficulty on Different Splits.}\label{appx:difficulty_robust}
We used the same 20\% split of $\mathcal{D}_U$ as used in SSL pre-training for measuring the few-shot difficulty, but with label information. This is because the dataset partition for calculating few-shot difficulty (which is the rest 80\%) should be matched with that for evaluating SL/SSL methods, for consistent analysis. However, few-shot difficulty can differ according to the dataset splits. To remedy this concern, we provide the few-shot performance using different splits when 20\% of target dataset are used for pre-training with label information. Table \ref{tab:difficulty_robust} shows that the ranks of few-shot difficulty between datasets do not change even if dataset splits are changed. 

\begin{table}[!h]
\caption{5-way $k$-shot performances are reported. These performances are converted to few-shot difficulty. To quantify the data difficulty, we designed the upper performance case, where we use SL pre-training with 20\% of target data as labeled data. To show the robustness of few-shot difficulty, accuracy is estimated three times using different splits for 20\% of target data. ResNet18 is used as a backbone.}\label{tab:difficulty_robust}
\vskip 0.10in
\centering
\footnotesize\addtolength{\tabcolsep}{-2pt}
\resizebox{\linewidth}{!}{
\begin{tabular}{cccccccccc}
    \toprule
    Split seed & $k$ & CropDisease & EuroSAT & ISIC & ChestX & Places & Plantae & Cars & CUB \\
    \midrule
    \multirow{3}{*}{1} & 1 & 96.92{\scriptsize$\pm$.32} & 90.51{\scriptsize$\pm$.55} & 42.83{\scriptsize$\pm$.80} & 31.00{\scriptsize$\pm$.60}  & 63.97{\scriptsize$\pm$.87} & 52.83{\scriptsize$\pm$.89} & 48.71{\scriptsize$\pm$.82} & 42.96{\scriptsize$\pm$.76} \\
     & 5 & 99.51{\scriptsize$\pm$.10} & 96.74{\scriptsize$\pm$.21} & 55.55{\scriptsize$\pm$.67} & 39.19{\scriptsize$\pm$.58} & 81.56{\scriptsize$\pm$.57} & 74.24{\scriptsize$\pm$.71} & 72.83{\scriptsize$\pm$.67} & 64.03{\scriptsize$\pm$.77} \\
     & 20 & 99.69{\scriptsize$\pm$.07} & 97.45{\scriptsize$\pm$.17} & 61.32{\scriptsize$\pm$.62} & 42.11{\scriptsize$\pm$.56} & 86.10{\scriptsize$\pm$.47} & 82.17{\scriptsize$\pm$.64} & 82.08{\scriptsize$\pm$.53} & 74.14{\scriptsize$\pm$.66} \\
     \midrule
    \multirow{3}{*}{2} & 1 & 96.52{\scriptsize$\pm$.36} & 90.84{\scriptsize$\pm$.50} & 43.05{\scriptsize$\pm$.74} & 30.03{\scriptsize$\pm$.62}  & 63.61{\scriptsize$\pm$.90} & 54.94{\scriptsize$\pm$.88} & 48.98{\scriptsize$\pm$.85} & 43.84{\scriptsize$\pm$.80} \\
     & 5 & 99.51{\scriptsize$\pm$.09} & 96.93{\scriptsize$\pm$.20} & 55.92{\scriptsize$\pm$.65} & 39.64{\scriptsize$\pm$.55}  & 81.65{\scriptsize$\pm$.56} & 75.73{\scriptsize$\pm$.72} & 72.95{\scriptsize$\pm$.63} & 63.76{\scriptsize$\pm$.78} \\
     & 20 & 99.78{\scriptsize$\pm$.05} & 97.67{\scriptsize$\pm$.16} & 63.34{\scriptsize$\pm$.56} & 45.96{\scriptsize$\pm$.54}  & 86.18{\scriptsize$\pm$.45} & 82.74{\scriptsize$\pm$.59} & 81.41{\scriptsize$\pm$.49} & 75.41{\scriptsize$\pm$.67} \\
     \midrule
    \multirow{3}{*}{3} & 1 & 96.54{\scriptsize$\pm$.34} & 89.23{\scriptsize$\pm$.56} & 42.94{\scriptsize$\pm$.78} & 29.36{\scriptsize$\pm$.60}  & 65.25{\scriptsize$\pm$.84} & 54.20{\scriptsize$\pm$.91} & 48.39{\scriptsize$\pm$.79} & 43.95{\scriptsize$\pm$.75} \\
     & 5 & 99.54{\scriptsize$\pm$.08} & 96.73{\scriptsize$\pm$.19} & 56.78{\scriptsize$\pm$.68} & 38.23{\scriptsize$\pm$.57}  & 82.44{\scriptsize$\pm$.54} & 74.51{\scriptsize$\pm$.79} & 71.71{\scriptsize$\pm$.70} & 64.39{\scriptsize$\pm$.75} \\
     & 20 & 99.81{\scriptsize$\pm$.05} & 97.60{\scriptsize$\pm$.16} & 63.16{\scriptsize$\pm$.62} & 44.60{\scriptsize$\pm$.55}  & 86.64{\scriptsize$\pm$.45} & 82.94{\scriptsize$\pm$.61} & 82.04{\scriptsize$\pm$.50} & 75.79{\scriptsize$\pm$.64} \\
    \bottomrule
\end{tabular}}
\end{table}
\clearpage
\section{Domain Similarity and Few-Shot Difficulty Visualizations}\label{appx:similarity_difficulty_variants}

\begin{figure}[h]
    \begin{subfigure}[h]{0.32\linewidth}
         \centering
         \includegraphics[width=\linewidth]{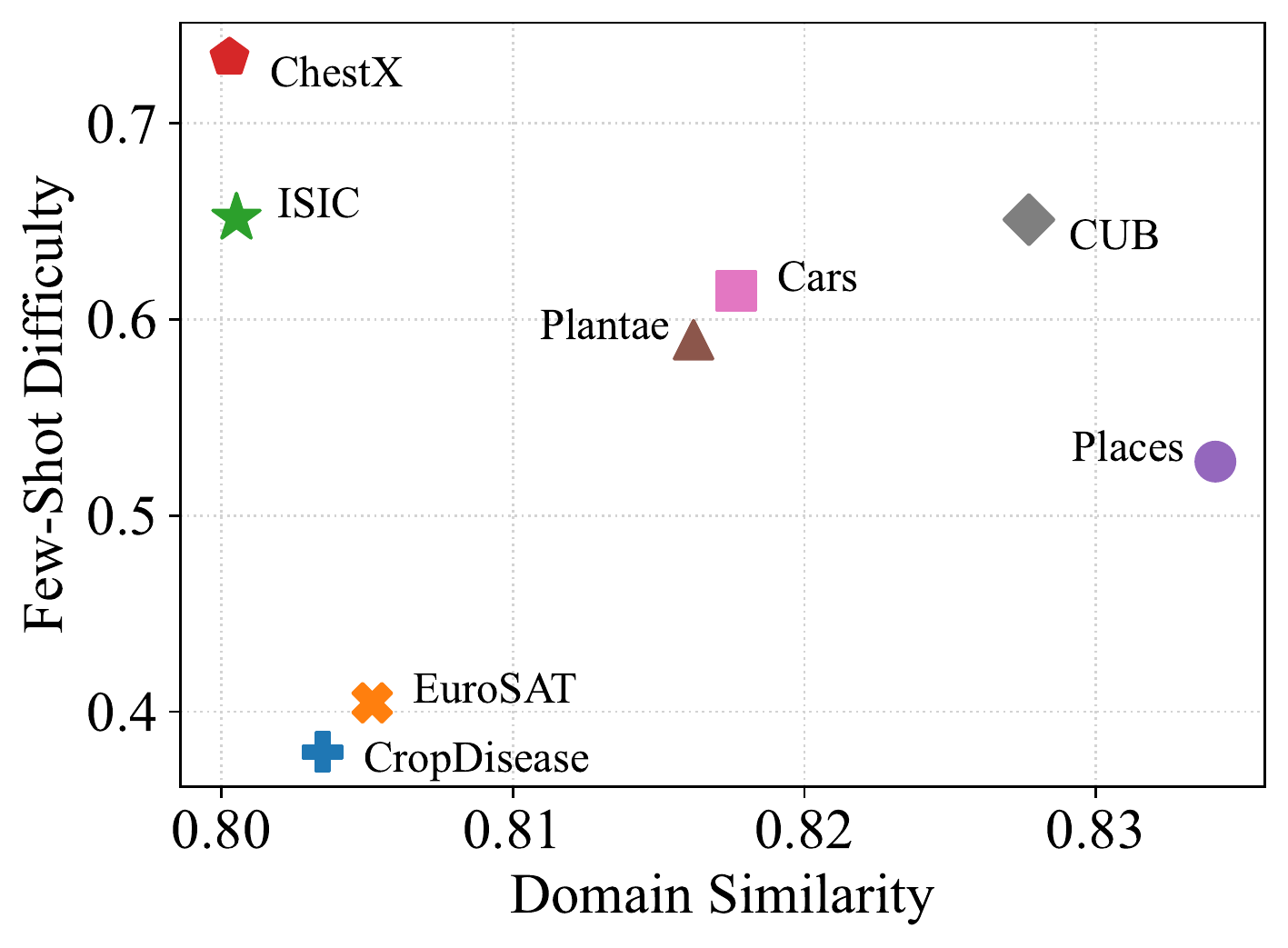}
         \caption{ImageNet (1-shot)}
         \label{fig:similarity_difficulty_imagenet_1shot}
     \end{subfigure}
     \hfill
     \begin{subfigure}[h]{0.32\linewidth}
         \centering
         \includegraphics[width=\linewidth]{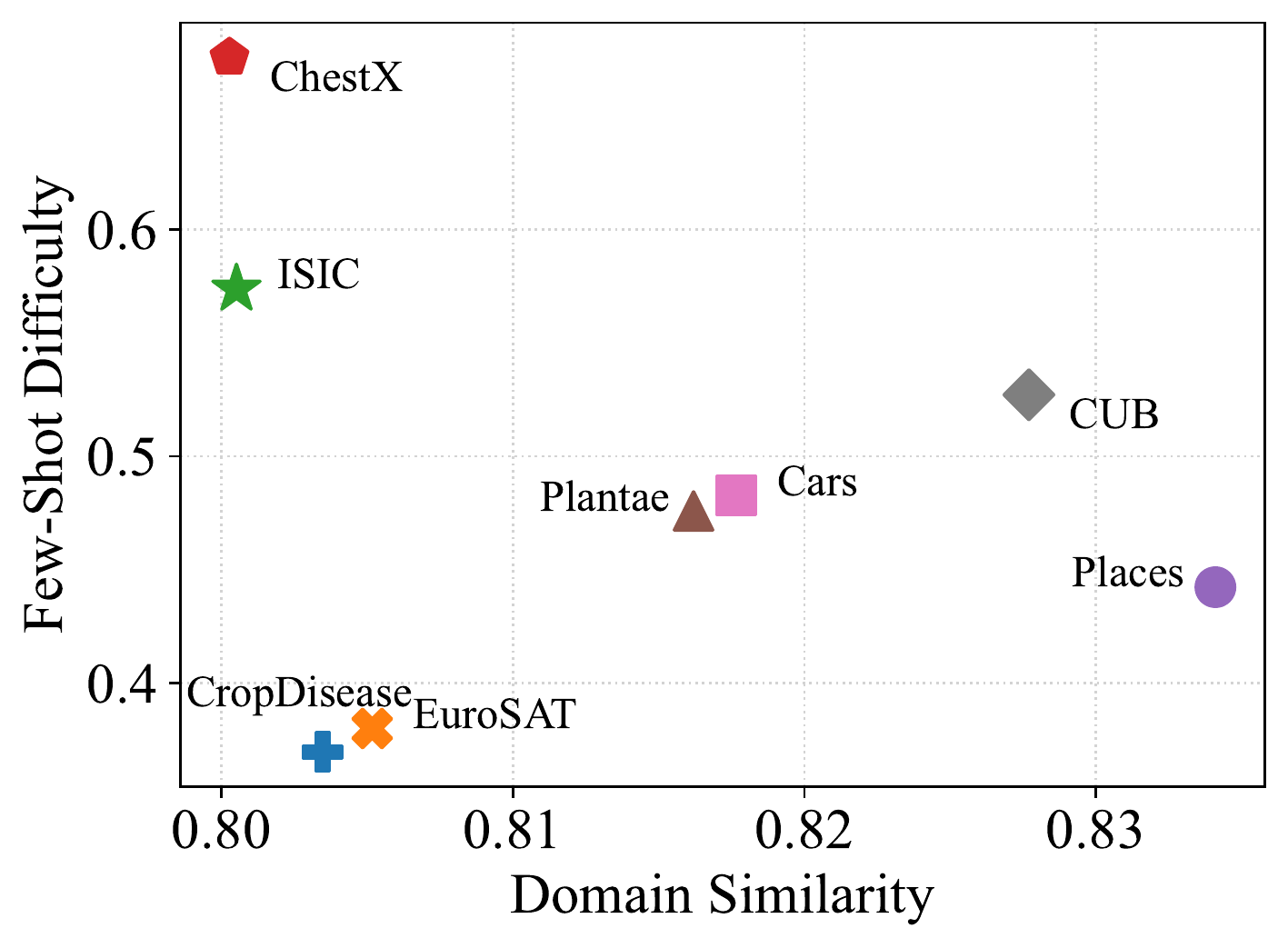}
         \caption{ImageNet (5-shot)}
         \label{fig:similarity_difficulty_imagenet_5shot}
     \end{subfigure}
     \hfill
     \begin{subfigure}[h]{0.32\linewidth}
         \centering
         \includegraphics[width=\linewidth]{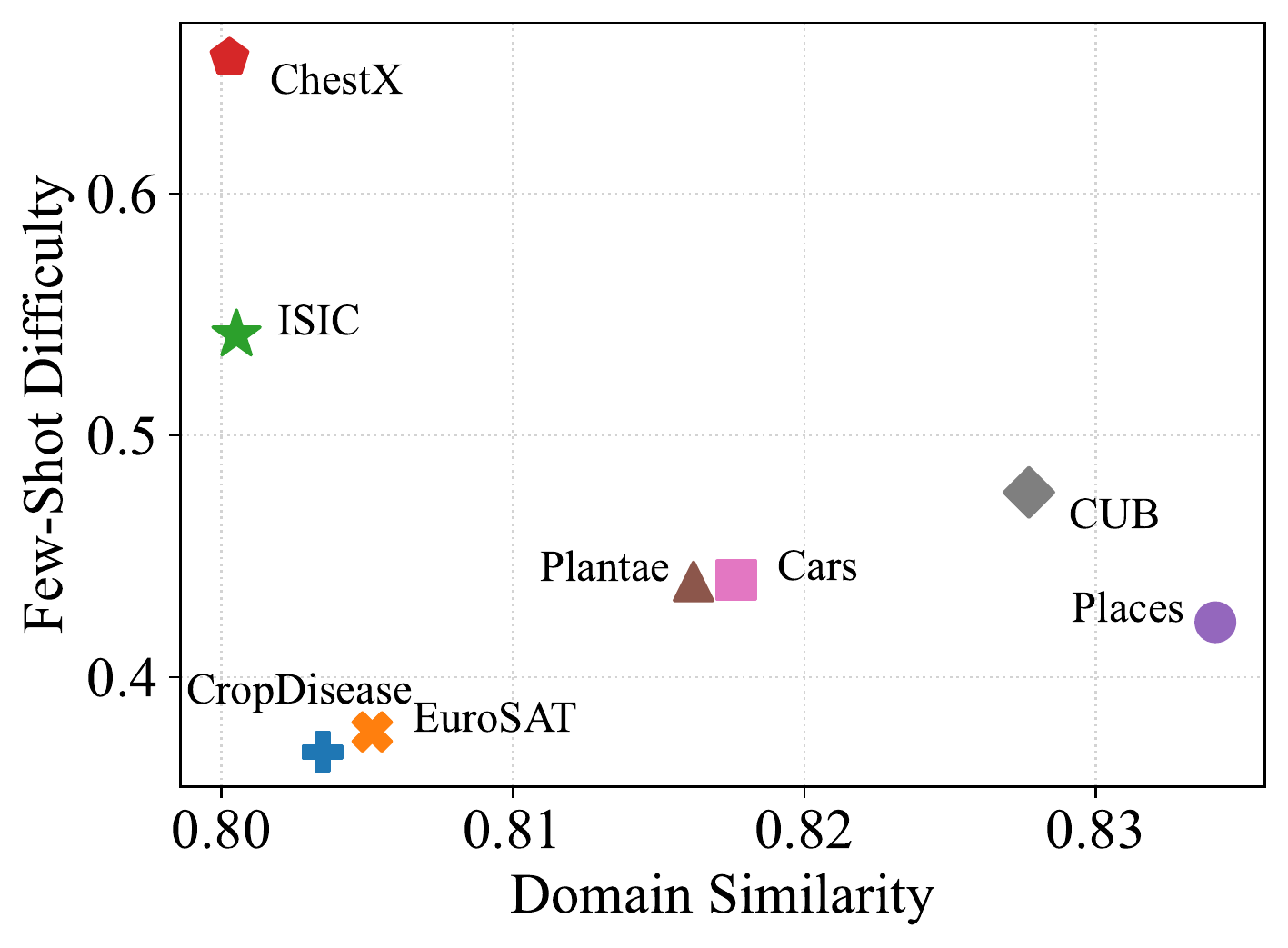}
         \caption{ImageNet (20-shot)}
         \label{fig:similarity_difficulty_imagenet_20shot}
     \end{subfigure}
     
     \begin{subfigure}[h]{0.32\linewidth}
         \centering
         \includegraphics[width=\linewidth]{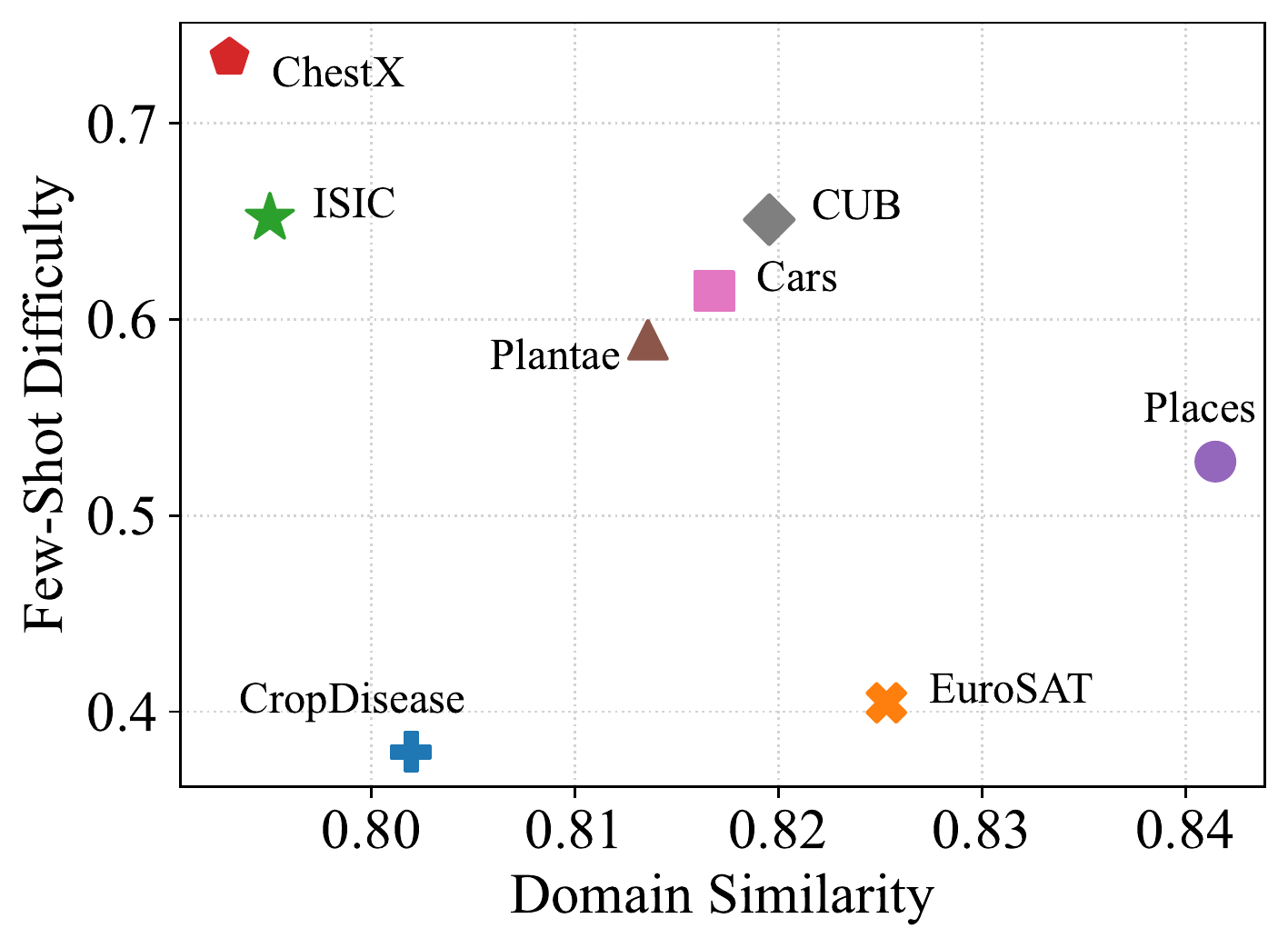}
         \caption{tieredImageNet (1-shot)}
         \label{fig:similarity_difficulty_tiered_1shot}
     \end{subfigure}
     \hfill
     \begin{subfigure}[h]{0.32\linewidth}
         \centering
         \includegraphics[width=\linewidth]{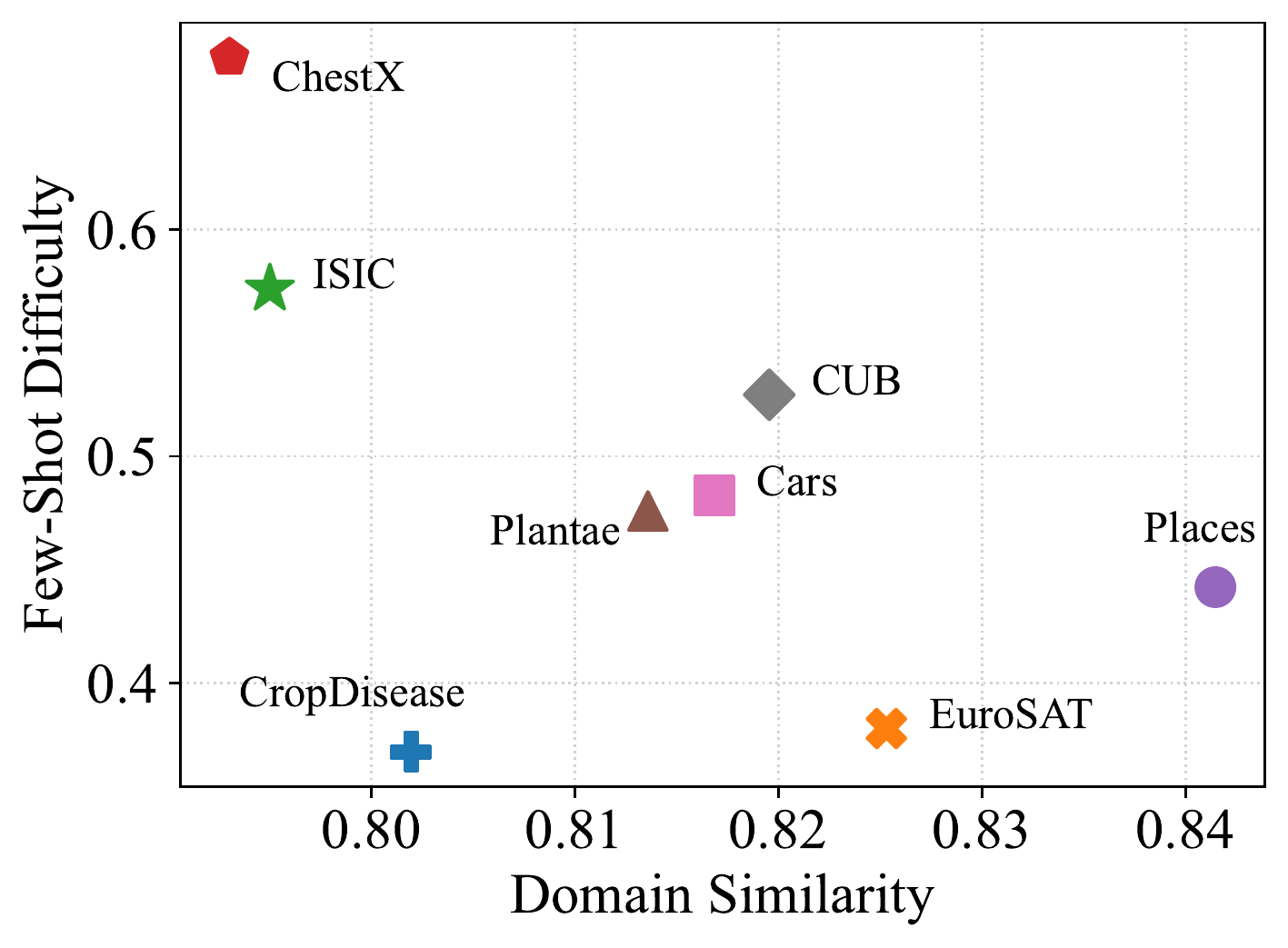}
         \caption{tieredImageNet (5-shot)}
         \label{fig:similarity_difficulty_tiered_5shot}
     \end{subfigure}
     \hfill
     \begin{subfigure}[h]{0.32\linewidth}
         \centering
         \includegraphics[width=\linewidth]{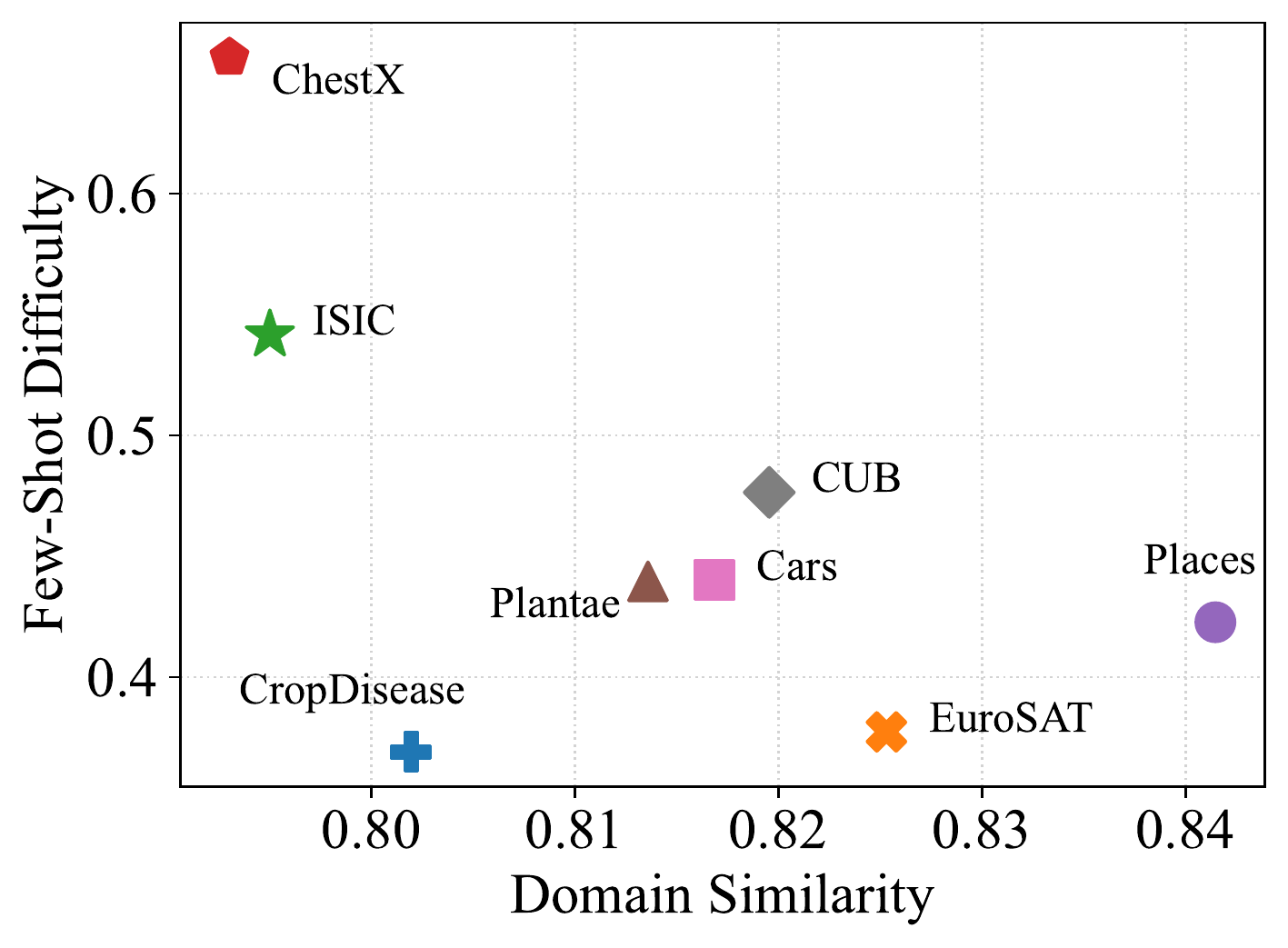}
         \caption{tieredImageNet (20-shot)}
         \label{fig:similarity_difficulty_tiered_20shot}
     \end{subfigure}
     
     \begin{subfigure}[h]{0.32\linewidth}
         \centering
         \includegraphics[width=\linewidth]{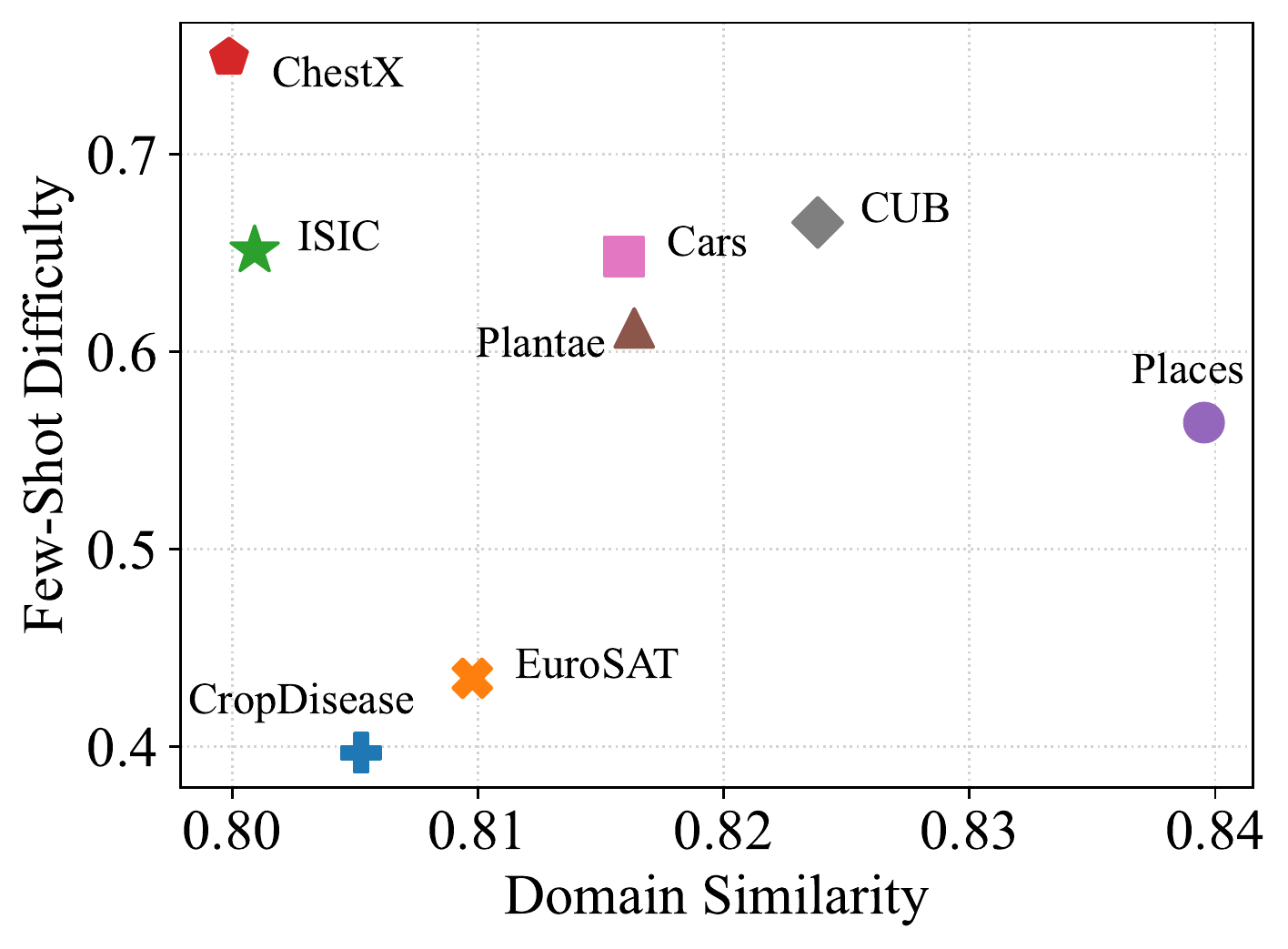}
         \caption{miniImageNet (1-shot)}
         \label{fig:similarity_difficulty_mini_1shot}
     \end{subfigure}
     \hfill
     \begin{subfigure}[h]{0.32\linewidth}
         \centering
         \includegraphics[width=\linewidth]{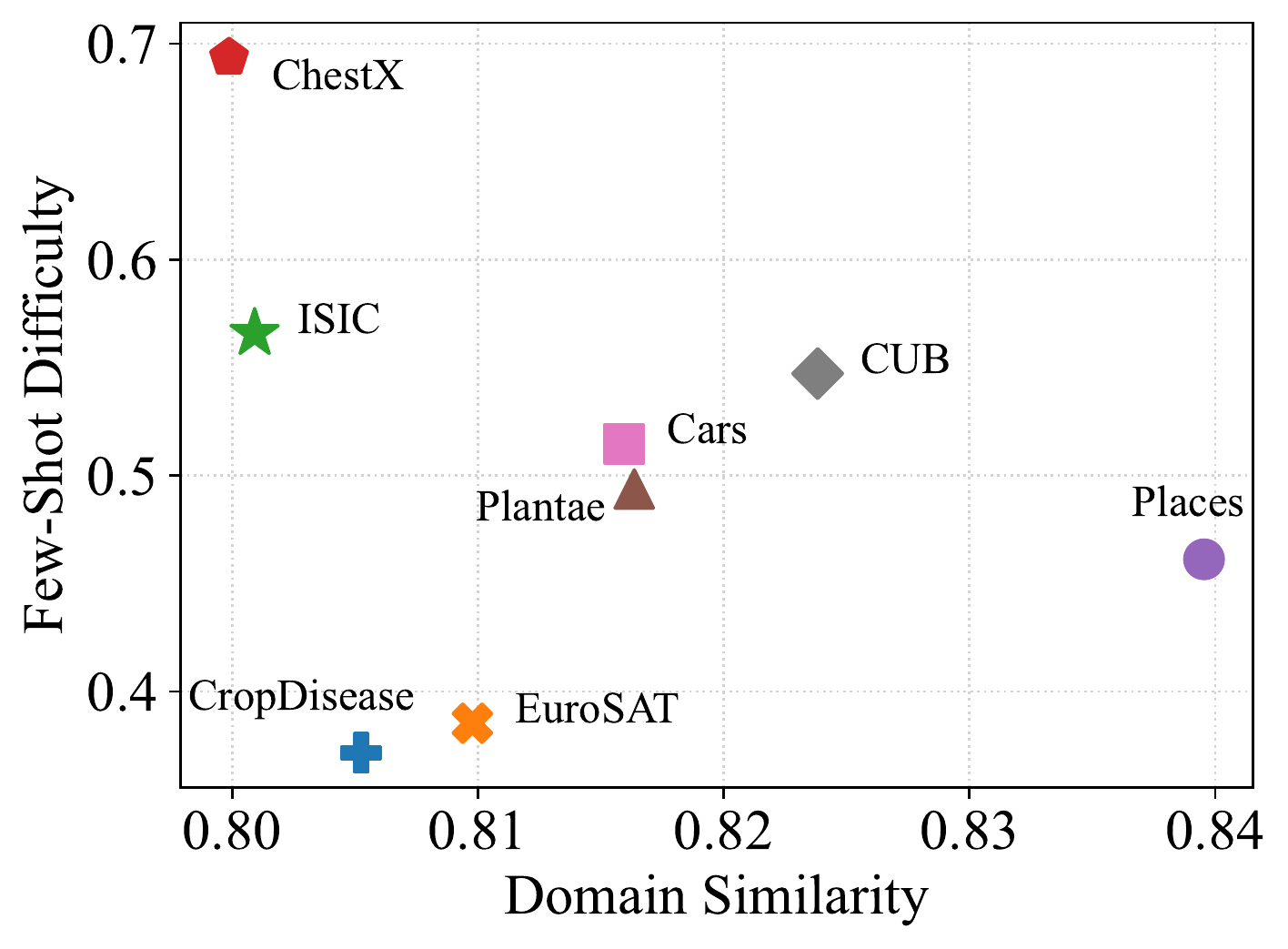}
         \caption{miniImageNet (5-shot)}
         \label{fig:similarity_difficulty_mini_5shot}
     \end{subfigure}
     \hfill
     \begin{subfigure}[h]{0.32\linewidth}
         \centering
         \includegraphics[width=\linewidth]{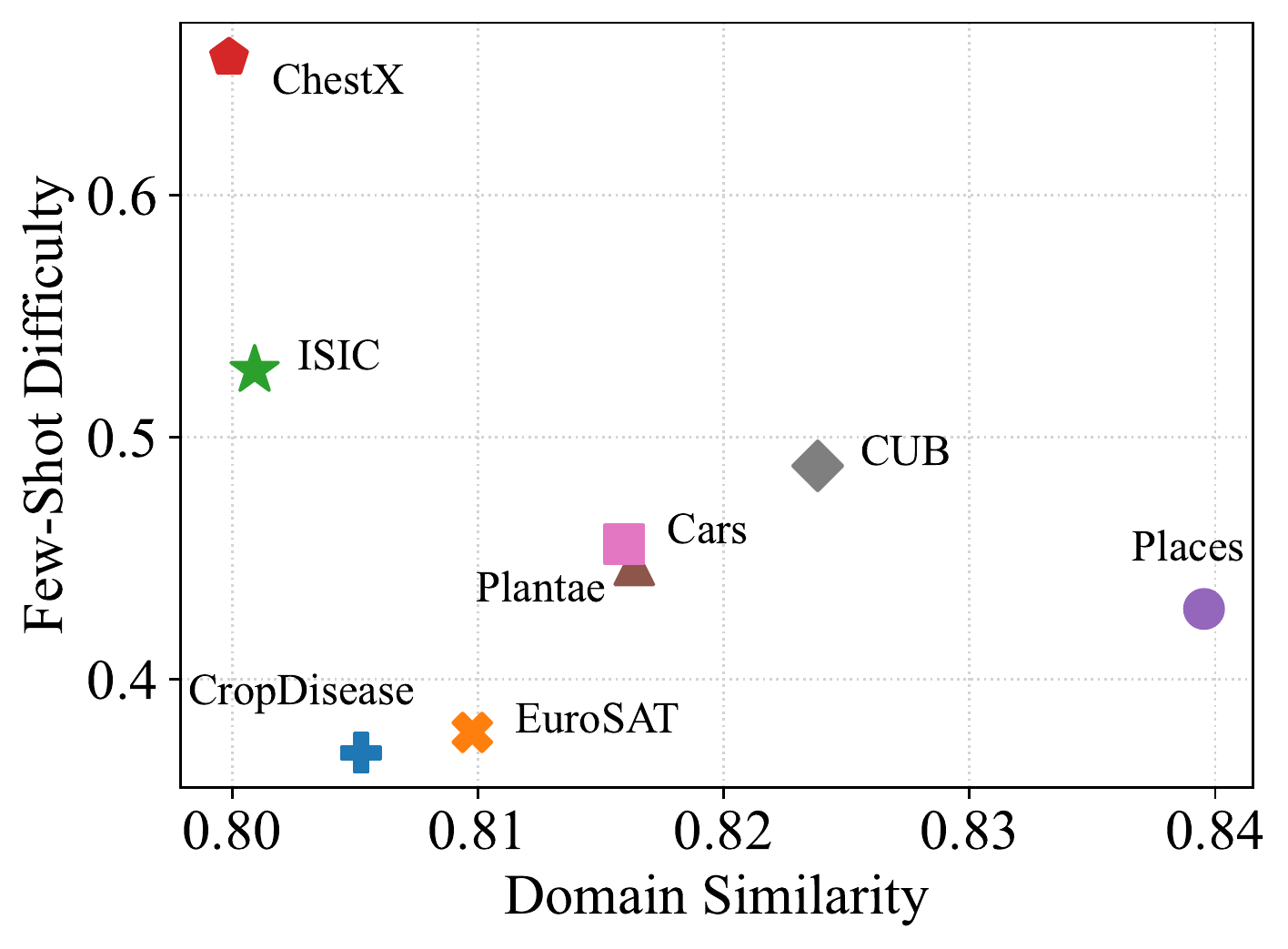}
         \caption{miniImageNet (20-shot)}
         \label{fig:similarity_difficulty_mini_20shot}
     \end{subfigure}

     \caption{Domain similarity and few-shot difficulty for eight benchmark datasets.}
     \label{fig:similarity_difficulty_all}
\end{figure}

In this section, we provide visualizations of domain similarity and few-shot difficulty. Domain similarity is dependent on the source dataset, and few-shot difficulty is dependent on backbone network (\emph{e.g.}, ResNet10 and ResNet18) and $k$. Figure \ref{fig:similarity_difficulty_all} visualizes domain similarity and few-shot difficulty for eight datasets, as depicted in Appendix \ref{appx:domain_similarity} and \ref{appx:difficulty}. Figure \ref{fig:similarity_difficulty_all}(a,b,c) have the same domain similarity, Figure \ref{fig:similarity_difficulty_all}(d,e,f) have the same, and Figure \ref{fig:similarity_difficulty_all}(g,h,i) have the same, because domain similarity is based on the source dataset. For few-shot difficulty, Figure \ref{fig:similarity_difficulty_all}(a,d) have the same difficulty, \ref{fig:similarity_difficulty_all}(b,e) have the same, and (c,f) have the same, because few-shot difficulty is based on backbone network and $k$.

\clearpage
\section{Performance of SSL according to the Ratio of Unlabeled Target Data}\label{appx:ratio_resnet10}

\begin{figure}[h]
     \centering
     \includegraphics[width=0.6\linewidth]{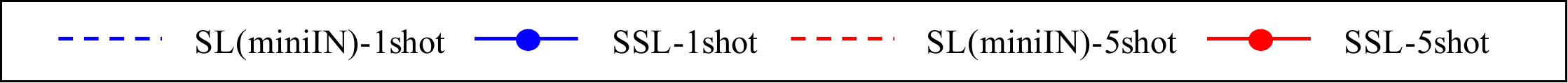}
     
     \vspace*{5pt}
     \begin{subfigure}[h]{0.24\linewidth}
         \centering
         \includegraphics[width=\linewidth]{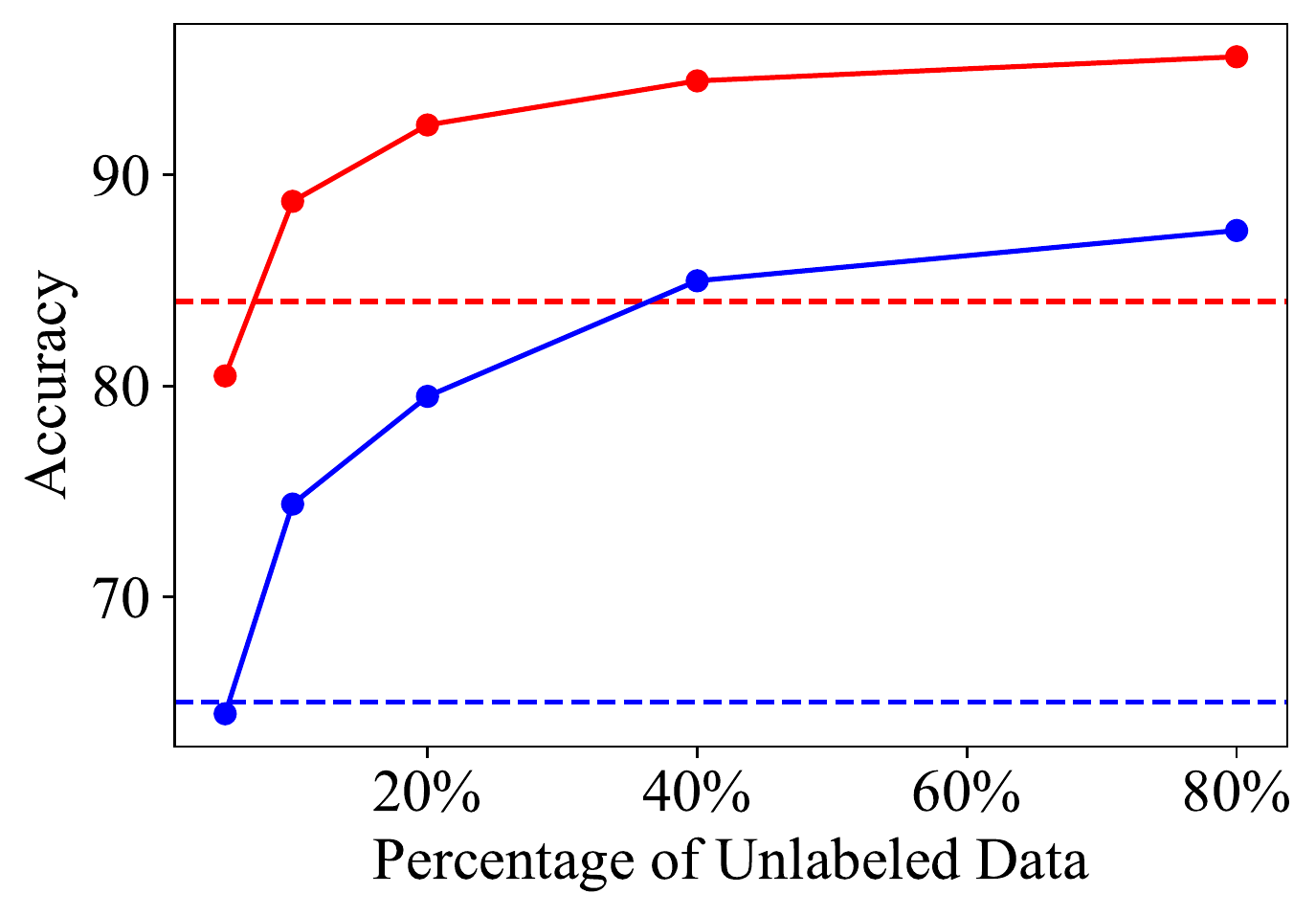}
         \caption{EuroSAT}
         \label{fig:resnet10_simclr_euro_ratio}
     \end{subfigure}
     \hfill
     \begin{subfigure}[h]{0.24\linewidth}
         \centering
         \includegraphics[width=\linewidth]{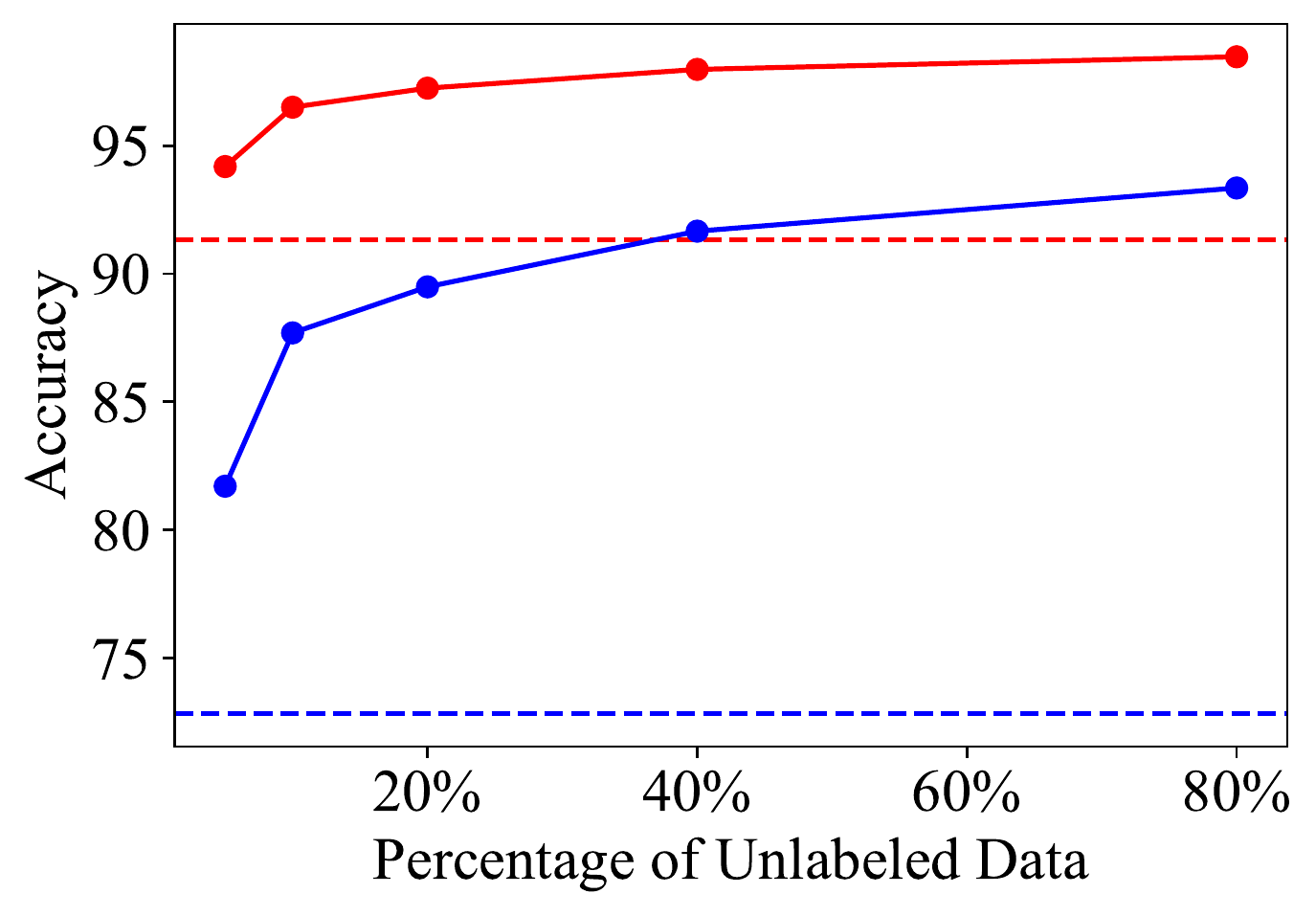}
         \caption{CropDisease}
         \label{fig:resnet10_simclr_crop_ratio}
     \end{subfigure}
     \hfill
     \begin{subfigure}[h]{0.24\linewidth}
         \centering
         \includegraphics[width=\linewidth]{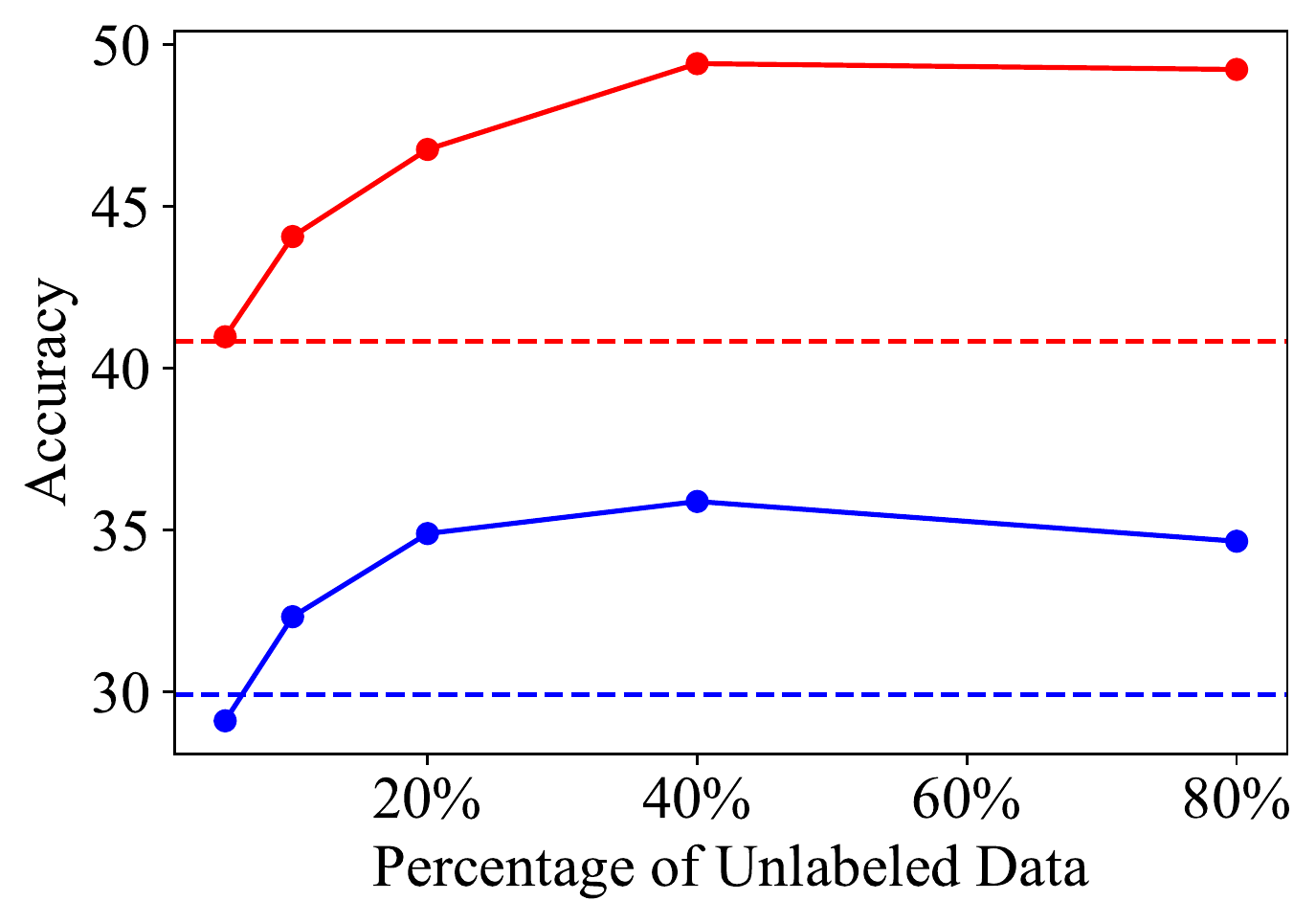}
         \caption{ISIC}
         \label{fig:resnet10_simclr_isic_ratio}
     \end{subfigure}
     \hfill
     \begin{subfigure}[h]{0.24\linewidth}
         \centering
         \includegraphics[width=\linewidth]{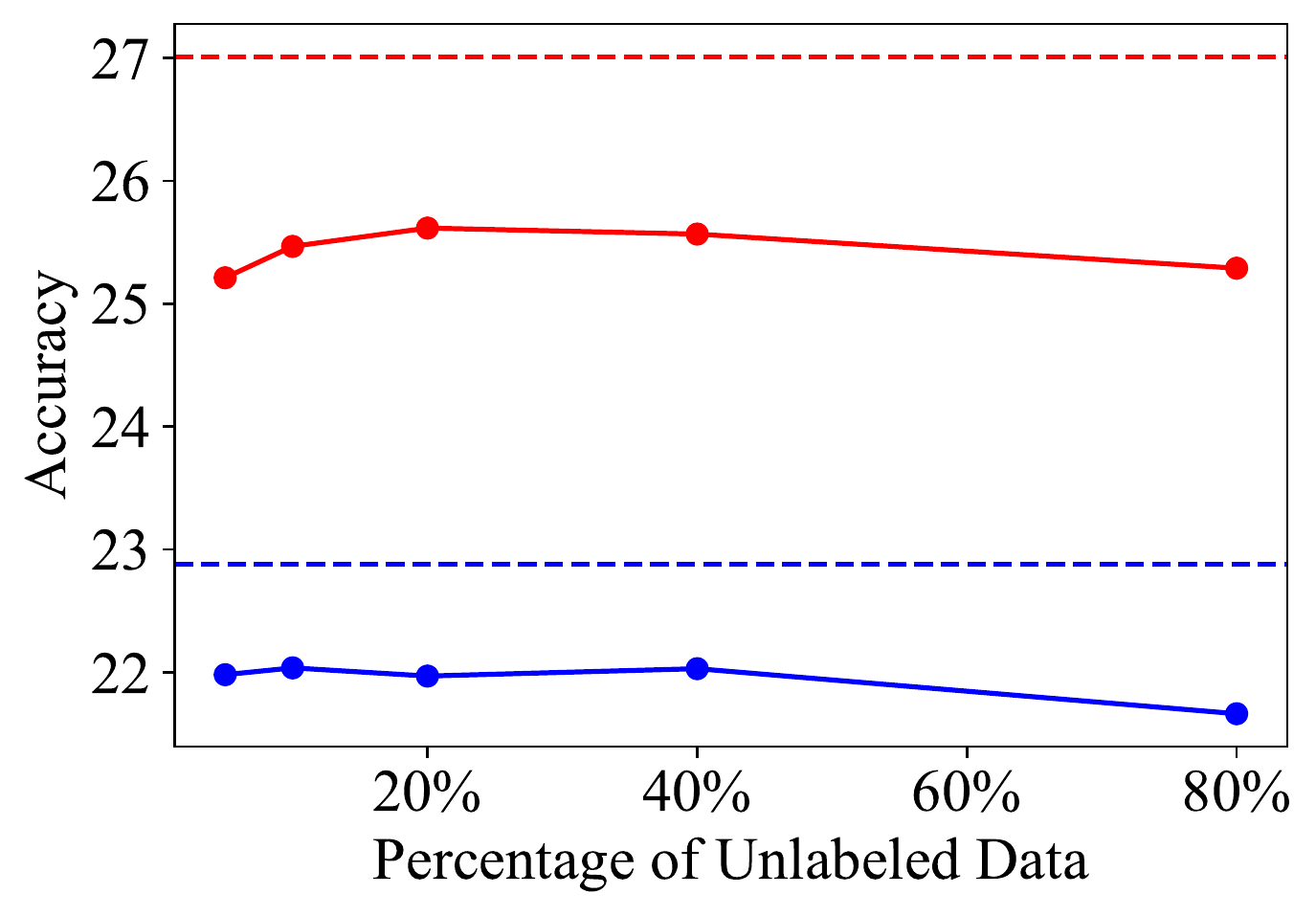}
         \caption{ChestX}
         \label{fig:resnet10_simclr_chest_ratio}
     \end{subfigure}
     \vspace*{-3pt}
     \caption{5way-$k$shot performance of SSL (SimCLR) according to the ratio of unlabeled target data and SL (Section \ref{sec:analysis1}). ResNet10 is used as a backbone. Blue and red lines indicate 1-shot and 5-shot accuracy, respectively. Dotted and solid lines are accuracy of SL (miniIN) and SSL (SimCLR), respectively.}
     \label{fig:resnet10_simclr_ratio}
\end{figure}

\begin{figure}[h]
     \centering
     \includegraphics[width=0.6\linewidth]{figure/legend_ratio_mini.pdf}
     
     \vspace*{5pt}
     \begin{subfigure}[h]{0.24\linewidth}
         \centering
         \includegraphics[width=\linewidth]{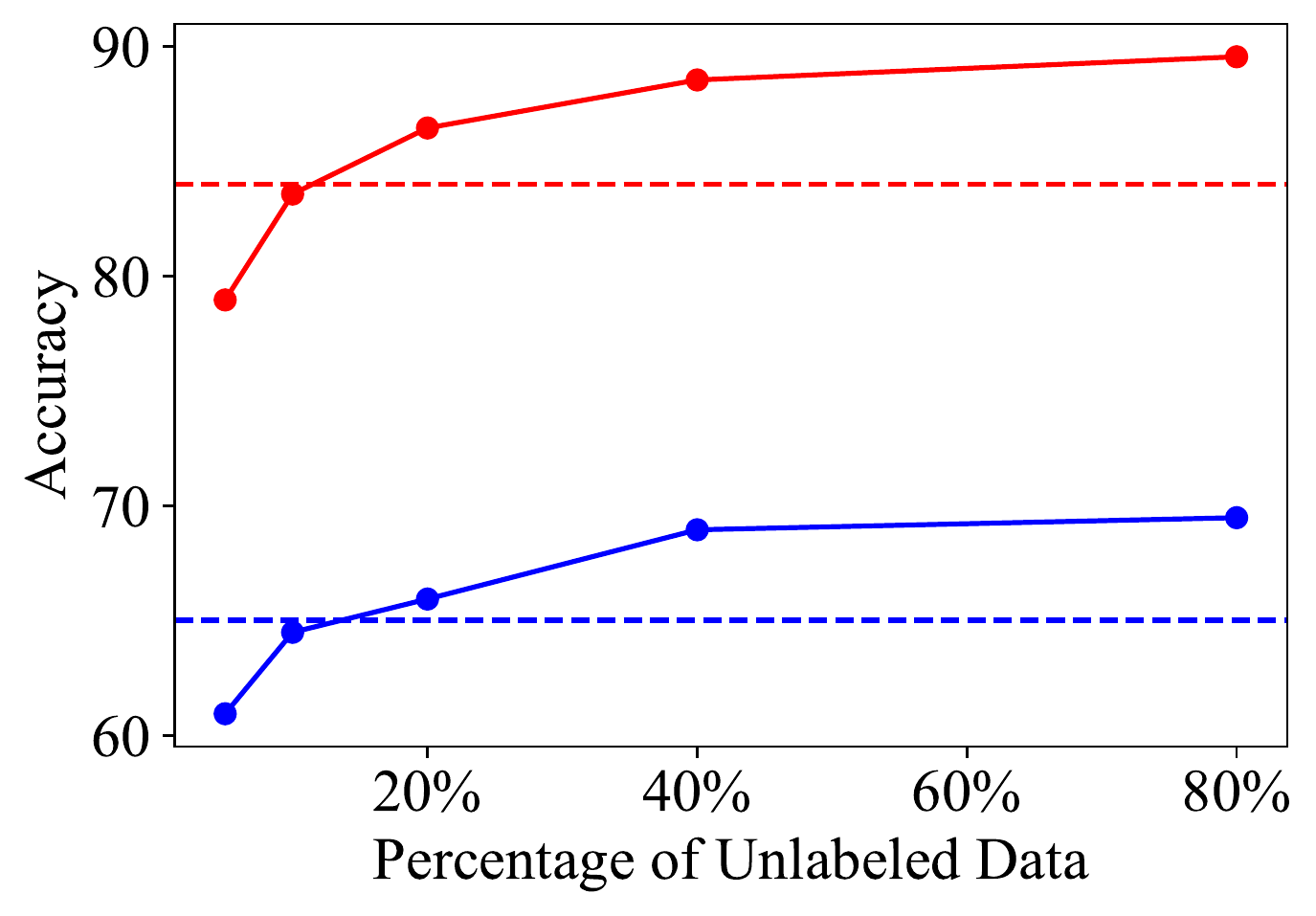}
         \caption{EuroSAT}
         \label{fig:resnet10_byol_euro_ratio}
     \end{subfigure}
     \hfill
     \begin{subfigure}[h]{0.24\linewidth}
         \centering
         \includegraphics[width=\linewidth]{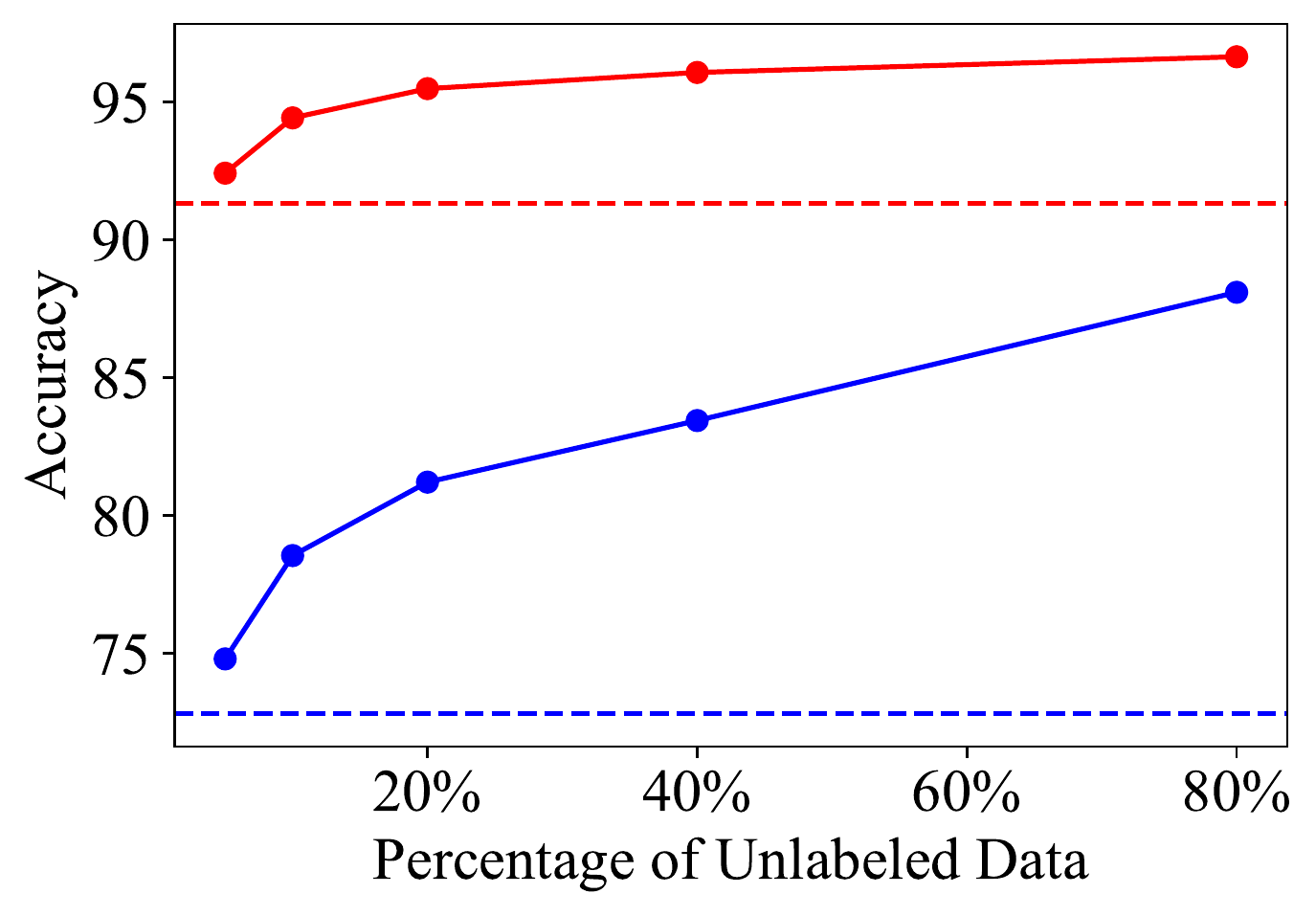}
         \caption{CropDisease}
         \label{fig:resnet10_byol_crop_ratio}
     \end{subfigure}
     \hfill
     \begin{subfigure}[h]{0.24\linewidth}
         \centering
         \includegraphics[width=\linewidth]{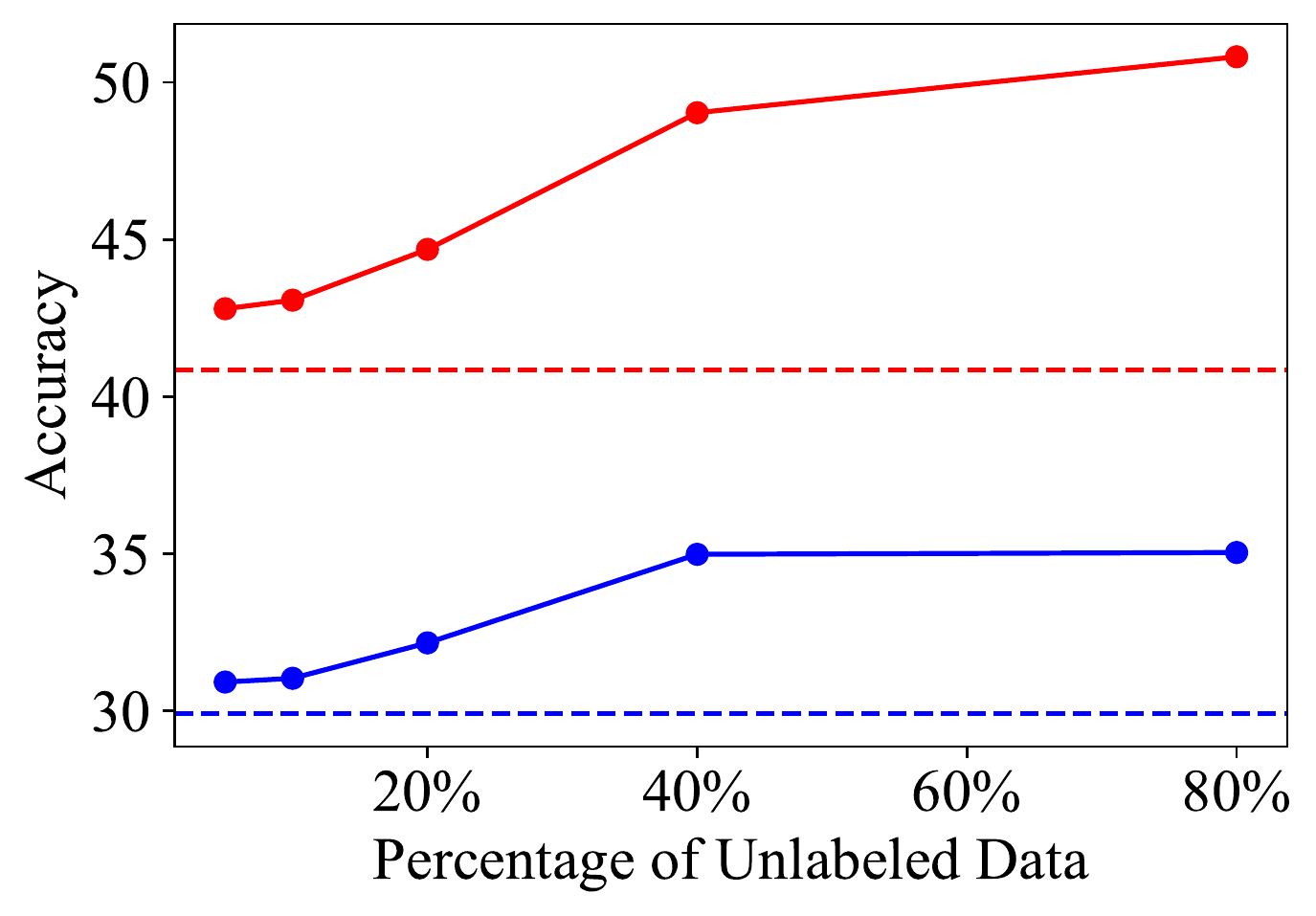}
         \caption{ISIC}
         \label{fig:resnet10_byol_isic_ratio}
     \end{subfigure}
     \hfill
     \begin{subfigure}[h]{0.24\linewidth}
         \centering
         \includegraphics[width=\linewidth]{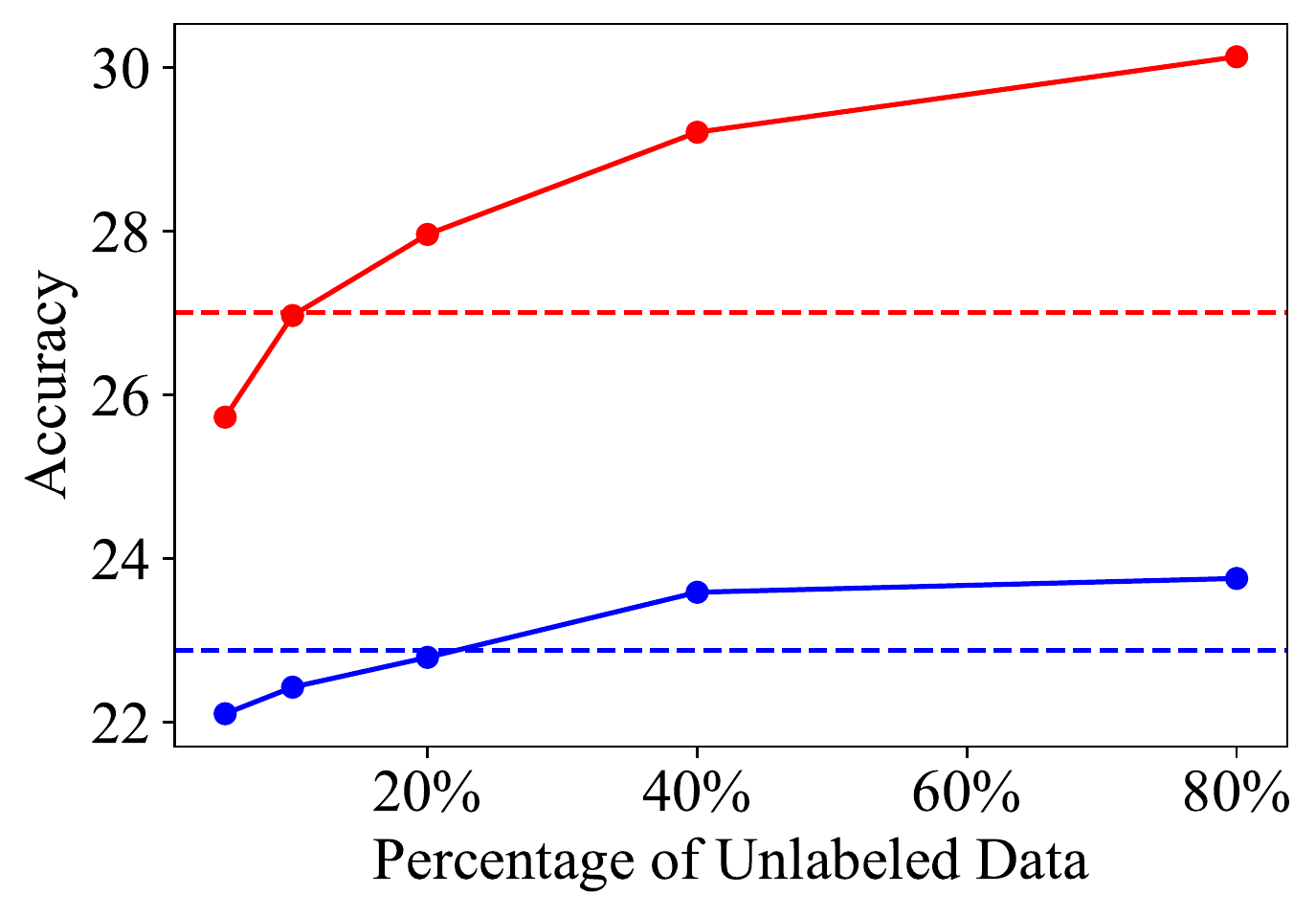}
         \caption{ChestX}
         \label{fig:resnet10_byol_chest_ratio}
     \end{subfigure}
     \vspace*{-3pt}
     \caption{5way-$k$shot performance of SSL (BYOL) according to the ratio of unlabeled target data and SL (Section \ref{sec:analysis1}). ResNet10 is used as a backbone. Blue and red lines indicate 1-shot and 5-shot accuracy, respectively. Dotted and solid lines are accuracy of SL (miniIN) and SSL (BYOL), respectively.}
     \label{fig:resnet10_byol_ratio}
\end{figure}

\begin{figure}[h]
     \centering
     \includegraphics[width=0.8\linewidth]{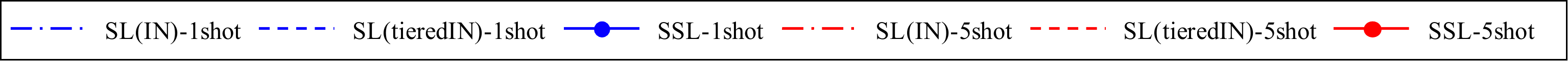}
     
     \vspace*{5pt}
     \begin{subfigure}[h]{0.24\linewidth}
         \centering
         \includegraphics[width=\linewidth]{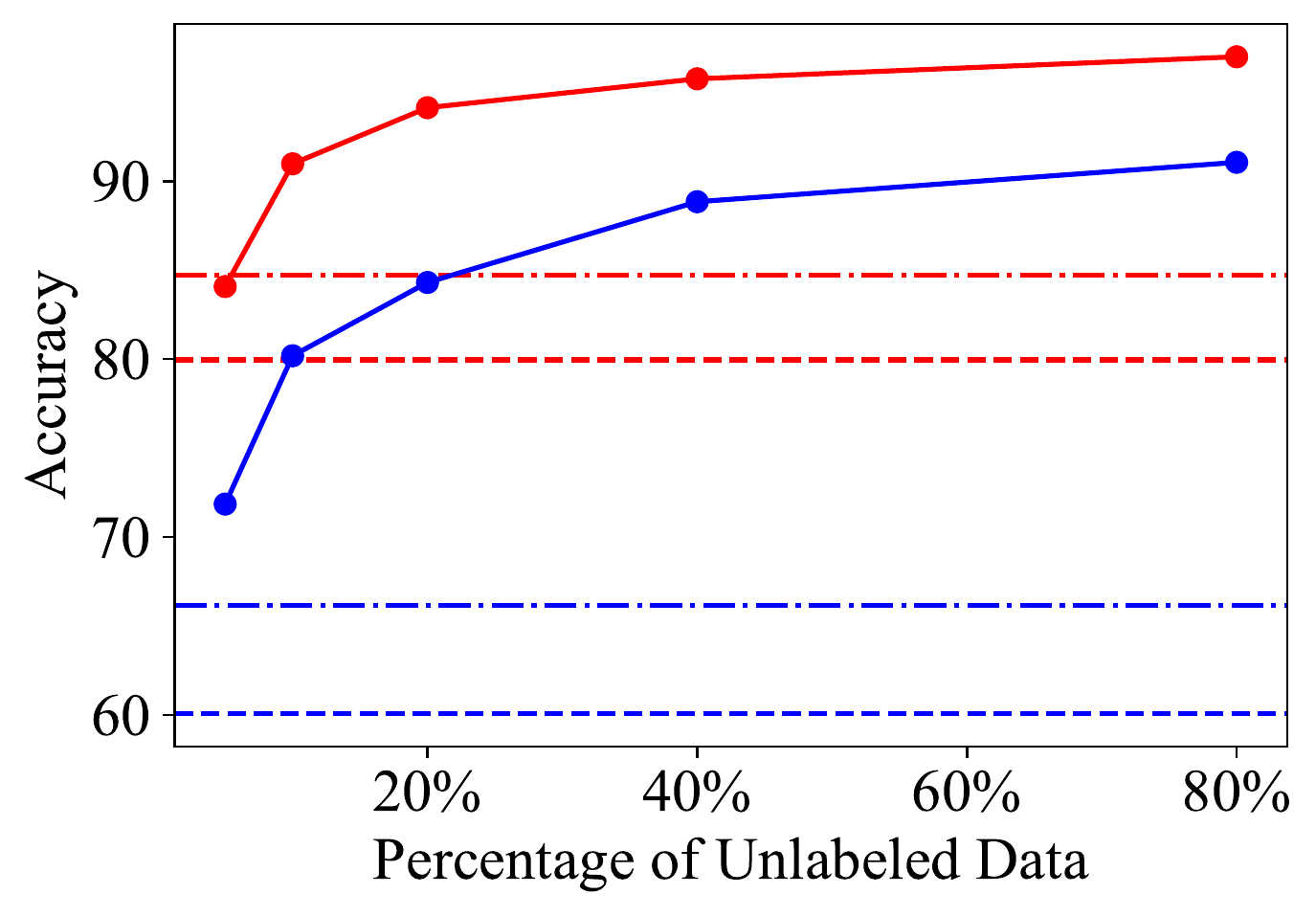}
         \caption{EuroSAT}
         \label{fig:resnet18_simclr_euro_ratio}
     \end{subfigure}
     \hfill
     \begin{subfigure}[h]{0.24\linewidth}
         \centering
         \includegraphics[width=\linewidth]{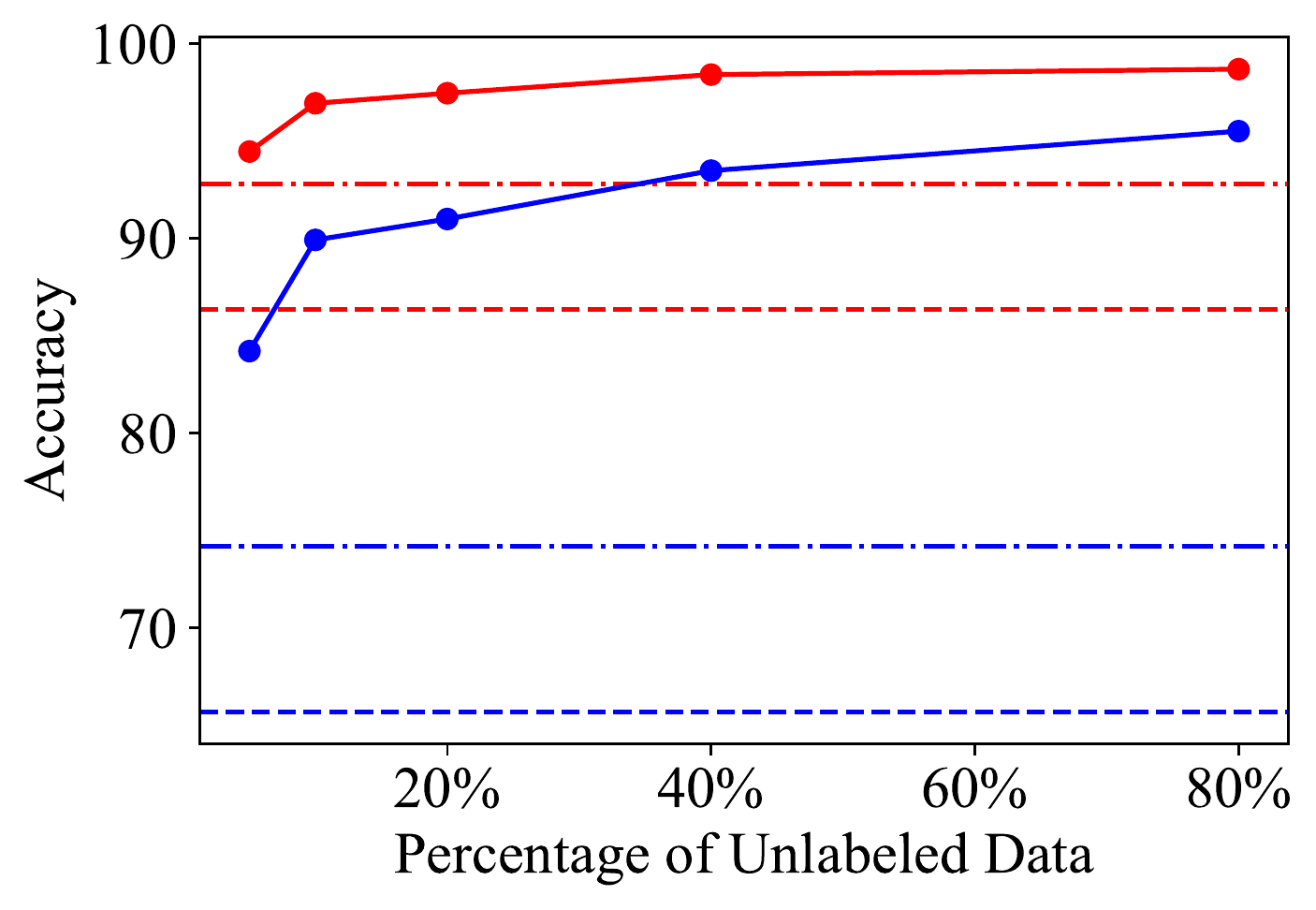}
         \caption{CropDisease}
         \label{fig:resnet18_simclr_crop_ratio}
     \end{subfigure}
     \hfill
     \begin{subfigure}[h]{0.24\linewidth}
         \centering
         \includegraphics[width=\linewidth]{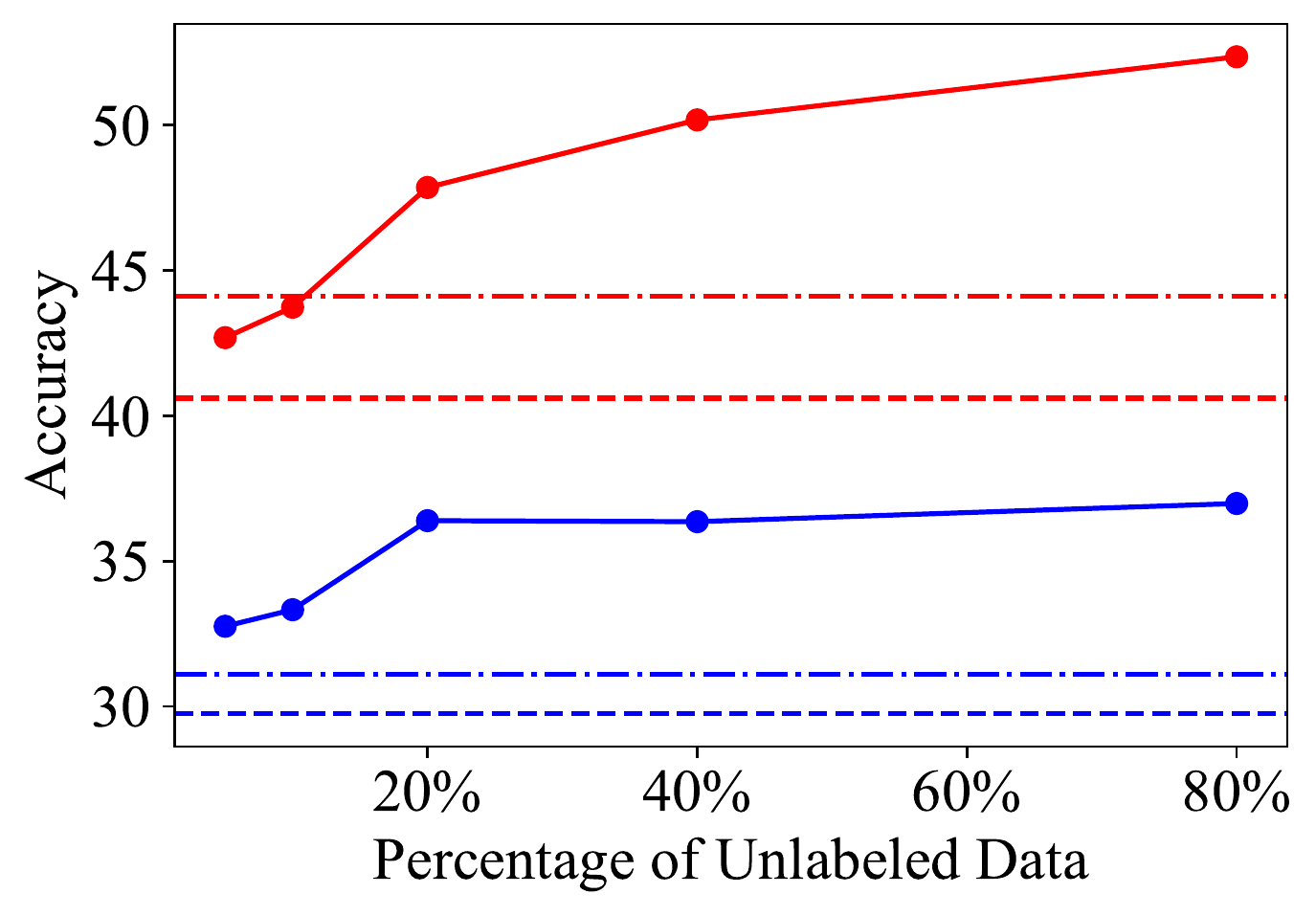}
         \caption{ISIC}
         \label{fig:resnet18_simclr_isic_ratio}
     \end{subfigure}
     \hfill
     \begin{subfigure}[h]{0.24\linewidth}
         \centering
         \includegraphics[width=\linewidth]{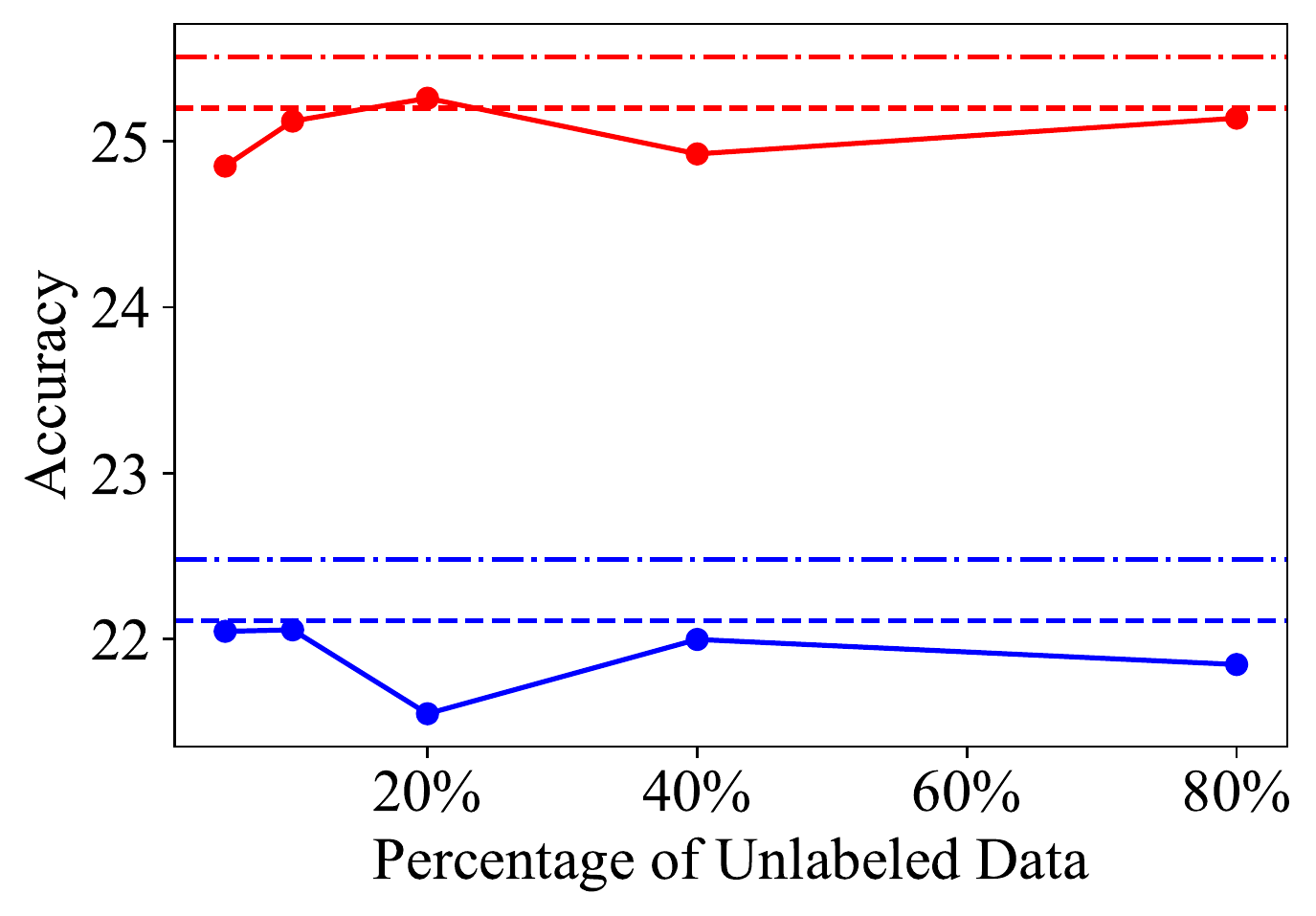}
         \caption{ChestX}
         \label{fig:resnet18_simclr_chest_ratio}
     \end{subfigure}
     \vspace*{-3pt}
     \caption{5way-$k$shot performance of SSL according to the ratio of unlabeled target data and SL (Section \ref{sec:analysis1}). ResNet18 is used as a backbone. SimCLR is used for the SSL method.}
     \label{fig:resnet18_simclr_ratio}
\end{figure}

In this section, we evaluate few-shot performance of SSL (SimCLR and BYOL) according to the ratio of unlabeled target data when ResNet10 is used as a backbone network. Figure \ref{fig:resnet10_simclr_ratio} and Figure \ref{fig:resnet10_byol_ratio} describe the few-shot performance according to the ratio of unlabeled target data when SimCLR and BYOL are used for SSL method, respectively. We control the ratio $\in$ \{5\%, 10\%, 20\%, 40\%, 80\%\}. We further evaluate few-shot performance of SSL (SimCLR) when ResNet18 is used as a backbone network, depicted as Figure \ref{fig:resnet18_simclr_ratio}.

It is observed that except for ChestX, SimCLR with a small portion (even 5\%) of target data as unlabeled data has better performance than SL that uses ImageNet, tieredImageNet, and miniImageNet. Note that ImageNet and tieredImageNet include around 1.3 million and 0.45 million samples with annotations, respectively. On the other hand, 5\% of EuroSAT, CropDisease, and ISIC unlabeled data include around 1.4k, 2.2k, and 0.5k samples. It implies that the consistency between source and target domains is much more important than the number of data for pre-training.

\vspace*{-10pt}

\vspace*{-10pt}

\section{Performance of SL and SSL for the Other Datasets}\label{appx:other_dataset_slssl}

Table \ref{tab:main_ssl_fwt} summarizes the few-shot performance of SL and SSL on non-BSCD-FSL datasets. Note that these datasets are known to be closer to the ImageNet than BSCD-FSL datasets \citep{ericsson2021well} and our estimated similarity shows the same trend. We would like to highlight that unlike BSCD-FSL sets, these four target domains take a big advantage from the ImageNet dataset. SL pre-trained on ImageNet has comparable or even better (in Cars and CUB) performance than SSL.

\begin{table*}[h]
\caption{$5$-way $k$-shot CD-FSL performance of the models pre-trained by SL and SSL, on four additional target datastes: Places, Plantae, Cars, and CUB. We report the average accuracy and its 95\% confidence interval over 600 few-shot episodes. B and S indicate base and strong augmentations, respectively. The best results are marked in bold and the second best are underlined. We include the result when using the model pre-trained on the entire ImageNet data, which also uses the ResNet18 backbone as tieredImageNet experiments.}\label{tab:main_ssl_fwt}
\centering
\footnotesize\addtolength{\tabcolsep}{-3.5pt}
\resizebox{\linewidth}{!}{
\begin{tabular}{c|c|c|c|cc|cc|cc|cc}
    \toprule
    \multirow{2}{*}{\makecell[l]{\!Source\!\\~Data\!}} & Pre-train & \multirow{2}{*}{Method} & \multirow{2}{*}{Aug.} & \multicolumn{2}{c|}{Places} & \multicolumn{2}{c|}{Plantae} & \multicolumn{2}{c|}{Cars} & \multicolumn{2}{c}{CUB} \\
    & Scheme & & & $k$=1 & $k$=5 & $k$=1 & $k$=5 & $k$=1 & $k$=5 & $k$=1 & $k$=5 \\
    \midrule
    ImageNet & SL          & Default                      & B & \underline{57.47}{\scriptsize$\pm$.86} & \underline{79.22}{\scriptsize$\pm$.64}  & \underline{43.66}{\scriptsize$\pm$.80} & \textbf{63.21}{\scriptsize$\pm$.82} & \textbf{45.82}{\scriptsize$\pm$.79} & \textbf{66.38}{\scriptsize$\pm$.80} & \textbf{65.24}{\scriptsize$\pm$.97} & \textbf{83.93}{\scriptsize$\pm$.66} \\
    \midrule
    \multirow{2}{*}{\makecell[l]{\!~~~~tiered\\\!ImageNet}} & \multirow{2}{*}{SL} & \multirow{2}{*}{Default}     & B & 52.07{\scriptsize$\pm$.86} & 72.12{\scriptsize$\pm$.69} & 38.63{\scriptsize$\pm$.74} & 54.76{\scriptsize$\pm$.82} & 31.23{\scriptsize$\pm$.65} & 42.59{\scriptsize$\pm$.70} & \underline{57.94}{\scriptsize$\pm$.93} & \underline{76.86}{\scriptsize$\pm$.78} \\
    &           &                              & S & 52.82{\scriptsize$\pm$.86} & 72.96{\scriptsize$\pm$.67} & 34.99{\scriptsize$\pm$.64} & 51.11{\scriptsize$\pm$.76}  & 31.05{\scriptsize$\pm$.63} & 42.32{\scriptsize$\pm$.69} & 54.18{\scriptsize$\pm$.91} & 74.14{\scriptsize$\pm$.80}  \\
    \midrule
    \multirow{8}{*}{\makecell[l]{\!Target\!\\~Data\!}} & \multirow{8}{*}{SSL} & \multirow{2}{*}{SimCLR}      & B & 45.82{\scriptsize$\pm$.85} & 62.07{\scriptsize$\pm$.78} & 38.52{\scriptsize$\pm$.74} & 53.89{\scriptsize$\pm$.80}  & 28.86{\scriptsize$\pm$.68} & 37.05{\scriptsize$\pm$.69} & 33.56{\scriptsize$\pm$.67} & 43.99{\scriptsize$\pm$.71} \\
    &                      &                              & S & \textbf{64.97}{\scriptsize$\pm$.94} & \textbf{80.43}{\scriptsize$\pm$.61} & \textbf{44.18}{\scriptsize$\pm$.85} & 60.07{\scriptsize$\pm$.84} & 32.46{\scriptsize$\pm$.70} & 44.55{\scriptsize$\pm$.74} & 36.15{\scriptsize$\pm$.76} & 47.36{\scriptsize$\pm$.79} \\
    &                      & \multirow{2}{*}{MoCo}        & B & 39.64{\scriptsize$\pm$.82} & 53.95{\scriptsize$\pm$.77} & 35.17{\scriptsize$\pm$.73} & 48.83{\scriptsize$\pm$.76} & 27.40{\scriptsize$\pm$.64} & 34.59{\scriptsize$\pm$.67} & 29.67{\scriptsize$\pm$.59} & 36.93{\scriptsize$\pm$.61}  \\
    &                      &                              & S & 55.53{\scriptsize$\pm$.74} & 71.50{\scriptsize$\pm$.73} & 36.49{\scriptsize$\pm$.73} & 49.15{\scriptsize$\pm$.76} & 29.36{\scriptsize$\pm$.67} & 38.44{\scriptsize$\pm$.70} &  31.76{\scriptsize$\pm$.66} & 40.81{\scriptsize$\pm$.72}  \\
    &                      & \multirow{2}{*}{BYOL}        & B & 40.38{\scriptsize$\pm$.72} & 60.06{\scriptsize$\pm$.73} & 38.60{\scriptsize$\pm$.72} & 57.81{\scriptsize$\pm$.81} & 31.04{\scriptsize$\pm$.66} & 41.79{\scriptsize$\pm$.72} &  35.27{\scriptsize$\pm$.67} & 49.61{\scriptsize$\pm$.71}  \\
    &                      &                              & S & 51.76{\scriptsize$\pm$.79} & 72.47{\scriptsize$\pm$.63} & 42.16{\scriptsize$\pm$.75} & \underline{61.02}{\scriptsize$\pm$.82} &  \underline{34.54}{\scriptsize$\pm$.70} & \underline{48.56}{\scriptsize$\pm$.76} & 36.50{\scriptsize$\pm$.68} & 51.31{\scriptsize$\pm$.78} \\
    &                      & \multirow{2}{*}{SimSiam}     & B & 35.27{\scriptsize$\pm$.68} & 48.12{\scriptsize$\pm$.69} & 36.11{\scriptsize$\pm$.76} & 48.63{\scriptsize$\pm$.79}  & 28.30{\scriptsize$\pm$.64} & 35.24{\scriptsize$\pm$.65} & 29.96{\scriptsize$\pm$.62} & 37.61{\scriptsize$\pm$.60}  \\
    &                      &                              & S & 52.56{\scriptsize$\pm$.92} & 68.29{\scriptsize$\pm$.74} & 36.19{\scriptsize$\pm$.69} & 50.23{\scriptsize$\pm$.76}  & 31.21{\scriptsize$\pm$.64} & 43.06{\scriptsize$\pm$.67} & 33.73{\scriptsize$\pm$.71} & 43.22{\scriptsize$\pm$.74}  \\
    \bottomrule
    \multicolumn{12}{c}{(a) ResNet18 is used as a backbone.}\\ \toprule
    \multirow{2}{*}{\makecell[l]{\!~~~~mini\\\!ImageNet}} & \multirow{2}{*}{SL} & \multirow{2}{*}{Default}      & B & 51.84{\scriptsize$\pm$.80} & 72.19{\scriptsize$\pm$.70} & 37.28{\scriptsize$\pm$.69} & 54.15{\scriptsize$\pm$.74} & 30.79{\scriptsize$\pm$.56} & 44.36{\scriptsize$\pm$.69} & \textbf{40.65}{\scriptsize$\pm$.78} & \textbf{58.54}{\scriptsize$\pm$.81}  \\
    &            &    & S & \underline{52.45}{\scriptsize$\pm$.78} & \underline{72.92}{\scriptsize$\pm$.66} & 36.72{\scriptsize$\pm$.67} & 53.26{\scriptsize$\pm$.73} &  30.20{\scriptsize$\pm$.54} & \underline{44.39}{\scriptsize$\pm$.66} & \underline{40.56}{\scriptsize$\pm$.78} & \underline{58.10}{\scriptsize$\pm$.78}  \\
    \midrule
    \multirow{8}{*}{\makecell[l]{\!Target\!\\~Data\!}} & \multirow{8}{*}{SSL} & \multirow{2}{*}{SimCLR}      & B & 44.06{\scriptsize$\pm$.78} & 62.86{\scriptsize$\pm$.78} & 38.43{\scriptsize$\pm$.77} & 54.68{\scriptsize$\pm$.80} & 28.59{\scriptsize$\pm$.66} & 38.24{\scriptsize$\pm$.73} & 33.88{\scriptsize$\pm$.68} & 45.31{\scriptsize$\pm$.73}  \\
    &                      &                              & S & \textbf{58.75}{\scriptsize$\pm$.93} & \textbf{78.39}{\scriptsize$\pm$.61} & \textbf{42.65}{\scriptsize$\pm$.80} & \textbf{59.77}{\scriptsize$\pm$.82} & \underline{30.89}{\scriptsize$\pm$.66} & \textbf{45.60}{\scriptsize$\pm$.72} & 35.49{\scriptsize$\pm$.73} & 47.69{\scriptsize$\pm$.77} \\
    &                      & \multirow{2}{*}{MoCo}        & B & 38.41{\scriptsize$\pm$.74} & 54.65{\scriptsize$\pm$.74} & 33.96{\scriptsize$\pm$.69} & 47.51{\scriptsize$\pm$.72} & 28.03{\scriptsize$\pm$.66} & 36.19{\scriptsize$\pm$.72} & 32.37{\scriptsize$\pm$.65} & 40.55{\scriptsize$\pm$.72}  \\
    &                      &                              & S & 52.05{\scriptsize$\pm$.90} & 71.57{\scriptsize$\pm$.70} & 36.36{\scriptsize$\pm$.73} & 50.37{\scriptsize$\pm$.78} & 28.25{\scriptsize$\pm$.61} & 38.89{\scriptsize$\pm$.69} & 33.53{\scriptsize$\pm$.72} & 42.87{\scriptsize$\pm$.74}  \\
    &                      & \multirow{2}{*}{BYOL}        & B & 40.60{\scriptsize$\pm$.69} & 59.28{\scriptsize$\pm$.71} & \underline{39.27}{\scriptsize$\pm$.73} & \underline{55.87}{\scriptsize$\pm$.79} & 30.11{\scriptsize$\pm$.62} & 41.21{\scriptsize$\pm$.69} & 34.74{\scriptsize$\pm$.65} & 49.10{\scriptsize$\pm$.74}  \\
    &                      &                              & S & 47.81{\scriptsize$\pm$.75} & 68.14{\scriptsize$\pm$.68}  & 39.12{\scriptsize$\pm$.71} & 55.31{\scriptsize$\pm$.79} & \textbf{31.53}{\scriptsize$\pm$.65} & 43.92{\scriptsize$\pm$.70} &  35.96{\scriptsize$\pm$.70} & 49.34{\scriptsize$\pm$.76} \\
    &                      & \multirow{2}{*}{SimSiam}     & B & 39.27{\scriptsize$\pm$.72} & 53.40{\scriptsize$\pm$.74} & 37.12{\scriptsize$\pm$.72} & 50.61{\scriptsize$\pm$.81} & 28.49{\scriptsize$\pm$.62} & 35.50{\scriptsize$\pm$.67} & 30.37{\scriptsize$\pm$.63} & 38.22{\scriptsize$\pm$.62}  \\
    &                      &                              & S & 51.62{\scriptsize$\pm$.81} & 69.77{\scriptsize$\pm$.66} & 38.49{\scriptsize$\pm$.73} & 53.10{\scriptsize$\pm$.78} & 30.00{\scriptsize$\pm$.59} & 40.92{\scriptsize$\pm$.67} & 34.25{\scriptsize$\pm$.71} & 44.85{\scriptsize$\pm$.74}  \\
    \bottomrule
    \multicolumn{12}{c}{(b) ResNet10 is used as a backbone.}\\
\end{tabular}}
\end{table*}

\clearpage
\section{Analyses on Other Source Datasets}\label{appx:other_source}

In this section, we expand our analyses of SL and SSL on ImageNet source in Section \ref{sec:analysis2}, onto two additional source datasets: tieredImageNet and miniImageNet.
Every observation that was previously identified on ImageNet is consistently made in the two additional source datasets.

\subsection{Limitations of Domain Similarity (Observation \ref{obs:obs5_2})} 
\begin{figure}[h]
    \centering
    \begin{subfigure}[h]{.49\linewidth}
        \centering
        \includegraphics[width=\linewidth]{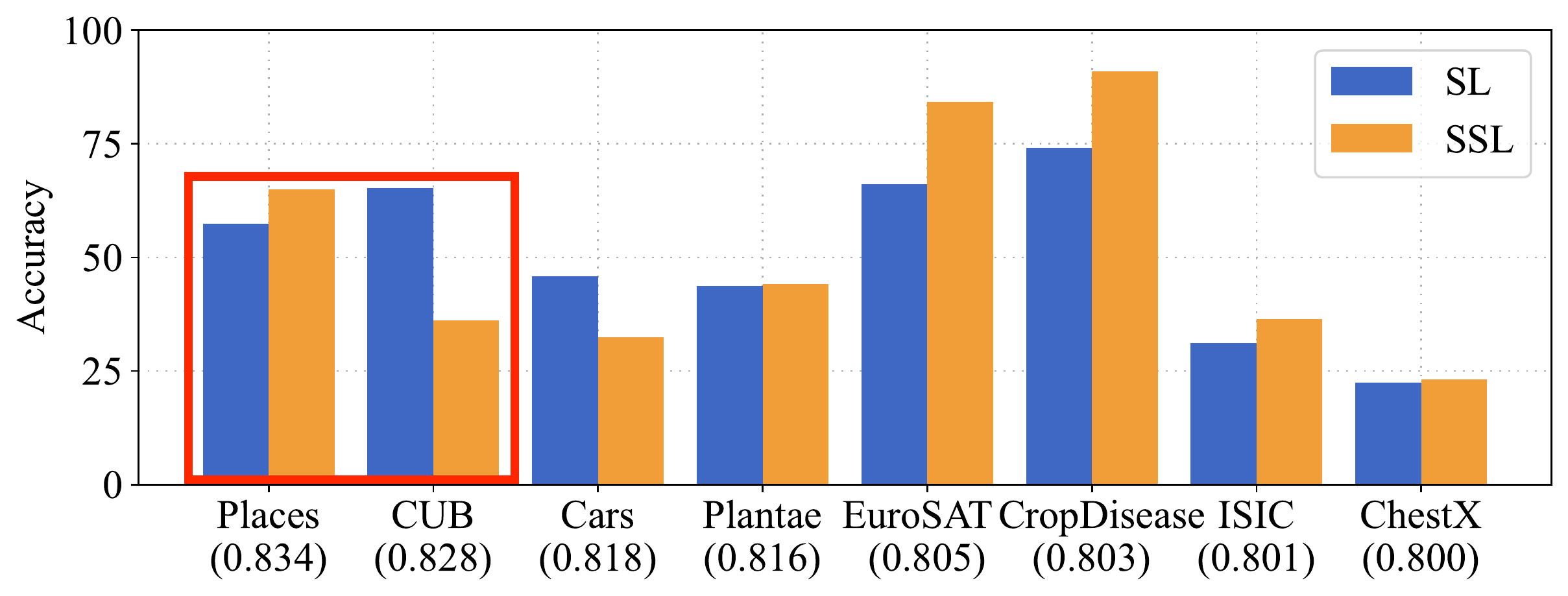}
        \vspace*{-0.6cm}
        \caption{ImageNet (5-way 1-shot)}
        \label{fig:sl_ssl_by_similarity_imagenet_1shot}
    \end{subfigure}
    \begin{subfigure}[h]{.49\linewidth}
        \centering
        \includegraphics[width=\linewidth]{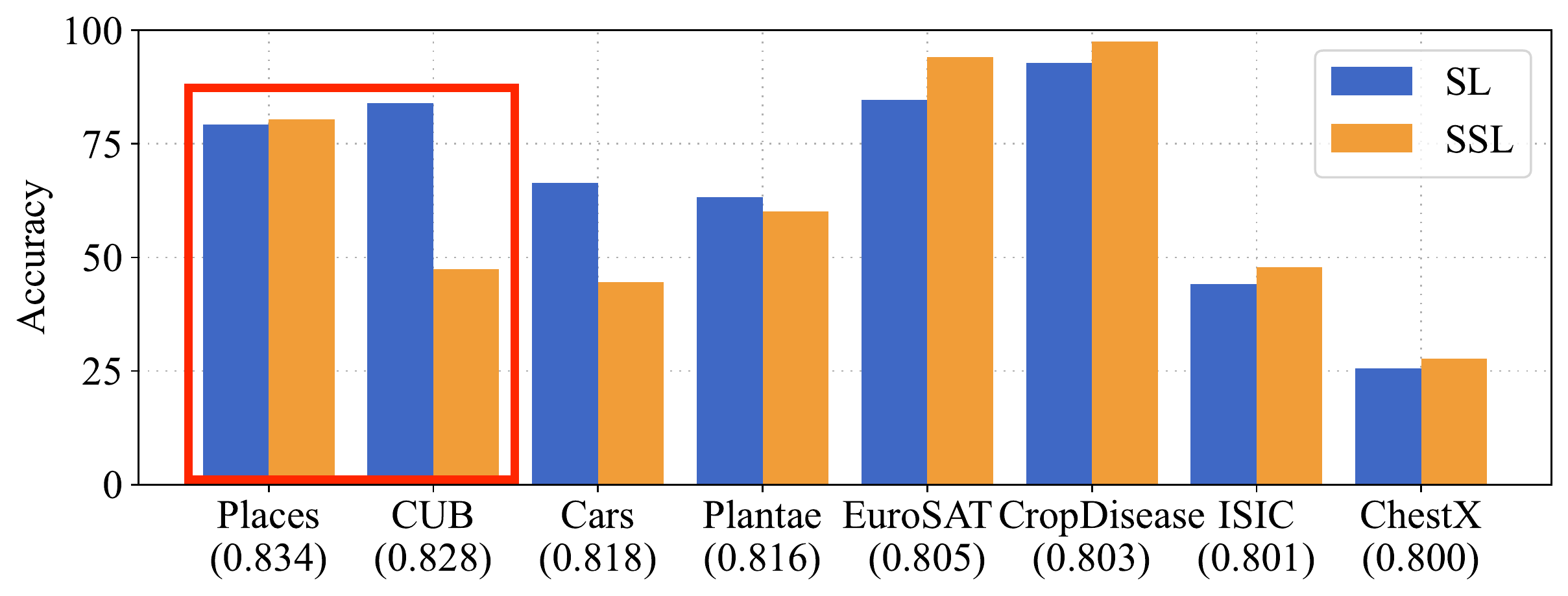}
         \vspace*{-0.6cm}
         \caption{ImageNet (5-way 5-shot)}
         \label{fig:sl_ssl_by_similarity_imagenet_5shot}
     \end{subfigure}
     
    \begin{subfigure}[h]{.49\linewidth}
        \centering
        \includegraphics[width=\linewidth]{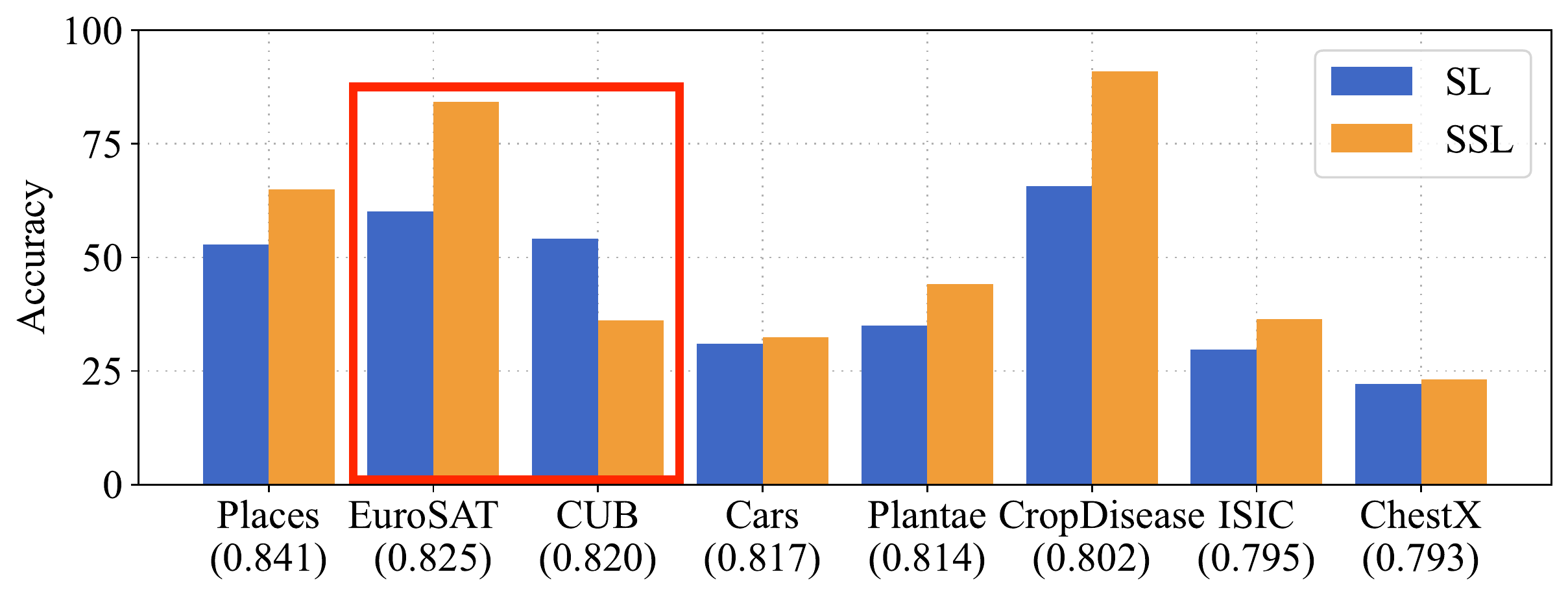}
        \vspace*{-0.6cm}
        \caption{tieredImageNet (5-way 1-shot)}
        \label{fig:sl_ssl_by_similarity_tiered_1shot}
    \end{subfigure}
    \begin{subfigure}[h]{.49\linewidth}
        \centering
        \includegraphics[width=\linewidth]{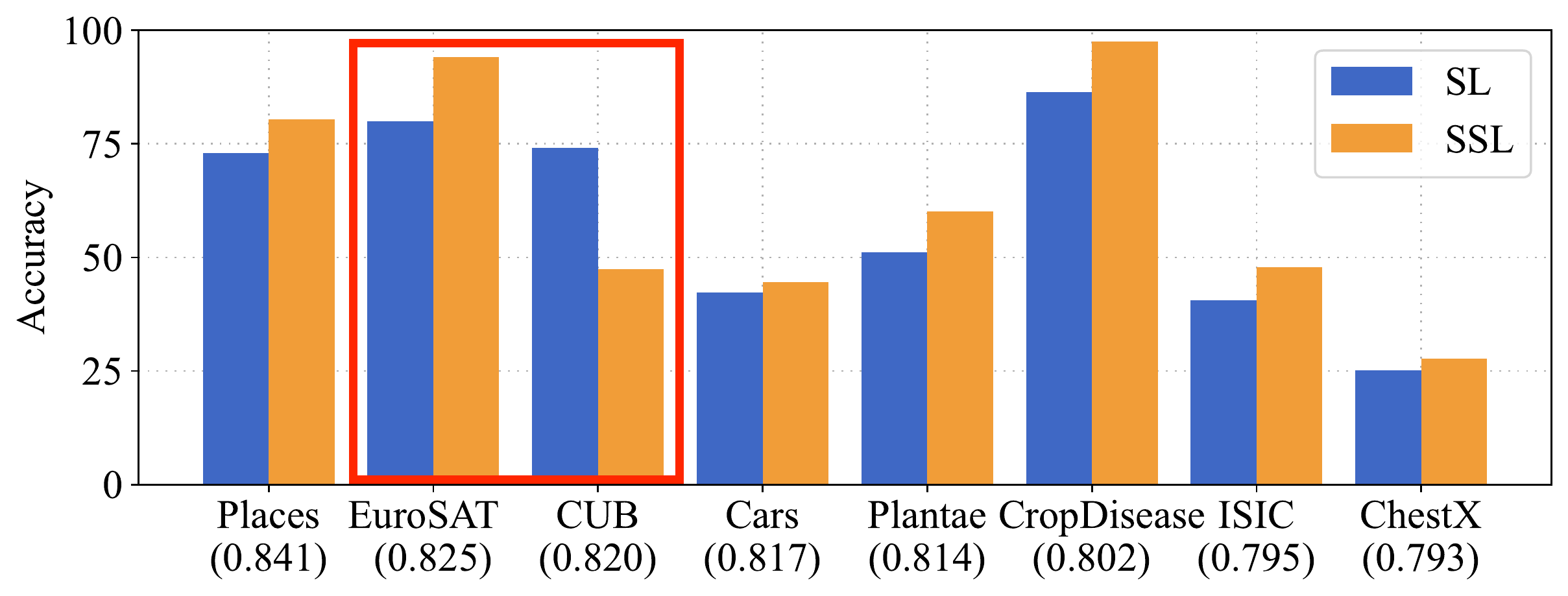}
         \vspace*{-0.6cm}
         \caption{tieredImageNet (5-way 5-shot)}
         \label{fig:sl_ssl_by_similarity_tiered_5shot}
     \end{subfigure}
     
    \begin{subfigure}[h]{.49\linewidth}
        \centering
        \includegraphics[width=\linewidth]{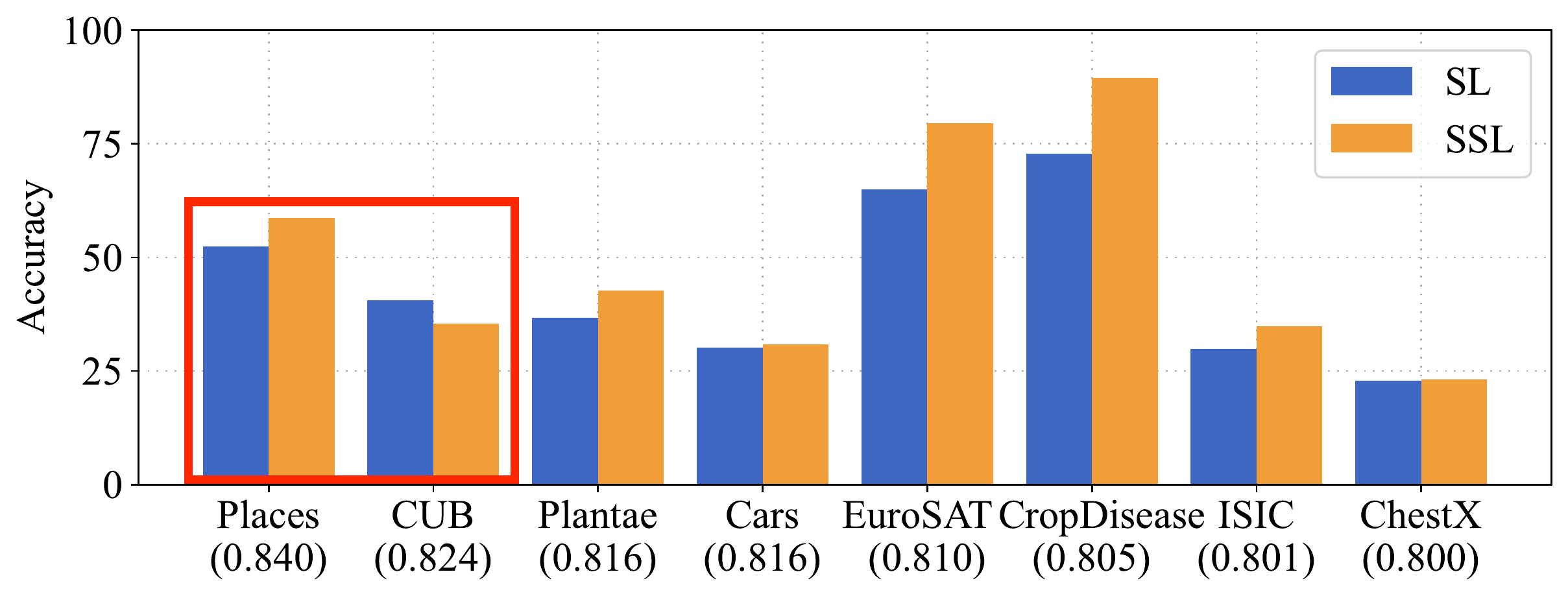}
        \vspace*{-0.6cm}
        \caption{miniImageNet (5-way 1-shot)}
        \label{fig:sl_ssl_by_similarity_mini_1shot}
    \end{subfigure}
    \begin{subfigure}[h]{.49\linewidth}
        \centering
        \includegraphics[width=\linewidth]{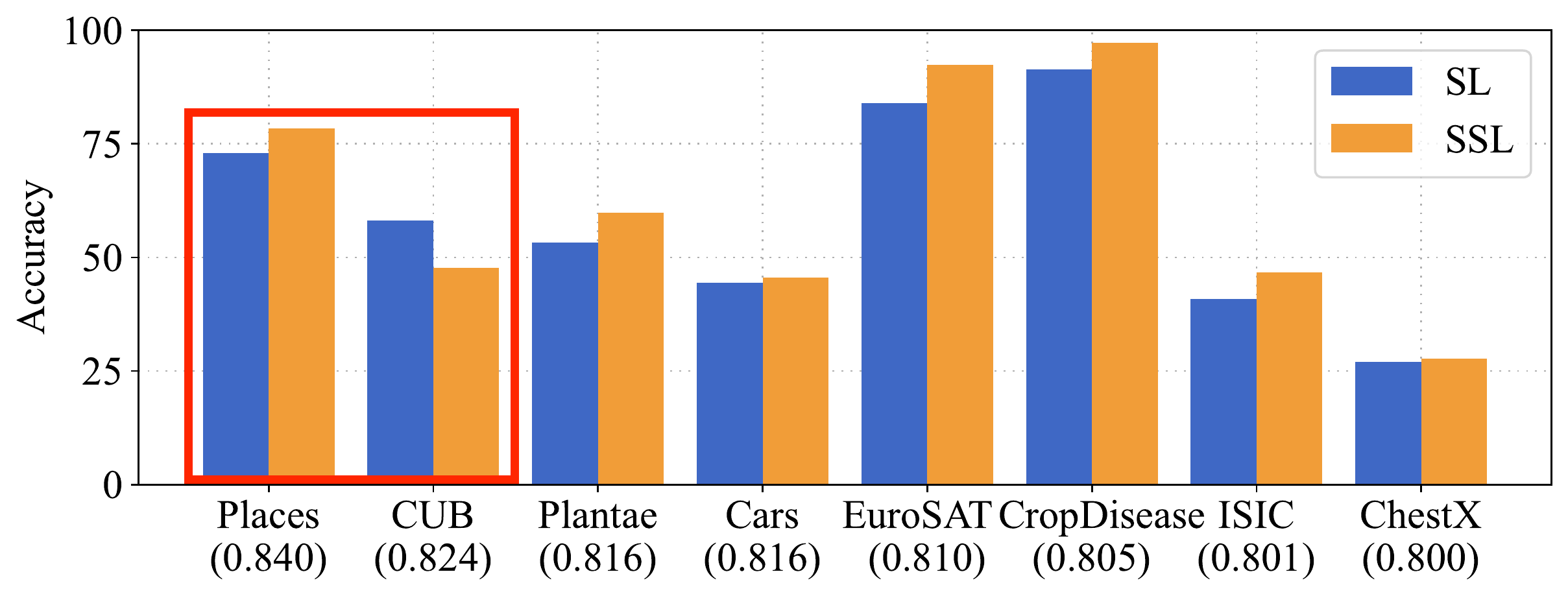}
        \vspace*{-0.6cm}
        \caption{miniImageNet (5-way 5-shot)}
        \label{fig:sl_ssl_by_similarity_mini_5shot}
    \end{subfigure}

     \vspace*{-0.1cm}
     \caption{$5$-way $k$-shot CD-FSL performance\,(\%) of SL and SSL according to domain similarity. Target datasets are shown in order of domain similarity\,(values in x-axis) to ImageNet, tieredImageNet and miniImageNet, respectively. For SSL, SimCLR is used for all datasets except ChestX, for which BYOL is used.
     }
     \label{fig:sl_ssl_by_similarity_all}
\end{figure}

Figure \ref{fig:sl_ssl_by_similarity_all} shows the performance of SL and SSL for three source datasets and eight target datasets, according to domain similarity. Across all source datasets, we consistently find that domain similarity alone is not sufficient to explain the relative performance of SL, compared to SSL. As mentioned in Section \ref{sec:analysis2}, we observe that SSL can outperform SL even when domain similarity is large, as highlighted by the difference between Places and CUB shown in Figures \ref{fig:sl_ssl_by_similarity_all}(a,b) for ImageNet, which are identical to Figures \ref{fig:sim_miniIN}(a,b) in the main paper. Similar observations are made between EuroSAT and CUB for tieredImageNet in Figures \ref{fig:sl_ssl_by_similarity_all}(c,d), and between Places and CUB for miniImageNet in Figures \ref{fig:sl_ssl_by_similarity_all}(e,f).

\clearpage
\subsection{When Does Performance Gain of SSL over SL Become Greater? (Observation \ref{obs:obs5_3})} 
\begin{figure}[h]
    \centering
    \begin{subfigure}[h]{.49\linewidth}
         \centering
         \includegraphics[width=\linewidth]{figure/sl_ssl_by_difficulty/sl_ssl_by_difficulty_imagenet_bscd_k05_kshot.pdf}
         \vspace*{-0.6cm}
         \caption{Small Similarity (ImageNet)}
         \label{fig:sl_ssl_by_difficulty_imagenet_bscd}
    \end{subfigure}
    \begin{subfigure}[h]{.49\linewidth}
         \centering
         \includegraphics[width=\linewidth]{figure/sl_ssl_by_difficulty/sl_ssl_by_difficulty_imagenet_other_k05_kshot.pdf}
         \vspace*{-0.6cm}
         \caption{Large Similarity (ImageNet)}
         \label{fig:sl_ssl_by_difficulty_imagenet_other}
    \end{subfigure}
    
    \begin{subfigure}[h]{.49\linewidth}
         \centering
         \includegraphics[width=\linewidth]{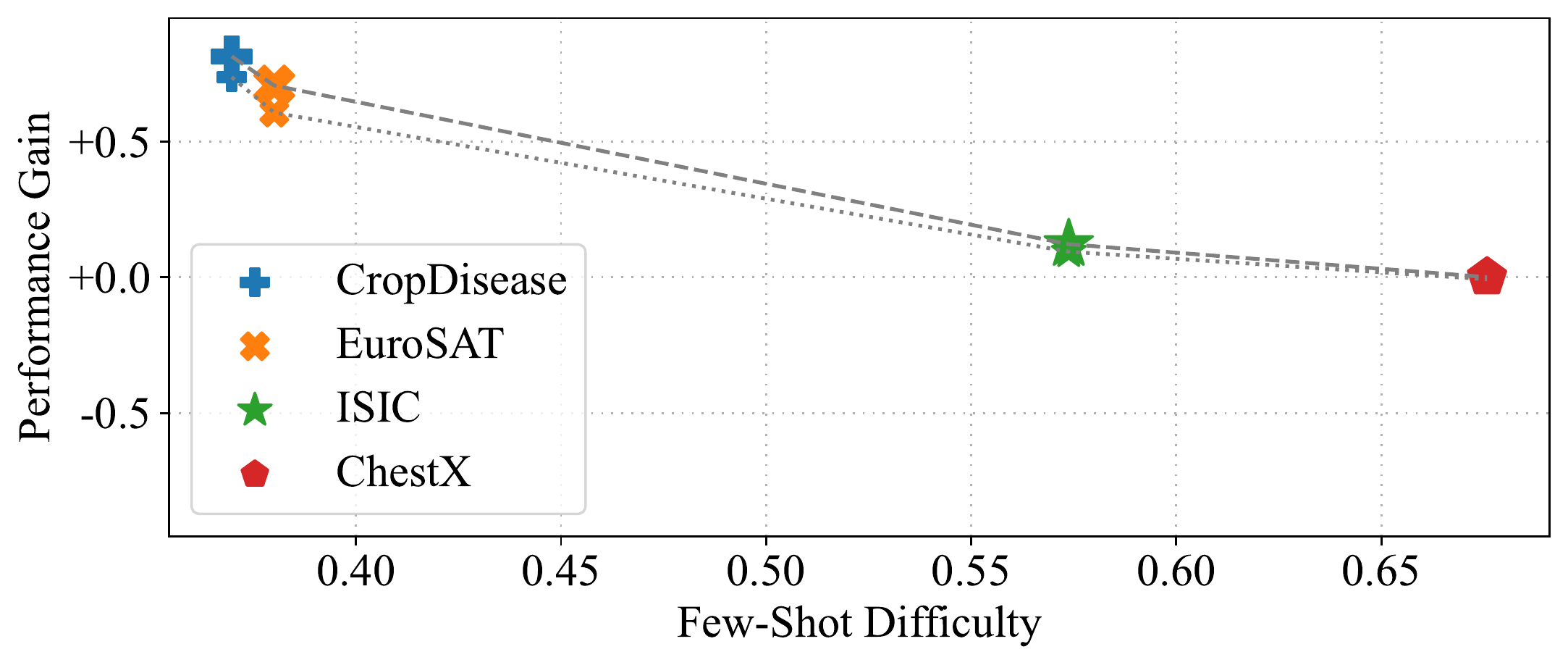}
         \vspace*{-0.6cm}
         \caption{Small Similarity (tieredImageNet)}
         \label{fig:sl_ssl_by_difficulty_tiered_bscd}
    \end{subfigure}
    \begin{subfigure}[h]{.49\linewidth}
         \centering
         \includegraphics[width=\linewidth]{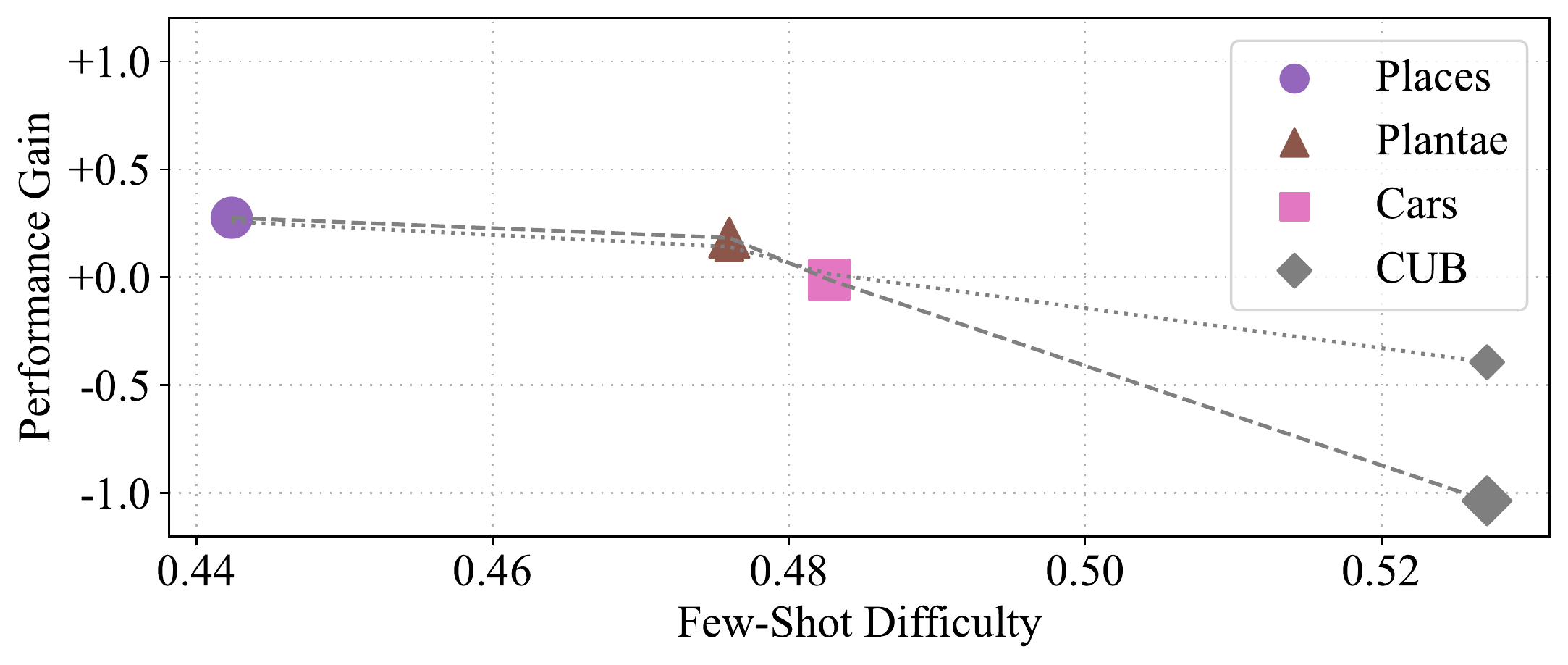}
         \vspace*{-0.6cm}
         \caption{Large Similarity (tieredImageNet)}
         \label{fig:sl_ssl_by_difficulty_tiered_other}
    \end{subfigure}
    
    \begin{subfigure}[h]{.49\linewidth}
         \centering
         \includegraphics[width=\linewidth]{figure/sl_ssl_by_difficulty/sl_ssl_by_difficulty_mini_bscd_k05_kshot.pdf}
         \vspace*{-0.6cm}
         \caption{Small Similarity (miniImageNet)}
         \label{fig:sl_ssl_by_difficulty_mini_bscd}
    \end{subfigure}
    \begin{subfigure}[h]{.49\linewidth}
         \centering
         \includegraphics[width=\linewidth]{figure/sl_ssl_by_difficulty/sl_ssl_by_difficulty_mini_other_k05_kshot.pdf}
         \vspace*{-0.6cm}
         \caption{Large Similarity (miniImageNet)}
         \label{fig:sl_ssl_by_difficulty_mini_other}
    \end{subfigure}
     
     \vspace*{-0.1cm}
     \caption{5-way $k$-shot performance gains of SSL over SL for the two dataset groups according to the few-shot difficulty (small: $k$=1, large: $k$=5). Results are shown for three source datasets: ImageNet, tieredImageNet, and miniImageNet, each with their corresponding backbones. SimCLR is used for SSL in all target datasets except ChestX, for which BYOL is used.
     }
     \label{fig:sl_ssl_by_difficulty_all}
\end{figure}

Figure \ref{fig:sl_ssl_by_difficulty_all} shows the performance gain of SSL over SL for three source datasets and eight target datasets, according to few-shot difficulty, for two groups with different levels of domain similarity.
Again, the identical observation is made for all three source datasets. When comparing the two groups (BSCD-FSL vs. others), larger performance gain is observed for the small domain similarity group (BSCD-FSL), compared to the latter (others). Within each group, the performance gain of SSL over SL increases with lower few-shot difficulty.

In addition, comparing between different source datasets, for target datasets with large similarity (Figure \ref{fig:sl_ssl_by_difficulty_all}(b,d,f)), the performance gain of SSL over SL decreases by larger source dataset size. For example, on the CUB dataset, the performance gain (for $k=5$) is $-0.249$, $-1.035$, and $-2.276$ for miniImageNet, tieredImageNet, and ImageNet, respectively. However, for target datasets with small similarity (Figure \ref{fig:sl_ssl_by_difficulty_all}(a,c,e)), the performance gain of SSL over SL does not have a consistent trend according to the source dataset size.

\clearpage
\section{Hyperparameter $\gamma$ in MSL Pre-Training}\label{appx:gamma_abla}

\subsection{Choice of Hyperparameter $\gamma$}
\begin{figure}[h]
     \centering
     \includegraphics[width=.80\linewidth]{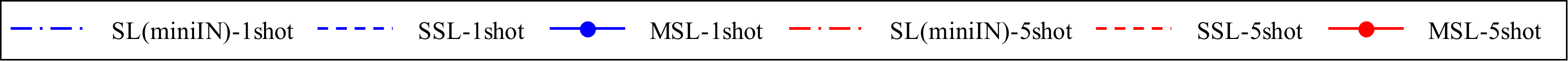}
     \vspace*{3pt}
     
     \begin{subfigure}[h]{0.24\linewidth}
         \centering
         \includegraphics[width=\linewidth]{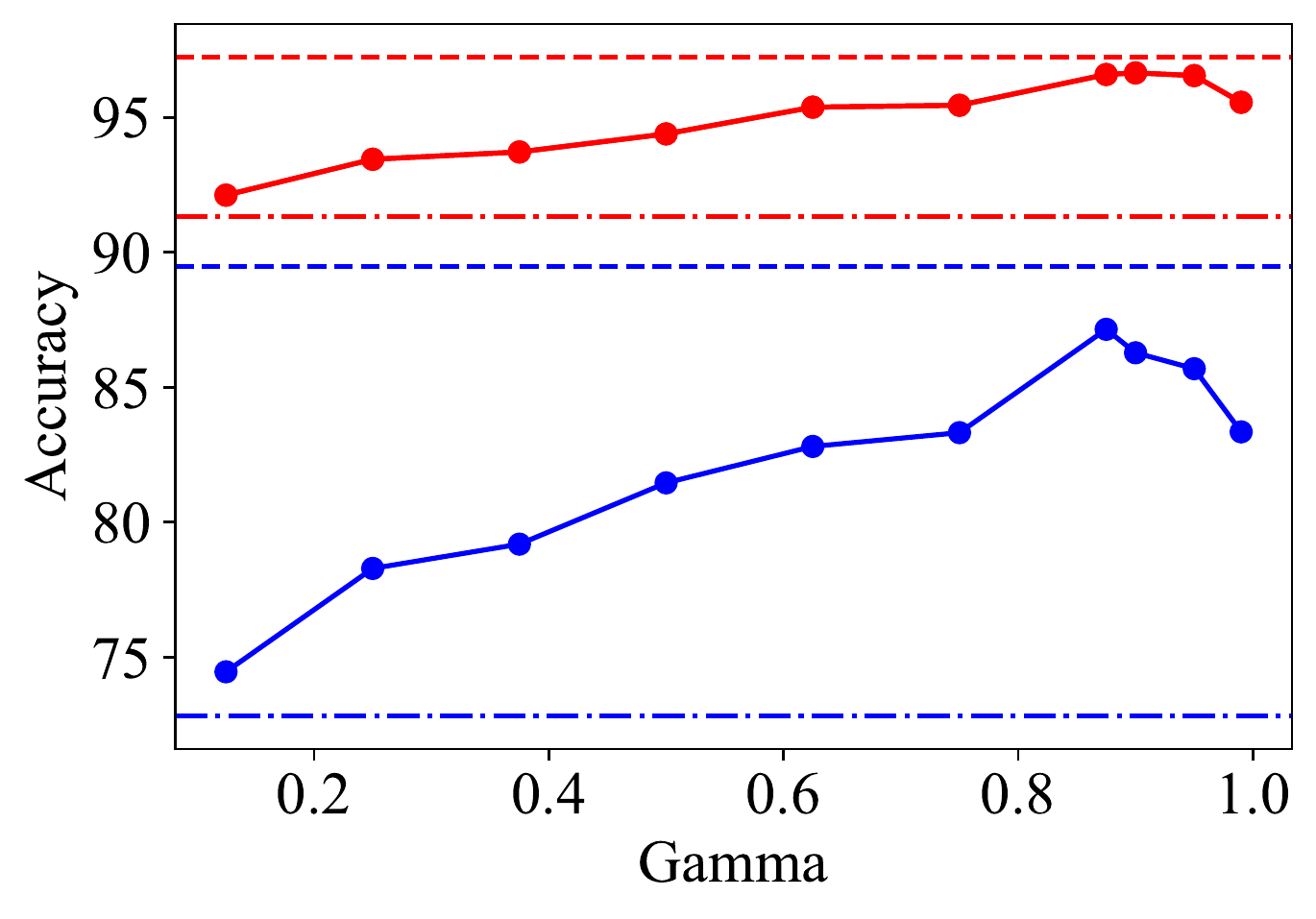}
         \caption{CropDisease}
         \label{fig:hdmtl_mini_crop}
     \end{subfigure}
     \hfill
     \begin{subfigure}[h]{0.24\linewidth}
         \centering
         \includegraphics[width=\linewidth]{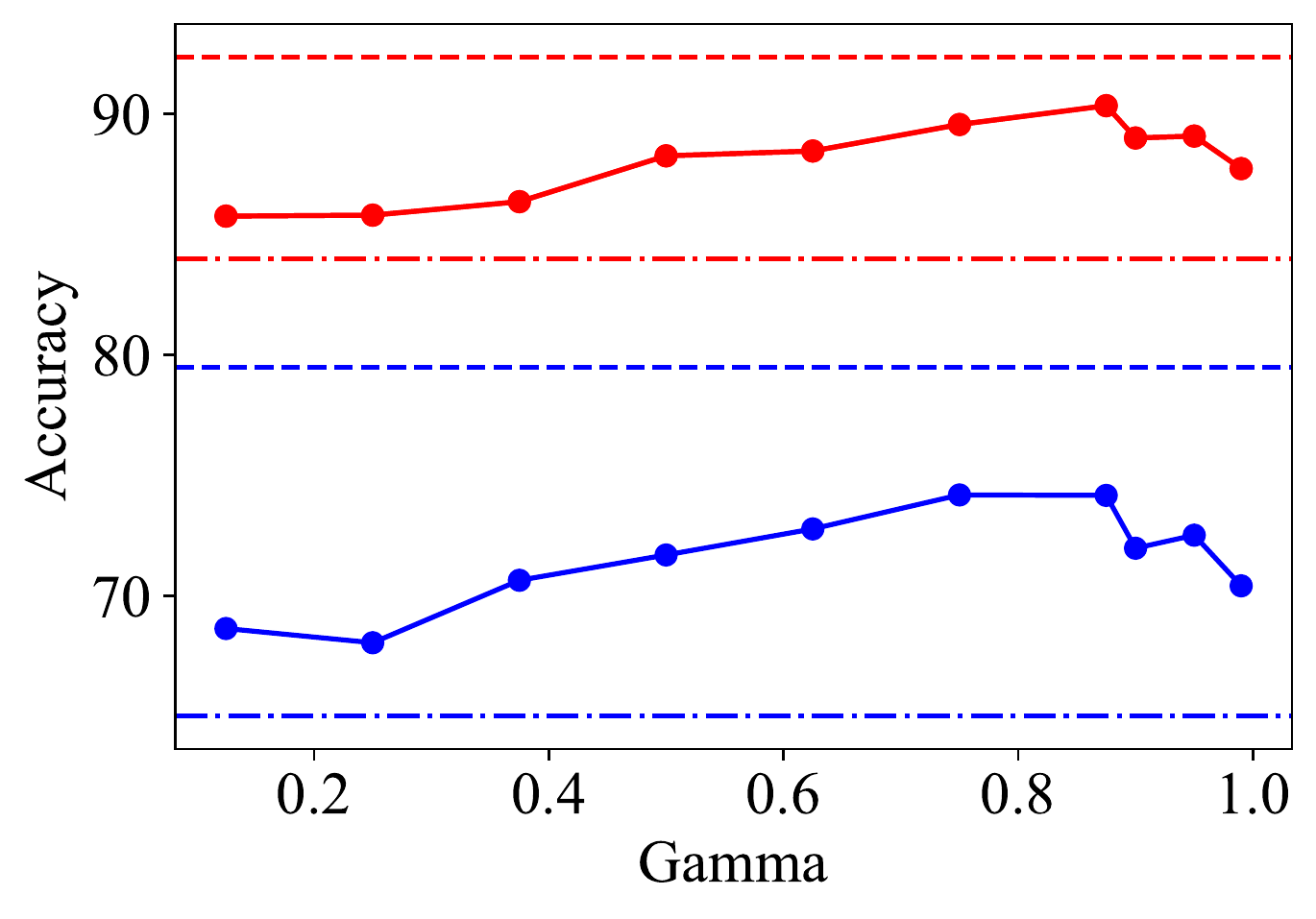}
         \caption{EuroSAT}
         \label{fig:hdmtl_mini_euro}
     \end{subfigure}
     \hfill
     \begin{subfigure}[h]{0.24\linewidth}
         \centering
         \includegraphics[width=\linewidth]{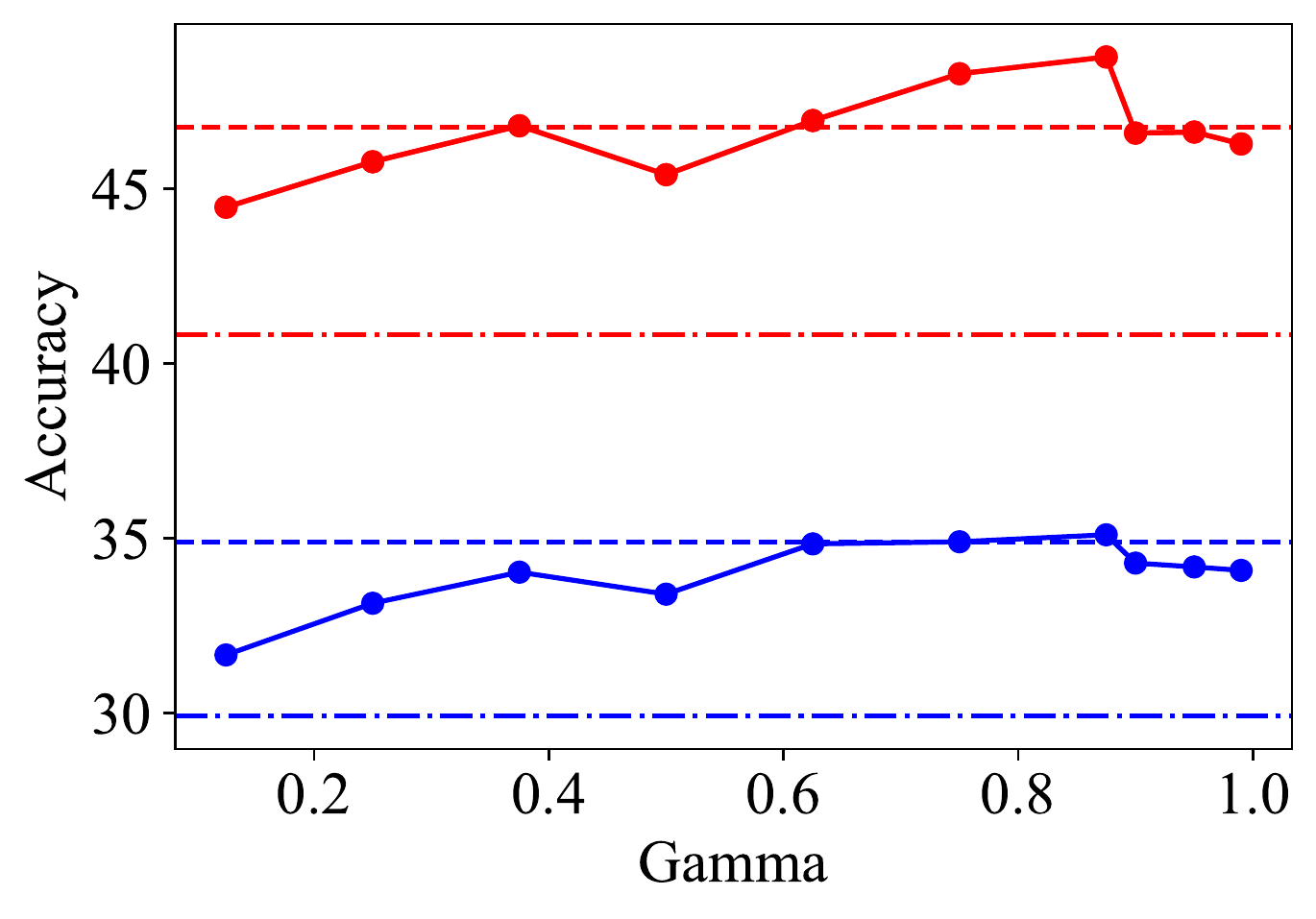}
         \caption{ISIC}
         \label{fig:hdmtl_mini_isic}
     \end{subfigure}
     \hfill
     \begin{subfigure}[h]{0.24\linewidth}
         \centering
         \includegraphics[width=\linewidth]{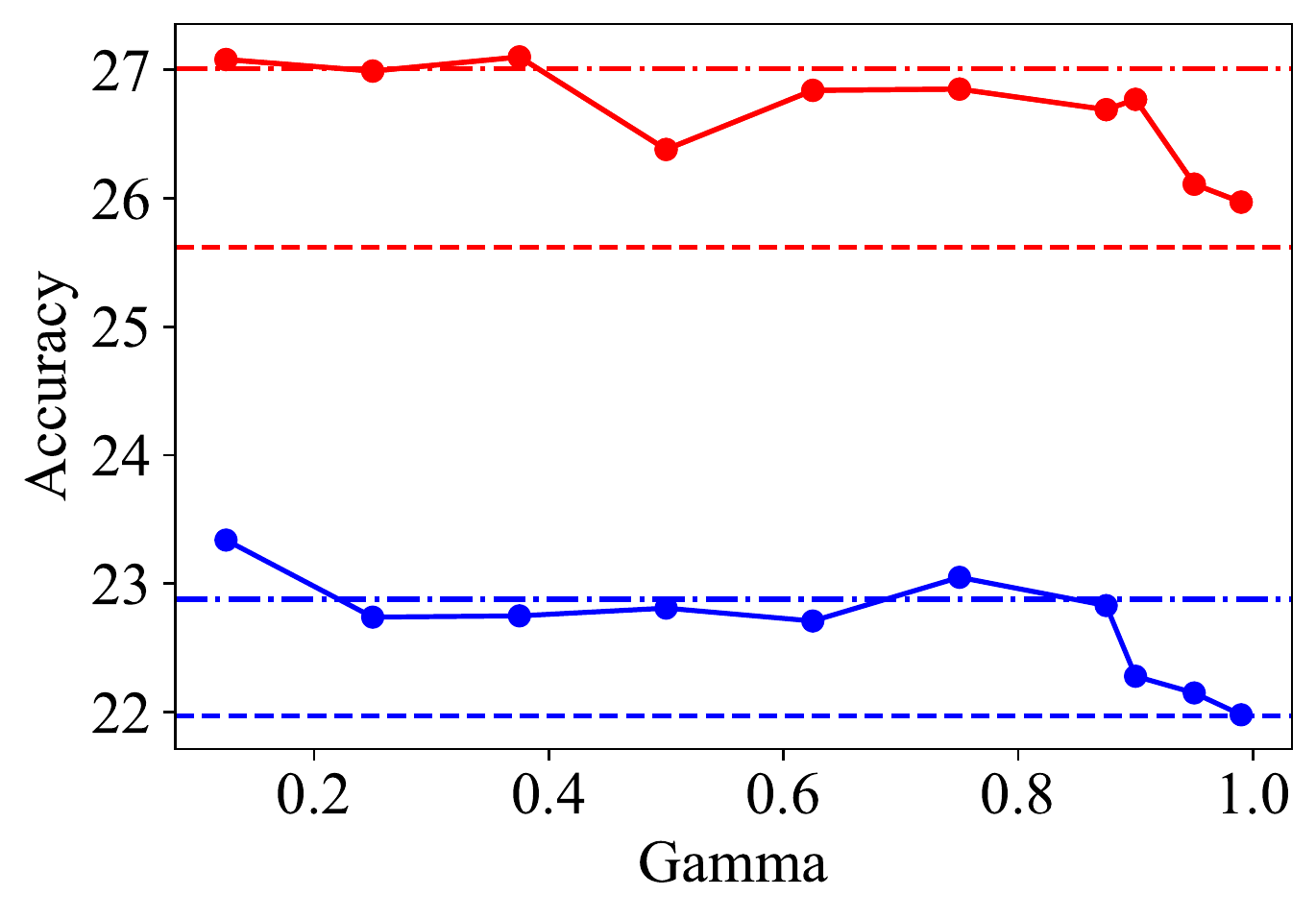}
         \caption{ChestX}
         \label{fig:hdmtl_mini_chest}
     \end{subfigure}
     
     \begin{subfigure}[h]{0.24\linewidth}
         \centering
         \includegraphics[width=\linewidth]{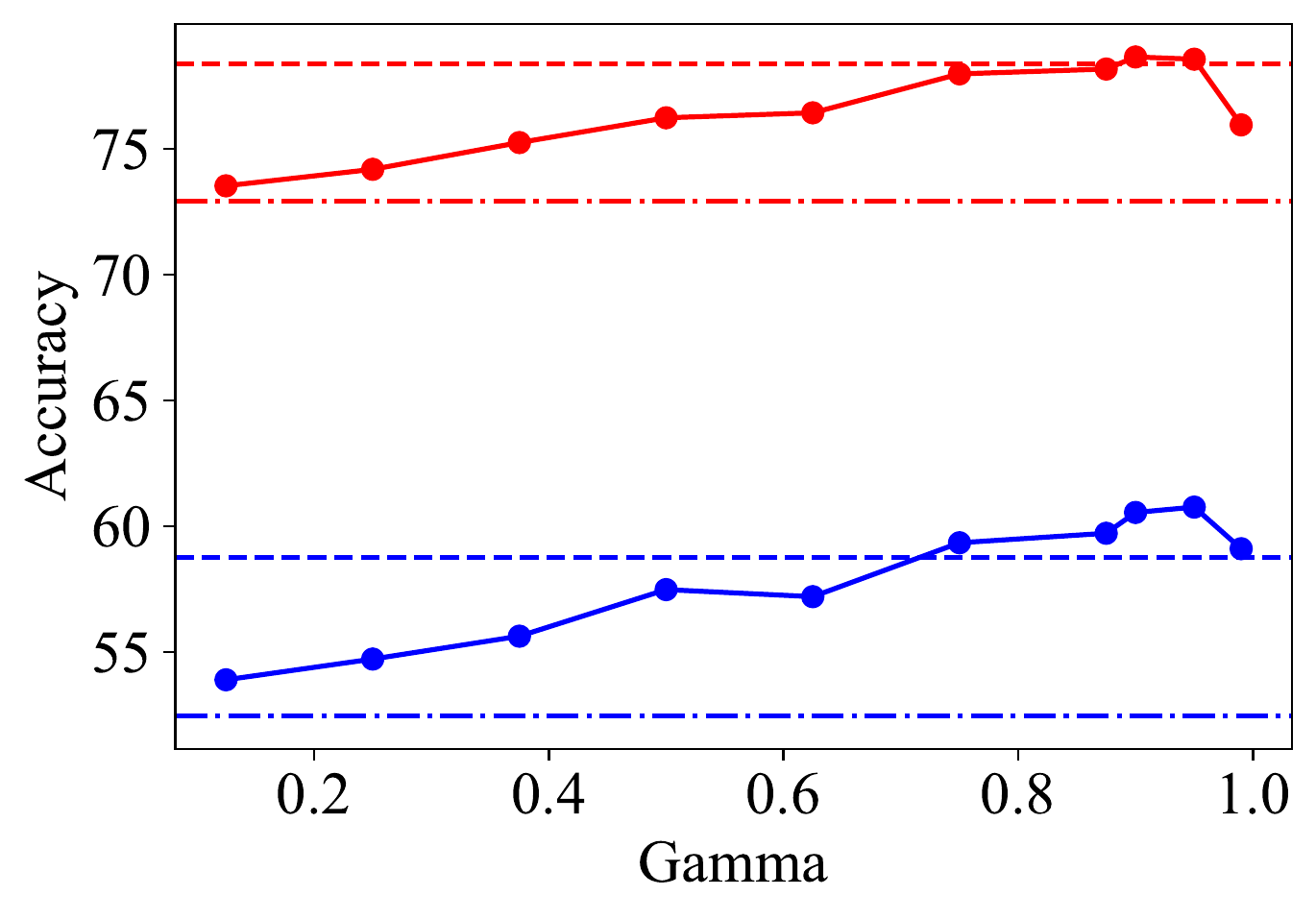}
         \caption{Places}
         \label{fig:hdmtl_mini_places}
     \end{subfigure}
     \hfill
     \begin{subfigure}[h]{0.24\linewidth}
         \centering
         \includegraphics[width=\linewidth]{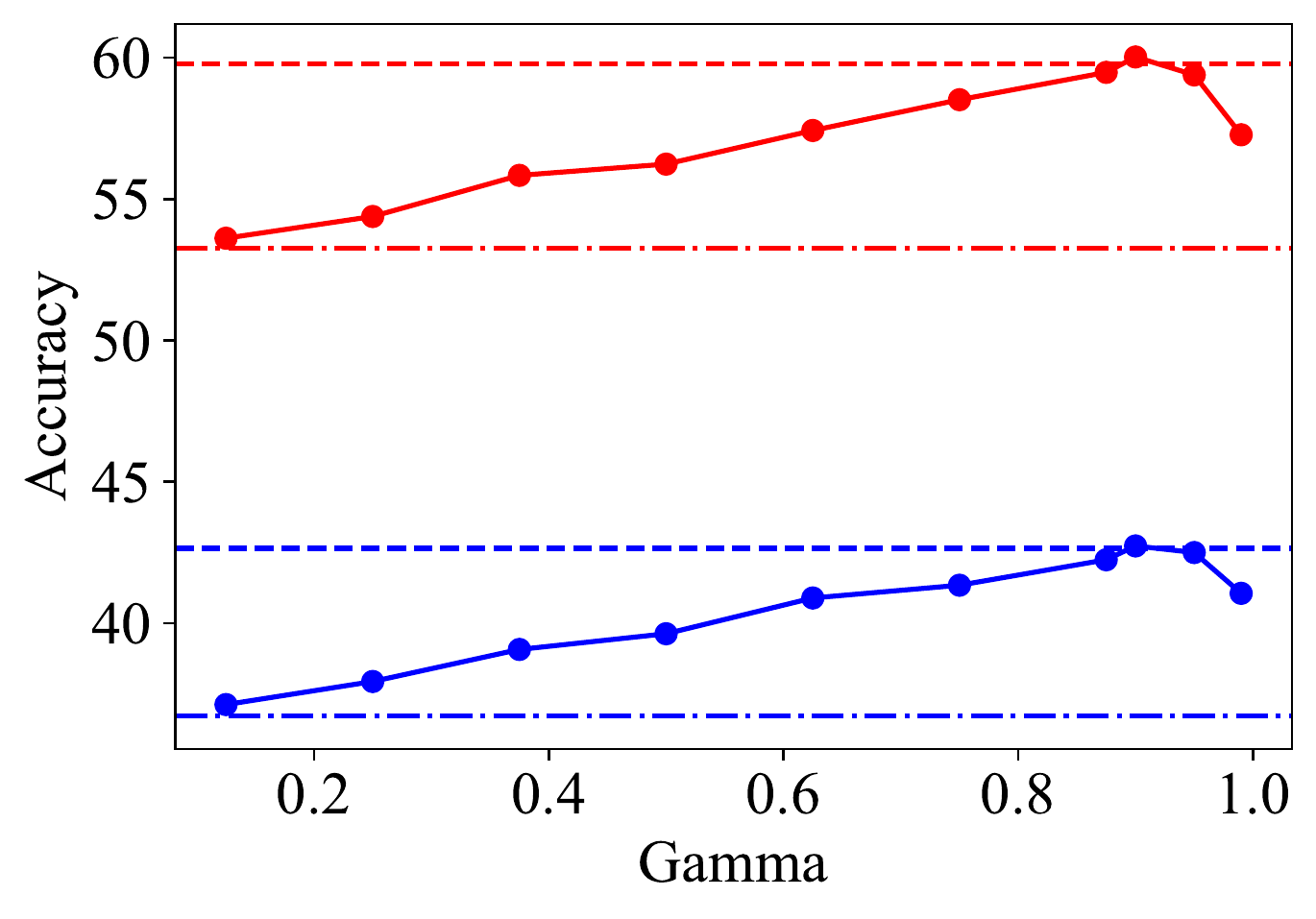}
         \caption{Plantae}
         \label{fig:hdmtl_mini_plantae}
     \end{subfigure}
     \hfill
     \begin{subfigure}[h]{0.24\linewidth}
         \centering
         \includegraphics[width=\linewidth]{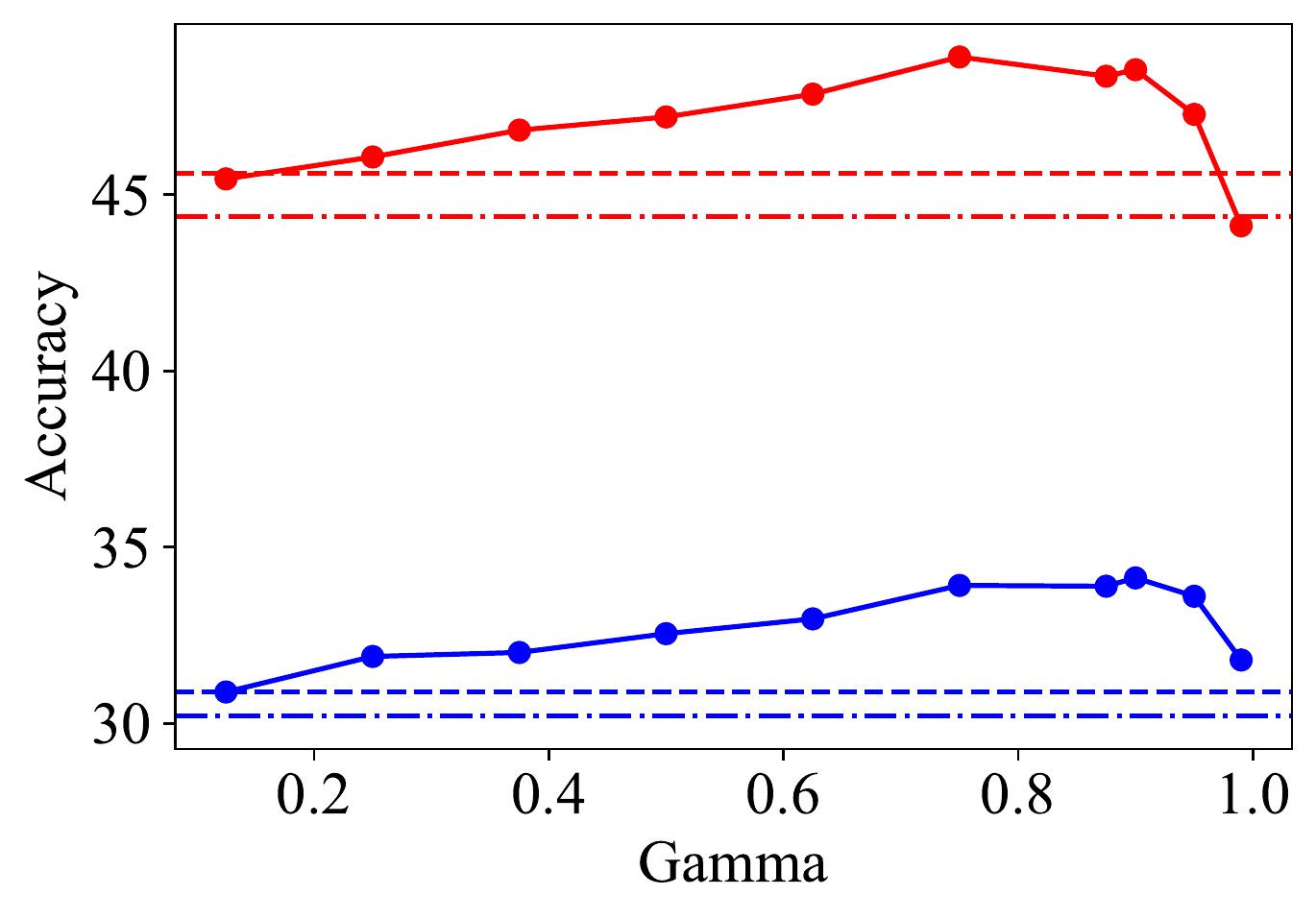}
         \caption{Cars}
         \label{fig:hdmtl_mini_cars}
     \end{subfigure}
     \hfill
     \begin{subfigure}[h]{0.24\linewidth}
         \centering
         \includegraphics[width=\linewidth]{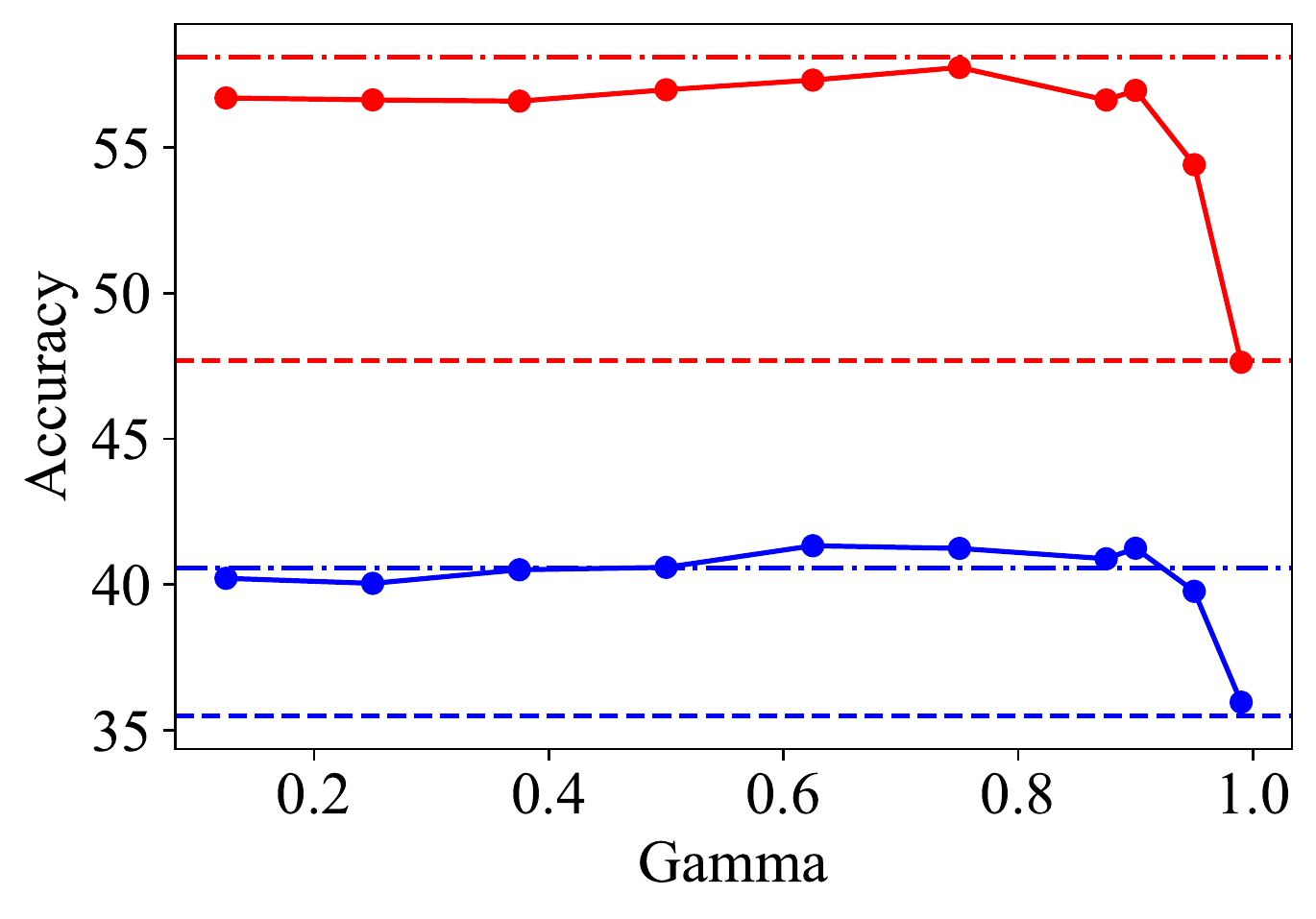}
         \caption{CUB}
         \label{fig:hdmtl_mini_cub}
     \end{subfigure}
     \vspace*{-3pt}
     \caption{5-way $k$-shot performance of MSL according to the balancing weight (i.e., $\gamma$) between SL and SSL (Section \ref{sec:analysis3}). ResNet10 is used as a backbone. SimCLR is used for the MSL and SSL method.}
     \label{fig:resnet10_simclr_gamma}
\end{figure}

\begin{figure}[h]
     \centering
     \includegraphics[width=.80\linewidth]{figure/legend2.pdf}
     \vspace*{3pt}

     \begin{subfigure}[h]{0.24\linewidth}
         \centering
         \includegraphics[width=\linewidth]{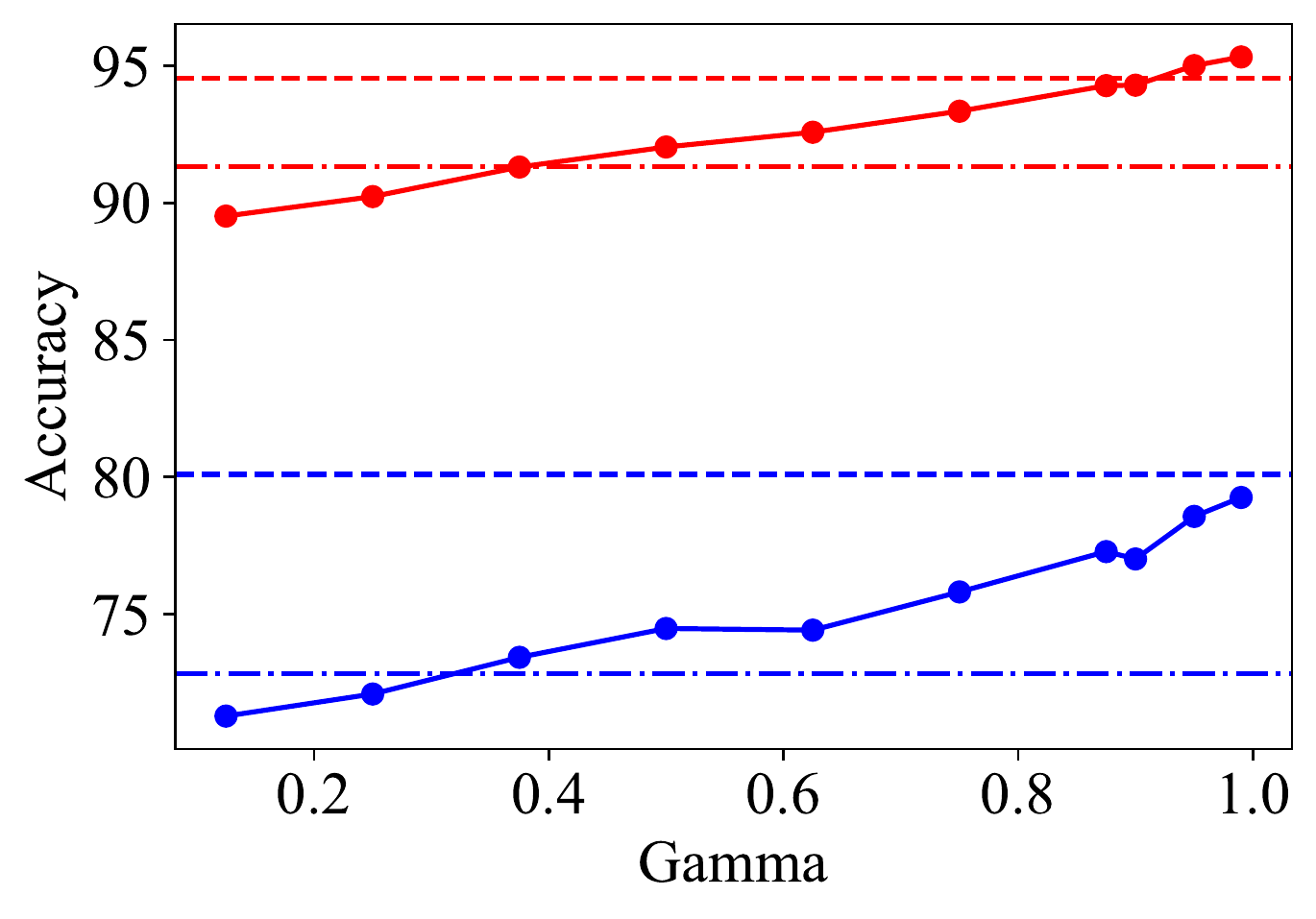}
         \caption{CropDisease}
         \label{fig:msl_mini_byol_crop}
     \end{subfigure}
     \hfill
     \begin{subfigure}[h]{0.24\linewidth}
         \centering
         \includegraphics[width=\linewidth]{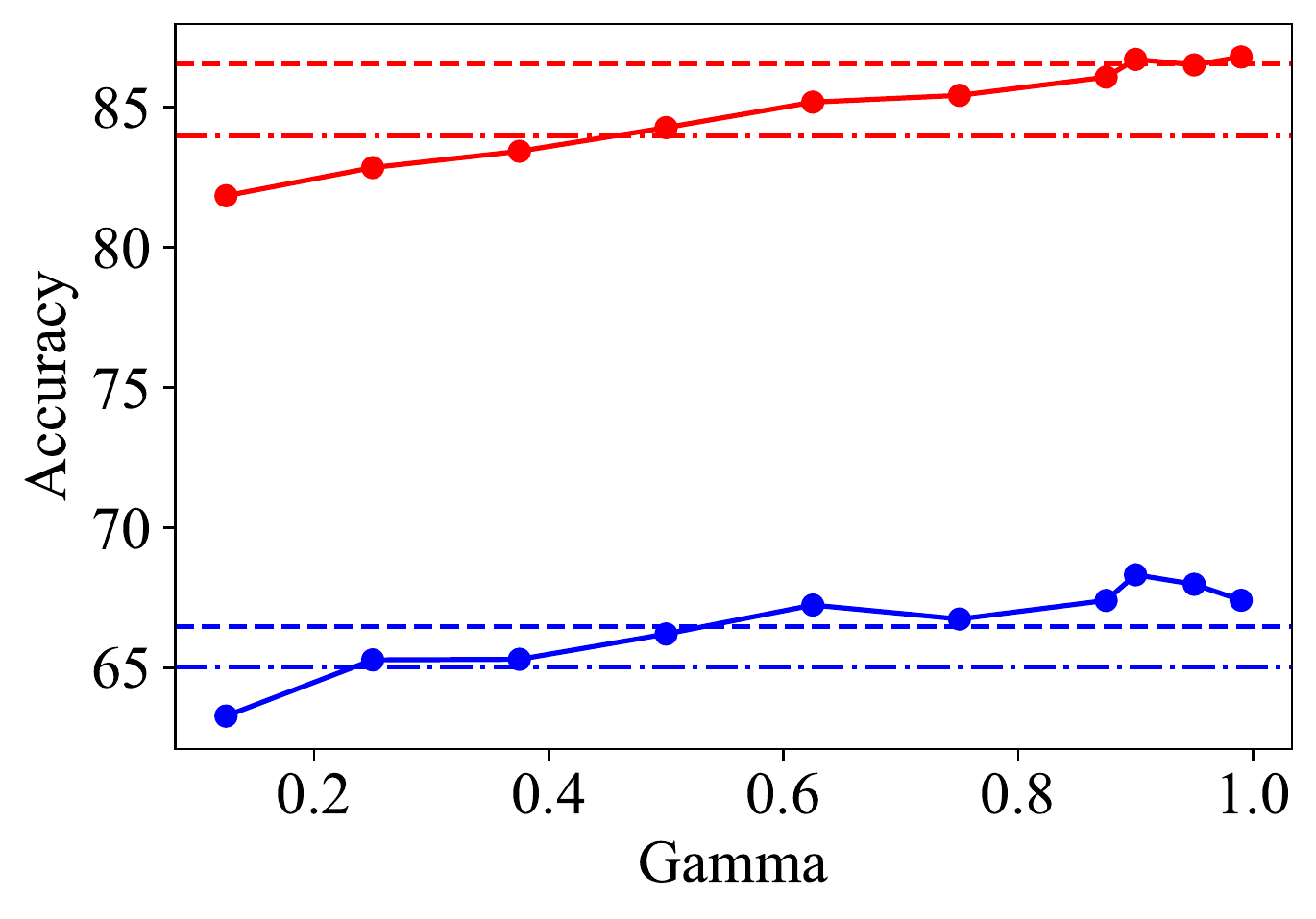}
         \caption{EuroSAT}
         \label{fig:msl_mini_byol_euro}
     \end{subfigure}
     \hfill
     \begin{subfigure}[h]{0.24\linewidth}
         \centering
         \includegraphics[width=\linewidth]{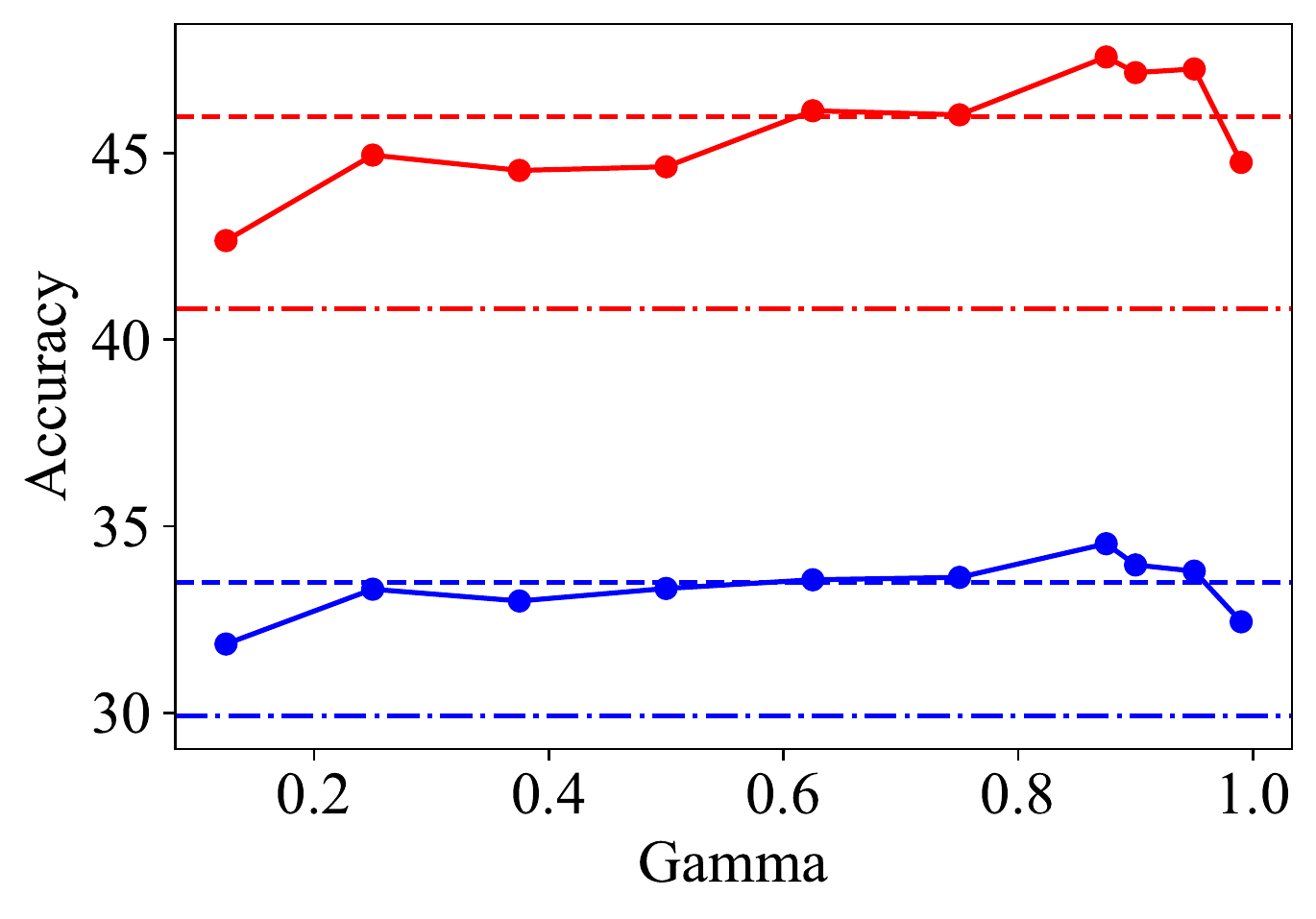}
         \caption{ISIC}
         \label{fig:msl_mini_byol_isic}
     \end{subfigure}
     \hfill
     \begin{subfigure}[h]{0.24\linewidth}
         \centering
         \includegraphics[width=\linewidth]{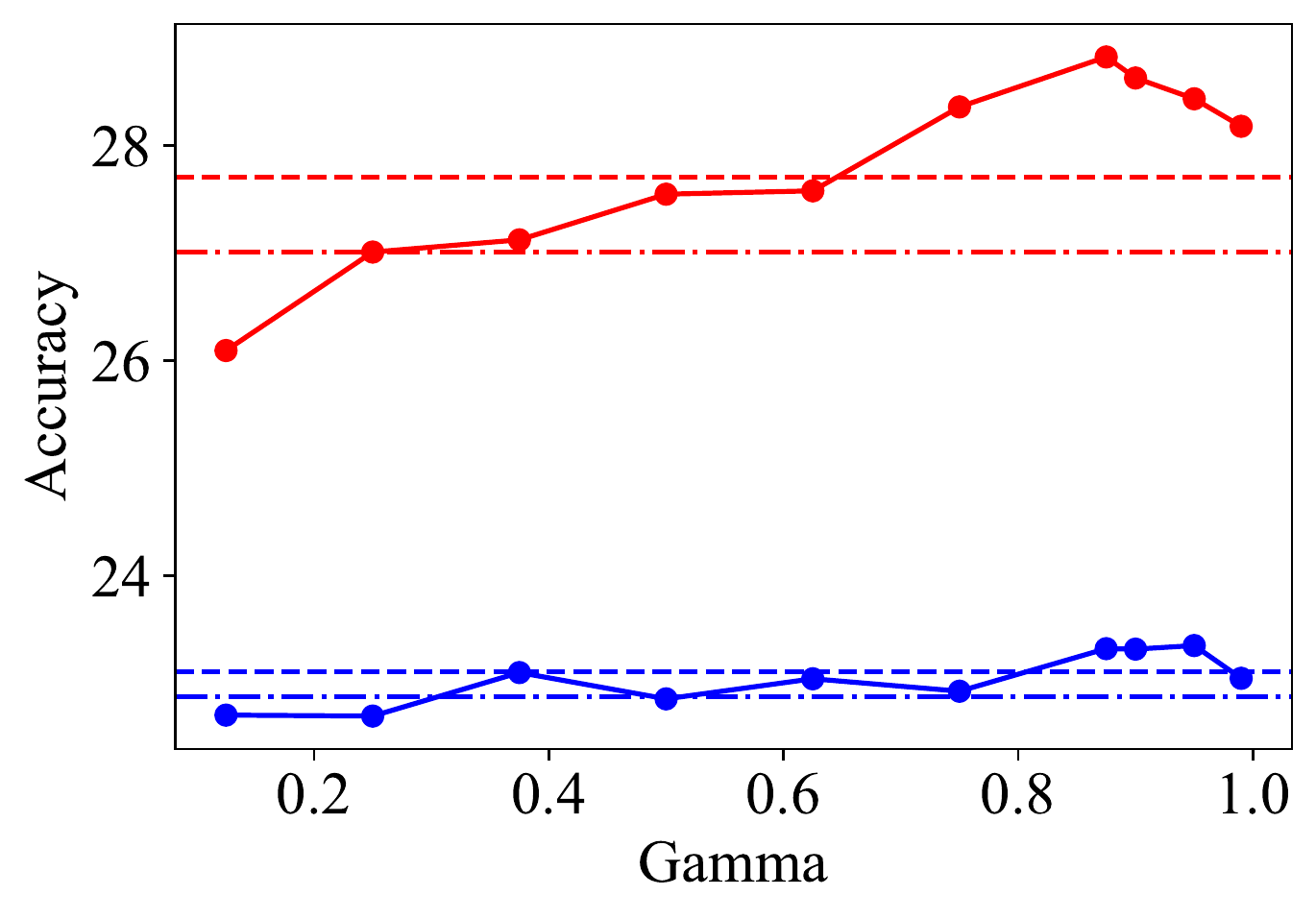}
         \caption{ChestX}
         \label{fig:msl_mini_byol_chest}
     \end{subfigure}
     
     \begin{subfigure}[h]{0.24\linewidth}
         \centering
         \includegraphics[width=\linewidth]{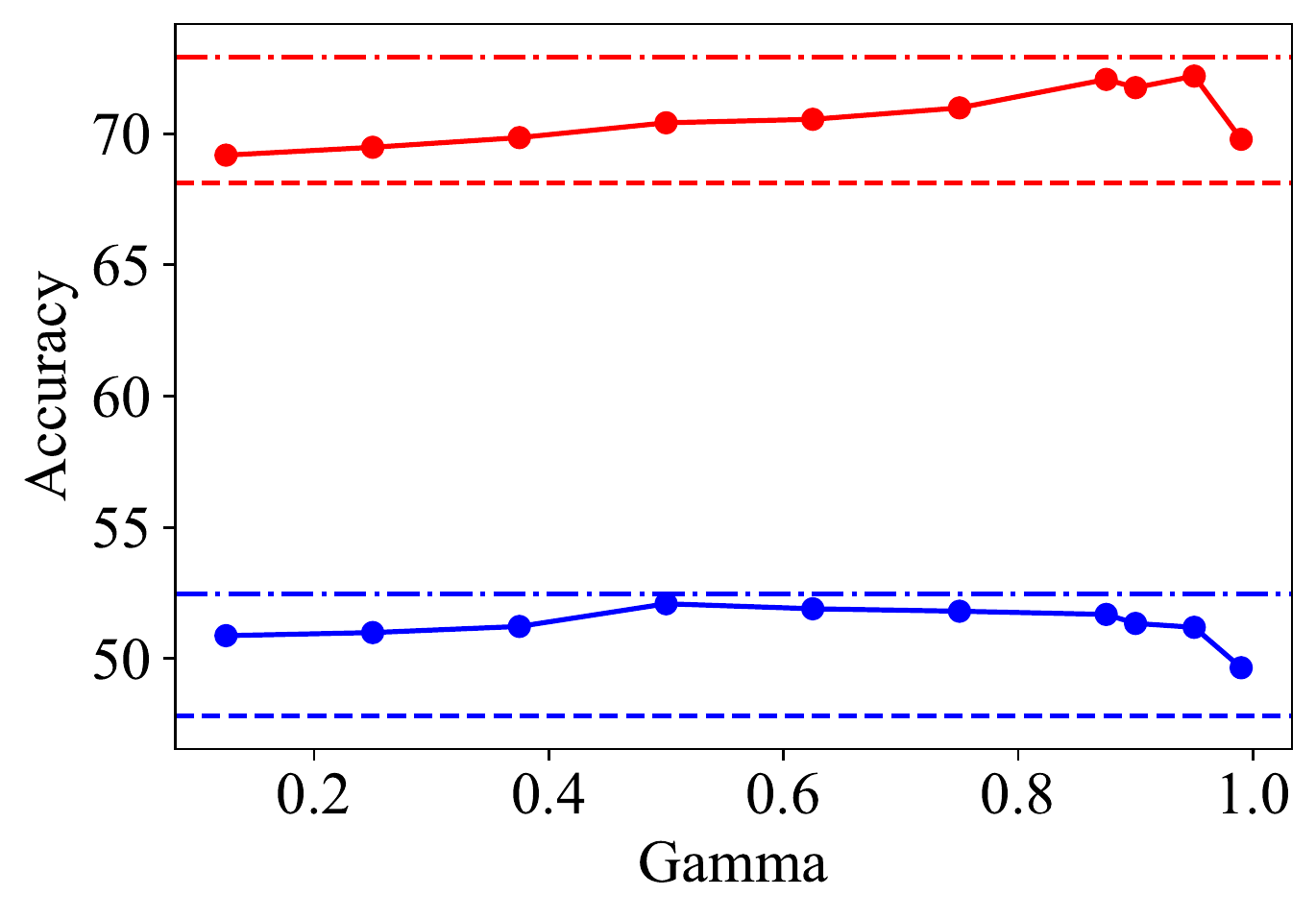}
         \caption{Places}
         \label{fig:msl_mini_byol_places}
     \end{subfigure}
     \hfill
     \begin{subfigure}[h]{0.24\linewidth}
         \centering
         \includegraphics[width=\linewidth]{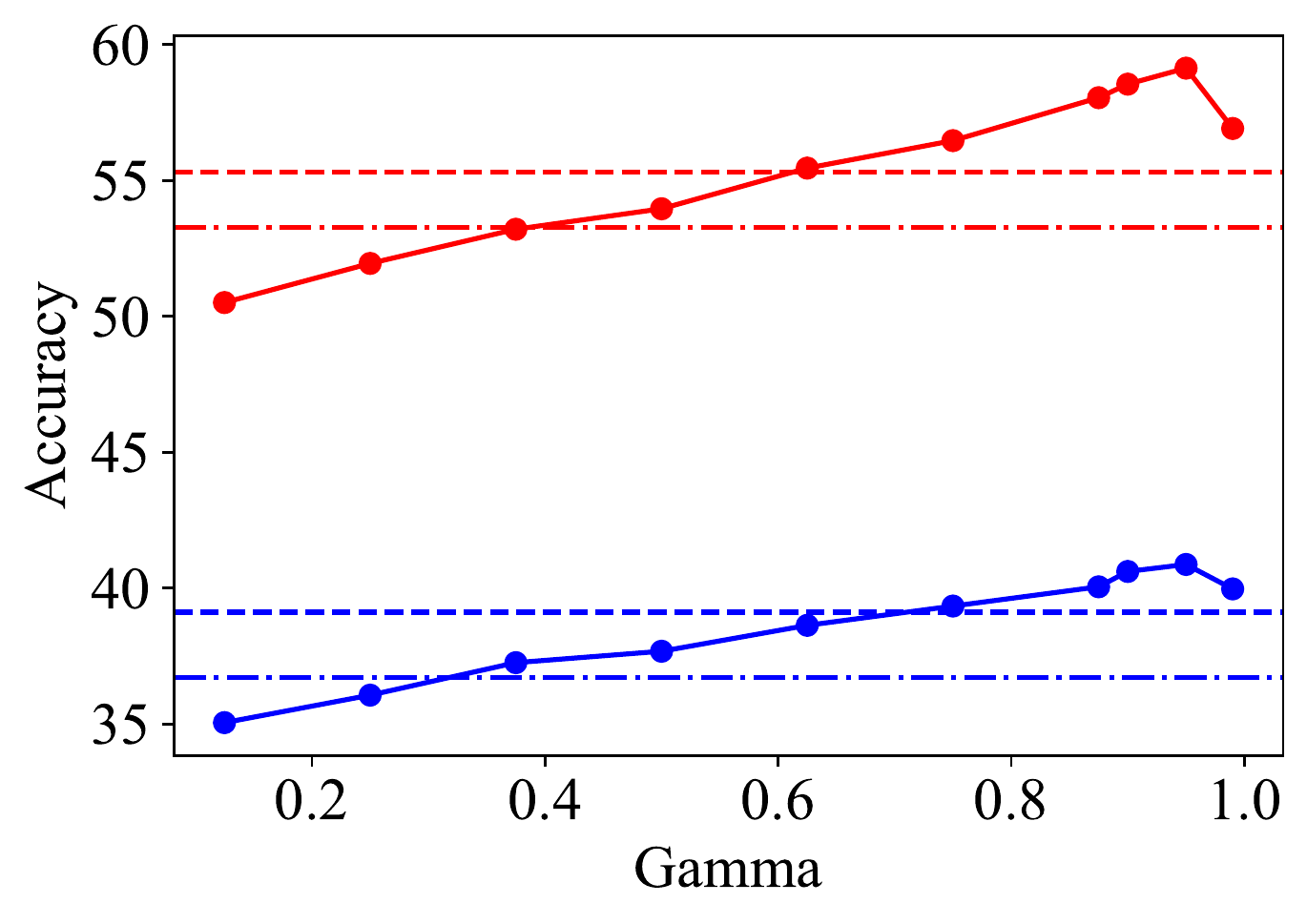}
         \caption{Plantae}
         \label{fig:msl_mini_byol_plantae}
     \end{subfigure}
     \hfill
     \begin{subfigure}[h]{0.24\linewidth}
         \centering
         \includegraphics[width=\linewidth]{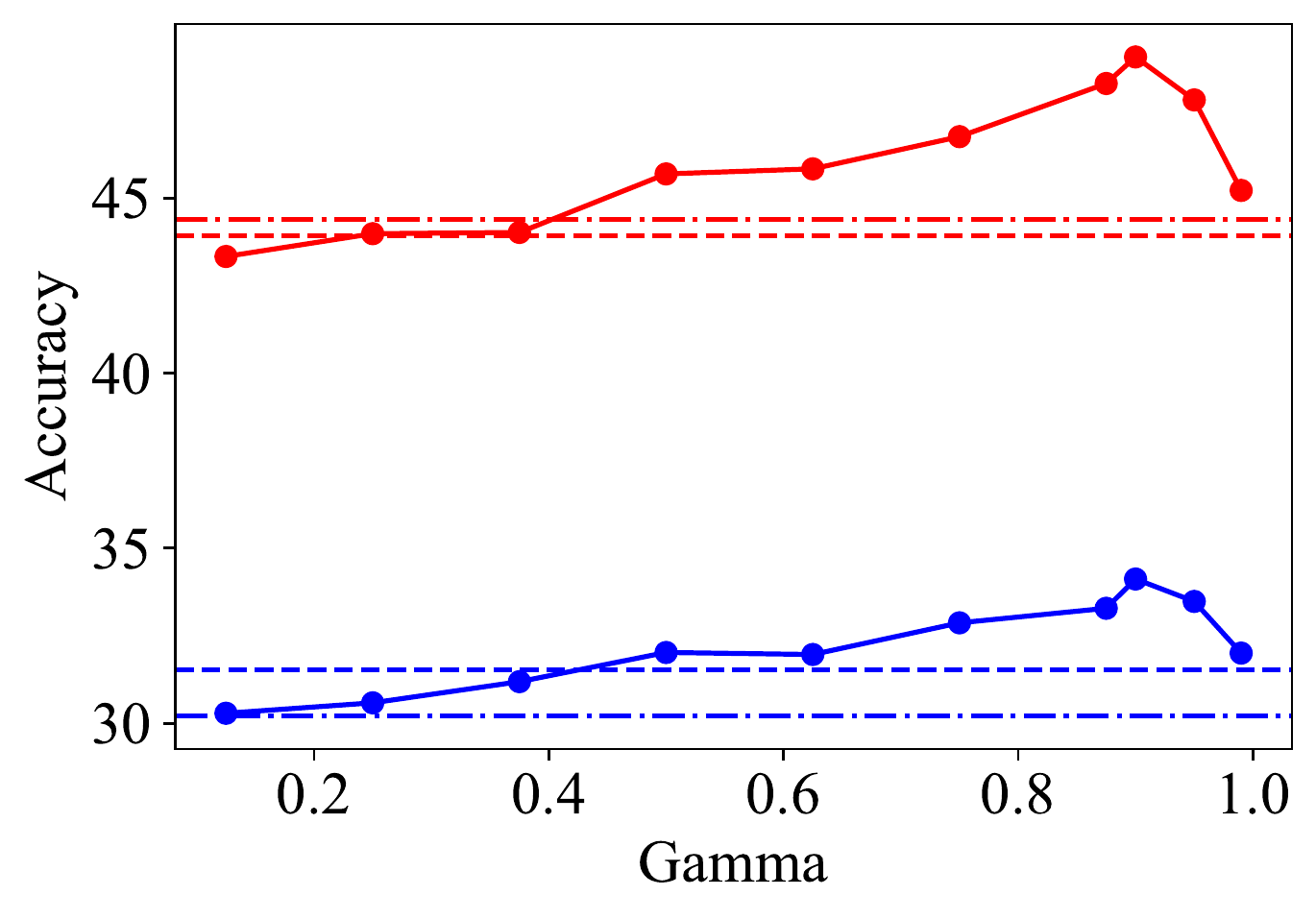}
         \caption{Cars}
         \label{fig:msl_mini_byol_cars}
     \end{subfigure}
     \hfill
     \begin{subfigure}[h]{0.24\linewidth}
         \centering
         \includegraphics[width=\linewidth]{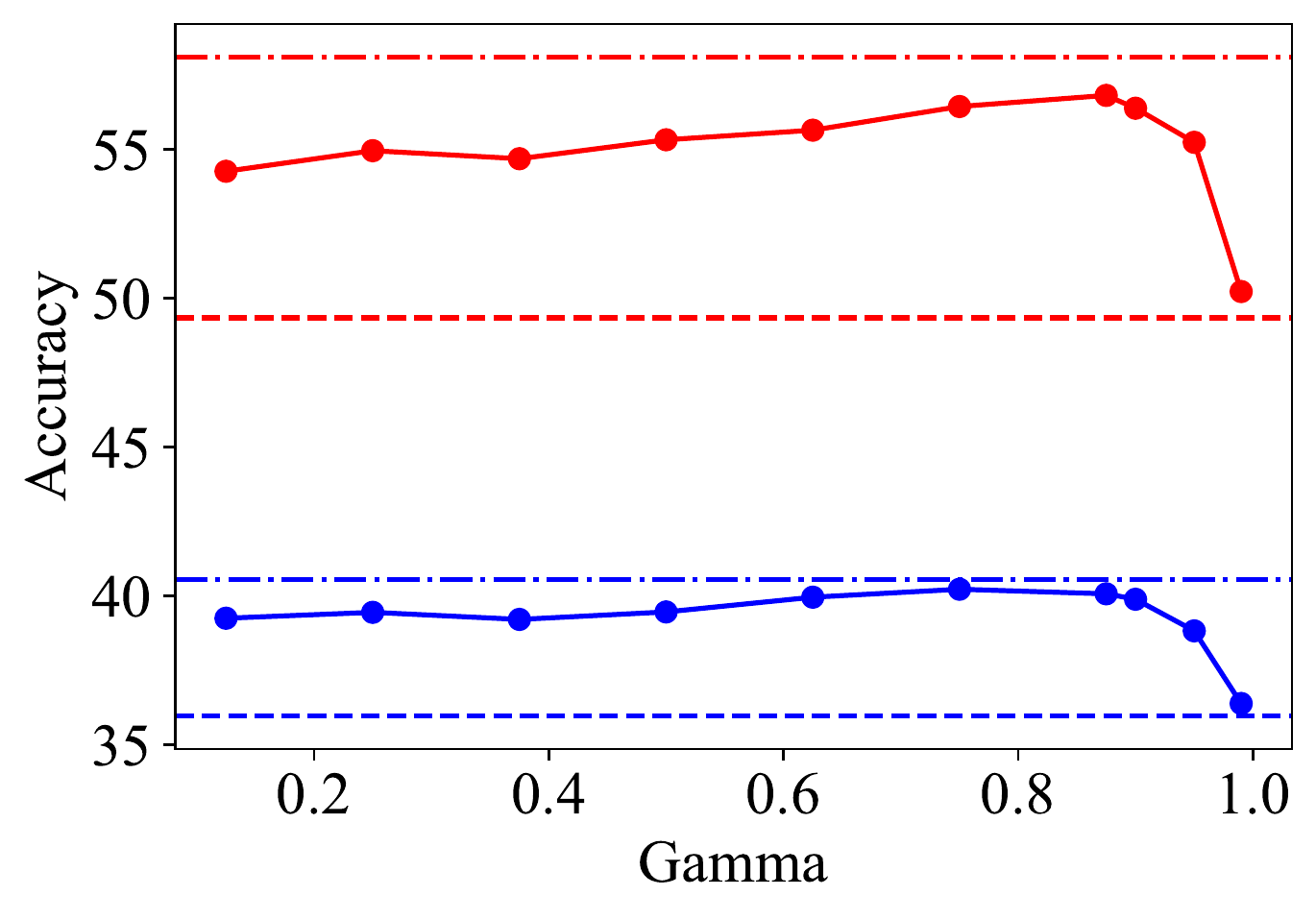}
         \caption{CUB}
         \label{fig:msl_mini_byol_cub}
     \end{subfigure}
     \vspace*{-3pt}
     \caption{5-way $k$-shot performance of MSL according to the balancing weight (i.e., $\gamma$) between SL and SSL (Section \ref{sec:analysis3}). ResNet10 is used as a backbone. BYOL is used for the MSL and SSL method.}
     \label{fig:resnet10_byol_gamma}
\end{figure}

One important hyperparameter in MSL is a balancing weight $\gamma$ (refer to Eq. \eqref{eq:loss_msl}). We investigated how we should choose $\gamma$ value. Figure \ref{fig:resnet10_simclr_gamma} and Figure \ref{fig:resnet10_byol_gamma} describe the few-shot performance of MSL according to the balancing weight $\gamma$ between SL and SSL when SimCLR or BYOL are used for SSL, respectively. In Figure \ref{fig:resnet10_simclr_gamma}, MSL performance (circle-marked solid lines) generally improves as $\gamma$ increases from 0.125 to 0.875, i.e., the weight for SSL is getting larger, except for ChestX. In Section \ref{sec:analysis1}, we found that non-contrastive SSL method nicely worked on ChestX. Figure \ref{fig:resnet10_byol_gamma} shows that MSL with BYOL loss guarantees good performance on ChestX in $\gamma=0.875$. We further increased $\gamma$ to \{0.9, 0.95, 0.99\}, but there was an overall decreasing trend of accuracy, so we fixed $\gamma$ to 0.875 in every MSL experiment in the paper.

\newpage
\subsection{Dynamic Hyperparameter $\gamma$}

Inspired by the two-stage pre-training schemes in Section \ref{sec:analysis3}, we investigate the effects of dynamically increasing $\gamma$ during the single-stage pre-training. Specifically, we investigate a simple pre-training scheme in which $\gamma$ linearly increases from 0 to 1 over the course of 1000 epochs (\emph{i.e.,} single-stage MSL with $\gamma = 0 \nearrow 1$).

Table \ref{tab:dynamic_gamma_1shot} and Table \ref{tab:dynamic_gamma_5shot} describe the few-shot performance of the devised method (in the lowermost row), with performances of other methods displayed for ease of comparison. We observe that the performance of the devised method lies between that of standalone SL and SSL except for ChestX. The devised method typically underperforms two-stage pre-training as well as standalone MSL, indicating that it is not an effective method to exploit both SL and SSL dynamically. 

\begin{table*}[!h]
\caption{5-way 1-shot CD-FSL performance\,(\%) of the models pre-trained with varying configurations of $\gamma$ in Eq.\,\eqref{eq:loss_msl} of MSL. ResNet18 is used as the backbone model, and ImageNet is used as the source data for SL. The {best} results are marked in bold.}\label{tab:dynamic_gamma_1shot}
\vspace*{0.1cm}
\centering
\footnotesize\addtolength{\tabcolsep}{-3.0pt}
\resizebox{\linewidth}{!}{
\begin{tabular}{c|c|c|cccc|cccc}
    \toprule
    Pre-train & \,\,\multirow{2}{*}{$\gamma$} \,\, & \multirow{2}{*}{Method} & \multicolumn{4}{c|}{Small Similarity} & \multicolumn{4}{c}{Large Similarity} \\
    Scheme & & & \!\!CropDisease\!\! & \!\!EuroSAT\!\! & ISIC & ChestX & Places & Plantae & Cars & CUB \\
    \midrule
    SL & 0 & Default & 74.18{\scriptsize$\pm$.82} &  66.14{\scriptsize$\pm$.83} &  31.11{\scriptsize$\pm$.55} &  22.48{\scriptsize$\pm$.39} & 57.47{\scriptsize$\pm$.86} &  43.66{\scriptsize$\pm$.80} &  45.82{\scriptsize$\pm$.79} & \textbf{65.24}{\scriptsize$\pm$.97} \\
    \midrule
    \multirow{2}{*}{SSL} & \multirow{2}{*}{1} & SimCLR & 91.00{\scriptsize$\pm$.76} &  84.30{\scriptsize$\pm$.73} &  {36.39}{\scriptsize$\pm$.66} &  21.55{\scriptsize$\pm$.41} & {64.97}{\scriptsize$\pm$.94} &  44.18{\scriptsize$\pm$.85} &  32.46{\scriptsize$\pm$.70} & 36.15{\scriptsize$\pm$.76} \\
    & & BYOL & 85.77{\scriptsize$\pm$.73} &  66.16{\scriptsize$\pm$.86} &  34.53{\scriptsize$\pm$.62} &  {22.75}{\scriptsize$\pm$.41} & 51.76{\scriptsize$\pm$.79} &  42.16{\scriptsize$\pm$.75} &  34.54{\scriptsize$\pm$.70} & 36.50{\scriptsize$\pm$.68} \\
    \midrule
    \multirow{2}{*}{MSL} & \multirow{2}{*}{0.875} & SimCLR & {88.38}{\scriptsize$\pm$.70} & {73.97}{\scriptsize$\pm$.79} & 34.02{\scriptsize$\pm$.62} & 22.04{\scriptsize$\pm$.40} & 65.13{\scriptsize$\pm$.88} & {47.47}{\scriptsize$\pm$.86} & {36.96}{\scriptsize$\pm$.77} & 47.35{\scriptsize$\pm$.87} \\
    & & BYOL & 86.47{\scriptsize$\pm$.74} & 73.18{\scriptsize$\pm$.83} & \textbf{37.10}{\scriptsize$\pm$.67} & 23.97{\scriptsize$\pm$.44}& 61.40{\scriptsize$\pm$.87} & 48.31{\scriptsize$\pm$.86} & 33.31{\scriptsize$\pm$.66} & {50.71}{\scriptsize$\pm$.87} \\
    \midrule
    \multirow{2}{*}{SL $\rightarrow$ SSL} & \multirow{2}{*}{0$\rightarrow$1} & SimCLR & \textbf{92.24}{\scriptsize$\pm$.70} & \textbf{86.51}{\scriptsize$\pm$.67} & 36.11{\scriptsize$\pm$.67} & 21.75{\scriptsize$\pm$.41} & \textbf{71.05}{\scriptsize$\pm$.92} & 49.02{\scriptsize$\pm$.91} & 37.43{\scriptsize$\pm$.79} & 42.40{\scriptsize$\pm$.85} \\
    & & BYOL & 87.64{\scriptsize$\pm$.70} & 74.05{\scriptsize$\pm$.84} & 35.62{\scriptsize$\pm$.65} & 23.01{\scriptsize$\pm$.43} & 58.12{\scriptsize$\pm$.87} & 48.28{\scriptsize$\pm$.88} & 38.23{\scriptsize$\pm$.75} & 42.48{\scriptsize$\pm$.82} \\
    \midrule
    \multirow{2}{*}{SL $\rightarrow$ MSL} & \multirow{2}{*}{0$\rightarrow$0.875} & SimCLR & {91.46}{\scriptsize$\pm$.66} & {77.62}{\scriptsize$\pm$.76} & 34.46{\scriptsize$\pm$.64} & 22.50{\scriptsize$\pm$.41} & {69.50}{\scriptsize$\pm$.87} & {51.27}{\scriptsize$\pm$.91} & 40.39{\scriptsize$\pm$.82} & 62.12{\scriptsize$\pm$.93} \\
    & & BYOL &  88.37{\scriptsize$\pm$.73} & 71.54{\scriptsize$\pm$.78} & {36.08}{\scriptsize$\pm$.63} & \textbf{24.42}{\scriptsize$\pm$.45} & 63.40{\scriptsize$\pm$.86} & \textbf{53.65}{\scriptsize$\pm$.88} & \textbf{46.62}{\scriptsize$\pm$.85} & {64.33}{\scriptsize$\pm$.93} \\
    \midrule
    \multirow{2}{*}{MSL} &\multirow{2}{*}{0 $\nearrow$ 1} & \,\,SimCLR\,\, & 81.32{\scriptsize$\pm$.79} & 70.68{\scriptsize$\pm$.82} & 32.70{\scriptsize$\pm$.60} & 22.77{\scriptsize$\pm$.41} & 61.36{\scriptsize$\pm$.84} & 44.50{\scriptsize$\pm$.83} & 36.27{\scriptsize$\pm$.69} & 50.40{\scriptsize$\pm$.86} \\
    & & BYOL & 77.37{\scriptsize$\pm$.83} & 67.84{\scriptsize$\pm$.82} & 34.70{\scriptsize$\pm$.64} & 23.38{\scriptsize$\pm$.41} & 59.18{\scriptsize$\pm$.82} & 45.37{\scriptsize$\pm$.83} & 36.18{\scriptsize$\pm$.71} & 51.00{\scriptsize$\pm$.85} \\
    \bottomrule
\end{tabular}\vspace*{0.05cm}}\\
\end{table*}

\begin{table*}[!h]
\caption{5-way 5-shot CD-FSL performance\,(\%) of the models pre-trained with varying configurations of $\gamma$ in Eq.\,\eqref{eq:loss_msl} of MSL. ResNet18 is used as the backbone model, and ImageNet is used as the source data for SL. The {best} results are marked in bold.}\label{tab:dynamic_gamma_5shot}
\vspace*{0.1cm}
\centering
\footnotesize\addtolength{\tabcolsep}{-3.0pt}
\resizebox{\linewidth}{!}{
\begin{tabular}{c|c|c|cccc|cccc}
    \toprule
    Pre-train & \,\,\multirow{2}{*}{$\gamma$} \,\, & \multirow{2}{*}{Method} & \multicolumn{4}{c|}{Small Similarity} & \multicolumn{4}{c}{Large Similarity} \\
    Scheme & & & \!\!CropDisease\!\! & \!\!EuroSAT\!\! & ISIC & ChestX & Places & Plantae & Cars & CUB \\
    \midrule
    SL & 0 & Default & 92.81{\scriptsize$\pm$.45} &  84.73{\scriptsize$\pm$.51} &  44.10{\scriptsize$\pm$.58} &  25.51{\scriptsize$\pm$.44} & 79.22{\scriptsize$\pm$.64} &  63.21{\scriptsize$\pm$.82} &  66.38{\scriptsize$\pm$.80} & 83.93{\scriptsize$\pm$.66} \\
    \midrule
    \multirow{2}{*}{SSL} & \multirow{2}{*}{1} & SimCLR & 97.46{\scriptsize$\pm$.34} &  94.12{\scriptsize$\pm$.32} &  {47.85}{\scriptsize$\pm$.65} &  25.26{\scriptsize$\pm$.44} & 80.43{\scriptsize$\pm$.61} &  60.07{\scriptsize$\pm$.84} &  44.55{\scriptsize$\pm$.74} & 47.36{\scriptsize$\pm$.79} \\
    & & BYOL & {96.93}{\scriptsize$\pm$.30} &  87.83{\scriptsize$\pm$.48} &  47.59{\scriptsize$\pm$.63} &  {28.36}{\scriptsize$\pm$.46} & 72.47{\scriptsize$\pm$.63} &  61.02{\scriptsize$\pm$.82} &  48.56{\scriptsize$\pm$.76} & 51.31{\scriptsize$\pm$.78} \\
    \midrule
    \multirow{2}{*}{MSL} & \multirow{2}{*}{0.875} & SimCLR & 96.50{\scriptsize$\pm$.35} & 90.11{\scriptsize$\pm$.40} & 45.38{\scriptsize$\pm$.63} & 26.05{\scriptsize$\pm$.44} & 82.56{\scriptsize$\pm$.58} & {64.76}{\scriptsize$\pm$.83} & {51.84}{\scriptsize$\pm$.79} & 64.53{\scriptsize$\pm$.80} \\
    & & BYOL & 96.74{\scriptsize$\pm$.31} & {90.82}{\scriptsize$\pm$.40} & 49.14{\scriptsize$\pm$.70} & 29.58{\scriptsize$\pm$.47}& {81.27}{\scriptsize$\pm$.59} & 67.39{\scriptsize$\pm$.81} & 46.76{\scriptsize$\pm$.73} & {69.67}{\scriptsize$\pm$.82} \\
    \midrule
    \multirow{2}{*}{SL $\rightarrow$ SSL} & \multirow{2}{*}{0$\rightarrow$1} & SimCLR & \textbf{97.88}{\scriptsize$\pm$.30} & \textbf{95.28}{\scriptsize$\pm$.27} & 48.38{\scriptsize$\pm$.60} & 25.25{\scriptsize$\pm$.44} & 84.40{\scriptsize$\pm$.53} & 66.35{\scriptsize$\pm$.82} & 51.31{\scriptsize$\pm$.84} & 57.11{\scriptsize$\pm$.88} \\
    & & BYOL & {97.58}{\scriptsize$\pm$.26} & {91.82}{\scriptsize$\pm$.39} & 49.32{\scriptsize$\pm$.63} & 28.27{\scriptsize$\pm$.48} & 78.87{\scriptsize$\pm$.60} & 67.83{\scriptsize$\pm$.82} & 54.70{\scriptsize$\pm$.84} & 60.60{\scriptsize$\pm$.82} \\
    \midrule
    \multirow{2}{*}{SL $\rightarrow$ MSL} & \multirow{2}{*}{0$\rightarrow$0.875} & SimCLR & 97.49{\scriptsize$\pm$.30} & 91.70{\scriptsize$\pm$.35} & 47.43{\scriptsize$\pm$.62} & 26.24{\scriptsize$\pm$.44} & \textbf{85.76}{\scriptsize$\pm$.52} & 69.24{\scriptsize$\pm$.81} & 58.97{\scriptsize$\pm$.82} & 81.51{\scriptsize$\pm$.72} \\
    & & BYOL & 97.09{\scriptsize$\pm$.31} & 90.89{\scriptsize$\pm$.40} & \textbf{50.72}{\scriptsize$\pm$.67} & 
    \textbf{30.20}{\scriptsize$\pm$.48} & 83.29{\scriptsize$\pm$.55} & \textbf{74.16}{\scriptsize$\pm$.77} & \textbf{68.87}{\scriptsize$\pm$.80} & \textbf{84.34}{\scriptsize$\pm$.67} \\
    \midrule
    \multirow{2}{*}{MSL} &\multirow{2}{*}{0 $\nearrow$ 1} & \,\,SimCLR\,\,  & 94.83{\scriptsize$\pm$.42} & 87.69{\scriptsize$\pm$.50} & 44.48{\scriptsize$\pm$.61} & 26.76{\scriptsize$\pm$.45} & 80.62{\scriptsize$\pm$.58} & 62.02{\scriptsize$\pm$.83} & 52.97{\scriptsize$\pm$.76} & 69.37{\scriptsize$\pm$.79} \\
    & & BYOL & 93.98{\scriptsize$\pm$.41} & 86.66{\scriptsize$\pm$.50} & 47.61{\scriptsize$\pm$.66} & 28.55{\scriptsize$\pm$.47} & 79.71{\scriptsize$\pm$.60} & 63.77{\scriptsize$\pm$.83} & 54.13{\scriptsize$\pm$.75} & 70.49{\scriptsize$\pm$.79} \\
    \bottomrule
\end{tabular}\vspace*{0.05cm}}\\
\end{table*}

\clearpage
\section{Increasing Batch Size on the Source Dataset for MSL}\label{appx:batchsize}

Naturally, the size of labeled source data and unlabeled target data differ greatly. For example, while ImageNet contains 1.3M training examples, Cars and CUB each contains 3,400 and 2,350 unlabeled examples when 20\% of the data is used. Thus, during 1,000 epochs of MSL pre-training (where each epoch corresponds to one pass through the unlabeled target data), only 2-3 passes are completed through ImageNet (refer to \textbf{MSL Pre-Training} setup in Appendix \ref{appx:training_setup}). Considering the effectiveness of SL when domain similarity is large, we posit that MSL under large domain similarity can benefit from higher batch size for on the source data, \textit{i.e.}, allowing more passes through the source dataset. In particular, we fix the batch size for the target data to 64, and increase the batch size for the source data.

Table \ref{tab:batchsize} describes CD-FSL performance according to the source batch size on Cars and CUB. It is shown that larger batch size for the source dataset can improve the MSL performance. We suppose that the MSL model with ImageNet obtains large generalization ability from large-scale data, gaining much larger benefit than miniImageNet or tieredImageNet source. This improvement is significant in Cars and CUB datasets because they are similar to ImageNet. In fact, ImageNet data already includes car types ($\sim$10 classes) and bird species ($\sim$59 classes).

\begin{table}[h]
\centering
\caption{5-way $k$-shot performance of MSL according to the source batch size when ImageNet is used as source.}
\vskip 0.1in
\small
\begin{tabular}{ccc|cc}
    \toprule
    Target Dataset & Method & Batch Size for Source Dataset & $k$=1 & $k$=5 \\
    \midrule
    \multirow{8}{*}{Cars} & \multirow{4}{*}{SimCLR} & 64 (default) & 36.96{\scriptsize$\pm$.77} & 51.84{\scriptsize$\pm$.79} \\
    & & 128 & 38.54{\scriptsize$\pm$.81} & 53.80{\scriptsize$\pm$.84} \\
    & & 256 & 38.24{\scriptsize$\pm$.78} & 54.18{\scriptsize$\pm$.81} \\
    & & 512 & 38.98{\scriptsize$\pm$.81} & 55.25{\scriptsize$\pm$.81} \\ \cmidrule{2-5}
    & \multirow{4}{*}{BYOL} & 64 (default) & 33.31{\scriptsize$\pm$.66} & 46.76{\scriptsize$\pm$.73} \\
    & & 128 & 39.85{\scriptsize$\pm$.81} & 58.01{\scriptsize$\pm$.80} \\
    & & 256 & 41.45{\scriptsize$\pm$.82} & 59.48{\scriptsize$\pm$.80} \\
    & & 512 & 40.98{\scriptsize$\pm$.80} & 59.48{\scriptsize$\pm$.81} \\ \midrule
    \multirow{8}{*}{CUB} & \multirow{4}{*}{SimCLR} & 64 (default) & 47.35{\scriptsize$\pm$.87} & 64.53{\scriptsize$\pm$.80} \\
    & & 128 & 49.91{\scriptsize$\pm$.87} & 68.01{\scriptsize$\pm$.81} \\
    & & 256 & 51.06{\scriptsize$\pm$.85} & 69.51{\scriptsize$\pm$.79} \\
    & & 512 & 51.48{\scriptsize$\pm$.88} & 70.13{\scriptsize$\pm$.79} \\ \cmidrule{2-5}
    & \multirow{4}{*}{BYOL} & 64 (default) & 50.71{\scriptsize$\pm$.87} & 69.67{\scriptsize$\pm$.82} \\
    & & 128 & 52.75{\scriptsize$\pm$.87} & 72.26{\scriptsize$\pm$.79} \\
    & & 256 & 54.17{\scriptsize$\pm$.86} & 73.50{\scriptsize$\pm$.79} \\
    & & 512 & 53.70{\scriptsize$\pm$.86} & 73.31{\scriptsize$\pm$.80} \\
    \bottomrule
\end{tabular}\label{tab:batchsize}
\end{table}

\clearpage
\section{Results Summary}\label{appx:summary}

\subsection{Source Dataset: ImageNet} \label{appx:summary_imagenet}
Table \ref{tab:summary_in_1shot} and Table \ref{tab:summary_in_5shot} describe 5-way 1-shot and 5-way 5-shot CD-FSL performance when ImageNet is used as the source dataset, respectively. Note that Table \ref{tab:summary_in_5shot} is added for convenience and this is the same with Table \ref{tab:main_msl} in the main paper. The results of STARTUP on BSCD-FSL (\textit{i.e.}, CropDisease, EuroSAT, ISIC, and ChestX) target datasets are from \citet{phoo2021selftraining}.
The results on the other four target datasets are our reimplementation with their official code.\footnote{\url{https://github.com/cpphoo/STARTUP}} Also, \citet{islam2021dynamic} did not provide the results of DynDistill on ImageNet source dataset, so we reimplemented it with their official code.\footnote{\url{https://github.com/asrafulashiq/dynamic-cdfsl}}

\begin{table*}[!h]
\caption{5-way 1-shot CD-FSL performance\,(\%) of the models pre-trained by SL, SSL, and MSL including their two-stage versions. ResNet18 is used as the backbone model, and ImageNet is used as the source data for SL. The balancing coefficient $\gamma$ in Eq.\,\eqref{eq:loss_msl} of MSL is set to be 0.875. The {best} results are marked in bold and the second best are underlined.}\label{tab:summary_in_1shot}
\vspace*{0.1cm}
\centering
\footnotesize\addtolength{\tabcolsep}{-3.0pt}
\resizebox{\linewidth}{!}{
\begin{tabular}{c|c|c|cccc|cccc}
    \toprule
    &\,\,Pre-train\,\, & \multirow{2}{*}{Method} & \multicolumn{4}{c|}{Small Similarity} & \multicolumn{4}{c}{Large Similarity} \\
    & Scheme & & \!\!CropDisease\!\! & \!\!EuroSAT\!\! & ISIC & ChestX & Places & Plantae & Cars & CUB \\
    \midrule
    & SL & Default & 74.18{\scriptsize$\pm$.82} &  66.14{\scriptsize$\pm$.83} &  31.11{\scriptsize$\pm$.55} &  22.48{\scriptsize$\pm$.39} & 57.47{\scriptsize$\pm$.86} &  43.66{\scriptsize$\pm$.80} &  \textbf{45.82}{\scriptsize$\pm$.79} & \textbf{65.24}{\scriptsize$\pm$.97} \\
    \cmidrule{2-11}
    &\multirow{2}{*}{SSL} & SimCLR & \textbf{91.00}{\scriptsize$\pm$.76} &  \textbf{84.30}{\scriptsize$\pm$.73} &  \underline{36.39}{\scriptsize$\pm$.66} &  21.55{\scriptsize$\pm$.41} & \underline{64.97}{\scriptsize$\pm$.94} &  44.18{\scriptsize$\pm$.85} &  32.46{\scriptsize$\pm$.70} & 36.15{\scriptsize$\pm$.76} \\
    \rotatebox[origin=c]{90}{\makecell[l]{\!\!\!\!\!\!\!\!\!\!\!\!\!\!Single-Stage}\!\!\!\!\!\!\!\!\!\!\!\!\!\!}&& BYOL & 85.77{\scriptsize$\pm$.73} &  66.16{\scriptsize$\pm$.86} &  34.53{\scriptsize$\pm$.62} &  \underline{22.75}{\scriptsize$\pm$.41} & 51.76{\scriptsize$\pm$.79} &  42.16{\scriptsize$\pm$.75} &  34.54{\scriptsize$\pm$.70} & 36.50{\scriptsize$\pm$.68} \\
    \cmidrule{2-11}
    &\multirow{2}{*}{MSL} & \,\,SimCLR\,\, &  \underline{88.38}{\scriptsize$\pm$.70} & \underline{73.97}{\scriptsize$\pm$.79} & 34.02{\scriptsize$\pm$.62} & 22.04{\scriptsize$\pm$.40} & \textbf{65.13}{\scriptsize$\pm$.88} & \underline{47.47}{\scriptsize$\pm$.86} & \underline{36.96}{\scriptsize$\pm$.77} & 47.35{\scriptsize$\pm$.87} \\
    && BYOL & 86.47{\scriptsize$\pm$.74} & 73.18{\scriptsize$\pm$.83} & \textbf{37.10}{\scriptsize$\pm$.67} & \textbf{23.97}{\scriptsize$\pm$.44}& 61.40{\scriptsize$\pm$.87} & \textbf{48.31}{\scriptsize$\pm$.86} & 33.31{\scriptsize$\pm$.66} & \underline{50.71}{\scriptsize$\pm$.87} \\
    \bottomrule
    \end{tabular}\vspace*{0.05cm}}\\
{\small (a) Performance comparison for single-stage schemes.} \\
\resizebox{\linewidth}{!}{
\begin{tabular}{c|c|c|p{12mm}p{12mm}p{12mm}p{13mm}|p{12mm}p{12mm}p{12mm}p{12mm}} \toprule
    &\multirow{2}{*}{\!SL$\rightarrow$SSL\!} & SimCLR & \textbf{92.24}{\scriptsize$\pm$.70} & \textbf{86.51}{\scriptsize$\pm$.67} & \textbf{36.11}{\scriptsize$\pm$.67} & 21.75{\scriptsize$\pm$.41} & \textbf{71.05}{\scriptsize$\pm$.92} & 49.02{\scriptsize$\pm$.91} & 37.43{\scriptsize$\pm$.79} & 42.40{\scriptsize$\pm$.85} \\
    && BYOL & 87.64{\scriptsize$\pm$.70} & 74.05{\scriptsize$\pm$.84} & 35.62{\scriptsize$\pm$.65} & 23.01{\scriptsize$\pm$.43} & 58.12{\scriptsize$\pm$.87} & 48.28{\scriptsize$\pm$.88} & 38.23{\scriptsize$\pm$.75} & 42.48{\scriptsize$\pm$.82} \\
    \cmidrule{2-11}
    \rotatebox[origin=c]{90}{\makecell[l]{\!\!\!\!\!\!\!\!\!\!\!\!\!\!Two-Stage\!\!\!\!\!\!\!\!\!\!\!\!\!\!}}&\multirow{2}{*}{\!SL$\rightarrow$MSL\!} & SimCLR & \underline{91.46}{\scriptsize$\pm$.66} & \underline{77.62}{\scriptsize$\pm$.76} & 34.46{\scriptsize$\pm$.64} & 22.50{\scriptsize$\pm$.41} & \underline{69.50}{\scriptsize$\pm$.87} & \underline{51.27}{\scriptsize$\pm$.91} & 40.39{\scriptsize$\pm$.82} & 62.12{\scriptsize$\pm$.93} \\
    && BYOL &  88.37{\scriptsize$\pm$.73} & 71.54{\scriptsize$\pm$.78} & \underline{36.08}{\scriptsize$\pm$.63} & \textbf{24.42}{\scriptsize$\pm$.45} & 63.40{\scriptsize$\pm$.86} & \textbf{53.65}{\scriptsize$\pm$.88} & \textbf{46.62}{\scriptsize$\pm$.85} & \underline{64.33}{\scriptsize$\pm$.93} \\
    \cmidrule{2-11}
    & \multirow{2}{*}{\!SL$\rightarrow$MSL$^+$\!\!\!} & STARTUP & 85.10{\scriptsize$\pm$.74} & 73.83{\scriptsize$\pm$.77} & 31.69{\scriptsize$\pm$.59} & 23.03{\scriptsize$\pm$.42} & 66.02{\scriptsize$\pm$.87} & 49.78{\scriptsize$\pm$.93} & 45.75{\scriptsize$\pm$.84} & \textbf{72.58}{\scriptsize$\pm$.93} \\
    && DynDistill & 87.53{\tiny$\pm$1.01}& 77.24{\tiny$\pm$1.06}& 34.55{\tiny$\pm$1.82} & \underline{24.02}{\tiny$\pm$1.59}& 60.84{\tiny$\pm$1.08}& 49.90{\tiny$\pm$1.22}& \underline{46.55}{\tiny$\pm$1.21}& 63.80{\tiny$\pm$1.32} \\
    \bottomrule
\end{tabular}
\vspace*{-0.05cm}}\\
{\small (b) Performance comparison for two-stage schemes.}
\end{table*}

\begin{table*}[!h]
\caption{5-way 5-shot CD-FSL performance\,(\%) of the models pre-trained by SL, SSL, and MSL including their two-stage versions. ResNet18 is used as the backbone model, and ImageNet is used as the source data for SL. The balancing coefficient $\gamma$ in Eq.\,\eqref{eq:loss_msl} of MSL is set to be 0.875. The {best} results are marked in bold and the second best are underlined.}\label{tab:summary_in_5shot}
\vspace*{0.1cm}
\centering
\footnotesize\addtolength{\tabcolsep}{-3.0pt}
\resizebox{\linewidth}{!}{
\begin{tabular}{c|c|c|cccc|cccc}
    \toprule
    &\,\,Pre-train\,\, & \multirow{2}{*}{Method} & \multicolumn{4}{c|}{Small Similarity} & \multicolumn{4}{c}{Large Similarity} \\
    & Scheme & & \!\!CropDisease\!\! & \!\!EuroSAT\!\! & ISIC & ChestX & Places & Plantae & Cars & CUB \\
    \midrule
    & SL & Default & 92.81{\scriptsize$\pm$.45} &  84.73{\scriptsize$\pm$.51} &  44.10{\scriptsize$\pm$.58} &  25.51{\scriptsize$\pm$.44} & 79.22{\scriptsize$\pm$.64} &  63.21{\scriptsize$\pm$.82} &  \textbf{66.38}{\scriptsize$\pm$.80} & \textbf{83.93}{\scriptsize$\pm$.66} \\
    \cmidrule{2-11}
    &\multirow{2}{*}{SSL} & SimCLR & \textbf{97.46}{\scriptsize$\pm$.34} &  \textbf{94.12}{\scriptsize$\pm$.32} &  \underline{47.85}{\scriptsize$\pm$.65} &  25.26{\scriptsize$\pm$.44} & 80.43{\scriptsize$\pm$.61} &  60.07{\scriptsize$\pm$.84} &  44.55{\scriptsize$\pm$.74} & 47.36{\scriptsize$\pm$.79} \\
    \rotatebox[origin=c]{90}{\makecell[l]{\!\!\!\!\!\!\!\!\!\!\!\!\!\!Single-Stage}\!\!\!\!\!\!\!\!\!\!\!\!\!\!}& & BYOL & \underline{96.93}{\scriptsize$\pm$.30} &  87.83{\scriptsize$\pm$.48} &  47.59{\scriptsize$\pm$.63} &  \underline{28.36}{\scriptsize$\pm$.46} & 72.47{\scriptsize$\pm$.63} &  61.02{\scriptsize$\pm$.82} &  48.56{\scriptsize$\pm$.76} & 51.31{\scriptsize$\pm$.78} \\
    \cmidrule{2-11}
    &\multirow{2}{*}{MSL} & \,\,SimCLR\,\, &  96.50{\scriptsize$\pm$.35} & 90.11{\scriptsize$\pm$.40} & 45.38{\scriptsize$\pm$.63} & 26.05{\scriptsize$\pm$.44} & \textbf{82.56}{\scriptsize$\pm$.58} & \underline{64.76}{\scriptsize$\pm$.83} & \underline{51.84}{\scriptsize$\pm$.79} & 64.53{\scriptsize$\pm$.80} \\
    && BYOL & 96.74{\scriptsize$\pm$.31} & \underline{90.82}{\scriptsize$\pm$.40} & \textbf{49.14}{\scriptsize$\pm$.70} & \textbf{29.58}{\scriptsize$\pm$.47}& \underline{81.27}{\scriptsize$\pm$.59} & \textbf{67.39}{\scriptsize$\pm$.81} & 46.76{\scriptsize$\pm$.73} & \underline{69.67}{\scriptsize$\pm$.82} \\
    \bottomrule
    \end{tabular}\vspace*{0.05cm}}\\
{\small (a) Performance comparison for single-stage schemes.} \\
\resizebox{\linewidth}{!}{
\begin{tabular}{c|c|c|cccc|p{12mm}p{12mm}p{12mm}p{12mm}} \toprule
    &\multirow{2}{*}{\!SL$\rightarrow$SSL\!} & SimCLR & \textbf{97.88}{\scriptsize$\pm$.30} & \textbf{95.28}{\scriptsize$\pm$.27} & 48.38{\scriptsize$\pm$.60} & 25.25{\scriptsize$\pm$.44} & 84.40{\scriptsize$\pm$.53} & 66.35{\scriptsize$\pm$.82} & 51.31{\scriptsize$\pm$.84} & 57.11{\scriptsize$\pm$.88} \\
    && BYOL & 97.58{\scriptsize$\pm$.26} & 91.82{\scriptsize$\pm$.39} & 49.32{\scriptsize$\pm$.63} & 28.27{\scriptsize$\pm$.48} & 78.87{\scriptsize$\pm$.60} & 67.83{\scriptsize$\pm$.82} & 54.70{\scriptsize$\pm$.84} & 60.60{\scriptsize$\pm$.82} \\
    \cmidrule{2-11}
    \rotatebox[origin=c]{90}{\makecell[l]{\!\!\!\!\!\!\!\!\!\!\!\!\!\!Two-Stage\!\!\!\!\!\!\!\!\!\!\!\!\!\!}}&\multirow{2}{*}{\!SL$\rightarrow$MSL\!} & SimCLR & 97.49{\scriptsize$\pm$.30} & 91.70{\scriptsize$\pm$.35} & 47.43{\scriptsize$\pm$.62} & 26.24{\scriptsize$\pm$.44} & \textbf{85.76}{\scriptsize$\pm$.52} & 69.24{\scriptsize$\pm$.81} & 58.97{\scriptsize$\pm$.82} & 81.51{\scriptsize$\pm$.72} \\
    && BYOL &  97.09{\scriptsize$\pm$.31} & 90.89{\scriptsize$\pm$.40} & \textbf{50.72}{\scriptsize$\pm$.67} & \textbf{30.20}{\scriptsize$\pm$.48} & 83.29{\scriptsize$\pm$.55} & \textbf{74.16}{\scriptsize$\pm$.77} & \underline{68.87}{\scriptsize$\pm$.80} & 84.34{\scriptsize$\pm$.67} \\
    \cmidrule{2-11}
    & \multirow{2}{*}{\!SL$\rightarrow$MSL$^+$\!\!\!} & STARTUP & 96.06{\scriptsize$\pm$.33} & 89.70{\scriptsize$\pm$.41} & 46.02{\scriptsize$\pm$.59} & 27.24{\scriptsize$\pm$.46} & \underline{85.00}{\scriptsize$\pm$.52} & 69.40{\scriptsize$\pm$.84} & 68.43{\scriptsize$\pm$.82} & \textbf{89.60}{\scriptsize$\pm$.55} \\
    && DynDistill & \underline{97.60}{\scriptsize$\pm$.35}& \underline{92.28}{\scriptsize$\pm$.46}& \underline{50.06}{\scriptsize$\pm$.86}& \underline{29.65}{\scriptsize$\pm$.67}& 82.22{\scriptsize$\pm$.81}& \underline{71.49}{\tiny$\pm$1.06}& \textbf{69.45}{\tiny$\pm$1.12}& \underline{86.54}{\tiny$\pm$1.88} \\
    \bottomrule
\end{tabular}
\vspace*{-0.05cm}}\\
{\small (b) Performance comparison for two-stage schemes.}
\vspace*{-0.4cm}
\end{table*}
    
\clearpage
\subsection{Source Dataset: tieredImageNet}
Table \ref{tab:summary_tiered_1shot} and Table \ref{tab:summary_tiered_5shot} describe 5-way 1-shot and 5-way 5-shot CD-FSL performance when tieredImageNet is used as the source dataset, respectively.
\citet{phoo2021selftraining} did not provide the results of STARTUP on tieredImageNet source dataset, so we reimplemented it with their official code. The results of DynDistill on BSCD-FSL are from \citet{islam2021dynamic}; however, note that DynDistill used a larger ResNet-18 backbone model than our setting, which is provided by \citet{tian2020rethinking}. Also, the results on the other four target datasets are our reimplementation with their official code.

The difference of the result of tieredImageNet from the result of ImageNet as the source dataset is that one-stage MSL can outperform SL on Cars and CUB datasets. It is considered that bigger source dataset makes SL stronger, as we have addressed this issue in Appendix \ref{appx:other_source}.

\begin{table*}[!h]
\caption{5-way 1-shot CD-FSL performance\,(\%) of the models pre-trained by SL, SSL, and MSL including their two-stage versions. ResNet18 is used as the backbone model, and tieredImageNet is used as the source data for SL. The balancing coefficient $\gamma$ in Eq.\,\eqref{eq:loss_msl} of MSL is set to be 0.875. The {best} results are marked in bold and the second best are underlined.}\label{tab:summary_tiered_1shot}
\vspace*{0.1cm}
\centering
\footnotesize\addtolength{\tabcolsep}{-3.0pt}
\resizebox{\linewidth}{!}{
\begin{tabular}{c|c|c|cccc|cccc}
    \toprule
    &\,\,Pre-train\,\, & \multirow{2}{*}{Method} & \multicolumn{4}{c|}{Small Similarity} & \multicolumn{4}{c}{Large Similarity} \\
    & Scheme & & \!\!CropDisease\!\! & \!\!EuroSAT\!\! & ISIC & ChestX & Places & Plantae & Cars & CUB \\
    \midrule
    &SL & Default & 65.70{\scriptsize$\pm$.94} &  60.07{\scriptsize$\pm$.88} &  29.75{\scriptsize$\pm$.56} &  22.11{\scriptsize$\pm$.42} & 52.82{\scriptsize$\pm$.86} &  34.99{\scriptsize$\pm$.64} &  31.38{\scriptsize$\pm$.61} &  \underline{54.18}{\scriptsize$\pm$.91} \\
    \cmidrule{2-11}
    &\multirow{2}{*}{SSL} & SimCLR & \textbf{91.00}{\scriptsize$\pm$.76} &  \textbf{84.30}{\scriptsize$\pm$.73} &  \underline{36.39}{\scriptsize$\pm$.66} &  21.55{\scriptsize$\pm$.41} & \textbf{64.97}{\scriptsize$\pm$.94} &  44.18{\scriptsize$\pm$.85} &  32.46{\scriptsize$\pm$.70} & 36.15{\scriptsize$\pm$.76} \\
    \rotatebox[origin=c]{90}{\makecell[l]{\!\!\!\!\!\!\!\!\!\!\!\!\!\!Single-Stage}\!\!\!\!\!\!\!\!\!\!\!\!\!\!}&& BYOL & 85.77{\scriptsize$\pm$.73} &  66.16{\scriptsize$\pm$.86} &  34.53{\scriptsize$\pm$.62} &  \underline{22.75}{\scriptsize$\pm$.41} & 51.76{\scriptsize$\pm$.79} &  42.16{\scriptsize$\pm$.75} &  34.54{\scriptsize$\pm$.70} & 36.50{\scriptsize$\pm$.68} \\
    \cmidrule{2-11}
    &\multirow{2}{*}{MSL} & \,\,SimCLR\,\, &  \underline{87.44}{\scriptsize$\pm$.72} & \underline{77.42}{\scriptsize$\pm$.77} & 35.47{\scriptsize$\pm$.64} & 21.95{\scriptsize$\pm$.40} & \underline{63.83}{\scriptsize$\pm$.93} & \underline{46.47}{\scriptsize$\pm$.87} & \underline{34.65}{\scriptsize$\pm$.74} & 50.41{\scriptsize$\pm$.90} \\
    && BYOL & 84.67{\scriptsize$\pm$.78} & 68.45{\scriptsize$\pm$.81} &  \textbf{37.30}{\scriptsize$\pm$.66} &  \textbf{24.41}{\scriptsize$\pm$.44} & 60.07{\scriptsize$\pm$.87} &  \textbf{46.49}{\scriptsize$\pm$.83} &  \textbf{37.88}{\scriptsize$\pm$.75} & \textbf{54.43}{\scriptsize$\pm$.88} \\
    \bottomrule
    \end{tabular}\vspace*{0.05cm}}\\
{\small (a) Performance comparison for single-stage schemes.} \\
\resizebox{\linewidth}{!}{
\begin{tabular}{c|c|c|cccc|p{12mm}p{12mm}p{12mm}p{12mm}} \toprule
    &\multirow{2}{*}{\!SL$\rightarrow$SSL\!} & SimCLR & \textbf{92.41}{\scriptsize$\pm$.70} & \textbf{86.61}{\scriptsize$\pm$.66} & 36.95{\scriptsize$\pm$.67} & 21.75{\scriptsize$\pm$.40} & \textbf{68.51}{\scriptsize$\pm$.94} & 47.92{\scriptsize$\pm$.88} & 35.37{\scriptsize$\pm$.77} & 44.74{\scriptsize$\pm$.86} \\
    && BYOL & 84.82{\scriptsize$\pm$.76} & 66.92{\scriptsize$\pm$.84} & \underline{37.19}{\scriptsize$\pm$.66} & \underline{24.23}{\scriptsize$\pm$.46} & 44.34{\scriptsize$\pm$.79} & 44.32{\scriptsize$\pm$.81} & 38.49{\scriptsize$\pm$.78} & 44.40{\scriptsize$\pm$.83} \\
    \cmidrule{2-11}
    \rotatebox[origin=c]{90}{\makecell[l]{\!\!\!\!\!\!\!\!\!\!\!\!\!\!Two-Stage\!\!\!\!\!\!\!\!\!\!\!\!\!\!}}&\multirow{2}{*}{\!SL$\rightarrow$MSL\!} & SimCLR & \underline{90.13}{\scriptsize$\pm$.69} & \underline{80.20}{\scriptsize$\pm$.78} & 35.32{\scriptsize$\pm$.64} & 22.18{\scriptsize$\pm$.38} & \underline{64.85}{\scriptsize$\pm$.92} & \underline{48.00}{\scriptsize$\pm$.86} & 35.83{\scriptsize$\pm$.75} & 60.87{\scriptsize$\pm$.90} \\
    && BYOL &  85.72{\scriptsize$\pm$.76} & 53.92{\scriptsize$\pm$.94} & \textbf{39.41}{\scriptsize$\pm$.68} & \textbf{24.31}{\scriptsize$\pm$.45} & 59.16{\scriptsize$\pm$.86} & \textbf{48.48}{\scriptsize$\pm$.83} & \textbf{41.02}{\scriptsize$\pm$.78} & \underline{61.98}{\scriptsize$\pm$.88} \\
    \cmidrule{2-11}
    & \multirow{2}{*}{\!SL$\rightarrow$MSL$^+$\!\!\!} & STARTUP & 77.67{\scriptsize$\pm$.83} & 69.60{\scriptsize$\pm$.86} & 33.90{\scriptsize$\pm$.63} & 23.13{\scriptsize$\pm$.40} & 59.14{\scriptsize$\pm$.87} & 41.80{\scriptsize$\pm$.85} & 34.45{\scriptsize$\pm$.66} & \textbf{63.83}{\scriptsize$\pm$.90} \\
    && DynDistill & 84.41{\scriptsize$\pm$.75} & 72.15{\scriptsize$\pm$.75} & 33.87{\scriptsize$\pm$.56} & 22.70{\scriptsize$\pm$.42} & 52.21{\tiny$\pm$1.15}& 43.06{\tiny$\pm$1.12}& \underline{38.51}{\tiny$\pm$1.03}& 58.67{\tiny$\pm$1.30} \\
    \bottomrule
\end{tabular}
\vspace*{-0.05cm}}\\
{\small (b) Performance comparison for two-stage schemes.}
\end{table*}

\begin{table*}[!h]
\caption{5-way 5-shot CD-FSL performance\,(\%) of the models pre-trained by SL, SSL, and MSL including their two-stage versions. ResNet18 is used as the backbone model, and tieredImageNet is used as the source data for SL. The balancing coefficient $\gamma$ in Eq.\,\eqref{eq:loss_msl} of MSL is set to be 0.875. The {best} results are marked in bold and the second best are underlined.}\label{tab:summary_tiered_5shot}
\vspace*{0.1cm}
\centering
\footnotesize\addtolength{\tabcolsep}{-3.0pt}
\resizebox{\linewidth}{!}{
\begin{tabular}{c|c|c|cccc|cccc}
    \toprule
    &\,\,Pre-train\,\, & \multirow{2}{*}{Method} & \multicolumn{4}{c|}{Small Similarity} & \multicolumn{4}{c}{Large Similarity} \\
    & Scheme & & \!\!CropDisease\!\! & \!\!EuroSAT\!\! & ISIC & ChestX & Places & Plantae & Cars & CUB \\
    \midrule
    &SL & Default & 86.34{\scriptsize$\pm$.60} &  79.95{\scriptsize$\pm$.66} &  40.60{\scriptsize$\pm$.58} &  25.20{\scriptsize$\pm$.41} & 72.96{\scriptsize$\pm$.67} &  51.11{\scriptsize$\pm$.76} &  45.18{\scriptsize$\pm$.68} &  \textbf{74.14}{\scriptsize$\pm$.80} \\
    \cmidrule{2-11}
    &\multirow{2}{*}{SSL} & SimCLR & \textbf{97.46}{\scriptsize$\pm$.34} &  \textbf{94.12}{\scriptsize$\pm$.32} &  \underline{47.85}{\scriptsize$\pm$.65} &  25.26{\scriptsize$\pm$.44} & \underline{80.43}{\scriptsize$\pm$.61} &  60.07{\scriptsize$\pm$.84} &  44.55{\scriptsize$\pm$.74} & 47.36{\scriptsize$\pm$.79} \\
    \rotatebox[origin=c]{90}{\makecell[l]{\!\!\!\!\!\!\!\!\!\!\!\!\!\!Single-Stage}\!\!\!\!\!\!\!\!\!\!\!\!\!\!}&& BYOL & \underline{96.93}{\scriptsize$\pm$.30} &  87.83{\scriptsize$\pm$.48} &  47.59{\scriptsize$\pm$.63} &  \underline{28.36}{\scriptsize$\pm$.46} & 72.47{\scriptsize$\pm$.63} &  61.02{\scriptsize$\pm$.82} &  48.56{\scriptsize$\pm$.76} & 51.31{\scriptsize$\pm$.78} \\
    \cmidrule{2-11}
    &\multirow{2}{*}{MSL} & \,\,SimCLR\,\, &  96.68{\scriptsize$\pm$.33} & \underline{91.72}{\scriptsize$\pm$.37} & 47.55{\scriptsize$\pm$.67} & 26.10{\scriptsize$\pm$.45} & \textbf{81.67}{\scriptsize$\pm$.58} & \underline{63.96}{\scriptsize$\pm$.82} & \underline{48.81}{\scriptsize$\pm$.77} & 68.78{\scriptsize$\pm$.82} \\
    && BYOL & 96.41{\scriptsize$\pm$.33} & 89.51{\scriptsize$\pm$.42} & \textbf{50.95}{\scriptsize$\pm$.69} & \textbf{30.04}{\scriptsize$\pm$.47} & 80.16{\scriptsize$\pm$.60} & \textbf{67.09}{\scriptsize$\pm$.80} & \textbf{54.75}{\scriptsize$\pm$.80} & \underline{73.03}{\scriptsize$\pm$.82} \\
    \bottomrule
    \end{tabular}\vspace*{0.05cm}}\\
{\small (a) Performance comparison for single-stage schemes.} \\
\resizebox{\linewidth}{!}{
\begin{tabular}{c|c|c|cccc|p{12mm}p{12mm}p{12mm}p{12mm}} \toprule
    &\multirow{2}{*}{\!SL$\rightarrow$SSL\!} & SimCLR & \textbf{97.88}{\scriptsize$\pm$.31} & \textbf{95.39}{\scriptsize$\pm$.26} & 50.28{\scriptsize$\pm$.61} & 25.31{\scriptsize$\pm$.44} & \textbf{83.51}{\scriptsize$\pm$.56} & 65.40{\scriptsize$\pm$.82} & 48.91{\scriptsize$\pm$.83} & 61.80{\scriptsize$\pm$.84} \\
    && BYOL & 96.25{\scriptsize$\pm$.31} & 89.39{\scriptsize$\pm$.45} & \underline{53.00}{\scriptsize$\pm$.64} & \underline{30.66}{\scriptsize$\pm$.48} & 71.57{\scriptsize$\pm$.63} & 63.06{\scriptsize$\pm$.79} & 55.04{\scriptsize$\pm$.82} & 62.78{\scriptsize$\pm$.80} \\
    \cmidrule{2-11}
    \rotatebox[origin=c]{90}{\makecell[l]{\!\!\!\!\!\!\!\!\!\!\!\!\!\!Two-Stage\!\!\!\!\!\!\!\!\!\!\!\!\!\!}}&\multirow{2}{*}{\!SL$\rightarrow$MSL\!} & SimCLR & \underline{97.43}{\scriptsize$\pm$.31} & \underline{93.09}{\scriptsize$\pm$.33} & 49.66{\scriptsize$\pm$.63} & 26.27{\scriptsize$\pm$.44} & \underline{83.03}{\scriptsize$\pm$.56} & \underline{65.78}{\scriptsize$\pm$.85} & 52.22{\scriptsize$\pm$.81} & 80.37{\scriptsize$\pm$.76} \\
    && BYOL &  96.64{\scriptsize$\pm$.32} & 85.97{\scriptsize$\pm$.50} & \textbf{53.67}{\scriptsize$\pm$.68} & \textbf{30.84}{\scriptsize$\pm$.51} & 80.76{\scriptsize$\pm$.56} & \textbf{69.77}{\scriptsize$\pm$.79} & \textbf{62.09}{\scriptsize$\pm$.78} & \underline{82.77}{\scriptsize$\pm$.69} \\
    \cmidrule{2-11}
    & \multirow{2}{*}{\!SL$\rightarrow$MSL$^+$\!\!\!} & STARTUP & 92.87{\scriptsize$\pm$.41} & 85.23{\scriptsize$\pm$.59} & 48.20{\scriptsize$\pm$.62} & 27.06{\scriptsize$\pm$.42} & 78.00{\scriptsize$\pm$.60} & 60.28{\scriptsize$\pm$.82} & 51.14{\scriptsize$\pm$.75} & \textbf{83.36}{\scriptsize$\pm$.66} \\
    && DynDistill & 95.90{\scriptsize$\pm$.34} & 89.44{\scriptsize$\pm$.42} & 47.21{\scriptsize$\pm$.56} & 27.67 {\scriptsize$\pm$.46} & 75.67{\scriptsize$\pm$.86}& 64.32{\tiny$\pm$1.08} & \underline{59.14}{\tiny$\pm$1.15}& 79.26{\scriptsize$\pm$.97} \\
    \bottomrule
\end{tabular}
\vspace*{-0.05cm}}\\
{\small (b) Performance comparison for two-stage schemes.}
\vspace*{-0.4cm}
\end{table*}

\clearpage
\subsection{Source Dataset: miniImageNet}\label{appx:summary_mini}
Table \ref{tab:summary_mini_1shot} and Table \ref{tab:summary_mini_5shot} describe 5-way 1-shot and 5-way 5-shot CD-FSL performance when miniImageNet is used as the source dataset, respectively.
The results of STARTUP and DynDistill on BSCD-FSL target datasets are from \citet{phoo2021selftraining} and \citet{islam2021dynamic}, respectively. The results on the other four target datasets are our reimplementation with their official codes.
Similar to the results when tieredImageNet is used as the source dataset, one-stage MSL can outperform SL on Cars and CUB datasets.
For a thorough comparison, we also report the results of meta-learning based approaches: MAML\,\citep{maml}, MatchingNet\,\citep{matchingnet}, and RelationNet\,\citep{relationnet}, where the numbers are from \cite{tseng2020cross,bscd_fsl,islam2021dynamic}. As previous studies on CD-FSL verified, meta-learning based algorithms are mostly outperformed by transfer learning based algorithms in the cross-domain setup.

\begin{table*}[!h]
\caption{5-way 1-shot CD-FSL performance\,(\%) of the models pre-trained by SL, SSL, and MSL including their two-stage versions. ResNet10 is used as the backbone model, and miniImageNet is used as the source data for SL. The balancing coefficient $\gamma$ in Eq.\,\eqref{eq:loss_msl} of MSL is set to be 0.875. The {best} results are marked in bold and the second best are underlined.}\label{tab:summary_mini_1shot}
\centering
\footnotesize\addtolength{\tabcolsep}{-3.0pt}
\resizebox{\linewidth}{!}{
\begin{tabular}{c|c|c|cccc|cccc}
    \toprule
    &\,\,Pre-train\,\, & \multirow{2}{*}{Method} & \multicolumn{4}{c|}{Small Similarity} & \multicolumn{4}{c}{Large Similarity} \\
    & Scheme & & \!\!CropDisease\!\! & \!\!EuroSAT\!\! & ISIC & ChestX & Places & Plantae & Cars & CUB \\
    \midrule
    & \multirow{3}{*}{-} & MAML & - & - & - & - & - & - & - & - \\
    &  & \!\!MatchingNet\!\! & 46.86{\scriptsize$\pm$.88} & 54.88{\scriptsize$\pm$.90} & 27.37{\scriptsize$\pm$.51} & 20.65{\scriptsize$\pm$.29} & 49.86{\scriptsize$\pm$.79} & 32.70{\scriptsize$\pm$.60} & 30.77{\scriptsize$\pm$.47} & 35.89{\scriptsize$\pm$.51} \\
    &  & RelationNet & - & - & - & - & 48.64{\scriptsize$\pm$.85} & 33.17{\scriptsize$\pm$.64} & 29.11{\scriptsize$\pm$.60} & \underline{42.44}{\scriptsize$\pm$.77} \\
    \cmidrule{2-11}
    &SL & Default & 72.82{\scriptsize$\pm$.87} &  65.03{\scriptsize$\pm$.88} &  29.91{\scriptsize$\pm$.54} &  22.88{\scriptsize$\pm$.42} & 52.45{\scriptsize$\pm$.78} &  36.72{\scriptsize$\pm$.67} &  30.20{\scriptsize$\pm$.54} &  40.56{\scriptsize$\pm$.78} \\ 
    \cmidrule{2-11}
    &\multirow{2}{*}{SSL} & SimCLR & \textbf{89.49}{\scriptsize$\pm$.74} &  \textbf{79.50}{\scriptsize$\pm$.78} &  34.90{\scriptsize$\pm$.64} &  21.97{\scriptsize$\pm$.41} & \underline{58.75}{\scriptsize$\pm$.93} & \underline{42.65}{\scriptsize$\pm$.80} &  30.89{\scriptsize$\pm$.66} & 35.49{\scriptsize$\pm$.73} \\
    \rotatebox[origin=c]{90}{\makecell[l]{\!\!\!\!\!\!Single-Stage\!\!\!\!\!\!\!\!\!\!\!\!\!\!\!\!\!\!\!\!\!\!\!\!\!}}&& BYOL & 80.10{\scriptsize$\pm$.76} &  66.45{\scriptsize$\pm$.80} &  33.50{\scriptsize$\pm$.59} & \underline{23.11}{\scriptsize$\pm$.42} & 47.81{\scriptsize$\pm$.75} &  39.12{\scriptsize$\pm$.71} & 31.53{\scriptsize$\pm$.65} & 35.96{\scriptsize$\pm$.70} \\
    \cmidrule{2-11}
    &\multirow{2}{*}{MSL} & \,\,SimCLR\,\, &  \underline{87.15}{\scriptsize$\pm$.75} &  \underline{74.18}{\scriptsize$\pm$.80} &  \underline{35.10}{\scriptsize$\pm$.64} &  22.83{\scriptsize$\pm$.41} & \textbf{59.72}{\scriptsize$\pm$.89} & 42.24{\scriptsize$\pm$.80} & \underline{33.89}{\scriptsize$\pm$.66} & 40.89{\scriptsize$\pm$.79} \\
    && BYOL & 74.16{\scriptsize$\pm$.82} & 66.64{\scriptsize$\pm$.81} & \textbf{35.63}{\scriptsize$\pm$.66} & \textbf{24.07}{\scriptsize$\pm$.47} & 53.60{\scriptsize$\pm$.82} & \textbf{43.94}{\scriptsize$\pm$.79} & \textbf{35.71}{\scriptsize$\pm$.68} & \textbf{42.73}{\scriptsize$\pm$.78} \\
    \bottomrule
    \end{tabular}\vspace*{0.05cm}}\\
{\small (a) Performance comparison for single-stage schemes.} \\
\resizebox{\linewidth}{!}{
\begin{tabular}{c|c|c|p{13mm}p{13mm}p{13mm}p{13mm}|p{12mm}p{12mm}p{12mm}p{12mm}} \toprule
    &\multirow{2}{*}{\!SL$\rightarrow$SSL\!} & SimCLR & \textbf{89.39}{\scriptsize$\pm$.82} & \textbf{82.64}{\scriptsize$\pm$.73} & 35.09{\scriptsize$\pm$.64} & 22.15{\scriptsize$\pm$.40} & \textbf{63.19}{\scriptsize$\pm$.92} & \textbf{46.30}{\scriptsize$\pm$.85} & 34.85{\scriptsize$\pm$.74} & 39.92{\scriptsize$\pm$.79} \\
    && BYOL & 82.61{\scriptsize$\pm$.76} & 67.67{\scriptsize$\pm$.77} & \textbf{35.92}{\scriptsize$\pm$.68} & \underline{23.76}{\scriptsize$\pm$.45} & 53.72{\scriptsize$\pm$.79} & \underline{45.02}{\scriptsize$\pm$.79} & \underline{37.40}{\scriptsize$\pm$.74} & 41.61{\scriptsize$\pm$.75} \\
    \cmidrule{2-11}
    \rotatebox[origin=c]{90}{\makecell[l]{\!\!\!\!\!\!\!\!\!\!\!\!\!\!Two-Stage\!\!\!\!\!\!\!\!\!\!\!\!\!\!}}&\multirow{2}{*}{\!SL$\rightarrow$MSL\!} & SimCLR & \underline{86.18}{\scriptsize$\pm$.77} & \underline{74.06}{\scriptsize$\pm$.85} & 33.91{\scriptsize$\pm$.65} & 22.13{\scriptsize$\pm$.40} & \underline{61.56}{\scriptsize$\pm$.86} & 43.47{\scriptsize$\pm$.79} & 35.78{\scriptsize$\pm$.72} & \underline{43.50}{\scriptsize$\pm$.82} \\
    && BYOL &  75.77{\scriptsize$\pm$.82} & 65.67{\scriptsize$\pm$.83} & \underline{35.23}{\scriptsize$\pm$.66} & \textbf{24.47}{\scriptsize$\pm$.44} & 54.86{\scriptsize$\pm$.81} & 44.68{\scriptsize$\pm$.78} & \textbf{38.20}{\scriptsize$\pm$.71} & \textbf{45.82}{\scriptsize$\pm$.79} \\
    \cmidrule{2-11}
    & \multirow{2}{*}{\!SL$\rightarrow$MSL$^+$\!\!\!} & STARTUP & 75.93{\scriptsize$\pm$.80} & 63.88{\scriptsize$\pm$.84} & 32.66{\scriptsize$\pm$.60} & 23.09{\scriptsize$\pm$.43} & 48.87{\scriptsize$\pm$.81} & 38.01{\scriptsize$\pm$.73} & 31.79{\scriptsize$\pm$.61} & 41.24{\scriptsize$\pm$.75} \\
    && DynDistill & 82.14{\scriptsize$\pm$.78} & 73.14{\scriptsize$\pm$.84} & 34.66{\scriptsize$\pm$.58} & 23.38{\scriptsize$\pm$.43} & 49.28{\tiny$\pm$1.11} & 40.60{\tiny$\pm$1.15} & 34.77{\scriptsize$\pm$.98} & 42.51{\tiny$\pm$1.11} \\
    \bottomrule
\end{tabular}
\vspace*{-0.05cm}}\\
{\small (b) Performance comparison for two-stage schemes.}
\end{table*}

\begin{table*}[!h]
\caption{5-way 5-shot CD-FSL performance\,(\%) of the models pre-trained by SL, SSL, and MSL including their two-stage versions. ResNet10 is used as the backbone model, and miniImageNet is used as the source data for SL. The balancing coefficient $\gamma$ in Eq.\,\eqref{eq:loss_msl} of MSL is set to be 0.875. The {best} results are marked in bold and the second best are underlined.}\label{tab:summary_mini_5shot}
\centering
\footnotesize\addtolength{\tabcolsep}{-3.0pt}
\resizebox{\linewidth}{!}{
\begin{tabular}{c|c|c|cccc|cccc}
    \toprule
    &\,\,Pre-train\,\, & \multirow{2}{*}{Method} & \multicolumn{4}{c|}{Small Similarity} & \multicolumn{4}{c}{Large Similarity} \\
    & Scheme & & \!\!CropDisease\!\! & \!\!EuroSAT\!\! & ISIC & ChestX & Places & Plantae & Cars & CUB \\
    \midrule
    & \multirow{3}{*}{-} & MAML & 78.05{\scriptsize$\pm$.68} & 71.70{\scriptsize$\pm$.72} & 40.13{\scriptsize$\pm$.58} & 23.48{\scriptsize$\pm$.96} & - & - & - & - \\
    &  & \!\!MatchingNet\!\! & 66.39{\scriptsize$\pm$.78} & 64.45{\scriptsize$\pm$.63} & 36.74{\scriptsize$\pm$.53} & 22.40{\scriptsize$\pm$.70} & 63.16{\scriptsize$\pm$.77} & 46.53{\scriptsize$\pm$.68} & 38.99{\scriptsize$\pm$.64} & 51.37{\scriptsize$\pm$.77} \\
    &  & RelationNet & 68.99{\scriptsize$\pm$.75} & 61.31{\scriptsize$\pm$.72} & 39.41{\scriptsize$\pm$.58} & 22.96{\scriptsize$\pm$.88} & 63.32{\scriptsize$\pm$.76} & 44.00{\scriptsize$\pm$.60} & 37.33{\scriptsize$\pm$.68} & 57.77{\scriptsize$\pm$.69} \\
    \cmidrule{2-11}
    &SL & Default & 91.32{\scriptsize$\pm$.49} & 84.00{\scriptsize$\pm$.56} & 40.84{\scriptsize$\pm$.56} & 27.01{\scriptsize$\pm$.44} & 72.92{\scriptsize$\pm$.66} & 53.26{\scriptsize$\pm$.73} & 44.39{\scriptsize$\pm$.66} &  \underline{58.10}{\scriptsize$\pm$.78} \\ \cmidrule{2-11}
    &\multirow{2}{*}{SSL} & SimCLR & \textbf{97.24}{\scriptsize$\pm$.33} &  \textbf{92.36}{\scriptsize$\pm$.37} & 46.76{\scriptsize$\pm$.61} &  25.62{\scriptsize$\pm$.43} &  \textbf{78.39}{\scriptsize$\pm$.61} & \underline{59.77}{\scriptsize$\pm$.82} &   45.60{\scriptsize$\pm$.72} &  47.69{\scriptsize$\pm$.77} \\
    \rotatebox[origin=c]{90}{\makecell[l]{\!\!\!\!\!\!Single-Stage\!\!\!\!\!\!\!\!\!\!\!\!\!\!\!\!\!\!\!\!\!\!\!\!\!}}&& BYOL & 94.53{\scriptsize$\pm$.41} &  86.55{\scriptsize$\pm$.50} &  45.99{\scriptsize$\pm$.63} &  \underline{27.71}{\scriptsize$\pm$.44} &  68.14{\scriptsize$\pm$.68} &  55.31{\scriptsize$\pm$.71} &  43.92{\scriptsize$\pm$.70} & 49.34{\scriptsize$\pm$.76} \\
    \cmidrule{2-11}
    &\multirow{2}{*}{MSL} & \,\,SimCLR\,\, &  \underline{96.59}{\scriptsize$\pm$.35} &  \underline{90.34}{\scriptsize$\pm$.34} & \textbf{48.78}{\scriptsize$\pm$.62} &  26.69{\scriptsize$\pm$.44}  &  \underline{78.17}{\scriptsize$\pm$.61} & 59.48{\scriptsize$\pm$.82} &  \underline{48.36}{\scriptsize$\pm$.75} & 56.63{\scriptsize$\pm$.78}\\
    && BYOL & 93.71{\scriptsize$\pm$.41} & 87.21{\scriptsize$\pm$.48} & \underline{48.63}{\scriptsize$\pm$.66} & \textbf{29.86}{\scriptsize$\pm$.47}  & 75.16{\scriptsize$\pm$.64} & \textbf{63.45}{\scriptsize$\pm$.81} & \textbf{53.33}{\scriptsize$\pm$.76} &  \textbf{60.66}{\scriptsize$\pm$.77}\\
    \bottomrule
    \end{tabular}\vspace*{0.05cm}}\\
{\small (a) Performance comparison for single-stage schemes.} \\
\resizebox{\linewidth}{!}{
\begin{tabular}{c|c|c|p{13mm}p{13mm}p{13mm}p{13mm}|p{12mm}p{12mm}p{12mm}p{12mm}} \toprule
    &\multirow{2}{*}{\!SL$\rightarrow$SSL\!} & SimCLR & \,\textbf{96.84}{\scriptsize$\pm$.40}\, & \textbf{94.51}{\scriptsize$\pm$.32} & 48.23{\scriptsize$\pm$.59} & 24.59{\scriptsize$\pm$.42} & \textbf{81.52}{\scriptsize$\pm$.56} & 64.37{\scriptsize$\pm$.81} & 50.72{\scriptsize$\pm$.80} & 55.06{\scriptsize$\pm$.84} \\
    && BYOL & 95.87{\scriptsize$\pm$.35} & 90.01{\scriptsize$\pm$.43} & \textbf{50.33}{\scriptsize$\pm$.67} & \underline{29.94}{\scriptsize$\pm$.48} & 75.83{\scriptsize$\pm$.62} & \underline{64.93}{\scriptsize$\pm$.79} & \underline{55.46}{\scriptsize$\pm$.79} & 59.78{\scriptsize$\pm$.81} \\
    \cmidrule{2-11}
    \rotatebox[origin=c]{90}{\makecell[l]{\!\!\!\!\!\!\!\!\!\!\!\!\!\!Two-Stage\!\!\!\!\!\!\!\!\!\!\!\!\!\!}}&\multirow{2}{*}{\!SL$\rightarrow$MSL\!} & SimCLR & \underline{96.67}{\scriptsize$\pm$.33} & \underline{90.18}{\scriptsize$\pm$.42} & 47.24{\scriptsize$\pm$.63} & 26.47{\scriptsize$\pm$.44} & \underline{79.95}{\scriptsize$\pm$.61} & 61.34{\scriptsize$\pm$.80} & 52.74{\scriptsize$\pm$.79} &  61.33{\scriptsize$\pm$.80} \\
    && BYOL &  94.50{\scriptsize$\pm$.39} & 87.96{\scriptsize$\pm$.48} & \underline{49.36}{\scriptsize$\pm$.66} & \textbf{30.23}{\scriptsize$\pm$.50} & 76.67{\scriptsize$\pm$.63} & \textbf{65.41}{\scriptsize$\pm$.78} & \textbf{58.62}{\scriptsize$\pm$.78} & \textbf{66.20}{\scriptsize$\pm$.78} \\
    \cmidrule{2-11}
    & \multirow{2}{*}{\!SL$\rightarrow$MSL$^+$\!\!\!} & STARTUP & 93.02{\scriptsize$\pm$.45} & 82.29{\scriptsize$\pm$.60} & 47.22{\scriptsize$\pm$.61} & 26.94{\scriptsize$\pm$.44} & 69.56{\scriptsize$\pm$.66} & 55.40{\scriptsize$\pm$.78} & 46.73{\scriptsize$\pm$.73} & 60.00{\scriptsize$\pm$.78} \\
    && DynDistill & 95.54{\scriptsize$\pm$.38} & 89.07{\scriptsize$\pm$.47} &  \underline{49.36}{\scriptsize$\pm$.59} & 28.31{\scriptsize$\pm$.46} & 70.98{\scriptsize$\pm$.94} & 58.63{\tiny$\pm$1.14} & 51.98{\tiny$\pm$1.18}& \underline{62.86}{\tiny$\pm$1.06} \\
    \bottomrule
\end{tabular}
\vspace*{-0.05cm}}\\
{\small (b) Performance comparison for two-stage schemes.}
\vspace*{-0.4cm}
\end{table*}

\clearpage
\subsection{Source Dataset: ImageNet (ResNet50)}\label{appx:summary_imagenet_resnet50}
Table \ref{tab:summary_in_1shot_resnet50} and Table \ref{tab:summary_in_5shot_resnet50} describe 5-way 1-shot and 5-way 5-shot CD-FSL performance when ResNet50 is used as a backbone and ImageNet is used as the source dataset, respectively. We find that the observations in our paper also hold for ResNet50.

\begin{table*}[!h]
\caption{5-way 1-shot CD-FSL performance\,(\%) of the models pre-trained by SL, SSL, and MSL including their two-stage versions. ResNet50 is used as the backbone model, and ImageNet is used as the source data for SL. The balancing coefficient $\gamma$ in Eq.\,\eqref{eq:loss_msl} of MSL is set to be 0.875. The {best} results are marked in bold and the second best are underlined.}\label{tab:summary_in_1shot_resnet50}
\vspace*{0.1cm}
\centering
\footnotesize\addtolength{\tabcolsep}{-0.8pt}
\resizebox{\linewidth}{!}{
\begin{tabular}{c|c|c|cccc|cccc}
    \toprule
    &\,\,Pre-train\,\, & \multirow{2}{*}{Method} & \multicolumn{4}{c|}{Small Similarity} & \multicolumn{4}{c}{Large Similarity} \\
    & Scheme & & \!\!CropDisease\!\! & \!\!EuroSAT\!\! & ISIC & ChestX & Places & Plantae & Cars & CUB \\
    \midrule
    &SL & Default &  73.74{\scriptsize$\pm$.90} &  67.38{\scriptsize$\pm$.88} &  29.00{\scriptsize$\pm$.51} & 22.03{\scriptsize$\pm$.38} & \underline{65.51}{\scriptsize$\pm$.91} &  \underline{46.52}{\scriptsize$\pm$.83} &  \textbf{51.25}{\scriptsize$\pm$.90} &  \textbf{71.83}{\scriptsize$\pm$.95} \\ \cmidrule{2-11}
    &\multirow{2}{*}{SSL} & SimCLR & \textbf{90.40}{\scriptsize$\pm$.77} &  \textbf{81.82}{\scriptsize$\pm$.77} &  \underline{35.26}{\scriptsize$\pm$.62} &  21.73{\scriptsize$\pm$.41} & 63.98{\scriptsize$\pm$.97} &  41.32{\scriptsize$\pm$.80} & 32.91{\scriptsize$\pm$.73} & 35.26{\scriptsize$\pm$.75} \\
    \rotatebox[origin=c]{90}{\makecell[l]{\!\!\!\!\!\!\!\!\!\!\!\!\!\!Single-Stage}\!\!\!\!\!\!\!\!\!\!\!\!\!\!}&& BYOL & 87.17{\scriptsize$\pm$.73} & \underline{72.71}{\scriptsize$\pm$.83} &  34.33{\scriptsize$\pm$.63} & \underline{22.67}{\scriptsize$\pm$.42} & 53.33{\scriptsize$\pm$.84} & 39.34{\scriptsize$\pm$.76} &  31.58{\scriptsize$\pm$.71} & 33.38{\scriptsize$\pm$.68} \\ \cmidrule{2-11}
    &\multirow{2}{*}{MSL} & \,\,SimCLR\,\, & 87.17{\scriptsize$\pm$.73} &  \underline{72.71}{\scriptsize$\pm$.83} &  34.33{\scriptsize$\pm$.63} &     \underline{22.67}{\scriptsize$\pm$.42} & \textbf{68.24}{\scriptsize$\pm$.90} &  45.85{\scriptsize$\pm$.85} &  36.52{\scriptsize$\pm$.77} & 47.53{\scriptsize$\pm$.88} \\
    && BYOL & \underline{87.25}{\scriptsize$\pm$.82} &  72.47{\scriptsize$\pm$.84} &  \textbf{36.68}{\scriptsize$\pm$.67} &  \textbf{23.54}{\scriptsize$\pm$.43} & 62.75{\scriptsize$\pm$.87} &  \textbf{49.20}{\scriptsize$\pm$.88} &  \underline{38.57}{\scriptsize$\pm$.77} & \underline{48.72}{\scriptsize$\pm$.87} \\
    \bottomrule
    \end{tabular}\vspace*{0.05cm}}\\
{\small (a) Performance comparison for single-stage schemes.} \\\vspace*{0.1cm}
\resizebox{\linewidth}{!}{
\begin{tabular}{c|c|c|cccc|cccc} \toprule
    &\multirow{2}{*}{\!SL$\rightarrow$SSL\!} & SimCLR & \underline{92.14}{\scriptsize$\pm$.72} &  \textbf{86.41}{\scriptsize$\pm$.65} &  \underline{36.31}{\scriptsize$\pm$.69} &  21.72{\scriptsize$\pm$.41} & \underline{70.58}{\scriptsize$\pm$.93} &  49.36{\scriptsize$\pm$.91} &  37.49{\scriptsize$\pm$.79} &  43.20{\scriptsize$\pm$.87} \\
    && BYOL & 84.44{\scriptsize$\pm$.97} &  69.11{\scriptsize$\pm$.94} &  35.90{\scriptsize$\pm$.69} &  21.62{\scriptsize$\pm$.40} & 49.05{\scriptsize$\pm$.89} &  37.40{\scriptsize$\pm$.89} &  35.96{\scriptsize$\pm$.75} &  36.95{\scriptsize$\pm$.74} \\ \cmidrule{2-11}
    \rotatebox[origin=c]{90}{\makecell[l]{\!\!\!\!\!\!\!\!\!\!Two-Stage}\!\!\!\!\!\!\!\!\!\!\!\!\!\!\!\!\!\!\!\!}&\multirow{2}{*}{\!SL$\rightarrow$MSL\!} & SimCLR & \textbf{92.62}{\scriptsize$\pm$.62} &  \underline{76.30}{\scriptsize$\pm$.79} &  35.51{\scriptsize$\pm$.67} &  \underline{22.48}{\scriptsize$\pm$.42} & \textbf{73.05}{\scriptsize$\pm$.88} &  \underline{54.08}{\scriptsize$\pm$.94} &  \underline{41.91}{\scriptsize$\pm$.85} &  \underline{61.51}{\scriptsize$\pm$.97} \\
    && BYOL & 91.04{\scriptsize$\pm$.68} &  73.78{\scriptsize$\pm$.81} &  \textbf{37.27}{\scriptsize$\pm$.67} &  \textbf{24.70}{\scriptsize$\pm$.42} & 65.55{\scriptsize$\pm$.86} &  \textbf{59.08}{\scriptsize$\pm$.94} &  \textbf{49.65}{\scriptsize$\pm$.89} &  \textbf{66.36}{\scriptsize$\pm$.90} \\
    \bottomrule
\end{tabular}
\vspace*{-0.05cm}}\\
{\small (b) Performance comparison for two-stage schemes.}
\end{table*}

\begin{table*}[!h]
\caption{5-way 5-shot CD-FSL performance\,(\%) of the models pre-trained by SL, SSL, and MSL including their two-stage versions. ResNet50 is used as the backbone model, and ImageNet is used as the source data for SL. The balancing coefficient $\gamma$ in Eq.\,\eqref{eq:loss_msl} of MSL is set to be 0.875. The {best} results are marked in bold and the second best are underlined.}\label{tab:summary_in_5shot_resnet50}
\vspace*{0.1cm}
\centering
\footnotesize\addtolength{\tabcolsep}{-0.8pt}
\resizebox{\linewidth}{!}{
\begin{tabular}{c|c|c|cccc|cccc}
    \toprule
    &\,\,Pre-train\,\, & \multirow{2}{*}{Method} & \multicolumn{4}{c|}{Small Similarity} & \multicolumn{4}{c}{Large Similarity} \\
    & Scheme & & \!\!CropDisease\!\! & \!\!EuroSAT\!\! & ISIC & ChestX & Places & Plantae & Cars & CUB \\
    \midrule
    &SL & Default & 92.65{\scriptsize$\pm$.47} &  85.95{\scriptsize$\pm$.54} &  42.00{\scriptsize$\pm$.60} &  25.04{\scriptsize$\pm$.43} & \textbf{86.46}{\scriptsize$\pm$.50} &  \underline{66.82}{\scriptsize$\pm$.80} &  \textbf{73.49}{\scriptsize$\pm$.80} &  \textbf{91.06}{\scriptsize$\pm$.54} \\ \cmidrule{2-11}
    &\multirow{2}{*}{SSL} & SimCLR & \underline{97.00}{\scriptsize$\pm$.37} &  \textbf{93.34}{\scriptsize$\pm$.34} &  \underline{46.16}{\scriptsize$\pm$.62} &  24.77{\scriptsize$\pm$.43} & 79.02{\scriptsize$\pm$.62} &  57.62{\scriptsize$\pm$.85} & 43.97{\scriptsize$\pm$.75} & 45.90{\scriptsize$\pm$.77} \\
    \rotatebox[origin=c]{90}{\makecell[l]{\!\!\!\!\!\!\!\!\!\!\!\!\!\!Single-Stage}\!\!\!\!\!\!\!\!\!\!\!\!\!\!}&& BYOL & 96.69{\scriptsize$\pm$.30} &  86.19{\scriptsize$\pm$.50} &  43.44{\scriptsize$\pm$.61} &  \underline{27.21}{\scriptsize$\pm$.45} & 73.73{\scriptsize$\pm$.65} &  58.40{\scriptsize$\pm$.81} &  41.87{\scriptsize$\pm$.71} & 48.33{\scriptsize$\pm$.74} \\ \cmidrule{2-11}
    &\multirow{2}{*}{MSL} & \,\,SimCLR\,\, & 96.83{\scriptsize$\pm$.34} &  89.02{\scriptsize$\pm$.46} &  45.49{\scriptsize$\pm$.61} &  26.15{\scriptsize$\pm$.44} & \underline{84.97}{\scriptsize$\pm$.54} &  63.69{\scriptsize$\pm$.83} &  51.11{\scriptsize$\pm$.81} & 65.23{\scriptsize$\pm$.80} \\
    && BYOL & \textbf{97.27}{\scriptsize$\pm$.31} &  \underline{90.46}{\scriptsize$\pm$.41} &  \textbf{49.06}{\scriptsize$\pm$.69} &  \textbf{28.98}{\scriptsize$\pm$.47} & 82.19{\scriptsize$\pm$.57} &  \textbf{68.57}{\scriptsize$\pm$.83} &  \underline{56.52}{\scriptsize$\pm$.81} & \underline{66.99}{\scriptsize$\pm$.78} \\
    \bottomrule
    \end{tabular}\vspace*{0.05cm}}\\
{\small (a) Performance comparison for single-stage schemes.} \\\vspace*{0.1cm}
\resizebox{\linewidth}{!}{
\begin{tabular}{c|c|c|cccc|cccc} \toprule
    &\multirow{2}{*}{\!SL$\rightarrow$SSL\!} & SimCLR & 97.54{\scriptsize$\pm$.34} &  \textbf{95.57}{\scriptsize$\pm$.27} &  48.95{\scriptsize$\pm$.63} &  24.75{\scriptsize$\pm$.44} & 85.39{\scriptsize$\pm$.51} &  66.46{\scriptsize$\pm$.85} &  50.54{\scriptsize$\pm$.85} &  58.48{\scriptsize$\pm$.89} \\
    && BYOL & 96.20{\scriptsize$\pm$.38} &  91.94{\scriptsize$\pm$.41} &  \underline{52.00}{\scriptsize$\pm$.66} &  26.42{\scriptsize$\pm$.46} & 77.78{\scriptsize$\pm$.61} &  66.95{\scriptsize$\pm$.85} &  52.90{\scriptsize$\pm$.81} &  54.37{\scriptsize$\pm$.78} \\ \cmidrule{2-11}
    \rotatebox[origin=c]{90}{\makecell[l]{\!\!\!\!\!\!\!\!\!\!Two-Stage}\!\!\!\!\!\!\!\!\!\!\!\!\!\!\!\!\!\!\!\!}&\multirow{2}{*}{\!SL$\rightarrow$MSL\!} & SimCLR & \textbf{98.18}{\scriptsize$\pm$.26} &  91.48{\scriptsize$\pm$.38} &  49.34{\scriptsize$\pm$.64} &  \underline{26.62}{\scriptsize$\pm$.45} & \textbf{87.87}{\scriptsize$\pm$.49} &  \underline{72.39}{\scriptsize$\pm$.81} &  \underline{59.86}{\scriptsize$\pm$.85} &  \underline{81.46}{\scriptsize$\pm$.76} \\
    && BYOL & \underline{97.80}{\scriptsize$\pm$.27} &  \underline{92.14}{\scriptsize$\pm$.35} &  \textbf{53.04}{\scriptsize$\pm$.67} &  \textbf{30.91}{\scriptsize$\pm$.51} & \underline{85.72}{\scriptsize$\pm$.52} &  \textbf{78.47}{\scriptsize$\pm$.73} &  \textbf{72.98}{\scriptsize$\pm$.80} &  \textbf{85.55}{\scriptsize$\pm$.68} \\
    \bottomrule
\end{tabular}}
\vspace*{-0.05cm}\\
{\small (b) Performance comparison for two-stage schemes.}
\vspace*{-0.4cm}
\end{table*}

\clearpage
\subsection{Source Dataset: ImageNet (20- and 50-shots)}\label{appx:summary_imagenet_}
Table \ref{tab:summary_in_20shot} and Table \ref{tab:summary_in_50shot} describe 5-way 20-shot and 5-way 50-shot CD-FSL performance when ResNet18 is used as a backbone and ImageNet is used as the source dataset, respectively. We find that the results are consistent with our main analysis.
For target datasets that have small similarity to the source dataset, it remains beneficial to perform SSL pre-training on the unlabeled target data to adapt to target domain features, compared to SL on source (Obs. \ref{obs:obs4_1}).
For target datasets with large similarity, the relative benefit of SSL on target data is larger when few-shot difficulty is low (Obs. \ref{obs:obs5_3}). We note that SL performance significantly benefits from large $k$ when similarity to the source domain is high.
Furthermore, the observations about joint synergy via MSL (Obs. \ref{obs:obs6_1}) and sequential synergy via two-stage pre-training (Obs. \ref{obs:obs6_2}) consistently hold.

\begin{table*}[!h]
\caption{5-way 20-shot CD-FSL performance\,(\%) of the models pre-trained by SL, SSL, and MSL including their two-stage versions. ResNet18 is used as the backbone model, and ImageNet is used as the source data for SL. The balancing coefficient $\gamma$ in Eq.\,\eqref{eq:loss_msl} of MSL is set to be 0.875. The {best} results are marked in bold and the second best are underlined.}\label{tab:summary_in_20shot}
\vspace*{0.1cm}
\centering
\footnotesize\addtolength{\tabcolsep}{-0.8pt}
\resizebox{\linewidth}{!}{
\begin{tabular}{c|c|c|cccc|cccc}
    \toprule
    &\,\,Pre-train\,\, & \multirow{2}{*}{Method} & \multicolumn{4}{c|}{Small Similarity} & \multicolumn{4}{c}{Large Similarity} \\
    & Scheme & & \!\!CropDisease\!\! & \!\!EuroSAT\!\! & ISIC & ChestX & Places & Plantae & Cars & CUB \\
    \midrule
    &SL & Default & 97.21{\scriptsize$\pm$.25} & 91.66{\scriptsize$\pm$.33} & 53.78{\scriptsize$\pm$.58} & 29.39{\scriptsize$\pm$.42} & \textbf{89.65}{\scriptsize$\pm$.40} & \underline{76.08}{\scriptsize$\pm$.72} & \textbf{82.79}{\scriptsize$\pm$.60} & \textbf{94.48}{\scriptsize$\pm$.38} \\ \cmidrule{2-11}
    &\multirow{2}{*}{SSL} & SimCLR & 97.88{\scriptsize$\pm$.29} & \textbf{96.02}{\scriptsize$\pm$.23} & 55.94{\scriptsize$\pm$.60} & 28.36{\scriptsize$\pm$.43} & 83.90{\scriptsize$\pm$.50} & 64.91{\scriptsize$\pm$.80} & 49.64{\scriptsize$\pm$.78} & 55.41{\scriptsize$\pm$.77} \\
    \rotatebox[origin=c]{90}{\makecell[l]{\!\!\!\!\!\!\!\!\!\!\!\!\!\!Single-Stage}\!\!\!\!\!\!\!\!\!\!\!\!\!\!}&& BYOL & 98.76{\scriptsize$\pm$.15} & 94.60{\scriptsize$\pm$.28} & \underline{58.52}{\scriptsize$\pm$.55} & \underline{35.26}{\scriptsize$\pm$.47} & 81.82{\scriptsize$\pm$.52} & 72.32{\scriptsize$\pm$.75} & 61.06{\scriptsize$\pm$.79} & 64.38{\scriptsize$\pm$.78} \\ \cmidrule{2-11}
    &\multirow{2}{*}{MSL} & SimCLR & \underline{98.78}{\scriptsize$\pm$.16} & 94.71{\scriptsize$\pm$.25} & 56.39{\scriptsize$\pm$.60} & 31.46{\scriptsize$\pm$.43} & \underline{88.64}{\scriptsize$\pm$.42} & 75.55{\scriptsize$\pm$.73} & \underline{66.05}{\scriptsize$\pm$.74} & 75.89{\scriptsize$\pm$.68} \\
    && BYOL & \textbf{98.97}{\scriptsize$\pm$.13} & \underline{95.03}{\scriptsize$\pm$.24} & \textbf{60.54}{\scriptsize$\pm$.62} & \textbf{37.35}{\scriptsize$\pm$.49} & 88.06{\scriptsize$\pm$.43} & \textbf{78.40}{\scriptsize$\pm$.69} & 59.74{\scriptsize$\pm$.72} & \underline{80.23}{\scriptsize$\pm$.64} \\
    \bottomrule
    \end{tabular}\vspace*{0.05cm}}\\
{\small (a) Performance comparison for single-stage schemes.} \\\vspace*{0.1cm}
\resizebox{\linewidth}{!}{
\begin{tabular}{c|c|c|cccc|cccc} \toprule
    &\multirow{2}{*}{\!SL$\rightarrow$SSL\!}  & SimCLR & 98.27{\scriptsize$\pm$.26} & \textbf{96.77}{\scriptsize$\pm$.21} & 58.28{\scriptsize$\pm$.57} & 28.57{\scriptsize$\pm$.42} & 87.88{\scriptsize$\pm$.42} & 71.52{\scriptsize$\pm$.77} & 59.40{\scriptsize$\pm$.82} & 67.79{\scriptsize$\pm$.80} \\
    && BYOL & \textbf{99.20}{\scriptsize$\pm$.12} & \underline{96.60}{\scriptsize$\pm$.19} & \underline{60.81}{\scriptsize$\pm$.60} & \underline{36.00}{\scriptsize$\pm$.48} & 86.67{\scriptsize$\pm$.43} & 78.91{\scriptsize$\pm$.67} & 70.90{\scriptsize$\pm$.76} & 73.95{\scriptsize$\pm$.71} \\ \cmidrule{2-11}
    \rotatebox[origin=c]{90}{\makecell[l]{\!\!\!\!\!\!\!\!\!\!Two-Stage}\!\!\!\!\!\!\!\!\!\!\!\!\!\!\!\!\!\!\!\!}&\multirow{2}{*}{\!SL$\rightarrow$MSL\!} & SimCLR & 98.96{\scriptsize$\pm$.15} & 95.58{\scriptsize$\pm$.22} & 58.87{\scriptsize$\pm$.59} & 31.78{\scriptsize$\pm$.43} & \textbf{90.69}{\scriptsize$\pm$.38} & \underline{79.85}{\scriptsize$\pm$.67} & \underline{76.54}{\scriptsize$\pm$.70} & \underline{90.25}{\scriptsize$\pm$.48} \\
    && BYOL & \underline{99.10}{\scriptsize$\pm$.14} & 95.82{\scriptsize$\pm$.21} & \textbf{62.43}{\scriptsize$\pm$.59} & \textbf{38.48}{\scriptsize$\pm$.47} & \underline{89.62}{\scriptsize$\pm$.40} & \textbf{84.00}{\scriptsize$\pm$.61} & \textbf{84.39}{\scriptsize$\pm$.60} & \textbf{91.59}{\scriptsize$\pm$.46} \\
    \bottomrule
\end{tabular}
\vspace*{-0.05cm}}\\
{\small (b) Performance comparison for two-stage schemes.}
\end{table*}

\begin{table*}[!h]
\caption{5-way 50-shot CD-FSL performance\,(\%) of the models pre-trained by SL, SSL, and MSL including their two-stage versions. ResNet18 is used as the backbone model, and ImageNet is used as the source data for SL. The balancing coefficient $\gamma$ in Eq.\,\eqref{eq:loss_msl} of MSL is set to be 0.875. The {best} results are marked in bold and the second best are underlined.}\label{tab:summary_in_50shot}
\vspace*{0.1cm}
\centering
\footnotesize\addtolength{\tabcolsep}{-0.8pt}
\resizebox{\linewidth}{!}{
\begin{tabular}{c|c|c|cccc|cccc}
    \toprule
    &\,\,Pre-train\,\, & \multirow{2}{*}{Method} & \multicolumn{4}{c|}{Small Similarity} & \multicolumn{4}{c}{Large Similarity} \\
    & Scheme & & \!\!CropDisease\!\! & \!\!EuroSAT\!\! & ISIC & ChestX & Places & Plantae & Cars & CUB \\
    \midrule
    &SL & Default & 98.19{\scriptsize$\pm$.17} & 93.70{\scriptsize$\pm$.27} & 60.42{\scriptsize$\pm$.55} & 33.27{\scriptsize$\pm$.44} & \textbf{91.49}{\scriptsize$\pm$.34} & \underline{80.96}{\scriptsize$\pm$.61} & \textbf{89.79}{\scriptsize$\pm$.44} & \textbf{98.18}{\scriptsize$\pm$.17} \\ \cmidrule{2-11}
    &\multirow{2}{*}{SSL} & SimCLR & 97.99{\scriptsize$\pm$.28} & \textbf{96.35}{\scriptsize$\pm$.22} & 58.40{\scriptsize$\pm$.56} & 30.61{\scriptsize$\pm$.43} & 84.44{\scriptsize$\pm$.47} & 66.60{\scriptsize$\pm$.74} & 53.17{\scriptsize$\pm$.71} & 62.66{\scriptsize$\pm$.73} \\
    \rotatebox[origin=c]{90}{\makecell[l]{\!\!\!\!\!\!\!\!\!\!\!\!\!\!Single-Stage}\!\!\!\!\!\!\!\!\!\!\!\!\!\!}&& BYOL & 99.04{\scriptsize$\pm$.13} & 95.63{\scriptsize$\pm$.24} & \underline{62.74}{\scriptsize$\pm$.55} & \underline{39.36}{\scriptsize$\pm$.49} & 84.26{\scriptsize$\pm$.46} & 77.20{\scriptsize$\pm$.68} & 70.37{\scriptsize$\pm$.68} & 81.13{\scriptsize$\pm$.57} \\ \cmidrule{2-11}
    &\multirow{2}{*}{MSL} & SimCLR & \underline{99.09}{\scriptsize$\pm$.13} & 95.93{\scriptsize$\pm$.21} & 60.33{\scriptsize$\pm$.57} & 35.52{\scriptsize$\pm$.44} & \underline{90.30}{\scriptsize$\pm$.35} & 80.41{\scriptsize$\pm$.61} & \underline{76.49}{\scriptsize$\pm$.66} & 89.97{\scriptsize$\pm$.44} \\
    && BYOL & \textbf{99.27}{\scriptsize$\pm$.10} & \underline{96.00}{\scriptsize$\pm$.21} & \textbf{64.64}{\scriptsize$\pm$.58} & \textbf{41.88}{\scriptsize$\pm$.50} & 90.02{\scriptsize$\pm$.36} & \textbf{82.68}{\scriptsize$\pm$.60} & 70.34{\scriptsize$\pm$.68} & \underline{92.40}{\scriptsize$\pm$.37} \\
    \bottomrule
    \end{tabular}\vspace*{0.05cm}}\\
{\small (a) Performance comparison for single-stage schemes.} \\\vspace*{0.1cm}
\resizebox{\linewidth}{!}{
\begin{tabular}{c|c|c|cccc|cccc} \toprule
    &\multirow{2}{*}{\!SL$\rightarrow$SSL\!} & SimCLR & 98.29{\scriptsize$\pm$.26} & \underline{97.04}{\scriptsize$\pm$.19} & 61.40{\scriptsize$\pm$.54} & 30.67{\scriptsize$\pm$.43} & 88.44{\scriptsize$\pm$.39} & 73.16{\scriptsize$\pm$.72} & 63.73{\scriptsize$\pm$.77} & 75.53{\scriptsize$\pm$.68} \\
    && BYOL & \textbf{99.44}{\scriptsize$\pm$.09} & \textbf{97.35}{\scriptsize$\pm$.16} & \underline{66.71}{\scriptsize$\pm$.54} & \underline{40.93}{\scriptsize$\pm$.50} & 88.74{\scriptsize$\pm$.37} & 82.58{\scriptsize$\pm$.60} & 81.17{\scriptsize$\pm$.62} & 88.85{\scriptsize$\pm$.45} \\ \cmidrule{2-11}
    \rotatebox[origin=c]{90}{\makecell[l]{\!\!\!\!\!\!\!\!\!\!Two-Stage}\!\!\!\!\!\!\!\!\!\!\!\!\!\!\!\!\!\!\!\!}&\multirow{2}{*}{\!SL$\rightarrow$MSL\!} & SimCLR & 99.34{\scriptsize$\pm$.12} & 96.66{\scriptsize$\pm$.18} & 64.44{\scriptsize$\pm$.57} & 35.61{\scriptsize$\pm$.45} & \textbf{92.09}{\scriptsize$\pm$.33} & \underline{83.88}{\scriptsize$\pm$.56} & \underline{85.50}{\scriptsize$\pm$.55} & \underline{96.82}{\scriptsize$\pm$.24} \\
    && BYOL & \underline{99.37}{\scriptsize$\pm$.10} & 96.81{\scriptsize$\pm$.18} & \textbf{67.90}{\scriptsize$\pm$.57} & \textbf{43.59}{\scriptsize$\pm$.49} & \underline{91.36}{\scriptsize$\pm$.34} & \textbf{87.59}{\scriptsize$\pm$.49} & \textbf{91.28}{\scriptsize$\pm$.43} & \textbf{97.15}{\scriptsize$\pm$.23} \\
    \bottomrule
\end{tabular}}
\vspace*{-0.05cm}\\
{\small (b) Performance comparison for two-stage schemes.}
\vspace*{-0.4cm}
\end{table*}

\newpage
\section{Additional Pre-training Schemes}\label{appx:additional_pre}

\subsection{Alternative Two-Stage Schemes}

In this section, we study alternative two-stage pre-training schemes. Namely, we consider MSL$\rightarrow$SSL and SSL$\rightarrow$MSL.
By default, we use $\gamma=0.875$ as the balancing hyperparameter for MSL within each scheme. However, for MSL$\rightarrow$SSL, we also consider $\gamma=0.125$ as a middle-ground between SL$\rightarrow$SSL and MSL$\rightarrow$SSL (with $\gamma=0.875$).

Table \ref{tab:two_stage_schemes_1shot} and Table \ref{tab:two_stage_schemes_5shot} describe the few-show performances of the additional pre-training schemes (in the three lowermost rows), with other methods displayed for ease of comparison.
We observe that MSL$\rightarrow$SSL with either choice of $\gamma$ can achieve the best performance for target datasets with small similarity to the source dataset, e.g., CropDisease, ISIC, ChestX.
On the other hand, SSL$\rightarrow$MSL generally underperforms other methods, unable to achieve best performance for any of the target datasets.
We posit that this is because the latter stage of pre-training (MSL) is closer to the source dataset compared to the former (SSL), thus learning undesirable source information before few-shot adaptation.
For the second stage of MSL$\rightarrow$SSL, we used the SGD optimizer for BYOL as well as SimCLR, to match the optimizer used in the first stage, MSL.

\begin{table*}[!h]
\caption{5-way 1-shot CD-FSL performance\,(\%) of the models according to different two-stage pre-training schemes. ResNet18 is used as the backbone model, and ImageNet is used as the source data. If not specified otherwise, the balancing coefficient $\gamma$ in Eq.\,\eqref{eq:loss_msl} of MSL is set to be 0.875. The {best} results are marked in bold.}\label{tab:two_stage_schemes_1shot}
\vspace*{0.1cm}
\centering
\footnotesize\addtolength{\tabcolsep}{-3.0pt}
\resizebox{\linewidth}{!}{
\begin{tabular}{c|c|cccc|cccc}
    \toprule
    \,\,Pre-train\,\, & \multirow{2}{*}{Method} & \multicolumn{4}{c|}{Small Similarity} & \multicolumn{4}{c}{Large Similarity} \\
    Scheme & & \!\!CropDisease\!\! & \!\!EuroSAT\!\! & ISIC & ChestX & Places & Plantae & Cars & CUB \\
    \midrule
    \multirow{2}{*}{SL $\rightarrow$ SSL} & SimCLR& 92.24{\scriptsize$\pm$.70} & \textbf{86.51}{\scriptsize$\pm$.67} & 36.11{\scriptsize$\pm$.67} & 21.75{\scriptsize$\pm$.41} & \textbf{71.05}{\scriptsize$\pm$.92} & 49.02{\scriptsize$\pm$.91} & 37.43{\scriptsize$\pm$.79} & 42.40{\scriptsize$\pm$.85} \\
    & BYOL & 87.64{\scriptsize$\pm$.70} & 74.05{\scriptsize$\pm$.84} & 35.62{\scriptsize$\pm$.65} & 23.01{\scriptsize$\pm$.43} & 58.12{\scriptsize$\pm$.87} & 48.28{\scriptsize$\pm$.88} & 38.23{\scriptsize$\pm$.75} & 42.48{\scriptsize$\pm$.82} \\
    \midrule
    \multirow{2}{*}{SL $\rightarrow$ MSL} & SimCLR & 91.46{\scriptsize$\pm$.66} & 77.62{\scriptsize$\pm$.76} & 34.46{\scriptsize$\pm$.64} & 22.50{\scriptsize$\pm$.41} & 69.50{\scriptsize$\pm$.87} & 51.27{\scriptsize$\pm$.91} & 40.39{\scriptsize$\pm$.82} & 62.12{\scriptsize$\pm$.93} \\
    & BYOL & 88.37{\scriptsize$\pm$.73} & 71.54{\scriptsize$\pm$.78} & 36.08{\scriptsize$\pm$.63} & 24.42{\scriptsize$\pm$.45} & 63.40{\scriptsize$\pm$.86} & \textbf{53.65}{\scriptsize$\pm$.88} & \textbf{46.62}{\scriptsize$\pm$.85} & \textbf{64.33}{\scriptsize$\pm$.93} \\
    \midrule
    {MSL $\rightarrow$ SSL} & SimCLR & 92.18{\scriptsize$\pm$.72} & 86.27{\scriptsize$\pm$.68} & 36.20{\scriptsize$\pm$.68} & 21.55{\scriptsize$\pm$.41} & 67.12{\scriptsize$\pm$.93} & 46.61{\scriptsize$\pm$.88} & 34.86{\scriptsize$\pm$.76} & 41.11{\scriptsize$\pm$.82} \\
    ($\gamma=0.125$) & BYOL & 84.33{\scriptsize$\pm$.75} & 65.69{\scriptsize$\pm$.83} & 36.13{\scriptsize$\pm$.66} & 24.60{\scriptsize$\pm$.44} & 48.03{\scriptsize$\pm$.73} & 45.15{\scriptsize$\pm$.78} & 40.62{\scriptsize$\pm$.77} & 41.00{\scriptsize$\pm$.76} \\
    \midrule
    {MSL $\rightarrow$ SSL} & SimCLR & \textbf{92.47}{\scriptsize$\pm$.69} & 85.03{\scriptsize$\pm$.69} & 36.22{\scriptsize$\pm$.67} & 21.73{\scriptsize$\pm$.41} & 67.74{\scriptsize$\pm$.94} & 46.80{\scriptsize$\pm$.86} & 35.08{\scriptsize$\pm$.77} & 38.94{\scriptsize$\pm$.80} \\
    ($\gamma=0.875$) & BYOL & 92.09{\scriptsize$\pm$.66} & 52.65{\scriptsize$\pm$.90} & \textbf{37.51}{\scriptsize$\pm$.65} & \textbf{24.73}{\scriptsize$\pm$.44} & 46.13{\scriptsize$\pm$.72} & 45.67{\scriptsize$\pm$.81} & 38.80{\scriptsize$\pm$.80} & 40.75{\scriptsize$\pm$.76} \\
    \midrule
    \multirow{2}{*}{SSL $\rightarrow$ MSL} & SimCLR & 84.42{\scriptsize$\pm$.75} & 73.75{\scriptsize$\pm$.82} & 33.80{\scriptsize$\pm$.60} & 22.60{\scriptsize$\pm$.42} & 62.63{\scriptsize$\pm$.88} & 45.21{\scriptsize$\pm$.82} & 38.51{\scriptsize$\pm$.73} & 53.52{\scriptsize$\pm$.89} \\
    & BYOL & 78.26{\scriptsize$\pm$.82} & 70.28{\scriptsize$\pm$.78} & 35.37{\scriptsize$\pm$.64} & 23.66{\scriptsize$\pm$.41} & 60.88{\scriptsize$\pm$.84} & 45.54{\scriptsize$\pm$.84} & 40.27{\scriptsize$\pm$.76} & 54.81{\scriptsize$\pm$.86} \\
    \bottomrule
    \end{tabular}}
    \vspace*{0.05cm}
\end{table*}

\begin{table*}[!h]
\caption{5-way 5-shot CD-FSL performance\,(\%) of the models according to different two-stage pre-training schemes. ResNet18 is used as the backbone model, and ImageNet is used as the source data. If not specified otherwise, the balancing coefficient $\gamma$ in Eq.\,\eqref{eq:loss_msl} of MSL is set to be 0.875. The {best} results are marked in bold.}\label{tab:two_stage_schemes_5shot}
\vspace*{0.1cm}
\centering
\footnotesize\addtolength{\tabcolsep}{-3.0pt}
\resizebox{\linewidth}{!}{
\begin{tabular}{c|c|cccc|cccc}
    \toprule
    \,\,Pre-train\,\, & \multirow{2}{*}{Method} & \multicolumn{4}{c|}{Small Similarity} & \multicolumn{4}{c}{Large Similarity} \\
    Scheme & & \!\!CropDisease\!\! & \!\!EuroSAT\!\! & ISIC & ChestX & Places & Plantae & Cars & CUB \\
    \midrule
    \multirow{2}{*}{SL $\rightarrow$ SSL} & SimCLR & 97.88{\scriptsize$\pm$.30} & \textbf{95.28}{\scriptsize$\pm$.27} & 48.38{\scriptsize$\pm$.60} & 25.25{\scriptsize$\pm$.44} & 84.40{\scriptsize$\pm$.53} & 66.35{\scriptsize$\pm$.82} & 51.31{\scriptsize$\pm$.84} & 57.11{\scriptsize$\pm$.88} \\
    & BYOL & 97.58{\scriptsize$\pm$.26} & 91.82{\scriptsize$\pm$.39} & 49.32{\scriptsize$\pm$.63} & 28.27{\scriptsize$\pm$.48} & 78.87{\scriptsize$\pm$.60} & 67.83{\scriptsize$\pm$.82} & 54.70{\scriptsize$\pm$.84} & 60.60{\scriptsize$\pm$.82} \\
    \midrule
    \multirow{2}{*}{SL $\rightarrow$ MSL} & SimCLR & 97.49{\scriptsize$\pm$.30} & 91.70{\scriptsize$\pm$.35} & 47.43{\scriptsize$\pm$.62} & 26.24{\scriptsize$\pm$.44} & \textbf{85.76}{\scriptsize$\pm$.52} & 69.24{\scriptsize$\pm$.81} & 58.97{\scriptsize$\pm$.82} & 81.51{\scriptsize$\pm$.72} \\
    & BYOL & 97.09{\scriptsize$\pm$.31} & 90.89{\scriptsize$\pm$.40} & 50.72{\scriptsize$\pm$.67} & 30.20{\scriptsize$\pm$.48} & 83.29{\scriptsize$\pm$.55} & \textbf{74.16}{\scriptsize$\pm$.77} & \textbf{68.87}{\scriptsize$\pm$.80} & \textbf{84.34}{\scriptsize$\pm$.67} \\
    \midrule
    {MSL $\rightarrow$ SSL} & SimCLR & 97.59{\scriptsize$\pm$.34} & 95.26{\scriptsize$\pm$.27} & 50.22{\scriptsize$\pm$.60} & 24.68{\scriptsize$\pm$.44} & 83.91{\scriptsize$\pm$.53} & 64.91{\scriptsize$\pm$.82} & 47.90{\scriptsize$\pm$.81} & 56.02{\scriptsize$\pm$.87} \\
    ($\gamma=0.125$) & BYOL & 96.89{\scriptsize$\pm$.30} & 88.96{\scriptsize$\pm$.43} & 52.26{\scriptsize$\pm$.65} & \textbf{31.19}{\scriptsize$\pm$.51} & 72.16{\scriptsize$\pm$.62} & 66.51{\scriptsize$\pm$.77} & 58.57{\scriptsize$\pm$.82} & 59.73{\scriptsize$\pm$.80} \\
    \midrule
    {MSL $\rightarrow$ SSL} & SimCLR & 97.76{\scriptsize$\pm$.32} & 94.87{\scriptsize$\pm$.28} & 48.98{\scriptsize$\pm$.61} & 24.92{\scriptsize$\pm$.43} & 83.84{\scriptsize$\pm$.54} & 64.35{\scriptsize$\pm$.82} & 48.07{\scriptsize$\pm$.82} & 52.65{\scriptsize$\pm$.84} \\
    ($\gamma=0.875$) & BYOL & \textbf{98.41}{\scriptsize$\pm$.24} & 85.96{\scriptsize$\pm$.49} & \textbf{53.22}{\scriptsize$\pm$.66} & 30.67{\scriptsize$\pm$.49} & 72.13{\scriptsize$\pm$.65} & 67.51{\scriptsize$\pm$.81} & 55.17{\scriptsize$\pm$.82} & 59.08{\scriptsize$\pm$.80} \\
    \midrule
    \multirow{2}{*}{SSL $\rightarrow$ MSL} & SimCLR & 95.93{\scriptsize$\pm$.37} & 89.86{\scriptsize$\pm$.41} & 46.38{\scriptsize$\pm$.63} & 26.69{\scriptsize$\pm$.45} & 82.00{\scriptsize$\pm$.58} & 63.38{\scriptsize$\pm$.82} & 56.54{\scriptsize$\pm$.79} & 73.12{\scriptsize$\pm$.76} \\
    & BYOL & 94.76{\scriptsize$\pm$.38} & 88.76{\scriptsize$\pm$.43} & 48.16{\scriptsize$\pm$.66} & 29.23{\scriptsize$\pm$.48} & 81.08{\scriptsize$\pm$.58} & 64.52{\scriptsize$\pm$.80} & 59.02{\scriptsize$\pm$.79} & 74.61{\scriptsize$\pm$.76} \\
    \bottomrule
    \end{tabular}}
    \vspace*{0.05cm}
\end{table*}

\clearpage
\subsection{Longer Training for SSL and MSL}

Table \ref{tab:longer_in_1shot} and Table \ref{tab:longer_in_5shot} describe 5-way 1-shot and 5-way 5-shot CD-FSL performance of SSL and MSL when we increase the number of pre-training epochs from 1000 to 2000.
We consider two schemes for this ablation: \textit{Extended} and \textit{Repeated}.
In the \textit{Extended} scheme, we simply adapt the milestones for the learning rate decay scheduler by doubling them from epochs 400, 600, 800 to 800, 1200, 1600.
For the \textit{Repeated} scheme, we apply the existing decay schedule in a cyclical manner; decaying the learning rate by a factor of 10 at epoch \{400, 600, 800\}, resetting the learning rate at epoch 1000, and again decaying at epoch \{1400, 1600, 1800\}.
This scheme is used to isolate the effects of the learning rate reset that occurs during two-stage pre-training.
Note that during two-stage pre-training, learning rate decay is applied to both stages independently, thus resulting in a jump in learning rate during the transition between stages. We follow the same implementation details as single-stage pre-training, unless explicitly stated.

We find that compared to standard single-stage pre-training, \textit{Extended} pre-training for SSL achieves comparable 1-shot performance and minor overall improvement in 5-shot performance--up to 3.41\% (Cars). On the other hand, MSL exhibits considerable performance increase overall under \textit{Extended} pre-training. This is magnified under large domain similarity, where 5-shot performance improves by up to 18.02\% (Cars). Considering this stark difference between SSL and MSL, we posit that longer training on MSL benefits from further extraction of source features, which are useful for similar target domains. However, we note that two-stage MSL still outperforms longer single-stage MSL, suggesting that SL pre-training is a more effective means of extracting source features. Comparing \textit{Extended} and \textit{Repeated} training, we find no major differences in performance, thus conclude that learning rate reset is not a major contributor in CD-FSL performance.

\begin{table*}[!h]
\caption{5-way 1-shot CD-FSL performance\,(\%) of the models pre-trained by SSL and MSL under two schemes of longer pre-training: \textit{Extended} and \textit{Repeated}. ResNet18 is used as the backbone model, and ImageNet is used as the source data for SL. For the standard single-stage and two-stage pre-training, refer to Table \ref{tab:summary_in_1shot} for comparison. The balancing coefficient $\gamma$ in Eq.\,\eqref{eq:loss_msl} of MSL is set to be 0.875.}\label{tab:longer_in_1shot}
\vspace*{0.1cm}
\centering
\footnotesize\addtolength{\tabcolsep}{-3.0pt}
\resizebox{\linewidth}{!}{
\begin{tabular}{c|c|c|cccc|cccc}
    \toprule
    &\,\,Pre-train\,\, & \multirow{2}{*}{Method} & \multicolumn{4}{c|}{Small Similarity} & \multicolumn{4}{c}{Large Similarity} \\
    & Scheme & & \!\!CropDisease\!\! & \!\!EuroSAT\!\! & ISIC & ChestX & Places & Plantae & Cars & CUB \\
    \midrule
    &\multirow{2}{*}{SSL} & SimCLR & 91.98{\scriptsize$\pm$.76} & 86.34{\scriptsize$\pm$.70} & 35.81{\scriptsize$\pm$.67} & 21.44{\scriptsize$\pm$.40} & 64.96{\scriptsize$\pm$.95} & 43.81{\scriptsize$\pm$.82} & 32.94{\scriptsize$\pm$.72} & 35.92{\scriptsize$\pm$.75} \\
    \rotatebox[origin=c]{90}{\makecell[l]{\!\!\!\!\!\!\!\!\!\!\!\!\!\!Extended}\!\!\!\!\!\!}&& BYOL & 87.13{\scriptsize$\pm$.74} & 64.33{\scriptsize$\pm$.81} & 35.50{\scriptsize$\pm$.62} & 23.11{\scriptsize$\pm$.42} & 50.46{\scriptsize$\pm$.82} & 41.92{\scriptsize$\pm$.75} & 35.78{\scriptsize$\pm$.70} & 36.67{\scriptsize$\pm$.68} \\
    \cmidrule{2-11}
    &\multirow{2}{*}{MSL} & \,\,SimCLR\,\, & 89.30{\scriptsize$\pm$.69} & 74.73{\scriptsize$\pm$.80} & 34.50{\scriptsize$\pm$.60} & 22.32{\scriptsize$\pm$.42} & 65.27{\scriptsize$\pm$.88} & 48.95{\scriptsize$\pm$.87} & 39.70{\scriptsize$\pm$.80} & 54.88{\scriptsize$\pm$.91} \\
    && BYOL & 90.90{\scriptsize$\pm$.67} & 62.86{\scriptsize$\pm$.86} & 36.22{\scriptsize$\pm$.64} & 24.37{\scriptsize$\pm$.42} & 50.94{\scriptsize$\pm$.79} & 42.82{\scriptsize$\pm$.76} & 44.35{\scriptsize$\pm$.85} & 57.18{\scriptsize$\pm$.91} \\
    \midrule
    &\multirow{2}{*}{SSL} & SimCLR & 90.99{\scriptsize$\pm$.77} & 85.96{\scriptsize$\pm$.71} & 35.78{\scriptsize$\pm$.64} & 21.70{\scriptsize$\pm$.41} & 66.26{\scriptsize$\pm$.96} & 44.53{\scriptsize$\pm$.85} & 33.89{\scriptsize$\pm$.74} & 36.86{\scriptsize$\pm$.77} \\
    \rotatebox[origin=c]{90}{\makecell[l]{\!\!\!\!\!\!\!\!\!\!\!\!\!\!Repeated}\!\!\!\!\!\!}&& BYOL & 86.36{\scriptsize$\pm$.78} & 64.95{\scriptsize$\pm$.80} & 35.15{\scriptsize$\pm$.63} & 23.48{\scriptsize$\pm$.43} & 50.21{\scriptsize$\pm$.78} & 41.58{\scriptsize$\pm$.75} & 35.73{\scriptsize$\pm$.72} & 38.74{\scriptsize$\pm$.70} \\
    \cmidrule{2-11}
    &\multirow{2}{*}{MSL} & \,\,SimCLR\,\, & 88.75{\scriptsize$\pm$.69} & 73.47{\scriptsize$\pm$.80} & 33.58{\scriptsize$\pm$.61} & 22.32{\scriptsize$\pm$.42} & 65.84{\scriptsize$\pm$.88} & 48.64{\scriptsize$\pm$.85} & 39.87{\scriptsize$\pm$.79} & 54.44{\scriptsize$\pm$.91} \\
    && BYOL & 89.70{\scriptsize$\pm$.71} & 68.43{\scriptsize$\pm$.82} & 36.76{\scriptsize$\pm$.65} & 24.17{\scriptsize$\pm$.44} & 53.47{\scriptsize$\pm$.81} & 46.20{\scriptsize$\pm$.80} & 43.80{\scriptsize$\pm$.84} & 54.94{\scriptsize$\pm$.84} \\
    \bottomrule
\end{tabular}}
\end{table*}

\begin{table*}[!h]
\caption{5-way 5-shot CD-FSL performance\,(\%) of the models pre-trained by SSL and MSL under two schemes of longer pre-training: \textit{Extended} and \textit{Repeated}. ResNet18 is used as the backbone model, and ImageNet is used as the source data for SL. For the standard single-stage and two-stage pre-training, refer to Table \ref{tab:summary_in_5shot} for comparison. The balancing coefficient $\gamma$ in Eq.\,\eqref{eq:loss_msl} of MSL is set to be 0.875.}\label{tab:longer_in_5shot}
\vspace*{0.1cm}
\centering
\footnotesize\addtolength{\tabcolsep}{-3.0pt}
\resizebox{\linewidth}{!}{
\begin{tabular}{c|c|c|cccc|cccc}
    \toprule
    &\,\,Pre-train\,\, & \multirow{2}{*}{Method} & \multicolumn{4}{c|}{Small Similarity} & \multicolumn{4}{c}{Large Similarity} \\
    & Scheme & & \!\!CropDisease\!\! & \!\!EuroSAT\!\! & ISIC & ChestX & Places & Plantae & Cars & CUB \\
    \midrule
    &\multirow{2}{*}{SSL} & SimCLR & 97.44{\scriptsize$\pm$.36} & 95.14{\scriptsize$\pm$.28} & 49.47{\scriptsize$\pm$.59} & 24.36{\scriptsize$\pm$.42} & 82.21{\scriptsize$\pm$.57} & 60.77{\scriptsize$\pm$.85} & 45.30{\scriptsize$\pm$.75} & 47.57{\scriptsize$\pm$.80} \\
    \rotatebox[origin=c]{90}{\makecell[l]{\!\!\!\!\!\!\!\!\!\!\!\!\!\!Extended}\!\!\!\!\!\!}&& BYOL & 97.29{\scriptsize$\pm$.28} & 89.54{\scriptsize$\pm$.42} & 50.17{\scriptsize$\pm$.64} & 29.16{\scriptsize$\pm$.49} & 74.13{\scriptsize$\pm$.65} & 62.91{\scriptsize$\pm$.81} & 51.97{\scriptsize$\pm$.78} & 53.35{\scriptsize$\pm$.78} \\
    \cmidrule{2-11}
    &\multirow{2}{*}{MSL} & \,\,SimCLR\,\, & 97.39{\scriptsize$\pm$.29} & 91.13{\scriptsize$\pm$.37} & 47.28{\scriptsize$\pm$.64} & 26.43{\scriptsize$\pm$.44} & 84.32{\scriptsize$\pm$.53} & 67.82{\scriptsize$\pm$.82} & 56.97{\scriptsize$\pm$.81} & 73.98{\scriptsize$\pm$.79} \\
    && BYOL & 97.87{\scriptsize$\pm$.26} & 89.27{\scriptsize$\pm$.43} & 50.18{\scriptsize$\pm$.67} & 30.82{\scriptsize$\pm$.49} & 77.55{\scriptsize$\pm$.60} & 66.32{\scriptsize$\pm$.78} & 64.78{\scriptsize$\pm$.81} & 77.18{\scriptsize$\pm$.75} \\
    \midrule
    &\multirow{2}{*}{SSL} & SimCLR & 97.33{\scriptsize$\pm$.36} & 94.99{\scriptsize$\pm$.28} & 49.43{\scriptsize$\pm$.59} & 24.74{\scriptsize$\pm$.43} & 82.54{\scriptsize$\pm$.56} & 61.30{\scriptsize$\pm$.84} & 46.21{\scriptsize$\pm$.76} & 48.43{\scriptsize$\pm$.81} \\
    \rotatebox[origin=c]{90}{\makecell[l]{\!\!\!\!\!\!\!\!\!\!\!\!\!\!Repeated}\!\!\!\!\!\!}&& BYOL & 97.24{\scriptsize$\pm$.27} & 89.92{\scriptsize$\pm$.41} & 49.95{\scriptsize$\pm$.63} & 29.64{\scriptsize$\pm$.47} & 74.25{\scriptsize$\pm$.63} & 62.86{\scriptsize$\pm$.80} & 52.05{\scriptsize$\pm$.79} & 55.73{\scriptsize$\pm$.77} \\
    \cmidrule{2-11}
    &\multirow{2}{*}{MSL} & \,\,SimCLR\,\, & 97.12{\scriptsize$\pm$.31} & 90.73{\scriptsize$\pm$.38} & 46.23{\scriptsize$\pm$.62} & 26.23{\scriptsize$\pm$.43} & 83.94{\scriptsize$\pm$.56} & 67.16{\scriptsize$\pm$.81} & 56.88{\scriptsize$\pm$.82} & 73.31{\scriptsize$\pm$.81} \\
    && BYOL & 97.71{\scriptsize$\pm$.27} & 86.99{\scriptsize$\pm$.52} & 49.60{\scriptsize$\pm$.69} & 30.37{\scriptsize$\pm$.50} & 78.41{\scriptsize$\pm$.59} & 69.56{\scriptsize$\pm$.80} & 64.23{\scriptsize$\pm$.82} & 75.00{\scriptsize$\pm$.78} \\
    \bottomrule
\end{tabular}}
\end{table*}

\clearpage
\section{t-SNE Visualization of Pre-trained Models on the Target Domains}\label{appx:tsne}
\begin{figure}[!h]
    \centering
    \begin{subfigure}[h]{.49\linewidth}
         \centering
         \includegraphics[width=\linewidth]{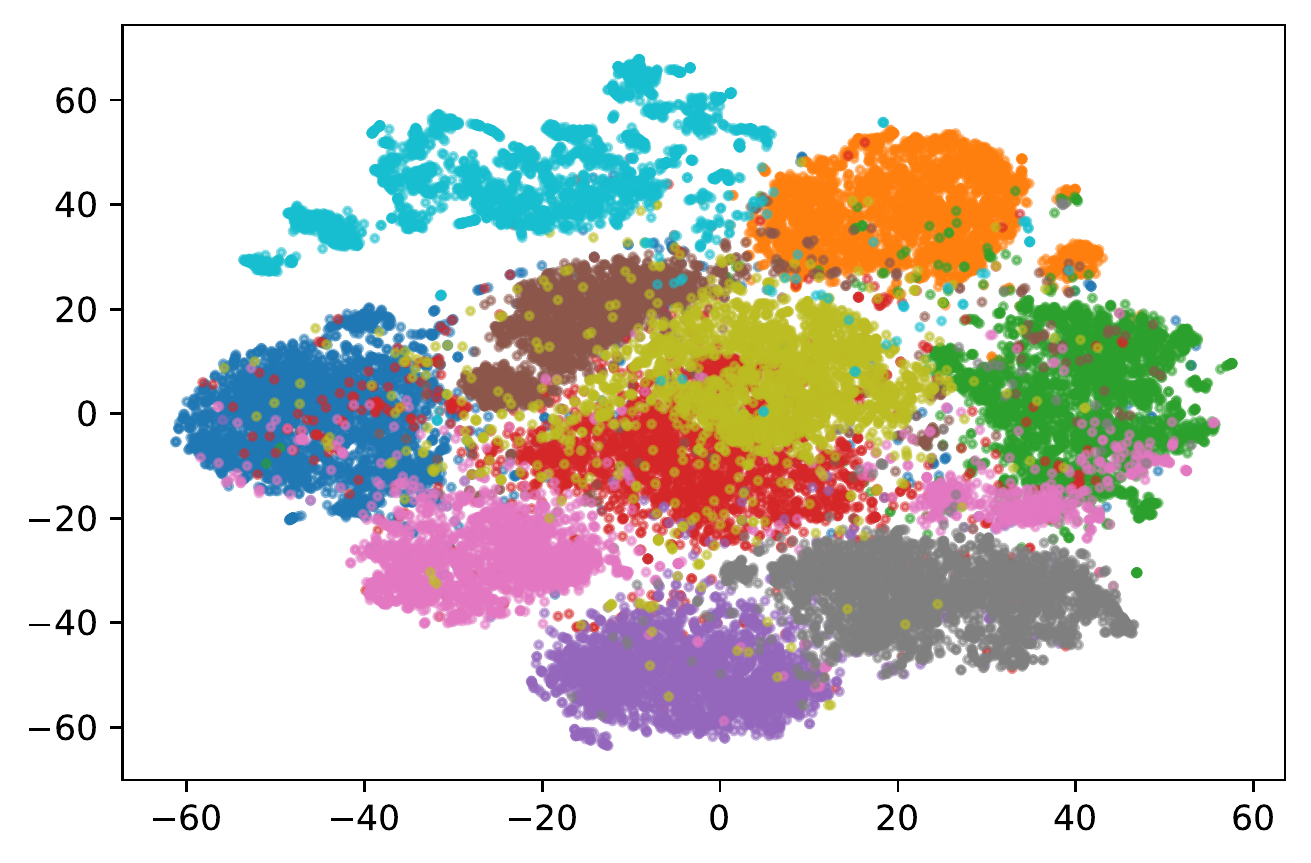}
         \vspace*{-0.6cm}
         \caption{EuroSAT (SL)}
         \label{fig:tsne_euro_sl}
    \end{subfigure}
    \begin{subfigure}[h]{.49\linewidth}
         \centering
         \includegraphics[width=\linewidth]{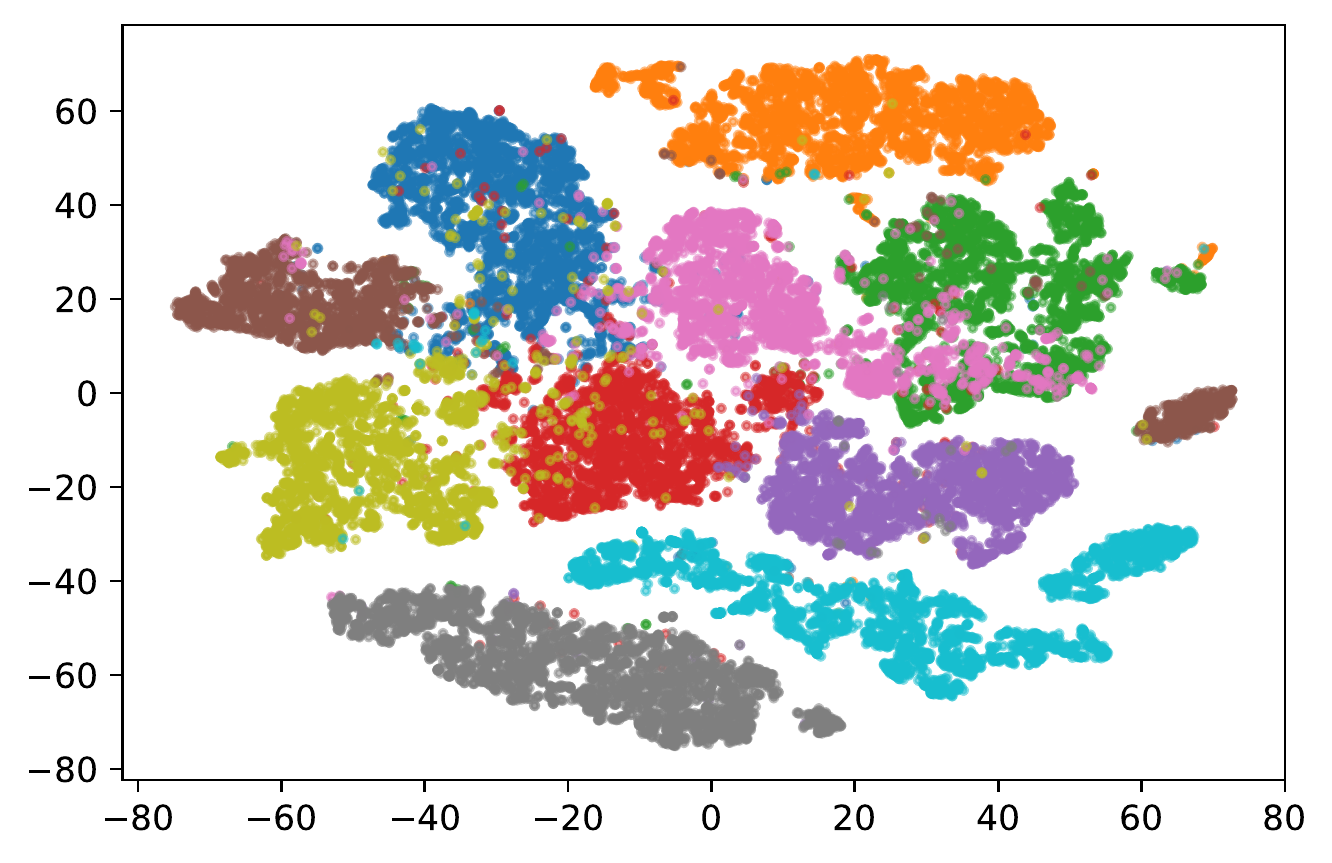}
         \vspace*{-0.6cm}
         \caption{EuroSAT (SSL)}
         \label{fig:tsne_euro_ssl}
    \end{subfigure}
    
    \begin{subfigure}[h]{.49\linewidth}
         \centering
         \includegraphics[width=\linewidth]{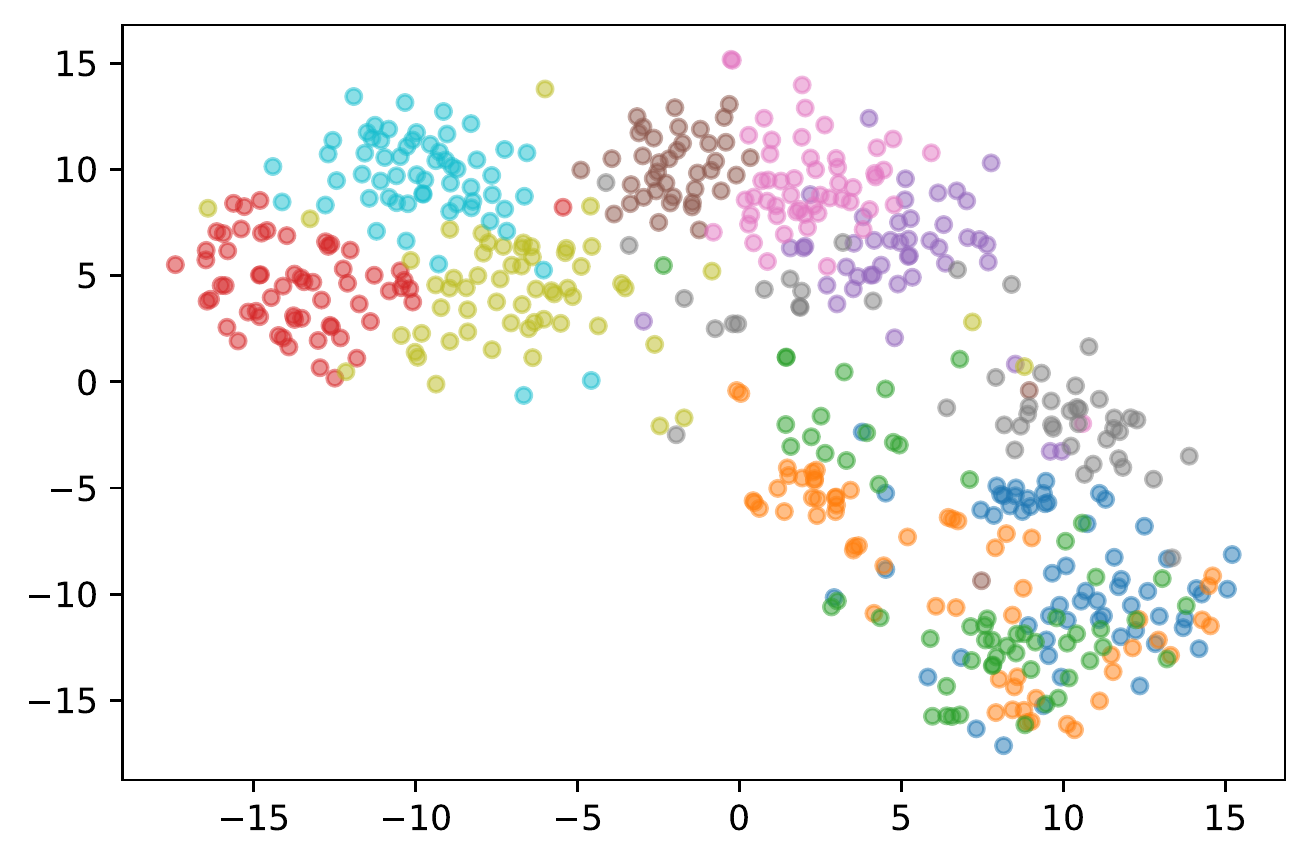}
         \vspace*{-0.6cm}
         \caption{CUB (SL)}
         \label{fig:tsne_cub_sl}
    \end{subfigure}
    \begin{subfigure}[h]{.49\linewidth}
         \centering
         \includegraphics[width=\linewidth]{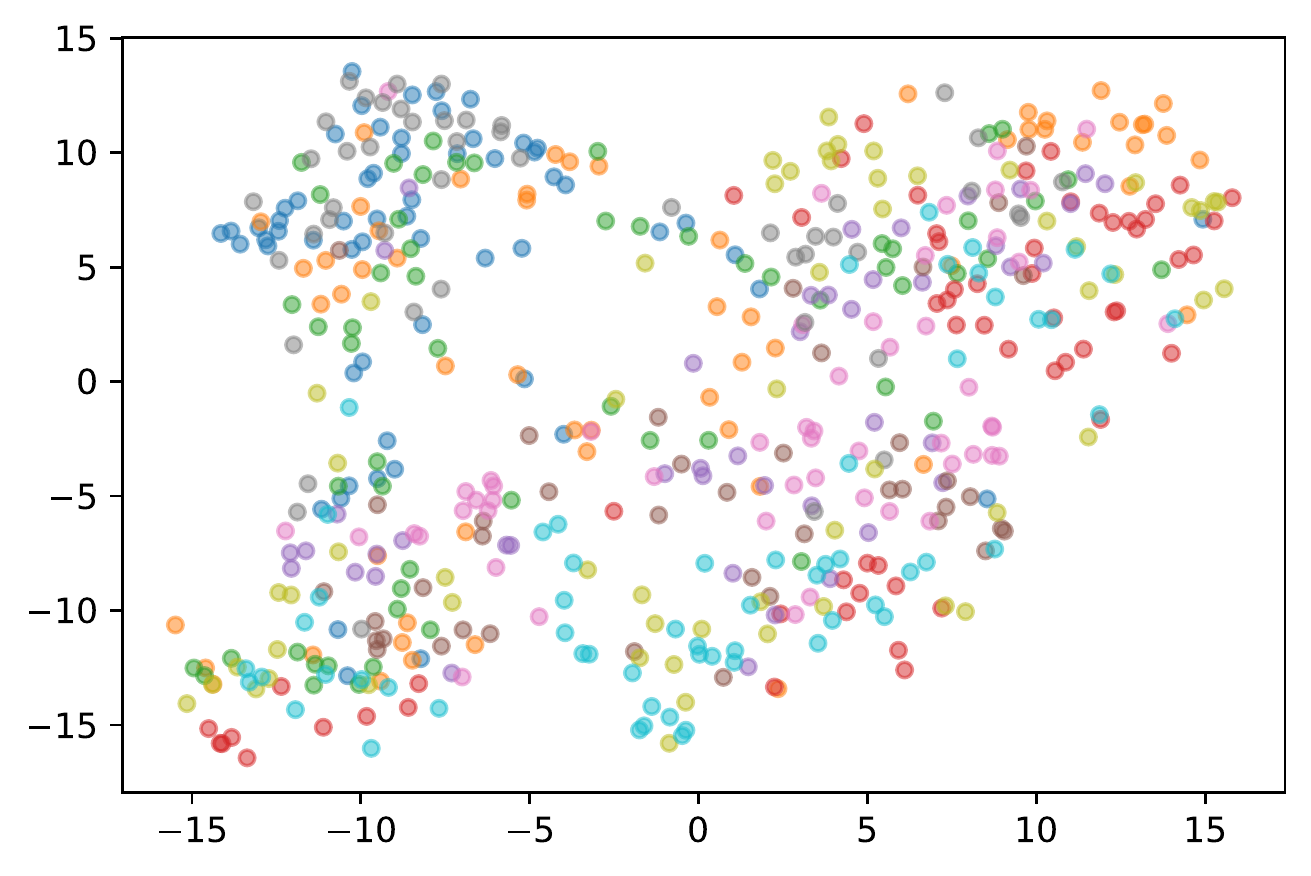}
         \vspace*{-0.6cm}
         \caption{CUB (SSL)}
         \label{fig:tsne_cub_ssl}
    \end{subfigure}
     
     \vspace*{-0.1cm}
     \caption{t-SNE visualization for each target dataset. ResNet18 is used as a backbone and ImageNet is used as the source dataset. Note that 10 classes are randomly sampled for the CUB dataset because CUB has 200 classes in total.}
     \label{fig:tsne}
\end{figure}

We provide t-SNE on the target datasets to visualize the difference of using the two types of pre-training models: supervised learning on the source domain and self-supervised learning on the target domain. Figure \ref{fig:tsne} describes t-SNE visualization of representations through SL/SSL models on the EuroSAT and CUB datasets. It is shown that on the EuroSAT dataset, representations through the SSL are clustered better than those through the SL; however, on the CUB dataset, representations through the SL are clustered better than those through the SSL. This implies that clustering ability of extractors trained through SL/SSL is related to the few-shot performance.

\clearpage
\section{Same Domain FSL Experiments}\label{appx:same-domain-fsl-results}

For the same domain FSL, although label space of the source and target datasets is still not shared, it is expected that SL is a better strategy and MSL improves the few-shot performance because MSL works like multi-task learning (MTL), improving the generalization ability. This is because the source and target datasets were collected in the same way.

To explain same-domain FSL experiments based on \emph{domain similarity} and \emph{few-shot difficulty}, we first provide them for the two datasets:

\begin{itemize}
    \item Domain Similarity
    \begin{itemize}
        \item miniImageNet $\leftrightarrow$ miniImageNet-test: 0.832
        \item tieredImageNet $\leftrightarrow$ tieredImageNet-test: 0.869
    \end{itemize}
    \item Few-shot Difficulty ($k$ = 5)
    \begin{itemize}
        \item miniImageNet-test: 0.467
        \item tieredImageNet-test: 0.414
    \end{itemize}
\end{itemize}

Interestingly, domain similarity of the miniImageNet-test dataset (0.832) is larger than that of every other benchmark, except for the Places (0.840). The Places dataset is found to be most similar to miniImageNet source. For the tieredImageNet-test dataset, domain similarity with tieredImageNet source (0.869) is larger than that of every other benchmark. Few-shot difficulty of miniImageNet-test is 0.467, which means slightly more difficult than Places and easier than Plantae. In addition, few-shot difficulty of tieredImageNet-test is 0.414, which means more difficult than EuroSAT and easier than Places. 

Table \ref{tab:homo_performance} describes the few-shot performance under the same domain; miniImageNet $\rightarrow$ miniImageNet-test and tieredImageNet $\rightarrow$ tieredImageNet-test. As expected, they have large similarity and high difficulty. Therefore, (1) SL is more powerful than SSL, (2) MSL is a better strategy than both SL and SSL, and (3) two-stage pre-training boosts performance.

\begin{table*}[h]
\caption{$5$-way $k$-shot FSL performance of the models pre-trained: miniImageNet $\rightarrow$ miniImageNet-test and tieredImageNet $\rightarrow$ tieredImageNet-test. We report the average accuracy and its 95\% confidence interval over 600 few-shot episodes. B and S indicate base and strong augmentations, respectively. For MSL and SL$\rightarrow$MSL, $\gamma$ is set to 0.875.}\label{tab:homo_performance}
\vskip 0.10in
\centering
\footnotesize\addtolength{\tabcolsep}{-1pt}
\begin{tabular}{c|c|c|cc|cc}
    \toprule
    Pre-train & \multirow{2}{*}{Method} & \multirow{2}{*}{Aug.} & \multicolumn{2}{c|}{miniImageNet} & \multicolumn{2}{c}{tieredImageNet} \\
    Scheme & & & $k$=1 & $k$=5 & $k$=1 & $k$=5 \\
    \midrule
    \multirow{2}{*}{SL} & \multirow{2}{*}{Default}      & B & 54.89{\scriptsize$\pm$.80} & 77.92{\scriptsize$\pm$.59} & 60.98{\scriptsize$\pm$.92} & 78.88{\scriptsize$\pm$.68} \\
    &    & S & 57.30{\scriptsize$\pm$.81} & 77.32{\scriptsize$\pm$.65} & 60.77{\scriptsize$\pm$.92} & 78.36{\scriptsize$\pm$.71} \\
    \midrule
    \multirow{4}{*}{SSL} & \multirow{2}{*}{SimCLR}      & B & 42.69{\scriptsize$\pm$.88} & 60.42{\scriptsize$\pm$.81} & 51.63{\scriptsize$\pm$.93} & 67.62{\scriptsize$\pm$.84} \\
    &                              & S &
    54.39{\scriptsize$\pm$.92} & 71.62{\scriptsize$\pm$.79} & 66.67{\scriptsize$\pm$1.02} &
    80.60{\scriptsize$\pm$.75} \\
    & \multirow{2}{*}{BYOL}        & B & 39.32{\scriptsize$\pm$.76} & 58.36{\scriptsize$\pm$.80} & 48.11{\scriptsize$\pm$.88} & 69.72{\scriptsize$\pm$.80} \\
    &                              & S & 44.71{\scriptsize$\pm$.80} & 63.66{\scriptsize$\pm$.77} & 59.00{\scriptsize$\pm$.97} & 78.59{\scriptsize$\pm$.70} \\
    \midrule
    \multirow{2}{*}{MSL} & SimCLR  & S & 63.15{\scriptsize$\pm$.85} & 80.03{\scriptsize$\pm$.63} & 74.24{\scriptsize$\pm$.94} & 86.90{\scriptsize$\pm$.64} \\
    & BYOL                        & S & 58.59{\scriptsize$\pm$.84} & 78.17{\scriptsize$\pm$.65} & 75.25{\scriptsize$\pm$.92} & 88.37{\scriptsize$\pm$.58} \\
    \midrule
    \multirow{2}{*}{\!\!\!SL$\rightarrow$SSL\!\!\!} & SimCLR  & S & 65.48{\scriptsize$\pm$.89} & 83.84{\scriptsize$\pm$.59} & 68.27{\scriptsize$\pm$1.04} & 81.22{\scriptsize$\pm$.75} \\
    & BYOL                        & S & 55.76{\scriptsize$\pm$.83} & 78.43{\scriptsize$\pm$.61} & 65.78{\scriptsize$\pm$.99} &
    80.93{\scriptsize$\pm$.66} \\
    \midrule
    \multirow{2}{*}{\!\!\!SL$\rightarrow$MSL\!\!\!} & SimCLR  & S & \textbf{67.24}{\scriptsize$\pm$.86} & \textbf{85.02}{\scriptsize$\pm$.52} & 74.14{\scriptsize$\pm$.96} & 87.38{\scriptsize$\pm$.62} \\
    & BYOL                        & S & 61.10{\scriptsize$\pm$.82} & 82.35{\scriptsize$\pm$.58} & \textbf{75.47}{\scriptsize$\pm$.90} & \textbf{88.72}{\scriptsize$\pm$.58} \\
    \bottomrule
\end{tabular}
\end{table*}

\end{document}